\documentclass[10pt,twocolumn,letterpaper]{article}

\usepackage[pagenumbers]{style/cvpr} 

\usepackage[accsupp]{axessibility}  

\usepackage{graphicx}
\usepackage{amsmath}
\usepackage{amssymb}
\usepackage{booktabs}
\usepackage[utf8]{inputenc} 
\usepackage[T1]{fontenc}    
\usepackage{url}            
\usepackage{booktabs}       
\usepackage{amsfonts}       
\usepackage{nicefrac}       
\usepackage{microtype}      
\usepackage{xcolor}         

\usepackage{graphicx}
\usepackage{comment}
\usepackage{color}
\usepackage{amsmath}
\usepackage{amssymb}
\usepackage[ruled,linesnumbered]{algorithm2e}
\usepackage{wrapfig}
\usepackage{multirow}
\usepackage{textcomp}

\usepackage[pagebackref,breaklinks,colorlinks]{hyperref}

\usepackage[capitalize]{cleveref}
\crefname{section}{Sec.}{Secs.}
\Crefname{section}{Section}{Sections}
\Crefname{table}{Table}{Tables}
\crefname{table}{Tab.}{Tabs.}

\newlength{\wid}
\newlength{\mrg}
\newlength{\mrgv}

\def\confName{CVPR}
\def\confYear{2023}

\begin{document}

\title{Sphere-Guided Training of Neural Implicit Surfaces}

\author{
    Andreea Dogaru$^{1, 2}$ \quad 
    Adrei Timotei Ardelean$^{1, 2}$ \quad 
    Savva Ignatyev$^1$ \\
    Egor Zakharov$^{1}$ \quad 
    Evgeny Burnaev$^{1, 3}$ \vspace{0.2cm} \\
    $^1$Skolkovo Institute of Science and Technology \quad $^2$Friedrich-Alexander-Universität Erlangen-Nürnberg \\
    $^3$Artificial Intelligence Research Institute
}

\twocolumn[{
\renewcommand\twocolumn[1][]{#1}
\maketitle
\begin{center}
    \centering
    \includegraphics[width=\textwidth]{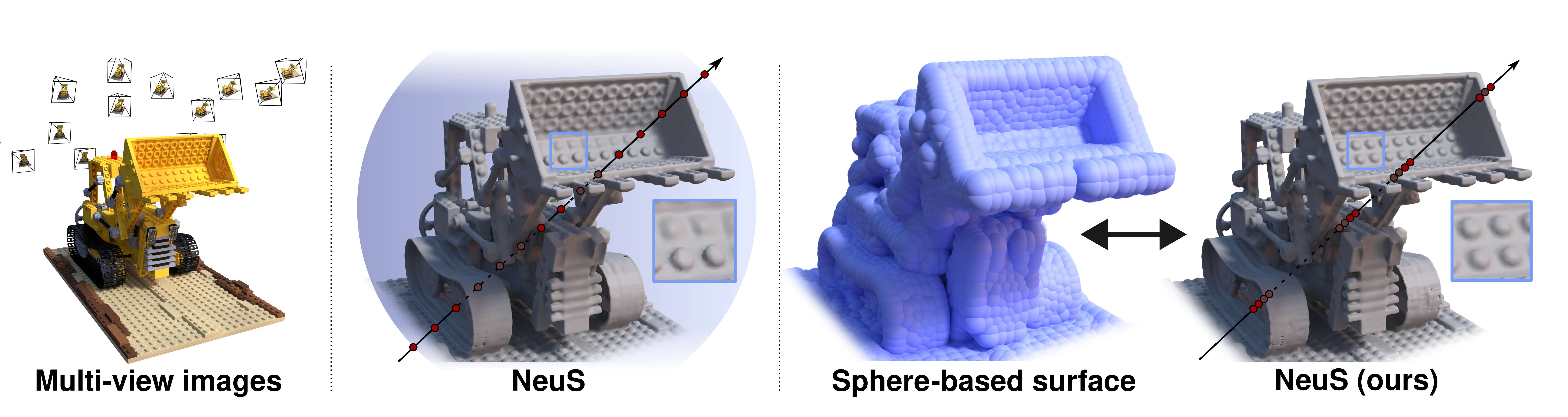}
    \captionof{figure}{We propose a new hybrid approach for learning neural implicit surfaces from multi-view images. In previous methods, the volumetric ray marching training procedure is applied for the whole bounding sphere of the scene (middle left). Instead, we train a coarse sphere-based surface reconstruction (middle right) alongside the neural surface to guide the ray sampling and ray marching. As a result, our method achieves an increased sampling efficiency by pruning empty scene space and better quality of reconstructions (right).}
    \label{fig:teaser}
\end{center}
}]


\makeatletter
\newcommand\blfootnote[1]{%
  \begingroup
  \renewcommand\thefootnote{}\footnote{#1}%
  \addtocounter{footnote}{-1}%
  \endgroup
}
\makeatother

\begin{abstract}
In recent years, neural distance functions trained via volumetric ray marching have been widely adopted for multi-view 3D reconstruction. These methods, however, apply the ray marching procedure for the entire scene volume, leading to reduced sampling efficiency and, as a result, lower reconstruction quality in the areas of high-frequency details. In this work, we address this problem via joint training of the implicit function and our new coarse sphere-based surface reconstruction. We use the coarse representation to efficiently exclude the empty volume of the scene from the volumetric ray marching procedure without additional forward passes of the neural surface network, which leads to an increased fidelity of the reconstructions compared to the base systems. We evaluate our approach by incorporating it into the training procedures of several implicit surface modeling methods and observe uniform improvements across both synthetic and real-world datasets.
Our codebase can be accessed via the project page$^{\dagger}$.
\end{abstract}

\blfootnote{$^\dagger$ \href{https://andreeadogaru.github.io/SphereGuided}{https://andreeadogaru.github.io/SphereGuided}}


\section{Introduction}

The task of multi-view 3D reconstruction remains the focus of modern computer vision and graphics research. 
It has major practical significance in AR/VR metaverses, synthetic media, medical imaging, and the special effects industry. This task is classically addressed via the multi-view stereo (MVS) reconstruction systems~\cite{Bleyer2011PatchMatchS, Campbell2008UsingMH, Esteban2003SilhouetteAS, Furukawa2010AccurateDA, Huang2018DeepMVSLM, Schnberger2016PixelwiseVS, zhang2020visibility}, which estimate the underlying scene geometry in the form of a point cloud using a photometric consistency between the different views. However, in recent years they have been largely phased out by the methods that represent the scene as neural implicit fields~\cite{darmon2022improving, Lassner2021PulsarES, Liu2020NeuralSV, Lombardi2019NeuralV, Lombardi2021MixtureOV, Mildenhall2020NeRFRS, Oechsle2021UNISURFUN, Sun2021DirectVG, Wang2021IsoPointsON, Wang2021NeuSLN, Xu2022PointNeRFPN, Yariv2021VolumeRO, yariv2020multiview, Yu2021PlenOctreesFR}. These approaches have multiple advantages compared to the classical MVS. For example, they can easily accommodate non-Lambertian and texture-less surfaces~\cite{Furukawa2015MultiViewSA}, are good at interpolating unseen parts of the geometry by leveraging regularization~\cite{kim2022infonerf}, and at the same time can achieve an impressive quality of renders~\cite{barron2021mipnerf}.

This work focuses on improving the subset of such methods specialized in opaque surface reconstruction~\cite{darmon2022improving, Oechsle2021UNISURFUN, Wang2021NeuSLN, Yariv2021VolumeRO}. Most of these approaches employ neural signed distance fields~\cite{park2019deepsdf} (SDFs) trained using volumetric ray marching~\cite{darmon2022improving, Wang2021NeuSLN, Yariv2021VolumeRO}. The training step of this procedure contains two stochastic elements: sampling a \emph{ray} corresponding to a training pixel and sampling a set of \emph{points} along the ray to approximate the color integral. The sampling efficiency at these steps largely determines the resulting quality of the reconstructions. While in the abovementioned methods the training rays are selected uniformly within the scene volume, their point sampling procedure typically employs a multi-stage importance~\cite{Wang2021NeuSLN} or uncertainty~\cite{darmon2022improving, Yariv2021VolumeRO} sampling to improve the accuracy of the reconstructions.

At the same time, it was shown~\cite{Wang2021IsoPointsON} that neural signed distance fields benefit from the surface-based sampling of rays for surface rendering methods, such as IDR~\cite{yariv2020multiview}, which the modern multi-view reconstruction systems do not incorporate. Additionally, some of the novel-view synthesis works~\cite{Liu2020NeuralSV, Lombardi2021MixtureOV, Yu2021PlenOctreesFR} successfully combined a simple two-stage coarse-to-fine sampling with explicitly defined surface guides to achieve a better rendering quality, as opposed to using the sophisticated multi-stage sampling procedures of the surface reconstruction methods. To guide the ray marching, they use explicit coarse surface approximations in the form of a set of volumetric primitives~\cite{Lombardi2021MixtureOV} or sparse octrees~\cite{Liu2020NeuralSV, Yu2021PlenOctreesFR}. However, these methods require a complete scene reconstruction to fit such an approximation~\cite{Lombardi2021MixtureOV, Yu2021PlenOctreesFR} or employ a heuristic optimization procedure~\cite{Liu2020NeuralSV} which we show performs poorly for the surface reconstruction task.

Inspired by these approaches, we improve the existing surface reconstruction methods' ray sampling and marching procedures using explicitly defined coarse representations. We propose training a coarse reconstruction as a sphere cloud which guides both sampling steps during volume rendering. We also propose a new optimization approach for coarse reconstruction based on gradient descent, which allows us to train it alongside the implicit surface field. Additionally, we introduce a point resampling scheme, which prevents the spheres from getting stuck in the local minima, and a repulsion mechanism that ensures high degrees of exploration of the reconstructed surface. Finally, we provide empirical evidence of the proposed method's applicability to different approaches for implicit surface modeling. Specifically, we pair our method with several modern systems~\cite{darmon2022improving, Oechsle2021UNISURFUN, Wang2021NeuSLN, Yariv2021VolumeRO} for surface reconstruction and observe uniform improvements in the resulting quality across multiple 3D reconstruction benchmarks.

\section{Related works}

\textbf{Implicit volumetric representations.} Neural implicit representations have gained much attention in recent years for the problem of multi-view 3D surface reconstruction. Their widespread adoption started after several works have introduced training approaches based on the differentiable rendering of implicit functions. They initially relied on the surface rendering~\cite{sitzmann2019scene, niemeyer2020differentiable, yariv2020multiview} procedure, where the pixel's color is approximated using the radiance of a single point in the volume. However, they were recently phased out by the training procedures based on the volume rendering with multiple samples via ray marching. Introduced in a seminal work on novel view synthesis, Neural Radiance Fields (NeRFs)~\cite{Mildenhall2020NeRFRS}, the volumetric ray marching has been later adapted~\cite{Oechsle2021UNISURFUN, Yariv2021VolumeRO, Wang2021NeuSLN} to the problem of surface modeling since it significantly improved the reconstruction quality. The ray marching procedure estimates the color along the ray using the volume rendering integral, approximated as a sum of the weighted radiances at multiple points throughout the volume. The aforementioned works employ methods based on importance~\cite{Wang2021NeuSLN}, uncertainty~\cite{Yariv2021VolumeRO} or surface intersection-based~\cite{Oechsle2021UNISURFUN} sampling to obtain this set of points, increasing the approximation accuracy compared to more simple strategies, such as uniform sampling. 

We propose a new hybrid surface representation that improves ray marching by limiting the sampling space to a volume coarsely bounding the scene. This is used in conjunction with the ray marching mechanism of the base neural reconstruction method which further optimizes the selection of samples around the reconstructed surface.
We also use this hybrid surface representation to guide the sampling of the training rays, improving the quality of reconstructions given the same training time. 

\textbf{Hybrid representations.} 
To improve the training efficiency and rendering frame rate, multiple hybrid representations~\cite{Aliev2020NeuralPG, fridovich2022plenoxels, Lombardi2019NeuralV, Mller2022InstantNG, Rakhimov2022NPBGAN, ruckert2022adop, Yu2021PlenOctreesFR} have been proposed, which jointly optimize the implicit and explicit representations. These methods employ point clouds~\cite{Aliev2020NeuralPG, Rakhimov2022NPBGAN, ruckert2022adop}, hash tables~\cite{Mller2022InstantNG}, sparse voxel grids~\cite{fridovich2022plenoxels, Lombardi2019NeuralV, Yu2021PlenOctreesFR}, and volumetric primitives~\cite{Lassner2021PulsarES} to improve both the training and rendering procedures in terms of either their speed or the resulting quality. Below we discuss the methods that are most closely related to our approach.

Iso-Points~\cite{Wang2021IsoPointsON} introduced joint optimization of the signed distance functions with a point-based surface representation. In our approach, we use a \emph{sphere-based} representation, which allows us to sample both the rays \emph{and} points along these rays to lie near the surface, thus modifying both the ray-sampling and the ray-marching procedures. Closely related to our work are Neural Sparse Voxel Fields~\cite{Liu2020NeuralSV} and Neural 3D Reconstruction in the Wild~\cite{sun2022neuconw} systems. They both employ sparse voxel grids to guide the ray marching. However, the method in~\cite{Liu2020NeuralSV} uses a greedy optimization strategy to train these representations, which, as we show, results in an inferior reconstruction quality compared to our gradient-based training. Compared to \cite{sun2022neuconw}, our method does not employ the initialization using a sparse point-cloud, and it trains the guiding reconstruction from scratch.

\def\P{\mathbf{P}}
\def\p{\mathbf{p}}
\def\x{\mathbf{x}}
\def\surf{\mathcal{S}}
\def\S{\mathbf{S}}
\def\v{\mathbf{v}}
\def\R{\mathbb{R}}
\def\o{\mathbf{o}}
\def\c{\mathbf{c}}

\section{Method}

Our approach addresses a multi-view 3D reconstruction problem. The goal is to estimate the surface of a scene, denoted as $\surf$, given a collection of images with the associated camera parameters. In our case, this surface is extracted as a level set of the learned implicit representations: either a signed distance function (SDF) or an occupancy field. In this section, we begin by describing the volume rendering approach utilized by most state-of-the-art methods. Then, we show how a learnable sphere cloud $\S$ could be used to improve the volume rendering-based training process and finally describe the optimization pipeline for the sphere cloud itself.

\begin{figure}
    \centering
    \includegraphics[width=0.45\textwidth]{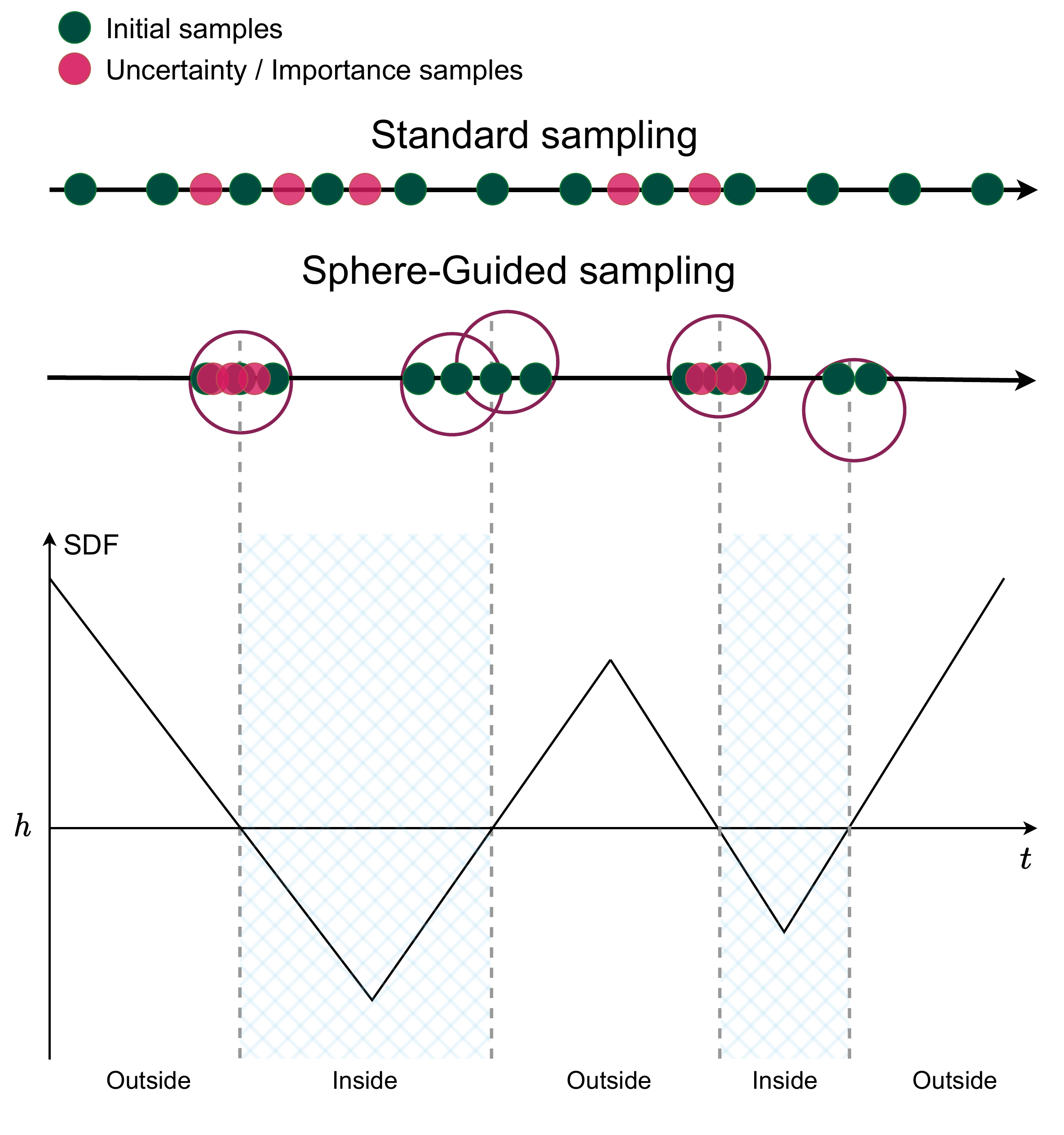}
    \caption{
    Our method works by filtering the samples along the ray that lie outside of the surface region, approximated by a trainable sphere cloud. Such filtering improves the sample efficiency in the optimization process and allows the implicit function to converge to a better optimum.}
    \label{fig:sampling_scheme}
    \vspace{-0.3cm}
\end{figure}

\subsection{Volume rendering}

We assume the underlying implicit model $f$ to represent the geometry of the surface, and that there is a transformation that maps it to a surface density function $\sigma : \R^3 \to \R^+$, defined at each point $\x$ in the volume.

In order to render the surface defined by $\sigma$ via volumetric rendering, we first need to consider a ray $\p(t) = \o + t\v,\ t \ge 0$, emanated from the camera origin $\o \in \R^3$ in the direction $\v \in \mathbb{S}^2$, and the corresponding color $C(\o, \v)$ of the pixel on the image plane of that camera. We also need to define a radiance field $c: \R^3 \times \mathbb{S}^2 \to \R^3$, which produces a view-dependent color at each point in the volume. The observed color $C(\o, \v)$ can then be expressed as the following integral along the ray:
\begin{equation}
    C(\o, \v) = \int_0^{+\infty} w(t) c(\p(t), \v) dt,\
    \label{eq:color}
\end{equation}
where $w(t)$ is the probability of a ray terminating at $\p(t)$, which can be derived from the density $\sigma$:
\begin{equation}
    w(t) = T(t) \sigma(t),\quad T(t) = \exp \bigg( - {\int_0^t \sigma(s) ds} \bigg).
    \label{eq:weight}
\end{equation}
In practice, the color integral is approximated by evaluating the density and radiance at a set of $n$ sampled points $\mathcal{P} = \{ \p_i = \o + t_i \v \}_{i=1}^n$ using the discretized version~\cite{Mildenhall2020NeRFRS} of the equations above:
\begin{equation}
    \hat{C}( \mathbf{o, v} ) = \sum_{i=1}^{N} T_i \alpha_i c_i,\quad T_i = \prod_{j=1}^{i-1} (1 - \alpha_j).
    \label{eq:render}
\end{equation}
Here $T_i$ denotes the accumulated transmittance and $\alpha_i$ --- the opacity value at point $\p_i$, which can also be estimated from the density function via the following formula:
\begin{equation}
    \alpha_i = 1 - \exp \bigg( - \int_{t_i}^{t_{i+1}} \sigma(t) dt \bigg).
\end{equation}

\subsection{Sphere-guided volume rendering}

The sampling strategy for the points $\mathcal{P}$ has a major impact on the resulting reconstructions since it directly affects the approximation quality of eq.~\ref{eq:color}. To improve it, some methods~\cite{Oechsle2021UNISURFUN} employ the root-finding procedure to obtain the first intersection with the surface along the ray and sample more points near it. Other methods~\cite{darmon2022improving, Wang2021NeuSLN, Yariv2021VolumeRO} are first estimating a dense set of proposals $\mathcal{T}$ via importance or uncertainty sampling. Then, $\mathcal{P}$ is obtained either via inverse transform sampling by evaluating the density $\sigma$ at the proposals $\mathcal{T}$ and normalizing it along the ray~\cite{darmon2022improving, Yariv2021VolumeRO}, or in some cases by using an entire set of proposals~\cite{Wang2021NeuSLN}.

\begin{algorithm}[ht]
    \caption{Sphere-guided sampling.}
    \SetAlgoLined
    \KwIn{ray $\p(t)$, spheres $\S$, $\#$samples $n$}
    Initialize a set of intervals $\mathcal{I} = \varnothing$ \\
    \For{$\S_i \in \S$}{
        Add sphere-ray intersection to $\mathcal{I}$:
        $\mathcal{I} := \mathcal{I}\, \cup\, \S_i \cap \p(t)$
    }
    Find a minimal set of intervals $\big\{ [s_k, t_k] \big\}:\, \bigcup_k [\p(s_k), \p(t_k)] = \mathcal{I}$ \\
    Initialize a set of points $\mathcal{T}_0 = \varnothing$ \\
    \For{$k = 1 \dots K$}{
        Set $n_k := \lfloor n / (t_k - s_k) \rfloor$ \\
        $\mathcal{T}_0 := \mathcal{T}_0\, \cup\, \text{linspace}(s_k, t_k, n_k)$
    }
    Obtain $\mathcal{T}$ using $\mathcal{T}_0$ and the sampling method of choice \\
    \Return $\mathcal{T}$
    \label{alg:sampling}
\end{algorithm}
\begin{figure*}[!t]
    \centering    
    \includegraphics[width=\textwidth]{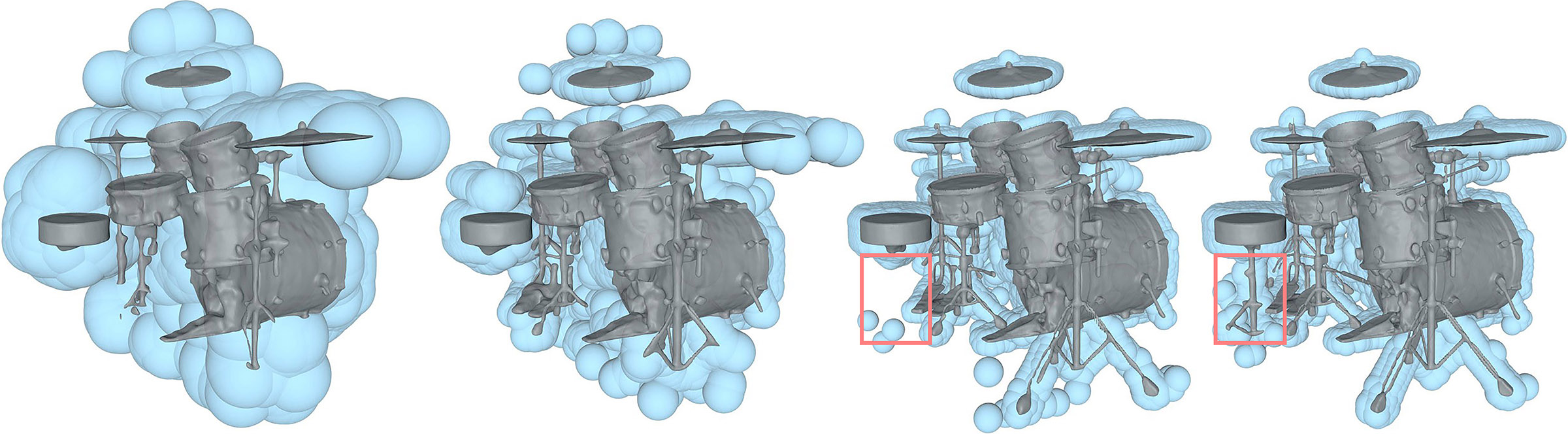}
    \begin{tabular}{cccc}
         \hspace{0.8cm} (a) $5 \cdot 10^{4}$ \text{iterations} \hspace{1cm} & (b) $7.5 \cdot 10^{4}$ \text{iterations} \hspace{0.6cm} & (c) $1.5 \cdot 10^{5}$ \text{iterations} \hspace{0.6cm} & (d) $3 \cdot 10^{5}$ \text{iterations} \hspace{0.8cm}
    \end{tabular}
    \caption{Visualization of the training process. Initially, we assign a large radius to all spheres in the cloud (a) and gradually reduce it during the optimization down to a minimum value (c). Our proposed repulsion loss prevents the clumping of the spheres and encourages exploration, which results in an improved reconstruction of the thin surfaces (d).}
    \label{fig:training_vis}
    \vspace{-0.3cm}
\end{figure*}

To further improve the efficiency of both proposal sampling and root-finding procedures, we utilize a set of guiding spheres $\S$ which cover the object's surface. They allow us to ensure that the training samples $\mathcal{P}$ are mainly generated from the areas of interest, making the implicit surface function converge to a better optimum, especially for the scenes with high-frequency details. We achieve that by applying both the root-finding and proposal sampling procedures only within the volume, defined by the sphere cloud $\S$, as illustrated in Figure \ref{fig:sampling_scheme}. For the sampling of proposals $\mathcal{T}$, our method is described in the Algorithm~\ref{alg:sampling}, while the details of the modified root-finding approach can be found in the supplementary materials. In short, the algorithm first intersects a given ray with the sphere cloud, yielding a set of intersections $\mathcal{I}$. Then it finds the minimum coverage of the intersections. That is, $\big\{[s_k, t_k] \big\}$ is the minimal set of segments with union $\mathcal{I}$. Then each of these segments is linearly sampled to obtain the initial set of proposals $\mathcal{T}_0$. This set is finally upsampled using the base method.

\subsection{Sphere cloud optimization}

At the beginning of training, we initialize the sphere cloud $\S$ of size $M$ with centers $\{ \c_i \}_{i=1}^M$, uniformly distributed across the volume of the scene, and set the radii of the spheres to an initial value $r_\text{max}$. The training then proceeds by alternating the updates of the sphere cloud and the implicit function. Importantly, we only update the sphere centers $\c_i$ via an optimization-based process and rely on scheduling their radii to decrease from the initial value $r_\text{max}$ to the minimum $r_\text{min}$ via a fixed schedule. Also, in our approach, all spheres in the cloud are assigned the same radius value. Figure \ref{fig:training_vis} illustrates the sphere cloud optimization during the training process. 

The main learning signal for the sphere centers comes from moving them towards the estimated surface $\hat\surf$, which is defined as an $h$-level set of the implicit function $f: \hat\surf = \{ \x \in \R^3 \ | \ f(\x) = h \}$, where $h$ depends on the type of the function (e.g., for SDF $h = 0$). This can be formulated as a following loss:
\begin{equation}
    \mathcal{L}_\text{surf} = \sum_{i=1}^{M} \big\| f(\c_i) - h \big\|_2.
\end{equation}

This objective ensures that the sphere centers lie in the proximity of the reconstructed surface, i.e., maximizes the precision. However, it does not guarantee that the point cloud covers an entire object's surface. To address that, we design a repulsion term that prevents the neighboring spheres from clumping together and encourages exploration of the entire surface region:
\begin{equation}
\label{eq:repulsion}
    \mathcal{L}_\text{rep} = \sum_{i=1}^{M}\sum_{j \in K(i)} r_n \dfrac{ \mathbb{I}{(||\c_j - \c_i||_2 < d)} }{ ||\c_j - \c_i||_2 },
\end{equation}
where $K(i)$ denotes the indices of the $k$-nearest spheres to $\S_i$, $r_n$ is the current radius of the spheres, and $d$ is a hyperparameter, which sets the maximum distance for the repulsion. Since the magnitude of this loss depends on the current radius of the spheres, the repulsion has more effect in the beginning of the training, encouraging better exploration of the scene volume.

Our final objective for optimization of the centers of the spheres is the following:
\begin{equation}
    \mathcal{L} = \mathcal{L}_\text{surf} + \lambda \mathcal{L}_\text{rep}.
\end{equation}
The radius scheduling in our method defines the exploration-exploitation trade-off and, in principle, could be picked separately for each scene. However, we found out that the following exponential schedule works well in most cases:
\begin{equation}
    r_n = \text{max}(r_{\text{max}} \, e^{-n \beta},\ r_{\text{min}}).
\end{equation}
Here, $n$ denotes the training iteration, and $\beta$ is a hyperparameter controlling the decay rate. We use the same $\beta$ value across datasets and set it so that the radius reaches the minimum value of $r_{min}$ in less than half of the training iterations.

\begin{figure*}[!ht]
    \centering    
    \setlength{\wid}{0.19\textwidth}
    \setlength{\mrg}{-0.45cm}
    \setlength{\mrgv}{-0.0cm}
    \begin{tabular}{c cc cc}
        \vspace{\mrgv}
        \includegraphics[width=\wid]{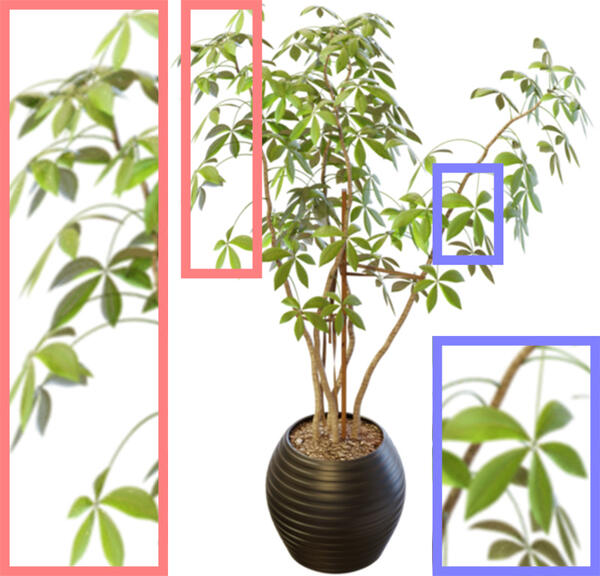} &
        \hspace{\mrg}
        \includegraphics[width=\wid]{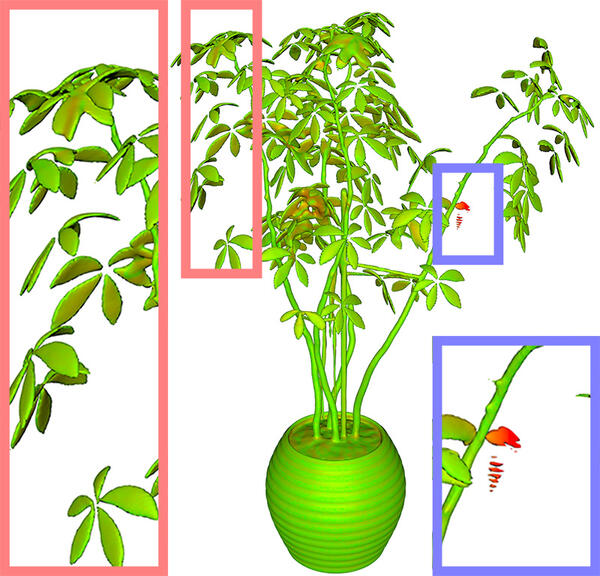} &
        \hspace{\mrg}
        \includegraphics[width=\wid]{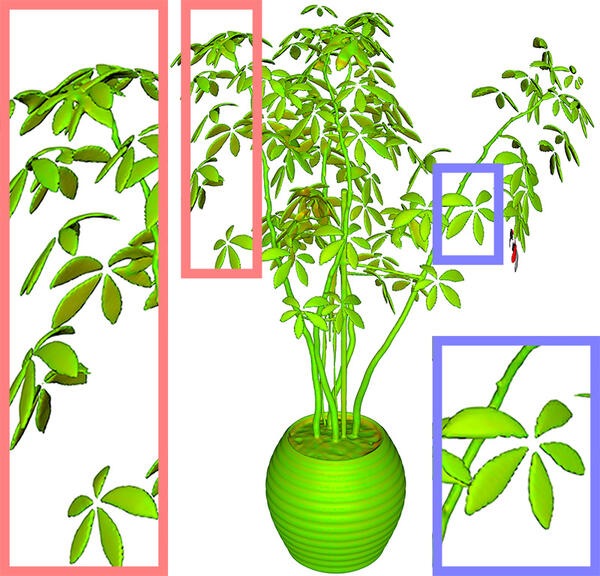} & 
        \hspace{\mrg}
        \includegraphics[width=\wid]{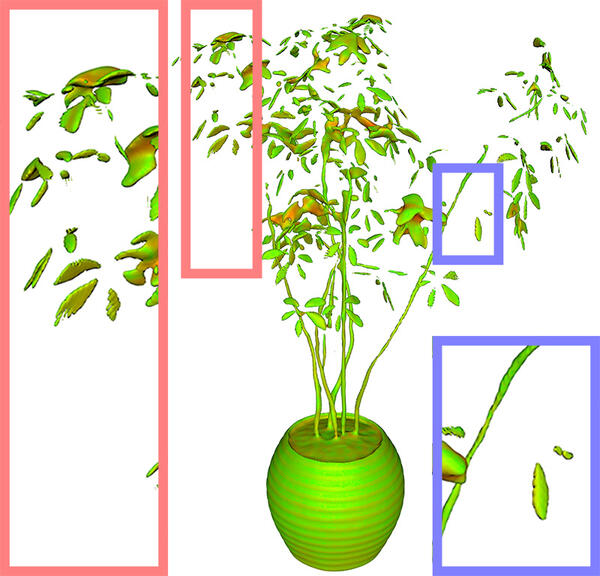} &
        \hspace{\mrg}
        \includegraphics[width=\wid]{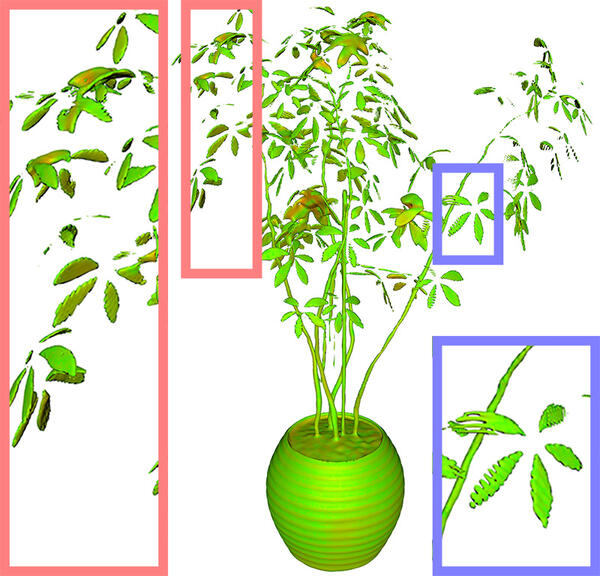} 
        \\ 
        \vspace{\mrgv}
        \includegraphics[width=\wid]{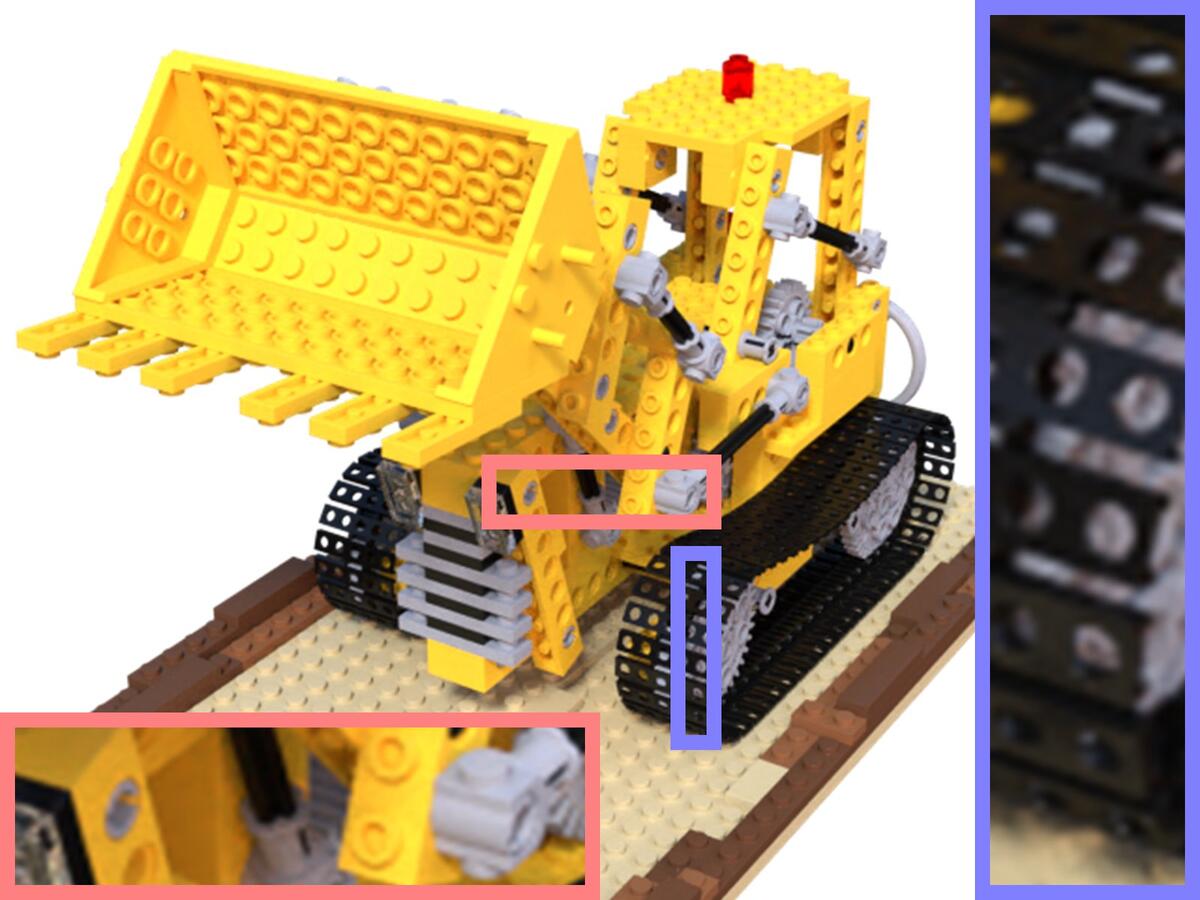} &
        \hspace{\mrg}
        \includegraphics[width=\wid]{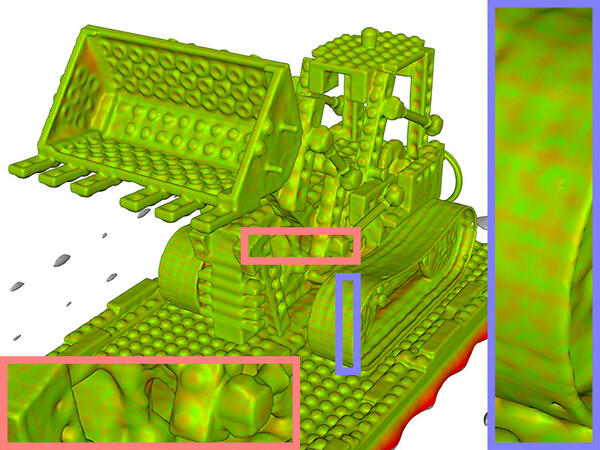} &
        \hspace{\mrg}
        \includegraphics[width=\wid]{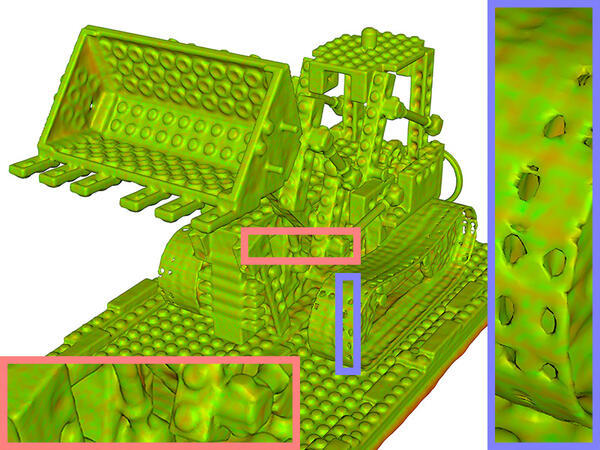} & 
        \hspace{\mrg}
        \includegraphics[width=\wid]{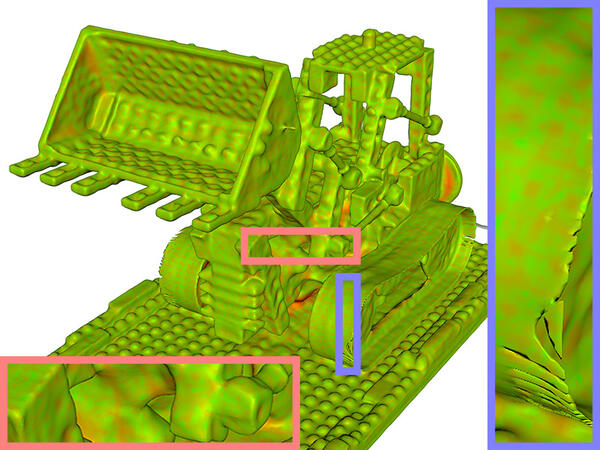} &
        \hspace{\mrg}
        \includegraphics[width=\wid]{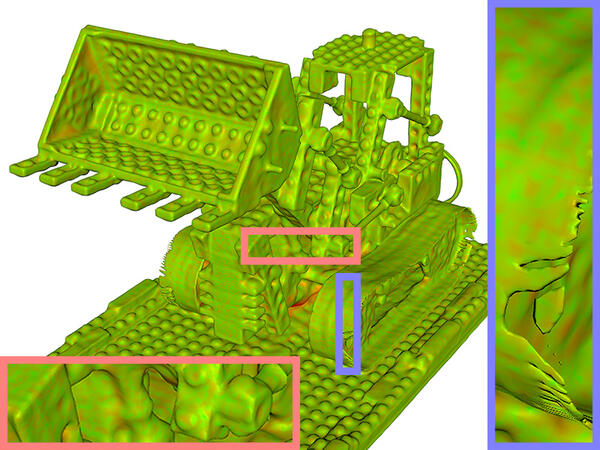} 
        \\
        \vspace{\mrgv}
        \includegraphics[width=\wid]{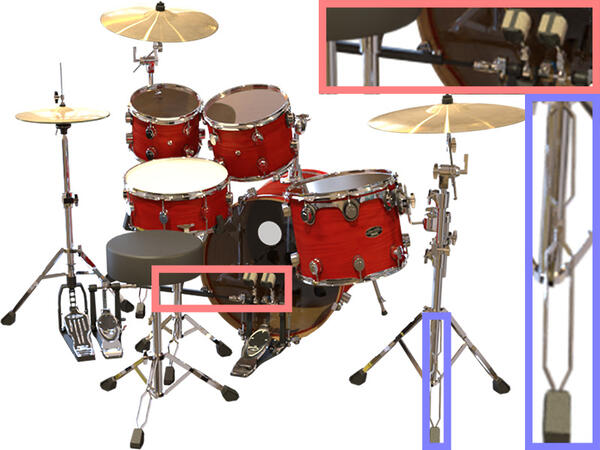} &
        \hspace{\mrg}
        \includegraphics[width=\wid]{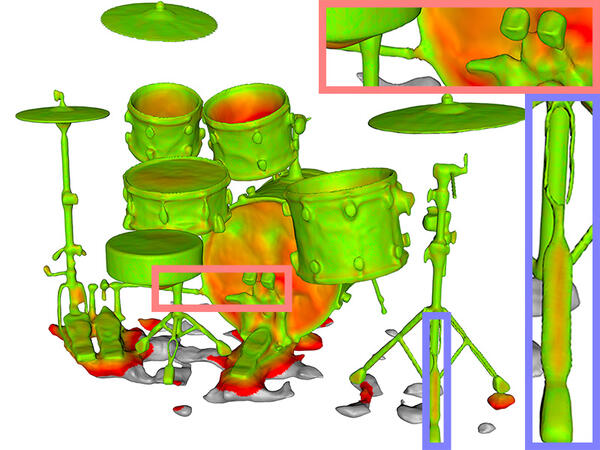} &
        \hspace{\mrg}
        \includegraphics[width=\wid]{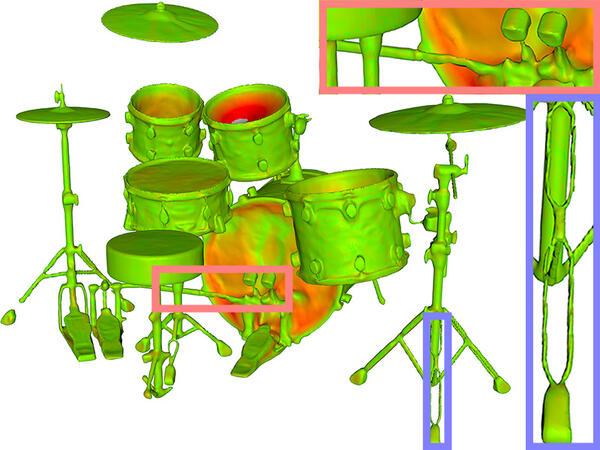} & 
        \hspace{\mrg}
        \includegraphics[width=\wid]{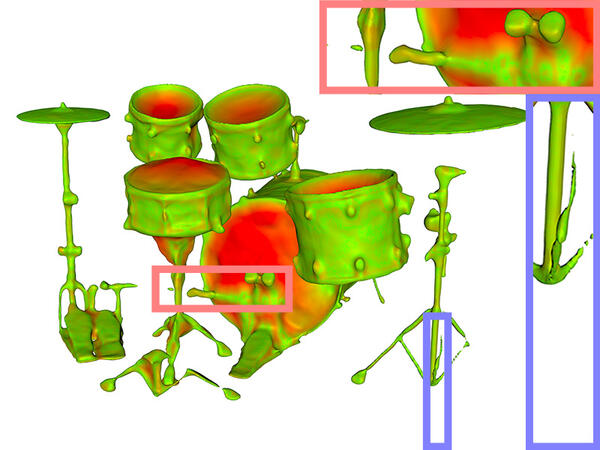} &
        \hspace{\mrg}
        \includegraphics[width=\wid]{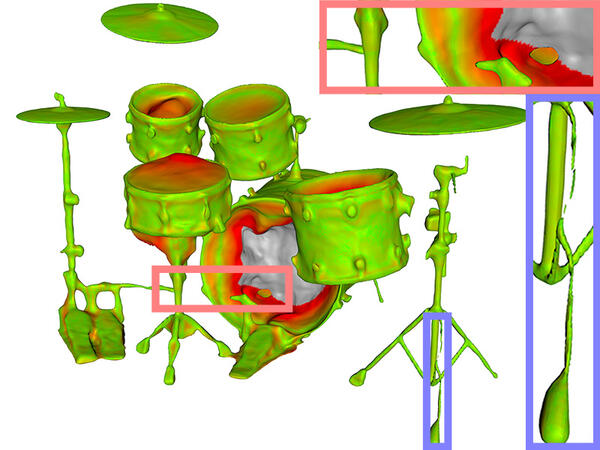} 
        \\
        \vspace{\mrgv}
        \includegraphics[width=\wid]{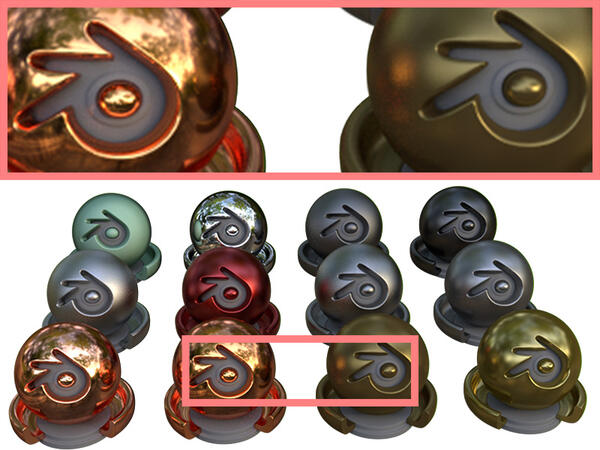} &
        \hspace{\mrg}
        \includegraphics[width=\wid]{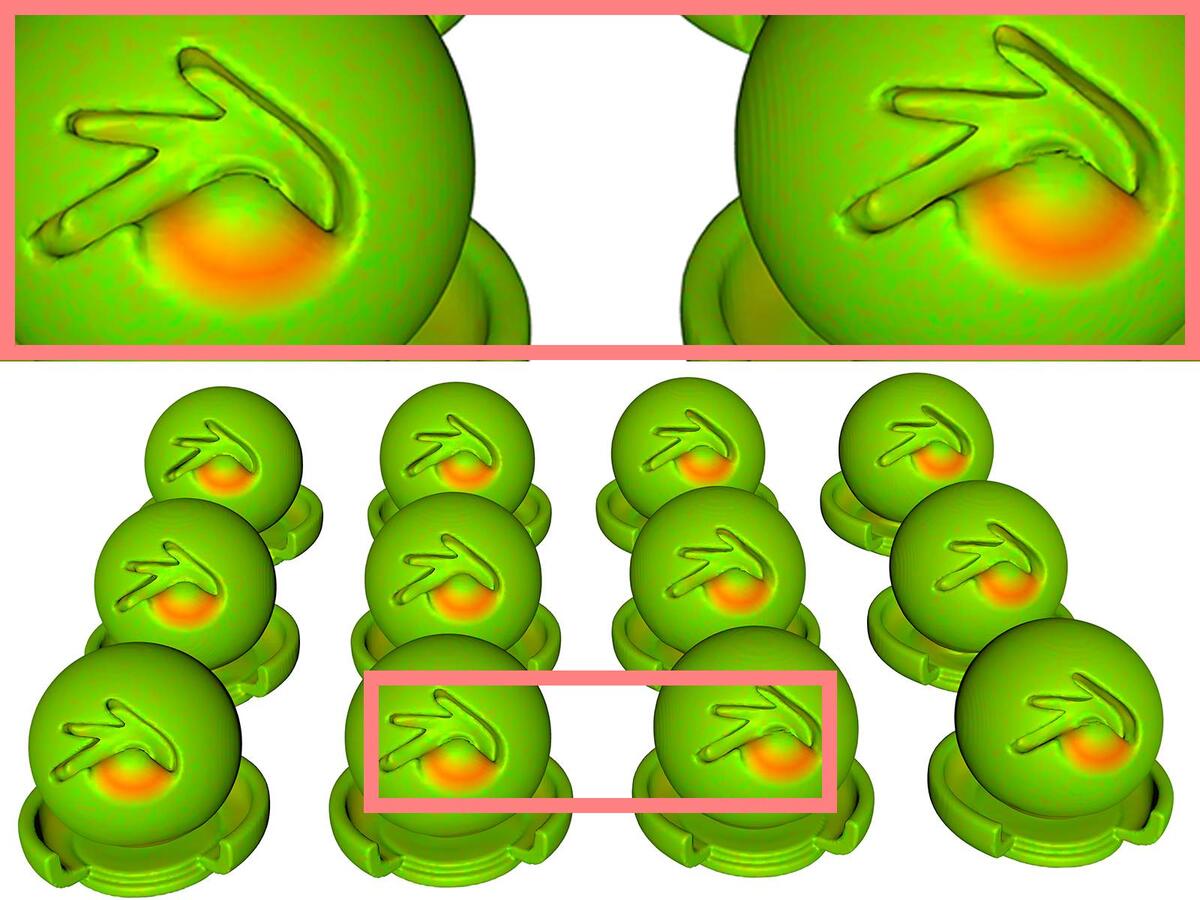} &
        \hspace{\mrg}
        \includegraphics[width=\wid]{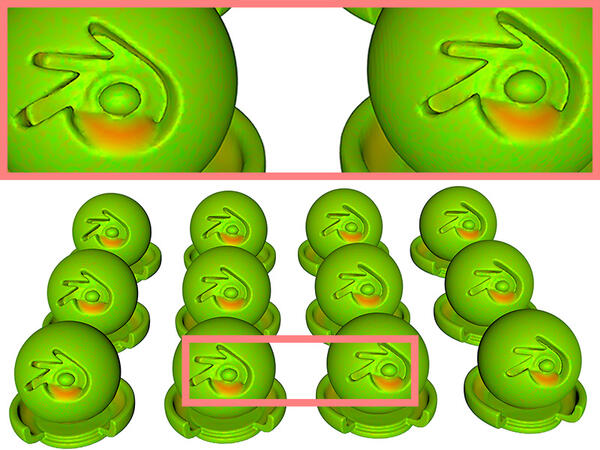} & 
        \hspace{\mrg}
        \includegraphics[width=\wid]{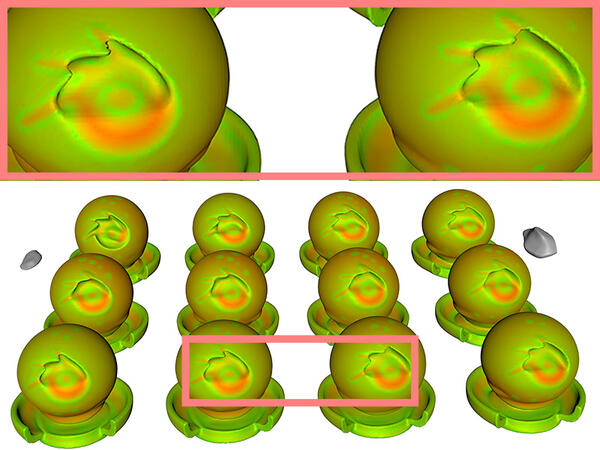} &
        \hspace{\mrg}
        \includegraphics[width=\wid]{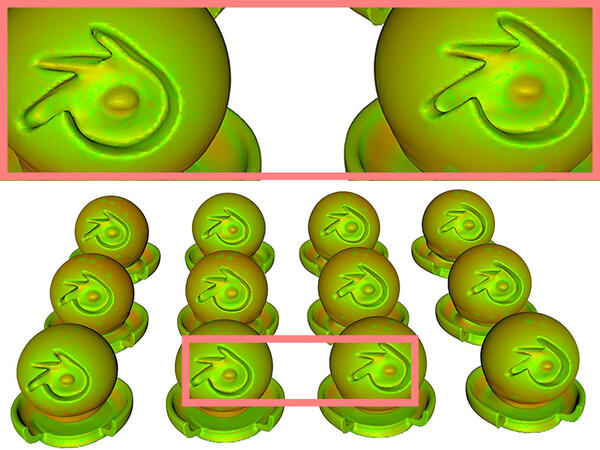} 
        \\ 
        \vspace{\mrgv}
        \textbf{Source} & \hspace{\mrg}
        \textbf{NeuS} & \hspace{\mrg}
        \textbf{NeuS (ours)} & \hspace{\mrg}
        \textbf{NeuralWarp} & \hspace{\mrg}
        \textbf{NeuralWarp (ours)}
    \end{tabular}
    \caption{Qualitative results on the \textbf{Realistic Synthetic 360} dataset~\cite{Mildenhall2020NeRFRS}. Our method improves upon the base \textbf{NeuS}~\cite{Wang2021NeuSLN} and \textbf{NeuralWarp}~\cite{darmon2022improving} surface modeling approaches, especially in the areas with thin details. The reconstructions are color-coded using the one-way Chamfer distance between the prediction and the ground truth, green color denotes a lower error.
    } 
    \label{fig:nerf_qual}
    \vspace{-0.1cm}
\end{figure*}

To avoid problems with the sphere cloud convergence, we apply a resampling procedure for the empty spheres which get stuck without reaching the surface. This process is typically applied up to 8 times during training, depending on the total number of iterations. Similarly to~\cite{Liu2020NeuralSV}, we sample $K$ points inside each sphere at which we evaluate the implicit function and find the spheres which have no surface inside them. We then resample these spheres near the ones which contain a surface region. 
Lastly, to avoid choosing training rays that do not intersect the surface of the object, we sample their endpoints uniformly from the volume bounded by the spheres.
For more details on the sphere resampling and sphere-guided ray sampling procedures, please refer to the supplementary materials.

\section{Experiments}

\begin{table*}[!ht]
    \centering
    \resizebox{0.8\linewidth}{!}{
    \begin{tabular}{l ccc ccc cc | c}
        & \multicolumn{8}{c}{Scene name}
        \\
        Method & Chair & Drums & Ficus & Hotdog & Lego & Mats & Mic & Ship & Mean 
        \\
        \hline
        COLMAP~\cite{schoenberger2016mvs} & 0.77 & 1.26 & 0.96 & 1.95 & 1.36 & 2.19 & 1.33 & 1.00 & 1.42
        \\
        \hline
        NeuS~\cite{Wang2021NeuSLN} & \textbf{0.38} & 1.88 & 0.51 & \textbf{0.52} & 0.68 & 0.40 & \textbf{0.60} & 0.60 & 0.70
        \\
        NeuS (ours) & 0.39 & \textbf{1.20} & \textbf{0.40} & 0.57 & \textbf{0.61} & \textbf{0.31} & 0.67 & \textbf{0.54} & \textbf{0.59}
        \\
        \hline
        NeuS w/ masks & \textbf{0.40} & \textbf{0.90} & 0.41 & 0.58 & 0.67 & 0.28 & 0.59 & 0.73 & 0.57
        \\
        NeuS w/ m. (ours) & 0.45 & 0.94 & \textbf{0.32} & \textbf{0.54} & 0.67 & \textbf{0.27} & \textbf{0.57} & \textbf{0.71} & \textbf{0.56}
        \\
        \hline
        NeuralWarp~\cite{darmon2022improving} & 0.43 & 3.00 & 0.94 & 1.65 & 0.81 & 1.02 & \textbf{0.75} & 1.27 & 1.23
        \\
        NeuralWarp (ours) & \textbf{0.41} & \textbf{2.67} & \textbf{0.61} & \textbf{1.44} & \textbf{0.76} & \textbf{0.92} & 0.80 & \textbf{1.07} & \textbf{1.09}
        \vspace{-0.05cm}
    \end{tabular}
    }
    \caption{Quantitative results on the Realistic Synthetic 360 dataset~\cite{Mildenhall2020NeRFRS}. We evaluate the Chamfer distance to compare the performance of different methods. We can see that base methods with our proposed modification achieve better performance than the original versions across most of the scenes.
    }\label{tab:nerf_quant}
    \vspace{-0.2cm}
\end{table*}
\begin{table*}
    \centering
    \resizebox{\linewidth}{!}{
    \begin{tabular}{l ccccc ccccc ccccc | c}
        & \multicolumn{15}{c}{Scene ID}
        \\
        Method & 24 & 37 & 40 & 55 & 63 & 65 & 69 & 83 & 97 & 105 & 106 & 110 & 114 & 118 & 122 & Mean 
        \\
        \hline
        UNISURF~\cite{Oechsle2021UNISURFUN} & 1.16 & 1.01 & 1.16 & \textbf{0.36} & 1.27 & 0.72 & \textbf{0.73} & 1.33 & 1.58 & 0.72 & 0.53 & 1.21 & 0.41 & 0.69 & 0.51 & 0.89
        \\
        UNISURF (ours) & \textbf{1.10} & \textbf{0.98} & \textbf{1.14} & 0.37 & \textbf{1.08} & \textbf{0.66} & 0.89 & 1.33 & \textbf{1.19} & \textbf{0.69} & \textbf{0.52} & \textbf{0.98} & \textbf{0.38} & \textbf{0.46} & \textbf{0.50} & \textbf{0.82}
        \\
        \hline
        VolSDF*~\cite{Yariv2021VolumeRO} & 0.94 & 1.73 & 1.04 & 0.47 & 0.87 & 0.74 & 0.84 & 1.24 & 1.31 & 0.71 & 0.78 & 1.61 & 0.61 & 0.71 & 0.54 & 0.94
        \\
        VolSDF* (ours) & \textbf{0.91} & \textbf{1.05} & \textbf{0.65} & \textbf{0.42} & \textbf{0.86} & \textbf{0.69} & \textbf{0.72} & \textbf{1.20} & \textbf{1.14} & \textbf{0.64} & \textbf{0.66} & \textbf{1.16} & \textbf{0.43} & \textbf{0.54} & \textbf{0.51} & \textbf{0.77}
        \\
        \hline
        NeuS~\cite{Wang2021NeuSLN} & 0.93 & 1.06 & 0.81 & 0.38 & 1.02 & 0.60 & \textbf{0.58} & 1.43 & 1.15 & \textbf{0.78} & 0.57 & 1.15 & 0.35 & 0.45 & 0.46 & 0.78
        \\
        NeuS (ours) & \textbf{0.83} & \textbf{0.91} & \textbf{0.69} & \textbf{0.36} & \textbf{0.95} & \textbf{0.54} & 0.65 & \textbf{1.37} & 1.15 & 0.82 & \textbf{0.55} & \textbf{0.86} & \textbf{0.33} & \textbf{0.42} & \textbf{0.42} & \textbf{0.72}
        \\
        \hline
        NeuS w/ masks & \textbf{0.83} & 0.98 & 0.56 & 0.37 & 1.13 & 0.59 & 0.60 & 1.45 & 0.95 & 0.78 & \textbf{0.52} & 1.43 & 0.36 & \textbf{0.45} & 0.45 & 0.76
        \\
        NeuS w/ m. (ours) & 0.85 & \textbf{0.92} & \textbf{0.46} & 0.37 & \textbf{1.05} & 0.59 & \textbf{0.58} & \textbf{1.27} & \textbf{0.88} & 0.78 & 0.53 & \textbf{0.93} & \textbf{0.33} & 0.46 & 0.45 & \textbf{0.70}
        \\
        														
        \hline
        NeuralWarp~\cite{darmon2022improving} & 0.49 & \textbf{0.71} & 0.38 & \textbf{0.38} & \textbf{0.79} & \textbf{0.81} & 0.82 & 1.20 & \textbf{1.06} & 0.68 & 0.66 & 0.74 & 0.41 & 0.63 & 0.51 & 0.68
        \\
        NeuralWarp (ours) & 0.49 & 0.77 & \textbf{0.37} & 0.40 & 0.81 & 0.87 & \textbf{0.72} & \textbf{1.19} & 1.07 & \textbf{0.66} & \textbf{0.64} & \textbf{0.70} & \textbf{0.37} & \textbf{0.58} & \textbf{0.48} & 0.68
        \\
        \multicolumn{17}{r}{* denotes unofficial implementation}
        \vspace{-0.2cm}
    \end{tabular}
    }
    \caption{We present a quantitative evaluation of our method on the DTU~\cite{Jensen2014LargeSM} dataset. We have combined our approach with four popular implicit surface reconstruction systems: UNISURF~\cite{Oechsle2021UNISURFUN}, NeuS~\cite{Wang2021NeuSLN}, NeuralWarp~\cite{darmon2022improving}, and VolSDF~\cite{Yariv2021VolumeRO}. In this comparison, we follow the previous works by measuring the Chamfer distance (lower the better) between the ground truth point cloud and the reconstructions, pre-processed by removing the regions outside the object via ground-truth segmentation masks. The models in this comparison were trained \emph{without} segmentation masks unless stated otherwise. We highlight the better score between the original approach and our modification.
    }
    \label{tab:dtu_quant}
\end{table*}

We conduct our main experiments using three popular 3D reconstruction benchmarks: DTU MVS~\cite{Jensen2014LargeSM}, BlendedMVS~\cite{yao2020blendedmvs} and Realistic Synthetic 360~\cite{Mildenhall2020NeRFRS} and evaluate our approach by combining it with four different methods for implicit surface reconstruction.

\subsection{Base methods}
Our approach acts as an addition to the 3D reconstruction systems that learn neural implicit surfaces through volume rendering. Therefore, we apply it to four representative systems to showcase its effectiveness and applicability. UNISURF \cite{Oechsle2021UNISURFUN} represents the geometry of the scene via an occupancy field that is learned through a combination of surface and volume rendering approaches. VolSDF  \cite{Yariv2021VolumeRO} and NeuS \cite{Wang2021NeuSLN} propose to train an SDF via volume rendering by transforming it into occupancy defined along the ray. NeuralWarp \cite{darmon2022improving} builds upon VolSDF by using an additional loss term that directly enforces the photo consistency of the learned geometry by warping patches across different views. Our approach can be seamlessly incorporated into all these systems, and we provide additional implementation details in the supplementary materials.

\subsection{Training process}

For each of the base models, we have used the official codebase, except for VolSDF, for which we use the code provided as part of the NeuralWarp method. Therefore, for the implementation aspects of the base methods, including the architectures and training details, we refer to the respective publications. We employ the same optimizer, scheduling, hyperparameters, number of iterations, and other technicalities as in the reference methods to train the implicit functions.

The sphere cloud optimization process is adapted for each system with regard to the implicit geometry type and the total number of iterations used for training. The sphere radius in the SDF-based methods decays exponentially from $r_\text{max}=0.4$ to $r_\text{min}=0.04$, while for UNISURF it ranges from $r_\text{max}=2.0$ to $r_\text{min}=0.1$. The repulsion penalty considers the $k=10$ nearest spheres that intersect with each sphere in the cloud, i.e. $d = 2 * r_n$, and its weight is $\lambda = 0.1$ in UNISURF and $\lambda = 0.0001$ in the other methods. We found a number of 15,000 spheres to be sufficient for representing most scenes, which are scaled to fit in the bounding sphere of radius one. We employ the Adam~\cite{KingmaB14} optimizer with a learning rate of $10^{-4}$ for the optimization of the centers of the spheres in all experiments. 

\begin{figure*}
    \centering    
    \setlength{\wid}{0.19\textwidth}
    \setlength{\mrg}{-0.45cm}
    \setlength{\mrgv}{0cm}
    \begin{tabular}{c cc cc}
        \vspace{\mrgv}
        \includegraphics[width=\wid]{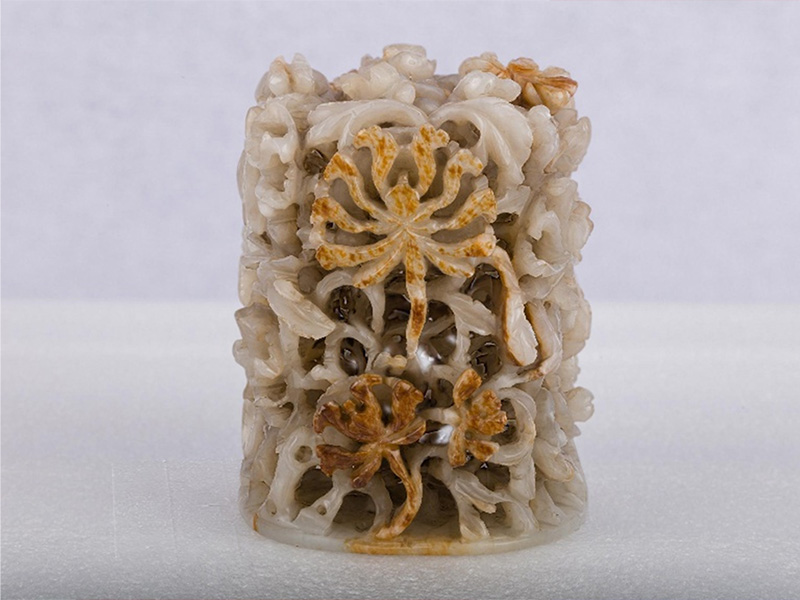} &
        \hspace{\mrg}
        \includegraphics[width=\wid]{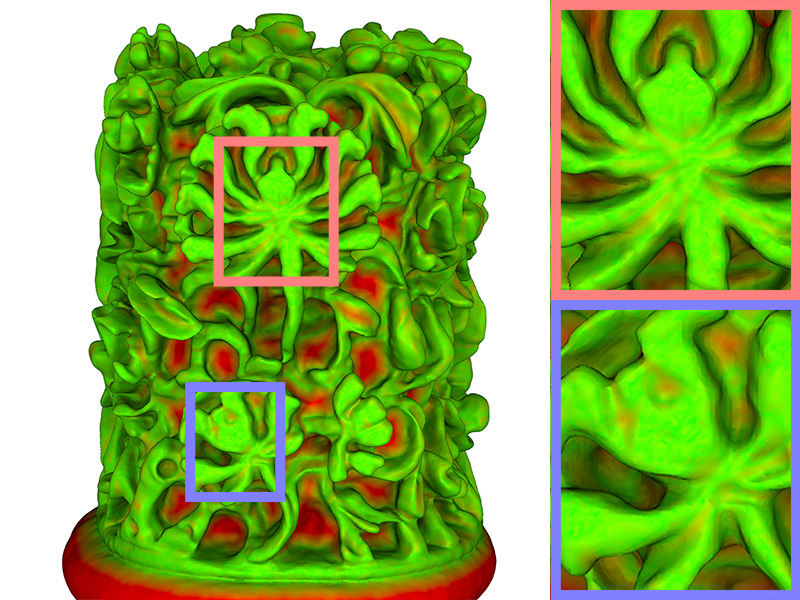} &
        \hspace{\mrg}
        \includegraphics[width=\wid]{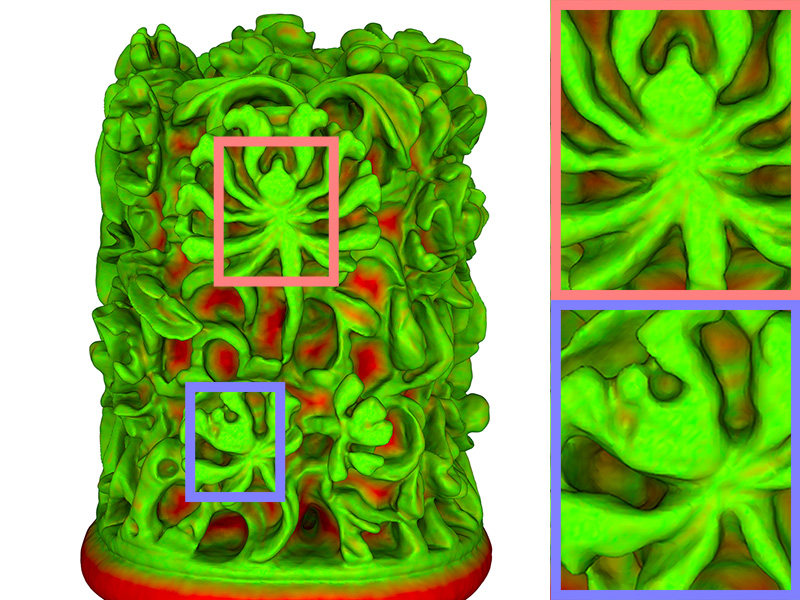} & 
        \hspace{\mrg}
        \includegraphics[width=\wid]{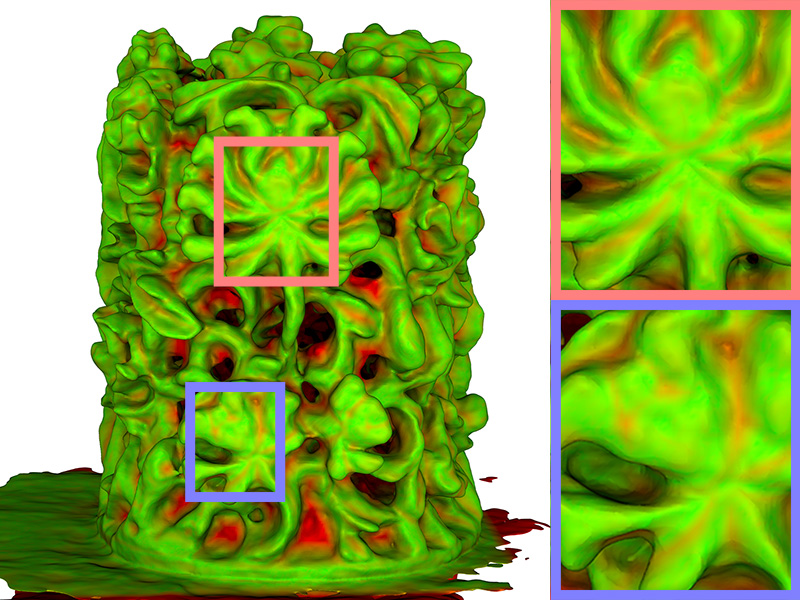} &
        \hspace{\mrg}
        \includegraphics[width=\wid]{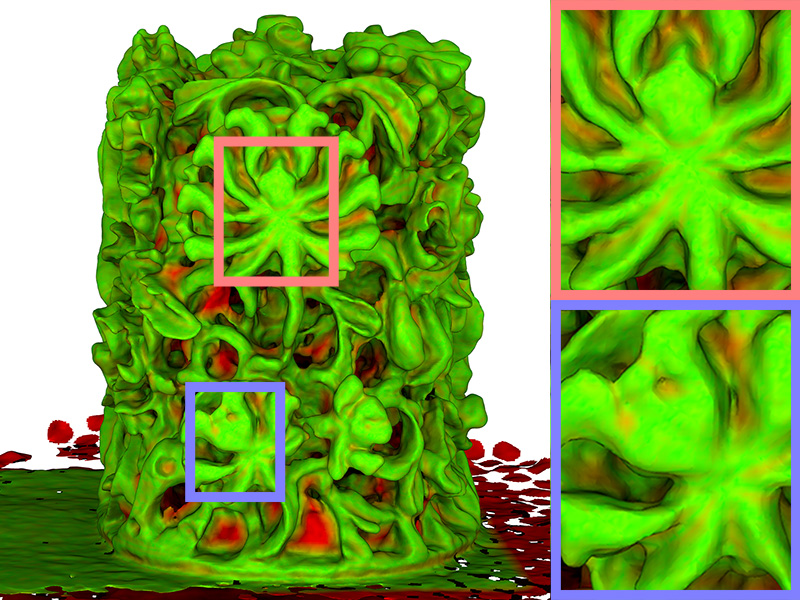} 
        \\ 
        \vspace{\mrgv}
        \includegraphics[width=\wid]{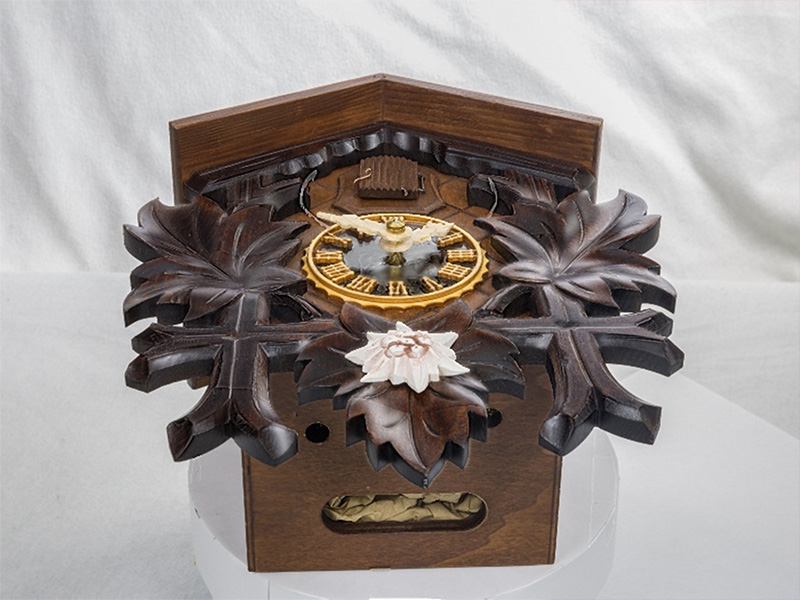} &
        \hspace{\mrg}
        \includegraphics[width=\wid]{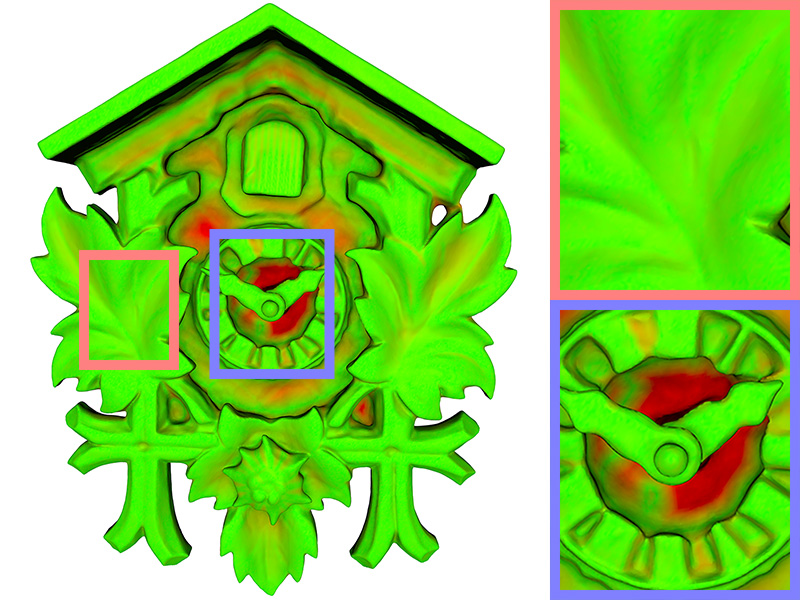} &
        \hspace{\mrg}
        \includegraphics[width=\wid]{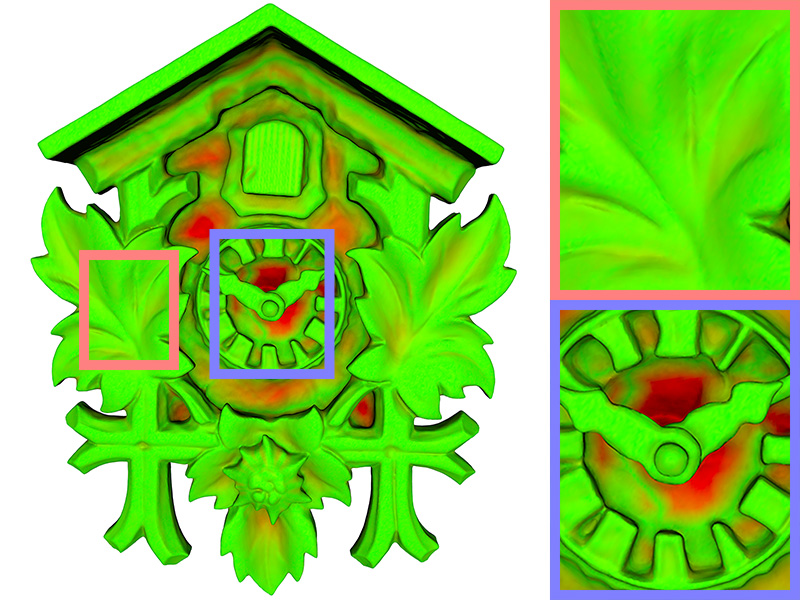} & 
        \hspace{\mrg}
        \includegraphics[width=\wid]{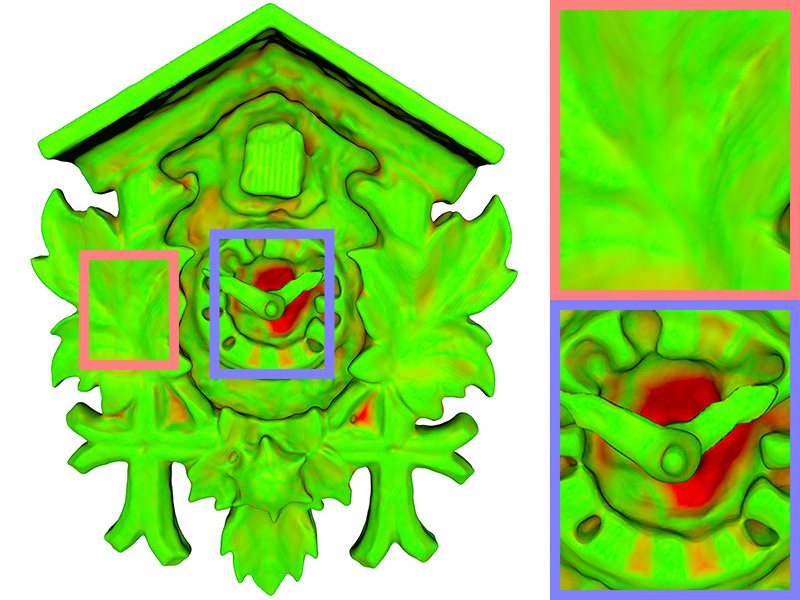} &
        \hspace{\mrg}
        \includegraphics[width=\wid]{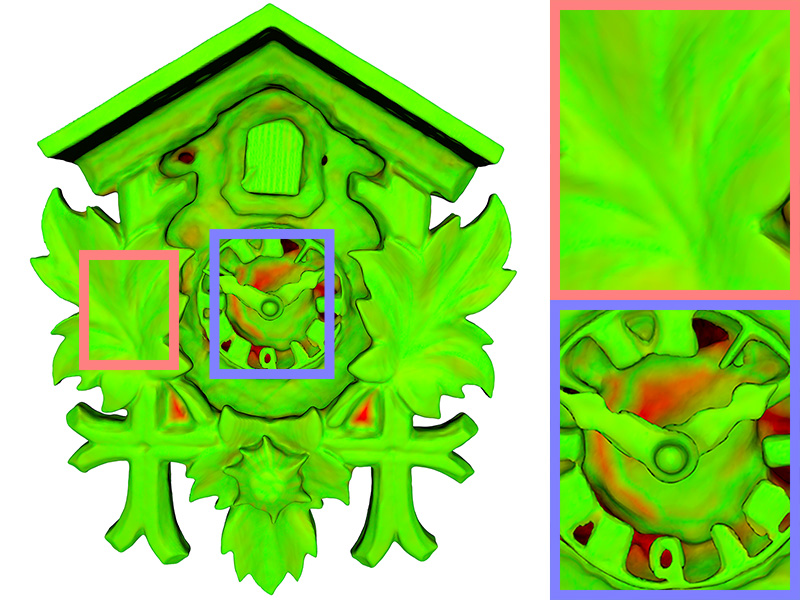} 
        \\
        \vspace{\mrgv}
        \includegraphics[width=\wid]{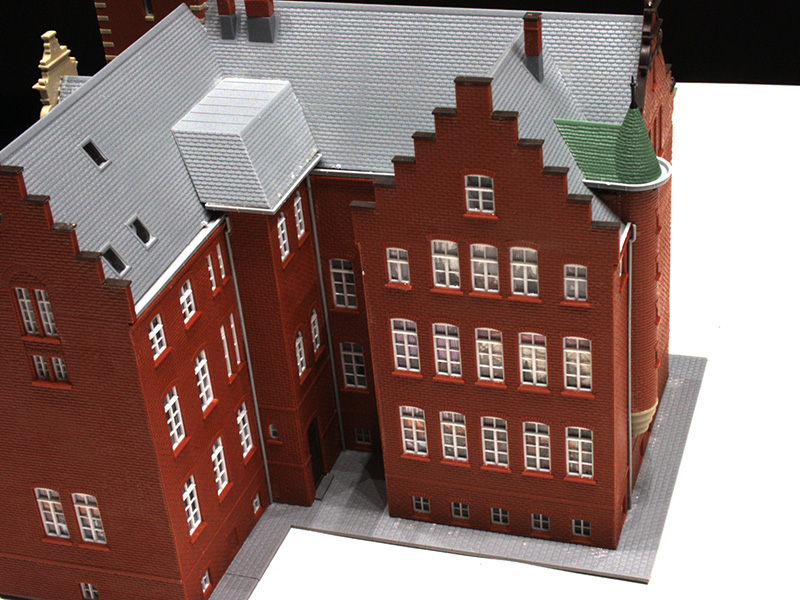} &
        \hspace{\mrg}
        \includegraphics[width=\wid]{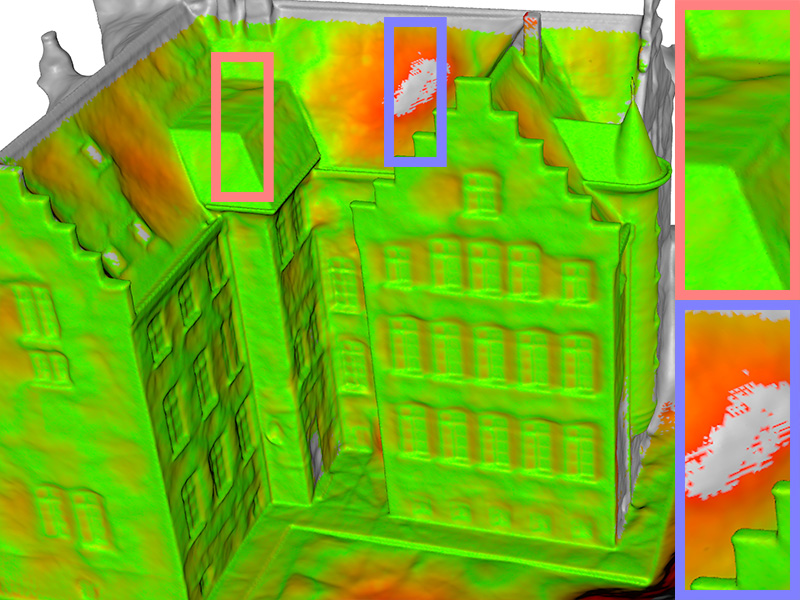} &
        \hspace{\mrg}
        \includegraphics[width=\wid]{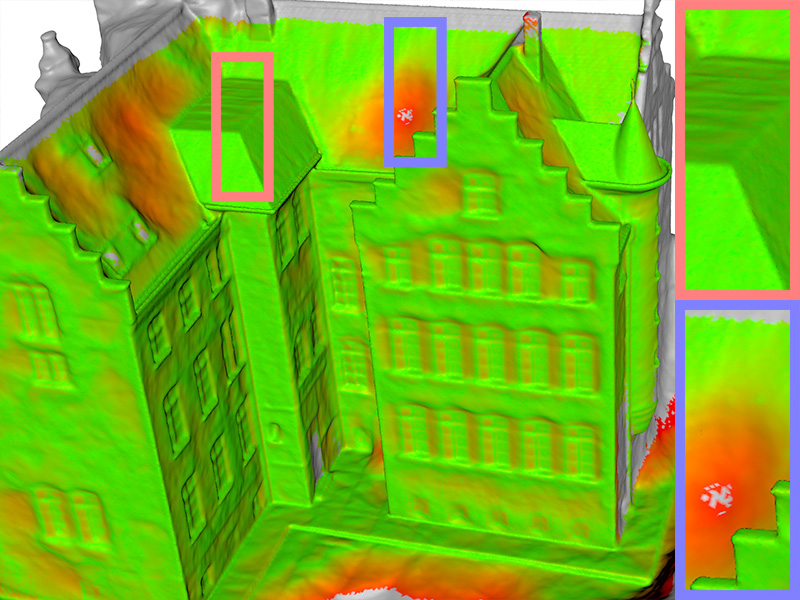} & 
        \hspace{\mrg}
        \includegraphics[width=\wid]{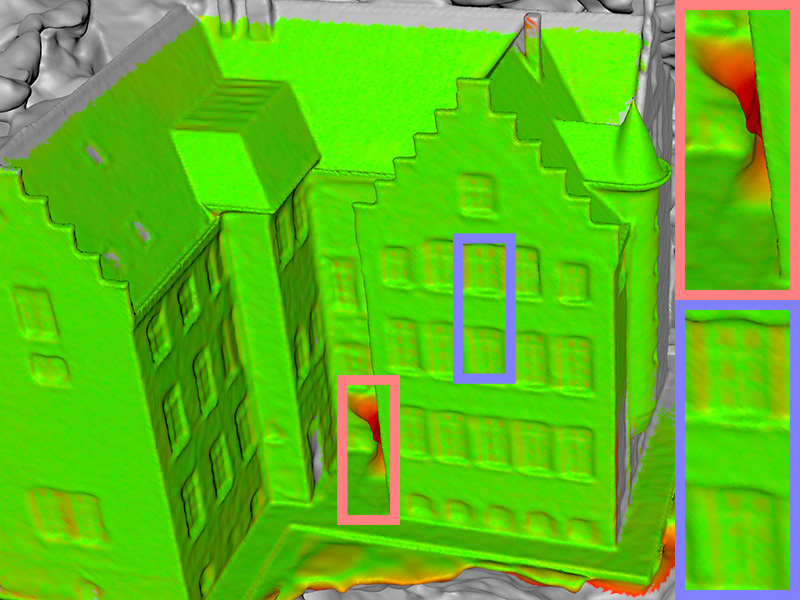} &
        \hspace{\mrg}
        \includegraphics[width=\wid]{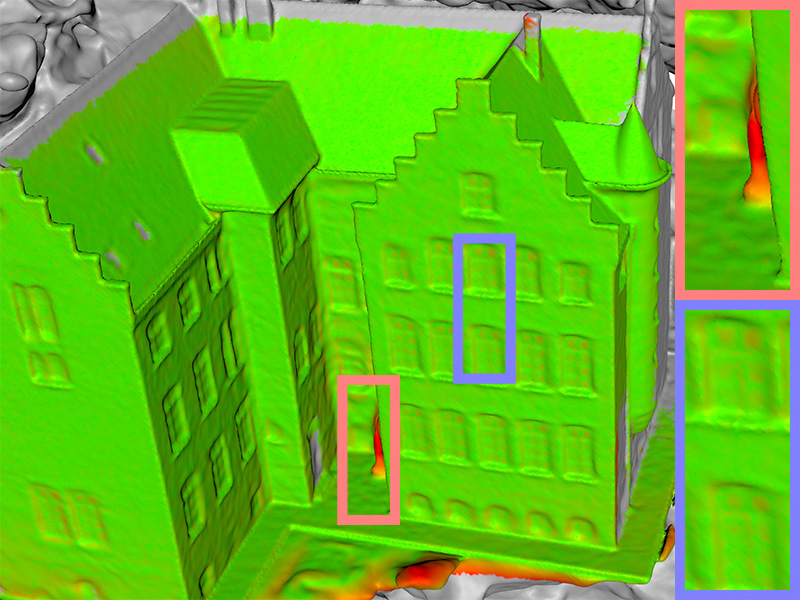} 
        \\ 
        \vspace{\mrgv}
        \includegraphics[width=\wid]{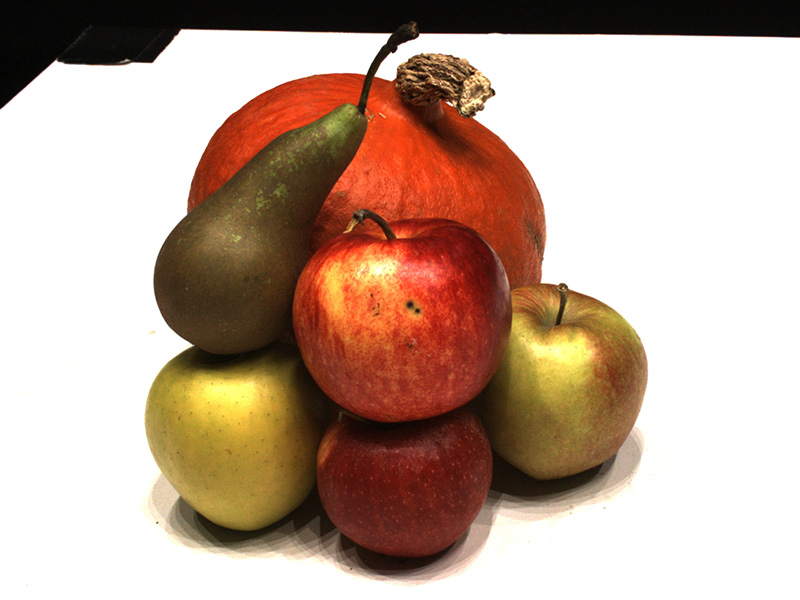} &
        \hspace{\mrg}
        \includegraphics[width=\wid]{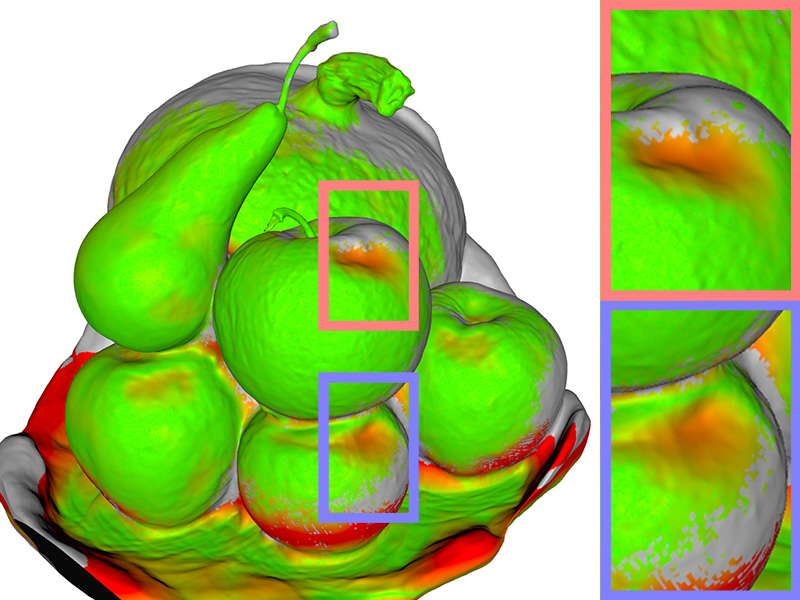} &
        \hspace{\mrg}
        \includegraphics[width=\wid]{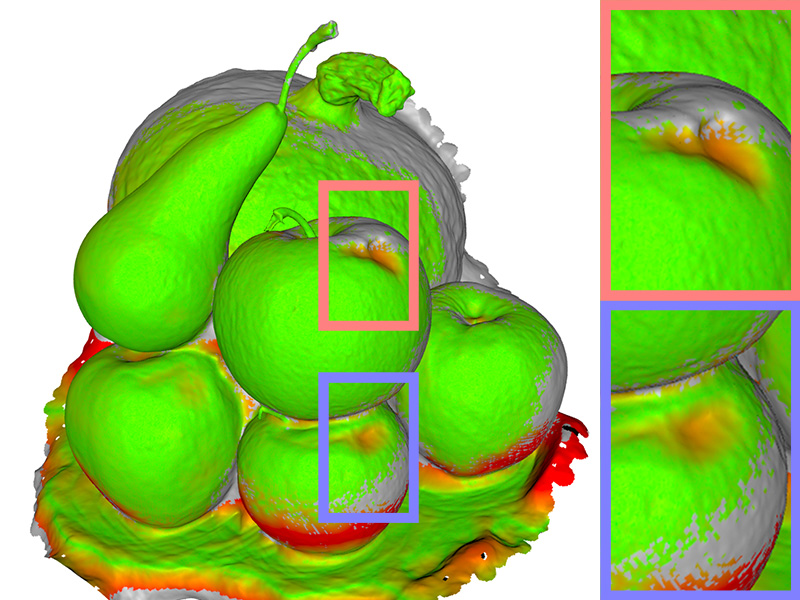} & 
        \hspace{\mrg}
        \includegraphics[width=\wid]{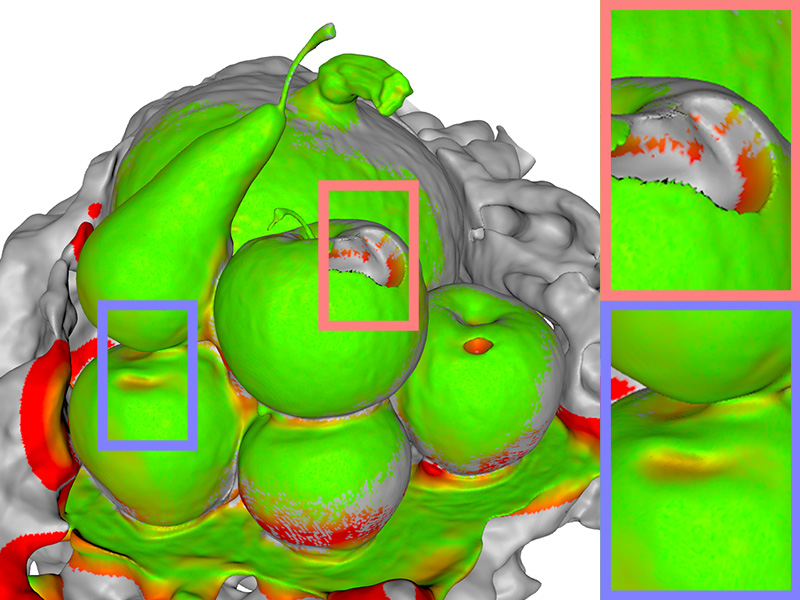} &
        \hspace{\mrg}
        \includegraphics[width=\wid]{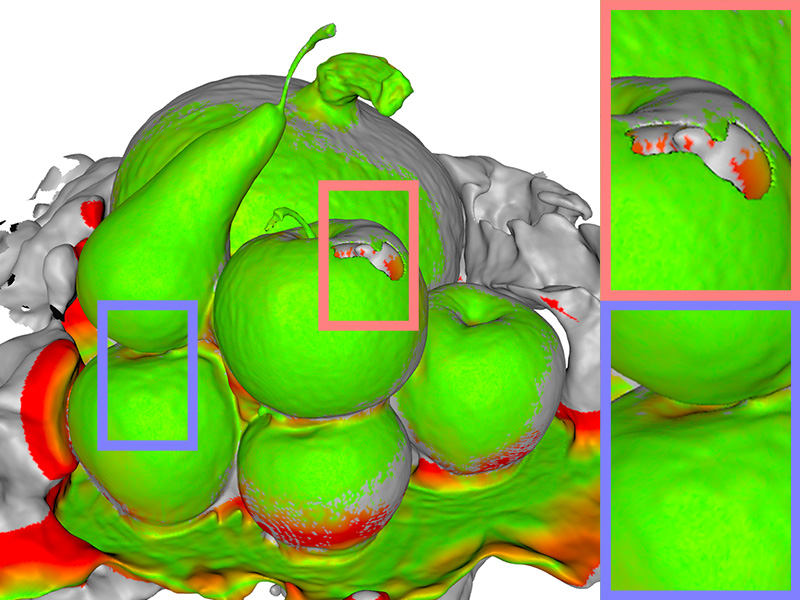} 
        \\ 
        \textbf{Source} & \hspace{\mrg}
        \textbf{NeuS} & \hspace{\mrg}
        \textbf{NeuS (ours)} & \hspace{\mrg}
        \textbf{NeuralWarp} & \hspace{\mrg}
        \textbf{NeuralWarp (ours)}
    \end{tabular}
    \vspace{-0.2cm}
    \caption{Qualitative results on the real-world \textbf{BlendedMVS}~\cite{yao2020blendedmvs} (rows 1-2) and \textbf{DTU}~\cite{Jensen2014LargeSM} (rows 3-4) datasets. Our method achieves more accurate reconstructions compared to the base approaches. Green color denotes a lower one-way Chamfer error. For the clock scene (row 2), we rendered the geometry from a frontal angle, not present in the training views, to highlight the differences in the reconstructions.
    }
    \vspace{-0.2cm}
    \label{fig:dtu_qual}
\end{figure*}

\subsection{Realistic Synthetic 360 evaluation}

The Realistic Synthetic 360 dataset was introduced in~\cite{Mildenhall2020NeRFRS} as a benchmark for the novel view synthesis task. Each of its eight scenes features an object realistically rendered from 100 training viewpoints and paired with a ground truth mesh. Though this dataset was not originally intended for the 3D reconstruction task, it contains objects with complex geometries and non-Lambertian materials, representing a challenge for classical 3D reconstruction systems. As some of the ground truth meshes contain internal surfaces that are not visible in any of the training views, we filter them by removing the non-visible parts. We perform a similar filtering step for the reconstructed meshes and compute the Chamfer distance between the cleaned meshes by sampling one million points on each surface. We report the distance computed at the original scale of the meshes multiplied by $10^2$ in Table~\ref{tab:nerf_quant}. We also report the qualitative results in Figure~\ref{fig:nerf_qual}.

We can see that our method achieves improvements across most of the scenes for all of the compared methods, which is especially noticeable in scenes such as ficus and materials. We hypothesize that this is the case because of the complex structure of the reconstructed surface. This complexity does not allow the standard methods to effectively estimate the location of the high-density regions, which hinders the optimization process, leading to both reduced qualities of reconstructions and renders.

\subsection{DTU and BlendedMVS evaluation}

The DTU MVS dataset~\cite{Jensen2014LargeSM} contains 49 or 64 images with fixed camera positions and ground truth point clouds of 80 scenes acquired using a structured light scanner. We follow recent works~\cite{darmon2022improving, Oechsle2021UNISURFUN, Wang2021NeuSLN, Yariv2021VolumeRO, yariv2020multiview} and perform the evaluation on the subset of 15 diverse scenes selected by~\cite{yariv2020multiview}. The authors also provide the corresponding segmentation masks for each of the chosen scenes, which are used in the ``w/ masks'' experiments. Additionally, since most of the objects in the DTU dataset have a relatively simple geometry, we conducted an additional qualitative evaluation for a subset of objects from the BlendedMVS dataset~\cite{yao2020blendedmvs}.

For a quantitative evaluation on the DTU dataset, we convert the trained implicit functions to triangle meshes using the Marching Cubes algorithm~\cite{lewiner2003}. We then measure the Chamfer distances between the ground-truth point clouds and the extracted meshes using the standard evaluation procedure and report the scores in Table~\ref{tab:dtu_quant}. These results are obtained without mask supervision during training, except for the ``NeuS w/ masks'' experiment. However, we follow other works in applying ground-truth masks for post-processing the meshes before calculating the metrics. Specifically, we filter the mesh using the visual hull obtained from the multi-view segmentations dilated with a radius of 12. The visual hull is calculated as the intersection of the silhouette cones emitted from each of the training cameras. Notably, we use the same dilation radius for all of the compared methods, hence the reported metrics may differ from the ones presented in the original publications. As we can see from the results, our method improves three out of four base methods uniformly across most of the scenes.

\begin{figure}
\centering
    \setlength{\wid}{0.32\linewidth}
    \setlength{\mrg}{-0.45cm}
    \setlength{\mrgv}{0cm}
    \footnotesize
    \vspace{\mrgv}
    \begin{tabular}{ccc}
        \vspace{\mrgv}
        &$10^5$ iterations&
        \\
        \vspace{\mrgv}
        \includegraphics[width=\wid]{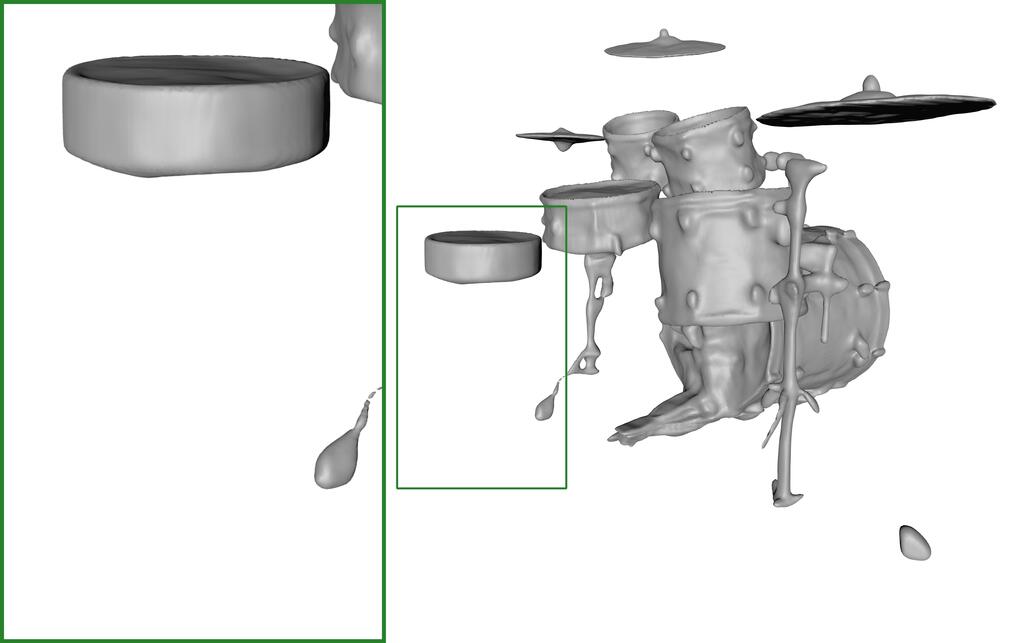} &
        \hspace{\mrg}
        \includegraphics[width=\wid]{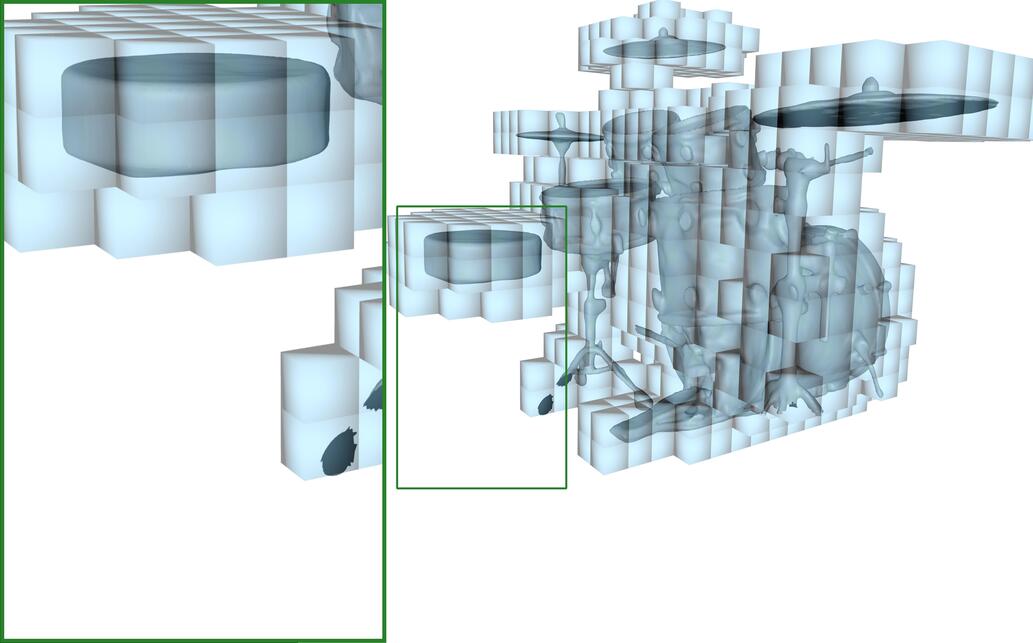} &
        \hspace{\mrg}
        \includegraphics[width=\wid]{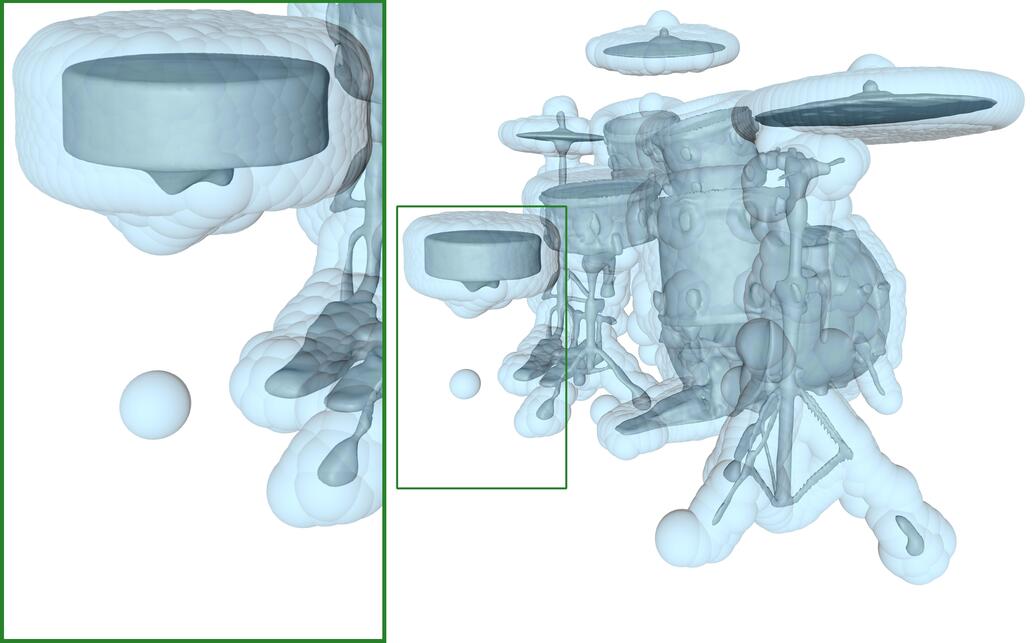}
        \\ 
        \vspace{\mrgv}
        &$3 \cdot 10^5$ iterations&
        \\
        \vspace{\mrgv}
        \includegraphics[width=\wid]{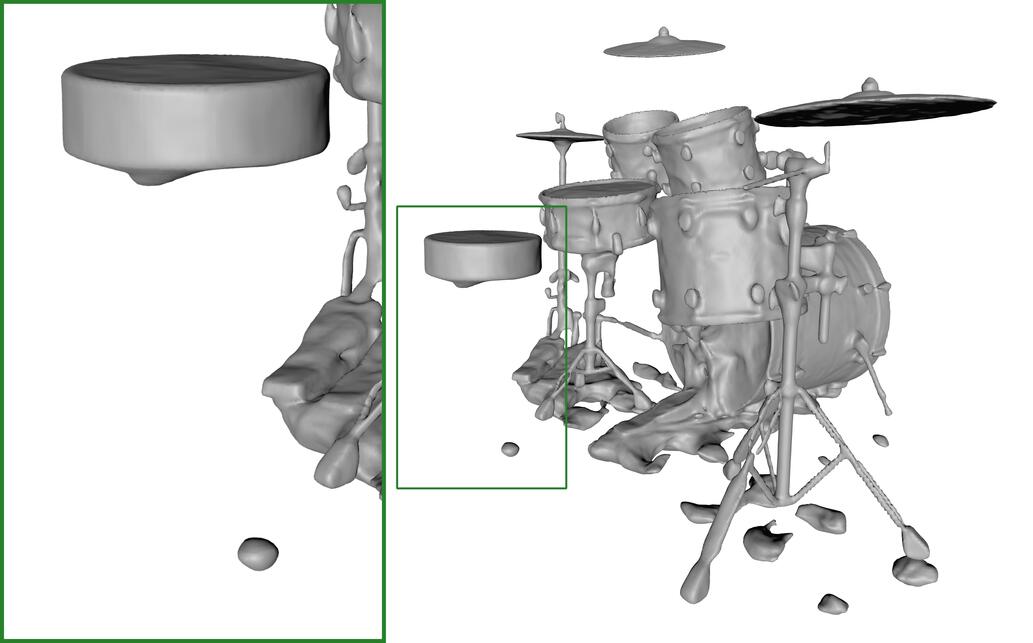} &
        \hspace{\mrg}
        \includegraphics[width=\wid]{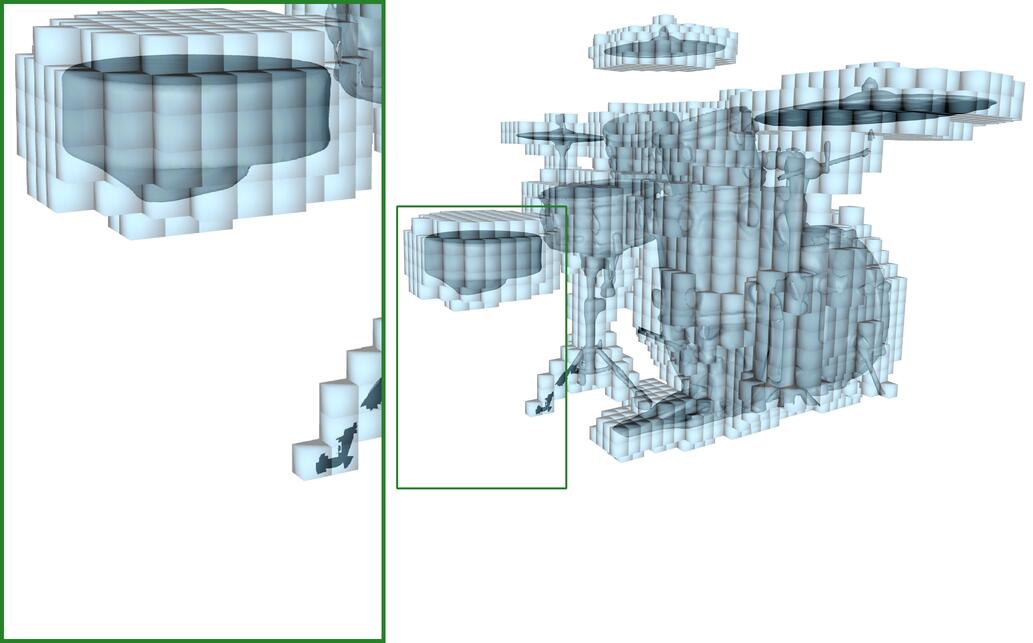} &
        \hspace{\mrg}
        \includegraphics[width=\wid]{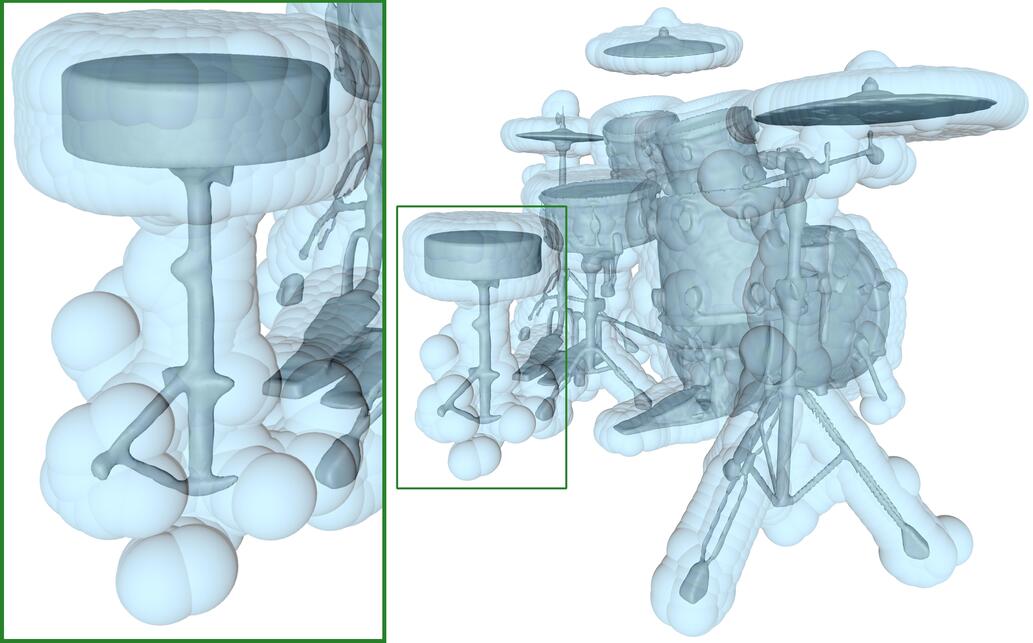}
        \\ 
        \vspace{\mrgv}
        \textbf{NeuS} & \hspace{\mrg}
        \textbf{NSVF ablation} & \hspace{\mrg}
        \textbf{NeuS (ours)}
    \end{tabular}
    \caption{
    Qualitative comparison of the geometry progression during optimization between our spheres and a NSVF-style explicit geometry. The models are trained without mask supervision to highlight the benefits of the proposed sphere guidance.
    }
    \label{fig:nsvf_ablation}
    \vspace{-0.2cm}
\end{figure}

\subsection{Ablation study}

\begin{table}[!ht]
    \centering
    \resizebox{0.99\linewidth}{!}{
    \begin{tabular}{l cccccccc }
        & \multicolumn{8}{c}{Scene name}
        \\
        Method & Chair & Drums & Ficus & Hotdog & Lego & Mats & Mic & Ship
        \\
        \hline
        NeuS & 0.38 & 1.88 & 0.51 & \textbf{0.52} & 0.68 & 0.40 & \textbf{0.60} & 0.60 
        \\
        NSVF ablation & \textbf{0.37} & 3.12 & 0.49 & \textbf{0.52} & 0.64 & 0.34 & 3.38 & 0.73 
        \\
        w/o $\mathcal{L}_\text{rep}$ & 0.76 & 2.37 & 0.79 & 2.71 & 1.03 & 0.95 & 1.87 & 4.30\\  
        NeuS (ours) & 0.39 & \textbf{1.20} & \textbf{0.40} & 0.57 & \textbf{0.61} & \textbf{0.31} & 0.67 & \textbf{0.54} 
        \\
        \hline
        \vspace{-0.3cm}
    \end{tabular}
    }
    \caption{Ablation study on the importance of our repulsion loss and the benefit of joint optimization of the guiding primitive.
    }\label{tab:nsvf_quant}
    \vspace{-0.35cm}
\end{table}

We have conducted an ablation study to evaluate the components of our method. We present the main results in Table~\ref{tab:nsvf_quant}, and include additional experiments in the supplementary material.

For the first experiment, we evaluate our gradient-based sphere cloud optimization scheme. To do that, we replace our sphere cloud with a sparse voxel octree data structure and use its optimization method proposed in Neural Sparse Voxel Fields~\cite{Liu2020NeuralSV} paper. We initialize our sparse voxel grid with 512 voxels and use the same scheduling for pruning and subdivision, as in the original NSFV approach, multiplied by a factor of two (since we use two times more iterations for training). For the pruning of voxels, we utilize the same resampling criteria as in our method but relaxed it to allow the voxels within an $\epsilon=0.01$ distance to the surface not to be pruned. We observe that NSVF underperforms due to the greedy voxel pruning optimization strategy. The errors in the coarse reconstruction become permanent after the empty voxels are pruned. In comparison, our gradient-based approach allows the exploration and inclusion of the neighboring areas initially missing from the coarse reconstruction. 

In the second experiment, we ablate the repulsion loss $\mathcal{L}_\text{rep}$ and observe that it leads to spheres clumping together in the learned sphere cloud, not covering the entire object, which eventually produces poor reconstructions. 
For additional ablation experiments, as well as rendered results, please refer to the supplementary material.

\subsection{Limitations}

\begin{figure}
\centering    
    \setlength{\wid}{0.22\textwidth}
    \setlength{\mrg}{-0.45cm}
    \setlength{\mrgv}{-0.2cm}
    \begin{tabular}{cc}
        \includegraphics[width=\wid]{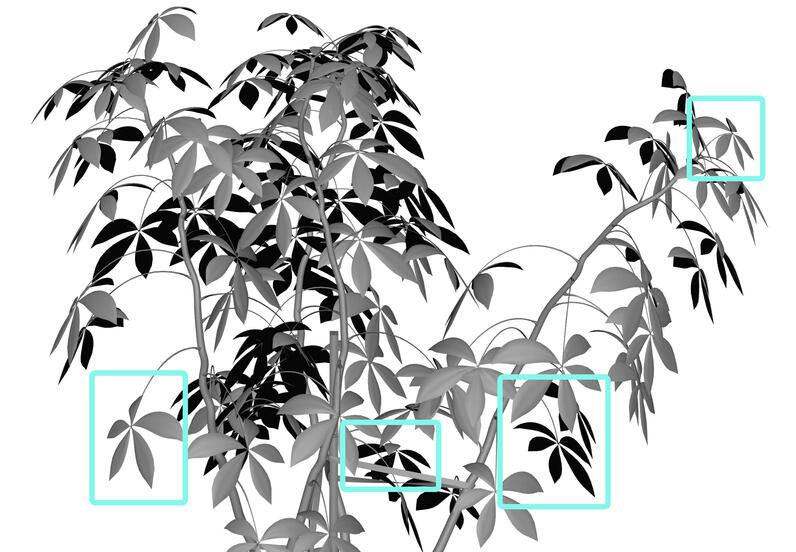} &
        \hspace{\mrg}
        \includegraphics[width=\wid]{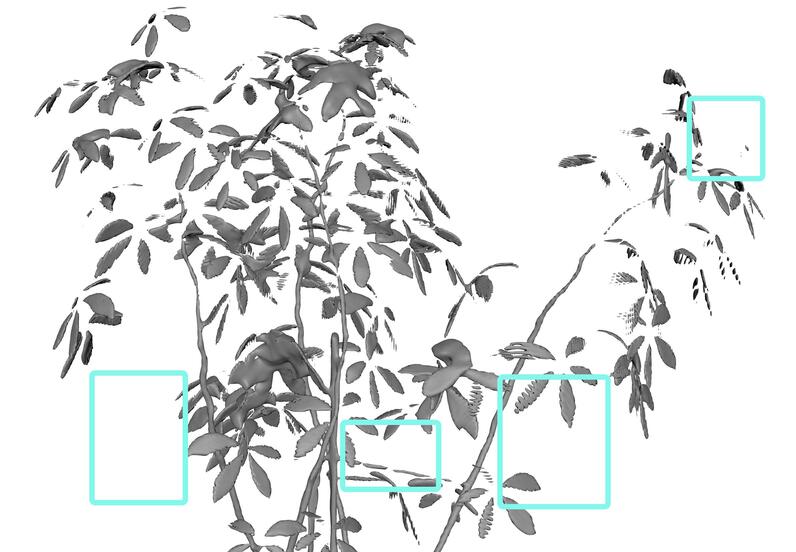} \\
        \includegraphics[width=\wid]{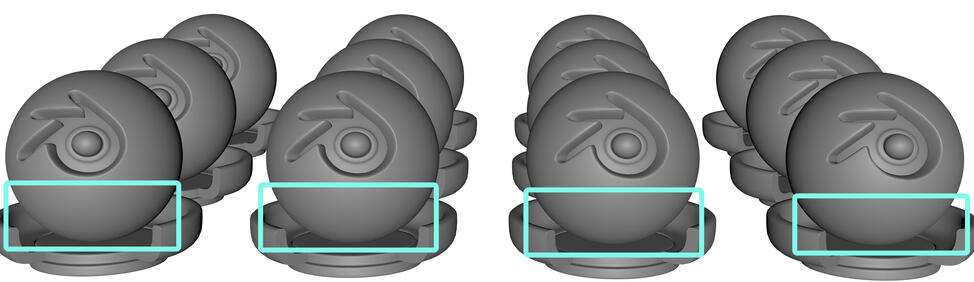} &
        \hspace{\mrg}
        \includegraphics[width=\wid]{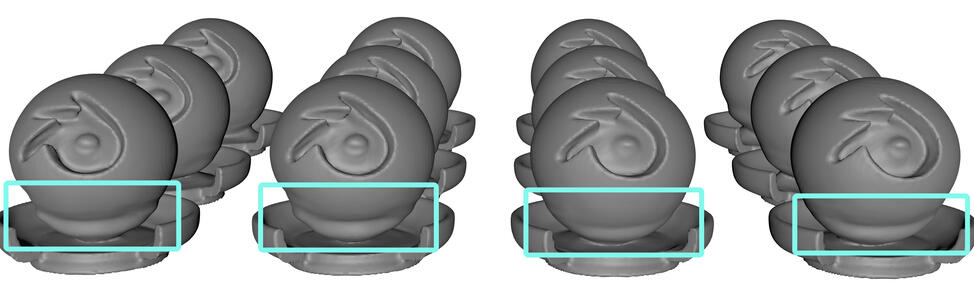} \\
        \textbf{Ground truth} & \hspace{\mrg} \textbf{NeuralWarp (ours)}
    \end{tabular}
        \caption{ The main limitation of our method is the reliance upon coarse geometry estimated relatively early in the training. Our method, by design, samples more from the regions with a large number of spheres, which may cause artifacts related to exploration. Also, systematic artifacts are not corrected and may be amplified by our algorithm.
    }
    \label{fig:limitations}
    \vspace{-0.3cm}
\end{figure}

While our approach improves the performance of base implicit surface reconstruction methods, it still inherits some of the limitations of the chosen framework. And, in some cases, even amplifies them by taking samples closer to the current approximation of the surface. In Figure~\ref{fig:limitations}, we show that the method fails to reconstruct the whole Ficus and creates artifacts on the Materials scene. In the first case, similarly to the baseline, the surface is not formed for several stems. Our spheres do not recover these regions since we skip rays that do not intersect the sphere cloud, leading to limited exploration capabilities after the initial convergence phase. On the materials scene, when using NeuralWarp, appearance variations on the base of the globes get incorrectly baked into the objects' surface. This artifact also appears and may be magnified in the case of our method.

\section{Conclusion}

We have presented a method for improving volume rendering of implicit surface representations by restricting the modeled volume to the region of interest, defined by the set of trainable spheres. We have shown that focusing the optimization process on the estimated surface region leads to an increased quality of reconstructions obtained by the implicit functions. At the same time, our proposed sphere cloud optimization approach ensures that the guiding representation closely follows the estimated surface of the object and accurately represents it at each step of the training. We have conducted an extensive evaluation of our method, which includes combining it with four base systems for implicit function training, and found it to improve their performance across multiple benchmark datasets. Our method also shows clear gains for modeling the real-world scenes when applied without \emph{any} additional supervision and post-filtering of the obtained reconstructions, which shows its suitability for in-the-wild applications.

\paragraph{Acknowledgements.} The work was supported by the Analytical center under the RF Government (subsidy agreement 000000D730321P5Q0002, Grant No. 70-2021-00145 02.11.2021).

{\small
\bibliographystyle{style/ieee_fullname}
\bibliography{refs}
}

\clearpage
\appendix
\section{Implementation details}

\paragraph{Sphere resampling.} We employ sphere resampling to ensure all the primitives model the area of interest. To identify the spheres that fail to reach the surface due to local minimum regions of the implicit geometry field, we uniformly sample $K=1000$ points in each sphere and evaluate the implicit function. Then, we mark the spheres that contain points either with all values lower, or all values higher than the surface level. Each marked sphere is replaced as follows: we randomly pick a sphere that is not marked and center a Gaussian distribution with a standard deviation of $\sigma = 2 * r_{min}$ around its origin, we sample the new origin of the sphere from this distribution; finally, we reinitialize the optimizer state for the considered primitive. A similar resampling strategy is applied periodically after every 5000 iterations for the spheres that are pushed by the repulsion outside of the scene bounds. 

We found the resampling of empty spheres to bring the most benefits to UNISURF~\cite{Oechsle2021UNISURFUN} model which uses an occupancy field for the implicit surface. This representation is more susceptible to local minimums than the SDF which has an additional Eikonal regularization~\cite{Gropp2020ImplicitGR}. Also, the number of spheres that require resampling for NeuralWarp~\cite{darmon2022improving} / VolSDF~\cite{Yariv2021VolumeRO} is around $10$ times higher in the initial steps than for NeuS~\cite{Wang2021NeuSLN}. We believe this is because the former imposes the Eikonal penalty only on two points on each ray, while the latter constrains the gradients of all points sampled.

\paragraph{Sphere-guided ray sampling.}{
Compared to base methods, the sphere-guided models have the relevant areas of the volume explicitly defined. We can exploit the sphere bounds into sampling more informative rays during training as follows: we randomly sample a point inside each sphere and project the points to the training views. We exclude the points that project outside of the image bounds and randomly sample from the remaining points a batch of corresponding camera rays which are used for a training step.}

\paragraph{Sphere-guided ray marching.}{
The proposed approach can be applied to the considered 3D reconstruction methods without altering their formulation and training process. However, the introduced sphere-guided volume rendering enhances the point sampling along the ray procedures of the base methods as described in \textbf{Algorithm 1}, which imposes the following modifications:
\begin{itemize}
    \item UNISURF~\cite{Oechsle2021UNISURFUN} - the root-finding procedure is adjusted by only searching for the surface within the volume covered by the spheres. More precisely, we sample points uniformly inside the intervals (as found at step 5 of the algorithm) to find the first sign change. We then apply the secant method on this segment as in the original method. Reducing the search to the area around the surface enables the algorithm to better estimate the ray-surface intersection. The rest of the sampling procedure follows the original model.

    \item NeuS~\cite{Wang2021NeuSLN} - the points sampled within the ray-sphere intersections are used to compute a coarse probability estimation along the ray. If the ray intersects multiple surfaces, the set computed at step 5 of the algorithm can have more than one interval. As the region between two such intervals is outside the sphere cloud, we do not want to include it in the importance sampling. Therefore, we set the probabilities of these regions to zero, ensuring that the added points through importance sampling will belong to the set of intervals. Similarly, we set the weights of the midpoints used in color computation that fall outside the sphere bounds to zero. In the experiments that do not have segmentation masks as input, NeuS samples a set of points outside the bounds of the scene for background modeling; we do not interfere with these samples.
    \item VolSDF~\cite{Yariv2021VolumeRO} / NeuralWarp~\cite{darmon2022improving} - We perform similar modifications as for the NeuS model. We set the uncertainty estimation of the ray segments between intervals to zero, so that the points added during the upsampling stage are contained within the intersections of the ray with the sphere cloud. Additionally, we consider the estimated opacity of the previously mentioned segments as zero before performing inverse transform sampling to ensure that the final set of points lies within the intervals computed at step 5 of the algorithm. We do not interfere with the points sampled in the base method for background modeling outside the scene bounds. 
\end{itemize}
}
\section{Additional results}
\begin{table*}[!ht]
    \centering
    \begin{tabular}{l ccc ccc cc | c}
        & \multicolumn{8}{c}{PSNR $\uparrow$}
        \\
        Method & Chair & Drums & Ficus & Hotdog & Lego & Mats & Mic & Ship & Mean 
        \\
        \hline
        NeRF~\cite{Mildenhall2020NeRFRS} & 33.00 & 25.01 & 30.13 & 36.18 & 32.54 & 29.62 & 32.91 & 28.65 & 31.01\\
        \hline
        NeuS~\cite{Wang2021NeuSLN} & 30.89 & 20.89 & 27.44 & 36.04 & 30.45 & 30.21 & 31.13 & 27.08 & 29.26
        \\
        NeuS (ours) & \textbf{32.21} & \textbf{21.81} & \textbf{30.40} & \textbf{36.76} & \textbf{31.80} & \textbf{31.08} & \textbf{31.44} & \textbf{28.63} & \textbf{30.52}
        \\
        \hline
        NeuralWarp~\cite{darmon2022improving} & 29.29 & 18.41 & 24.50 & 32.32 & 27.90 & \textbf{27.45} & \textbf{29.15} & 24.07 & 26.64
        \\
        NeuralWarp (ours) & \textbf{30.17} & \textbf{18.82} & \textbf{26.79} & \textbf{32.73} & \textbf{28.81} & 24.76 & 28.67 & \textbf{25.22} & \textbf{27.00} \\
        \hline
        \vspace{-0.2cm}
    \end{tabular}
    \begin{tabular}{l ccc ccc cc | c}
        & \multicolumn{8}{c}{SSIM $\uparrow$}
        \\
        Method & Chair & Drums & Ficus & Hotdog & Lego & Mats & Mic & Ship & Mean 
        \\
        \hline
        NeRF~\cite{Mildenhall2020NeRFRS} & 0.967 & 0.925 & 0.964 & 0.974 & 0.961 & 0.949 & 0.980 & 0.856 & 0.947\\
        \hline
        NeuS~\cite{Wang2021NeuSLN} & 0.948 & 0.897 & 0.957 & 0.973 & 0.952 & 0.960 & 0.971 & 0.864 & 0.940
        \\
        NeuS (ours) & \textbf{0.960} & \textbf{0.909} & \textbf{0.973} & \textbf{0.978} & \textbf{0.962} & \textbf{0.965} & \textbf{0.974} & \textbf{0.883} & \textbf{0.951}
        \\
        \hline
        NeuralWarp~\cite{darmon2022improving} & 0.936 & 0.872 & 0.933 & 0.960 & 0.928 & 0.938 & \textbf{0.962} & 0.817 & 0.918
        \\
        NeuralWarp (ours) & \textbf{0.947} & \textbf{0.878} & \textbf{0.950} & \textbf{0.964} & \textbf{0.939} & \textbf{0.941} & 0.960 & \textbf{0.828} & \textbf{0.926} \\
        \hline
        \vspace{-0.2cm}
    \end{tabular}
    \begin{tabular}{l ccc ccc cc | c}
        & \multicolumn{8}{c}{LPIPS $\downarrow$}
        \\
        Method & Chair & Drums & Ficus & Hotdog & Lego & Mats & Mic & Ship & Mean 
        \\
        \hline
        NeRF~\cite{Mildenhall2020NeRFRS} & 0.046 & 0.091 & 0.044 & 0.121 & 0.050 & 0.063 & 0.028 & 0.206 & 0.081\\
        \hline
        NeuS~\cite{Wang2021NeuSLN} & 0.064 & 0.118 & 0.047 & 0.045 & 0.060 & 0.053 & 0.030 & 0.184 & 0.075
        \\
        NeuS (ours) & \textbf{0.050} & \textbf{0.103} & \textbf{0.029} & \textbf{0.037} & \textbf{0.047} & \textbf{0.048} & \textbf{0.026} & \textbf{0.165} & \textbf{0.063}
        \\
        \hline
        NeuralWarp~\cite{darmon2022improving} & 0.076 & 0.161 & 0.072 & 0.068 & 0.091 & \textbf{0.074} & \textbf{0.043} & 0.243 & 0.104
        \\
        NeuralWarp (ours) & \textbf{0.065} & \textbf{0.142} & \textbf{0.053} & \textbf{0.061} & \textbf{0.076} & 0.076 & 0.047 & \textbf{0.228} & \textbf{0.094} \\
        \hline
        \vspace{-0.2cm}
    \end{tabular}
    \caption{Quantitative image results on the Realistic Synthetic 360 dataset~\cite{Mildenhall2020NeRFRS}. We evaluate PSNR, SSIM (higher is better), and LPIPS \cite{Zhang2018TheUE} (lower is better). The proposed approach improves the rendering quality of the NeuS and NeuralWarp models.
    }\label{tab:nerf_quant_render}
    \vspace{-0.2cm}
\end{table*}

\def\S{\mathbf{S}}
\paragraph{Further ablation study.} We perform additional experiments to highlight the benefits of the proposed sphere-guided training. 
We compare the baseline NeuS model and our improved model in three different setups by varying the number of points sampled per ray: 128 points (default setting), 64 points (32 linearly spaced + 32 importance sampled), and 32 points (16 linearly spaced + 16 importance sampled). The results indicate that our approach can better handle the diminished number of input points, as can be seen in Figure~\ref{fig:ablation_appendix}.

We further evaluate independently the effect of the proposed ray-sampling and ray-marching procedures based on the optimized sphere cloud. We consider both the base setting with 128 points per ray and also using 64 points per ray. We show in Table \ref{tab:ablation_app} that both components contribute to our final results. Sphere-based ray sampling has a larger and more consistent effect by itself in the default setting, whereas the ray-marching procedure becomes more important when the number of samples is low. This is because it is harder to approximate the integral with fewer points per ray, and an improved sampling strategy has more obvious benefits. The effect is especially visible in scenes with thin structures, such as Ficus (Figure \ref{fig:ablation_appendix}) and Ship. We note that the results on Drums are not always representative because the scene contains semi-transparent surfaces, which are, by design, not handled well by the base methods.

\begin{figure*}
    \centering    
    \setlength{\wid}{0.20\textwidth}
    \setlength{\mrg}{-0.7cm}
    \begin{tabular}{cccc}
        \includegraphics[width=\wid]{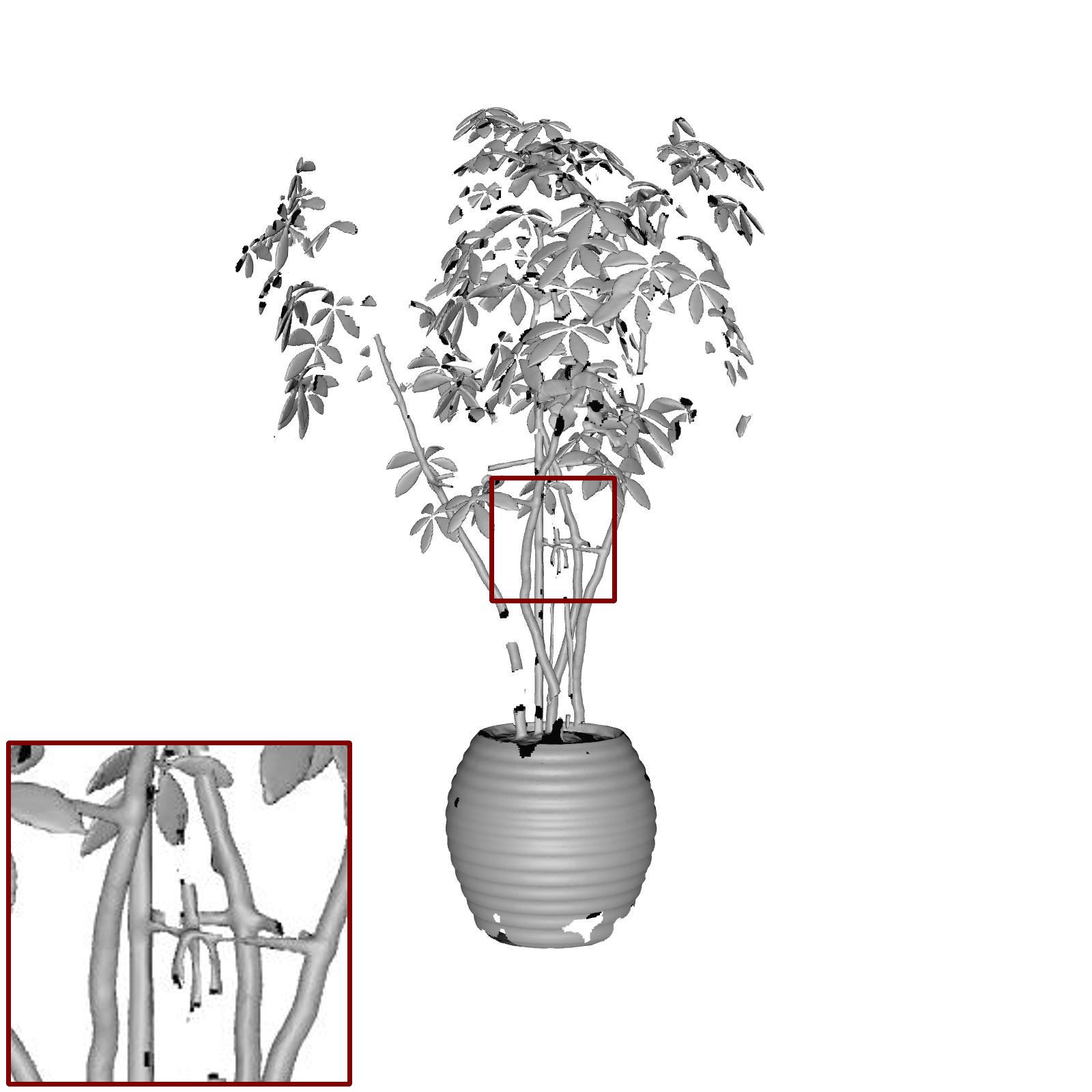} &
        \hspace{\mrg}
        \includegraphics[width=\wid]{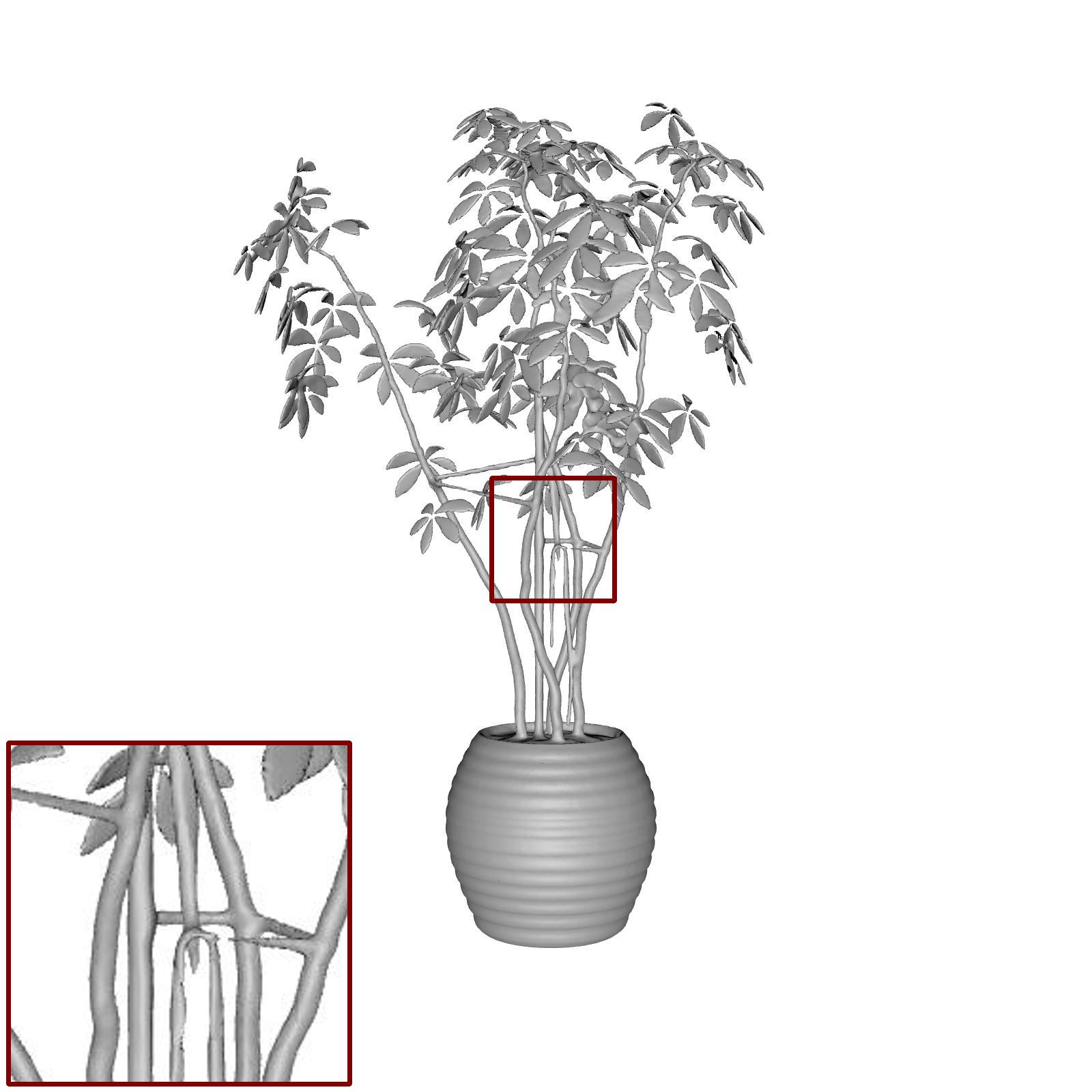} &
        \hspace{\mrg}
        \includegraphics[width=\wid]{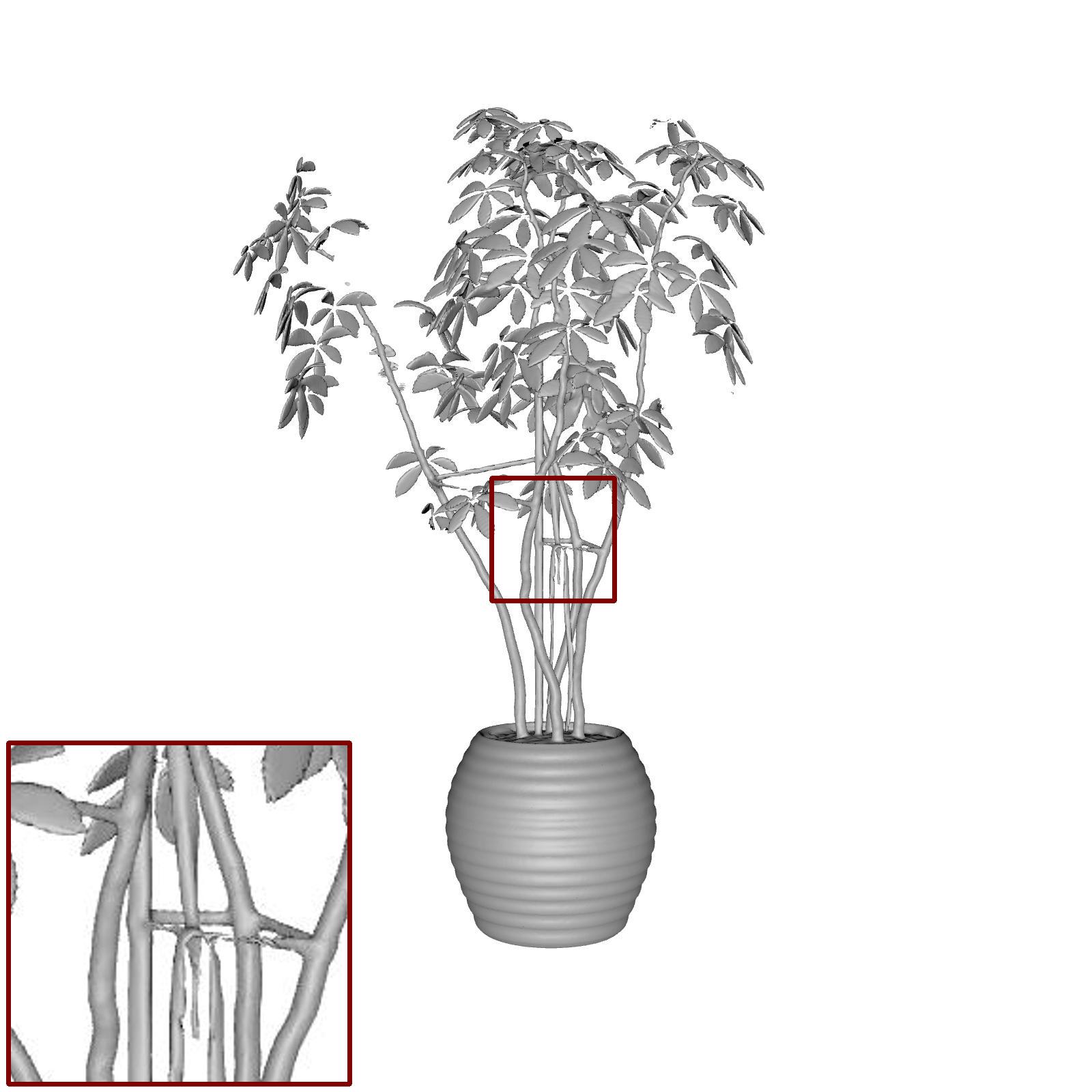} &
        \hspace{\mrg}
        \includegraphics[width=\wid]{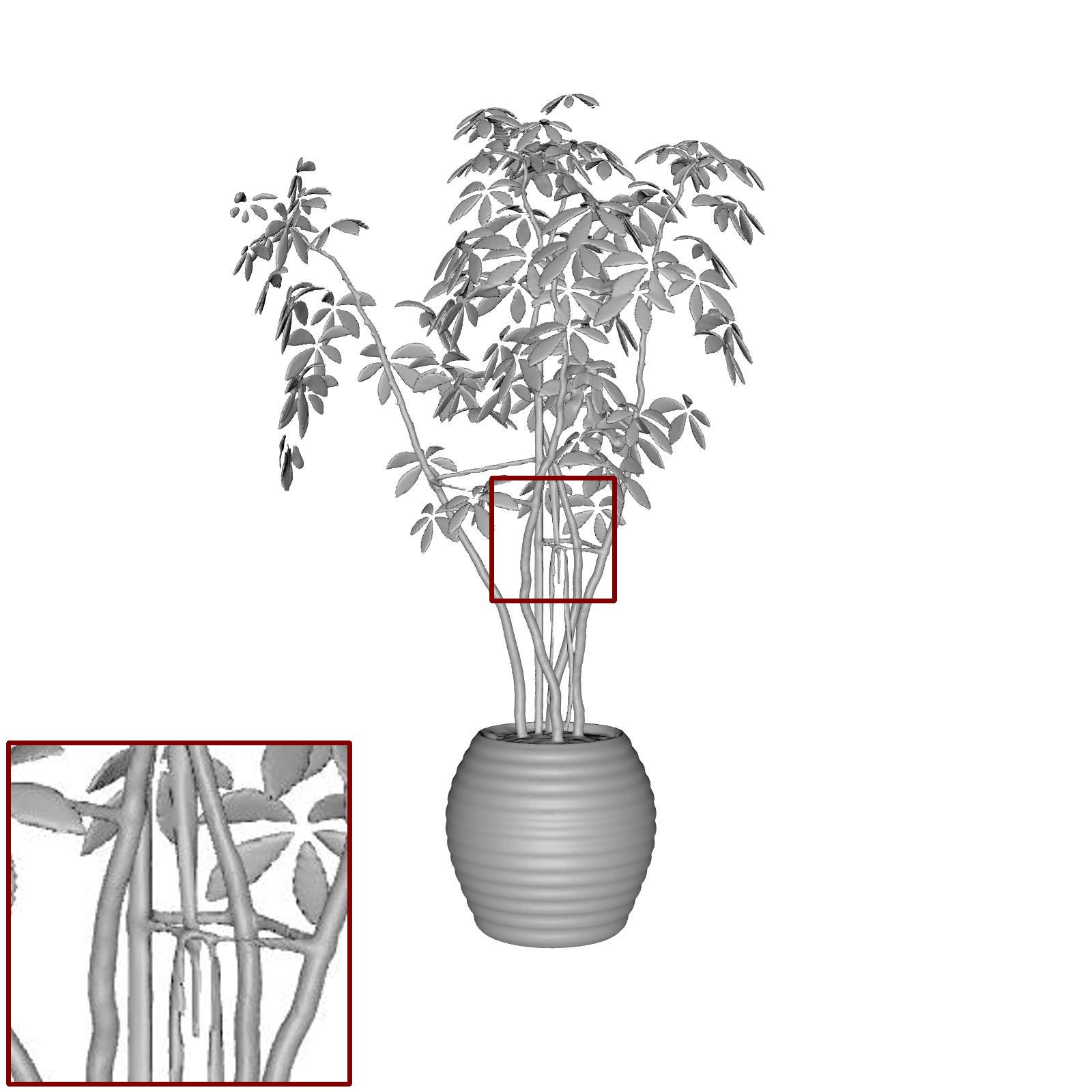}
        \vspace{0.1cm}
        \\
        \vspace{0.1cm}
        0.79 & \hspace{\mrg} 0.49 & \hspace{\mrg} 0.49 & \hspace{\mrg} 0.40
        \\
        \textbf{w/o $\mathcal{L}_\text{rep}$} & \hspace{\mrg}
        \textbf{\begin{tabular}{c} w/o $\S$-guided \\ ray sampling \end{tabular}} & \hspace{\mrg}
        \textbf{\begin{tabular}{c} NSVF ablation \end{tabular}} & \hspace{\mrg}
        \textbf{Ours}
    \end{tabular}
    \vspace{-0.2cm}
    \caption{Ablation study. We have used the NeuS base system, trained without mask supervision, as a base model. We show both the resulting qualitative results, as well as obtained Chamfer distances for the shown scene. As reference, the baseline NeuS model obtains a Chamfer distance of $0.51$.
    }
    \label{fig:ablation}
    \vspace{-0.2cm}
\end{figure*}

We also include a visual comparison between our complete model and three ablations in Figure~\ref{fig:ablation}. For this comparison, we use NeuS as the base system and train all versions without mask supervision.

\begin{figure}
    \centering    
    \setlength{\wid}{0.45\textwidth}
    \begin{tabular}{c c c}
        \includegraphics[width=\wid]{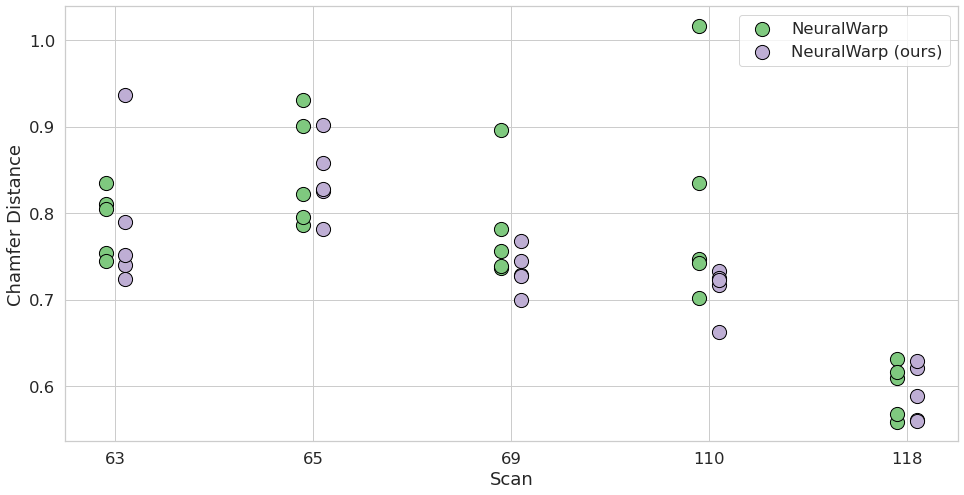}
    \end{tabular}
    \caption{Quantitative results for the subset of five scenes of the DTU~\cite{Jensen2014LargeSM} dataset. We separately train each method for each scene five times with different starting random seeds. The results for our approach are reported in green, while for the base method they are in purple. 
    The $x$-axis represents the scene ID, and the $y$-axis shows the obtained Chamfer distances. This comparison utilizes a masked Chamfer distance, following the same setting as in Table~1 in the experiments section. Our approach achieves a noticeably reduced variance of the results compared to the base system, and outperforms it on average, obtaining mean Chamfer distance of $0.73$, averaged across five seeds and five scans, against the NeuralWarp's $0.76$. We also have better worst- and best-case performance, with our method having mean best-case metrics of $0.69$ against NeuralWarp's $0.71$ and mean worst-case of $0.79$ against $0.86$.}
    \label{fig:dtu_quant_appendix}
\end{figure}

\begin{figure*}
    \centering    
    \setlength{\wid}{0.28\textwidth}
    \setlength{\mrg}{-0.45cm}
    \begin{tabular}{cccc}
        & \hspace{\mrg} \textbf{\begin{tabular}{c} 128 points \\ per ray \end{tabular}} & \hspace{\mrg} \textbf{\begin{tabular}{c} 64 points \\ per ray \end{tabular}} & \hspace{\mrg} \textbf{\begin{tabular}{c} 32 points \\ per ray \end{tabular}} \\
        \textbf{NeuS}&
        \hspace{\mrg}
        \includegraphics[width=\wid]{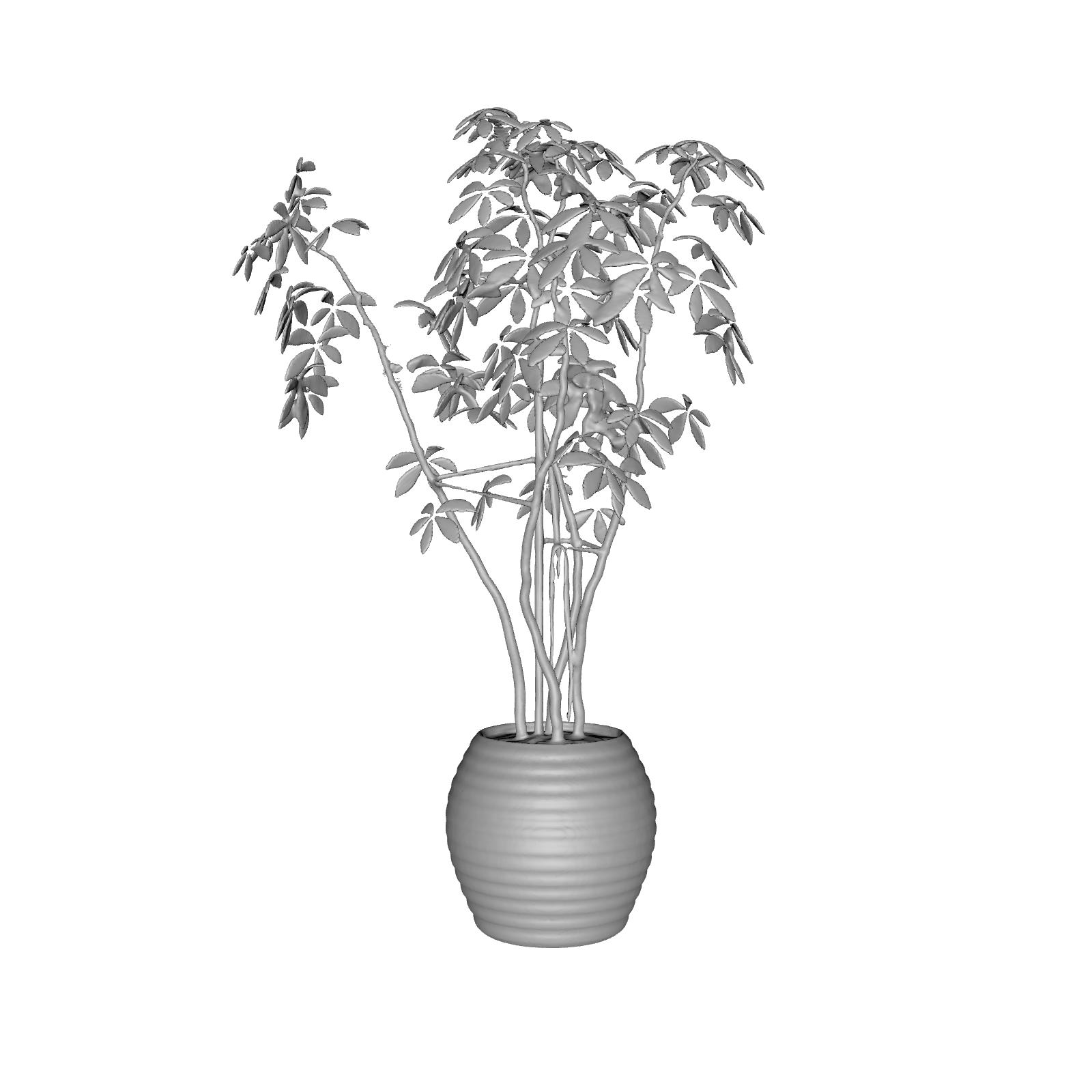} &
        \hspace{\mrg}
        \includegraphics[width=\wid]{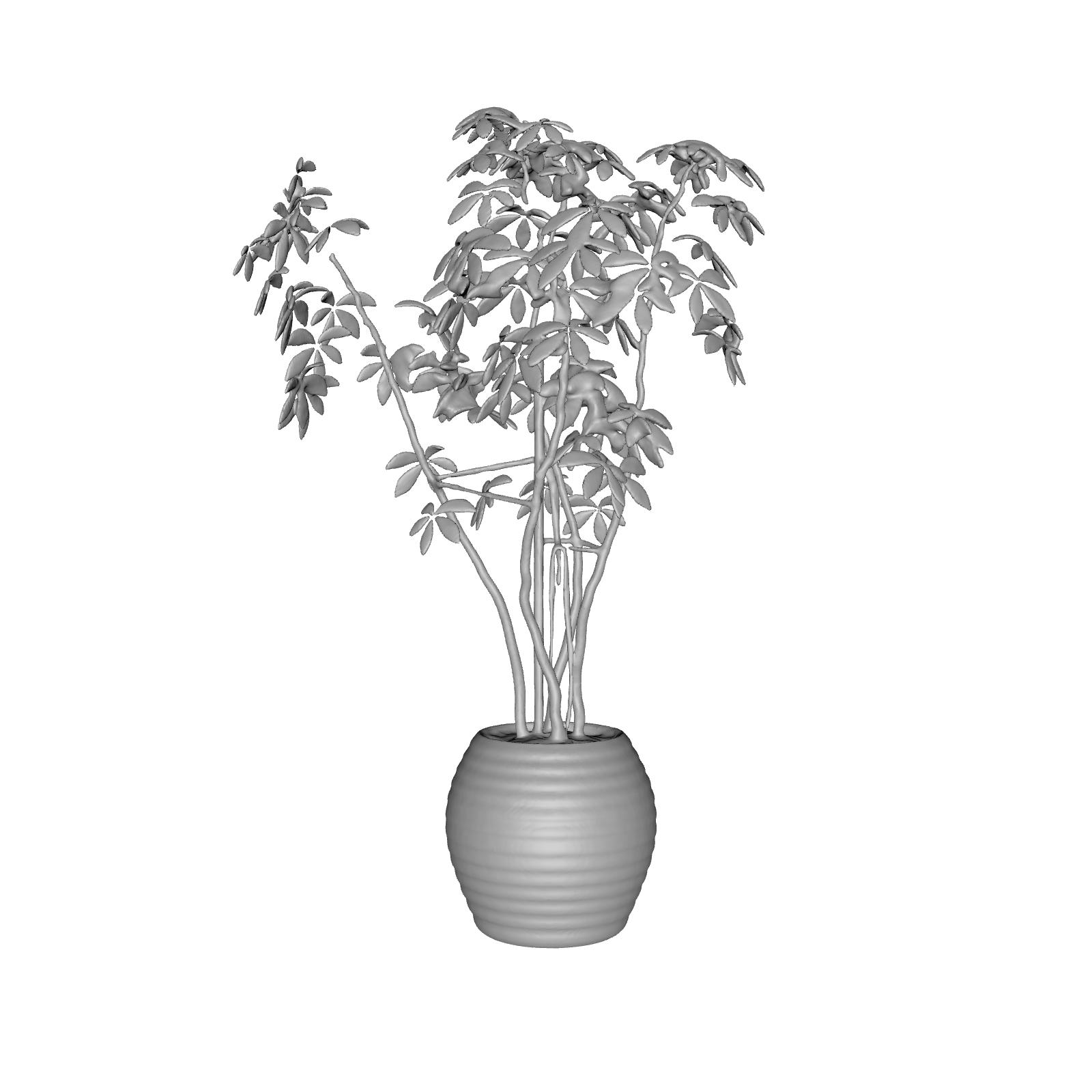} &
        \hspace{\mrg}
        \includegraphics[width=\wid]{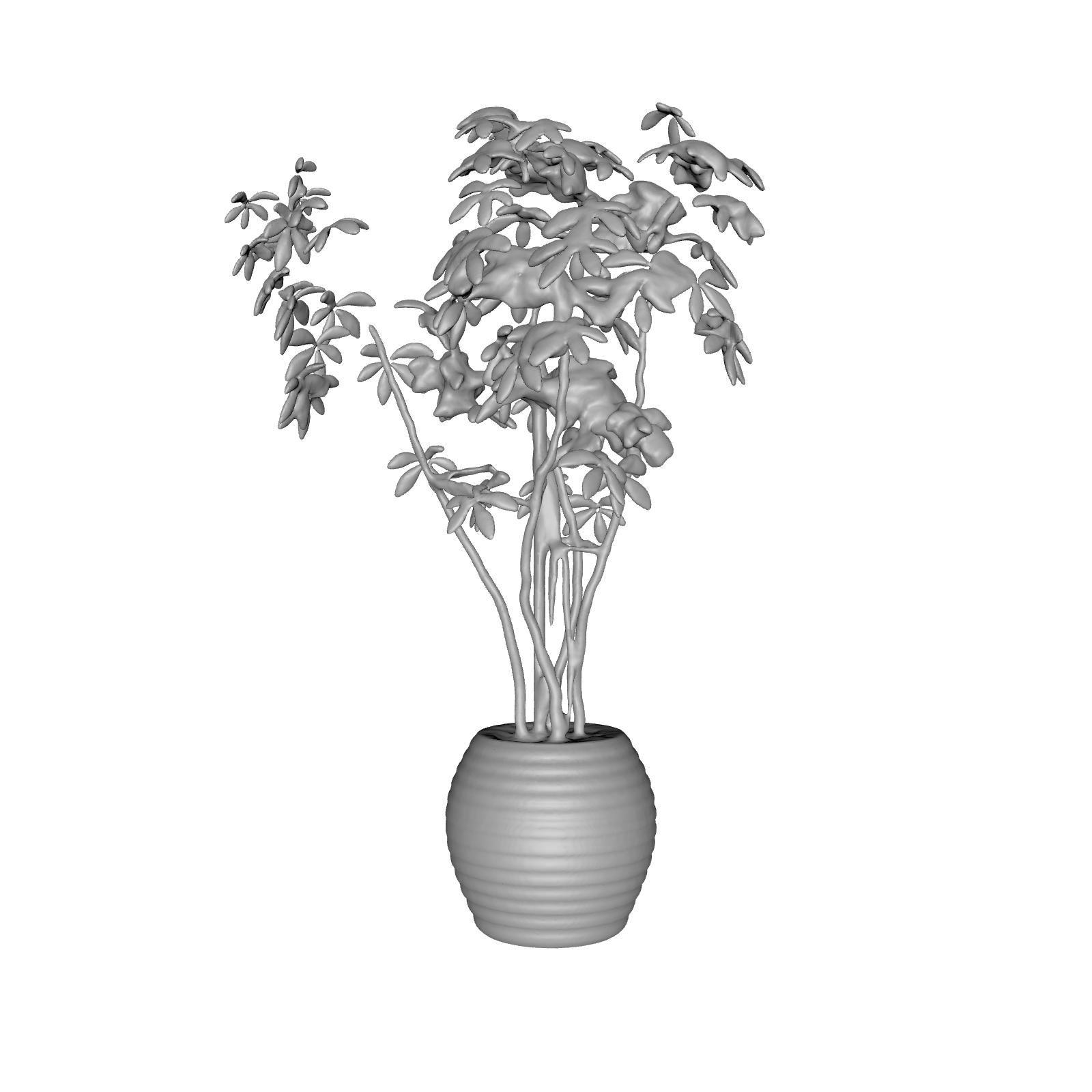} 
        \\
        & \hspace{\mrg} 0.50  & \hspace{\mrg} 0.61 & \hspace{\mrg} 0.87
        \\
        \textbf{\begin{tabular}{c} w/o $\S$-guided \\ ray marching \end{tabular}} &
        \hspace{\mrg}
        \includegraphics[width=\wid]{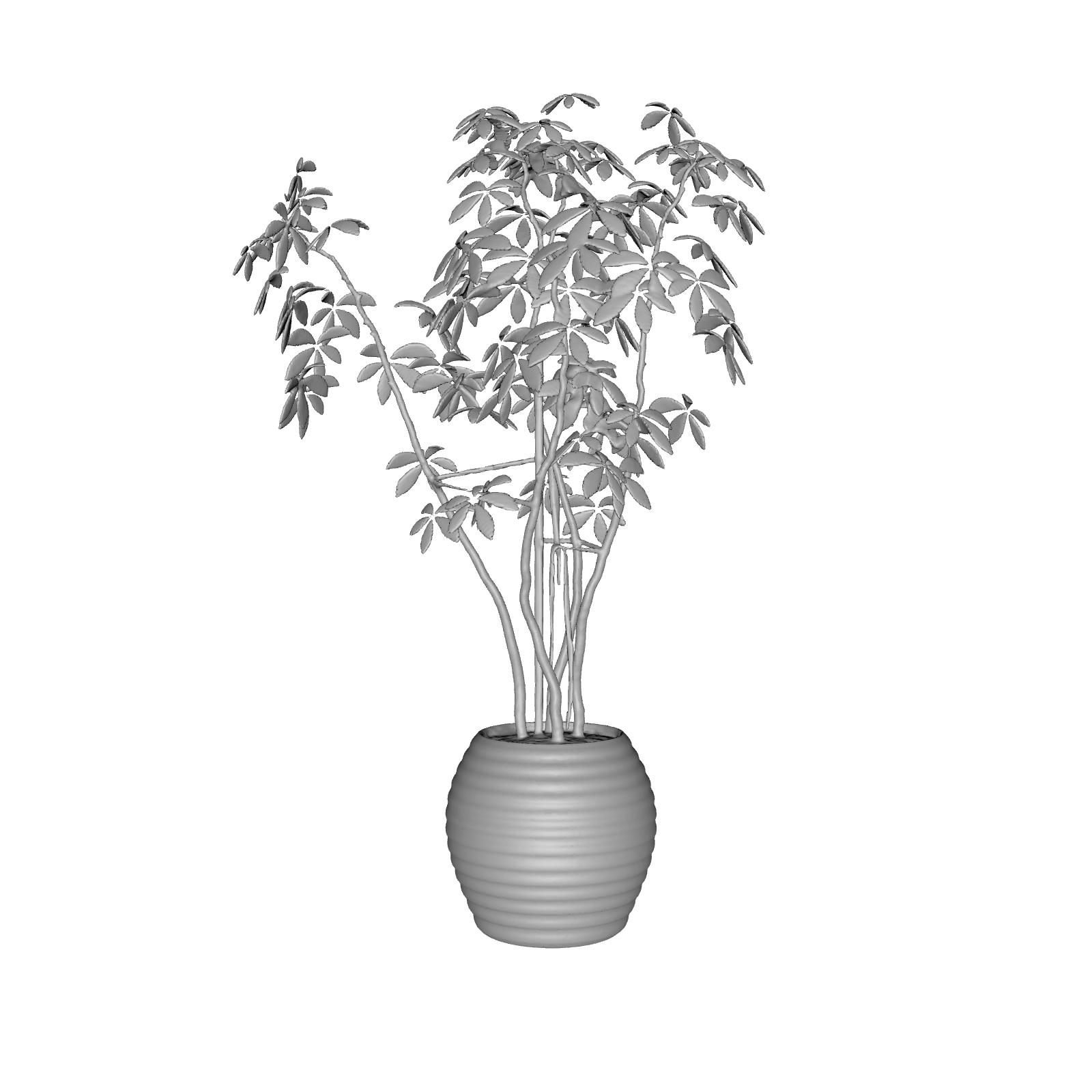} &
        \hspace{\mrg}
        \includegraphics[width=\wid]{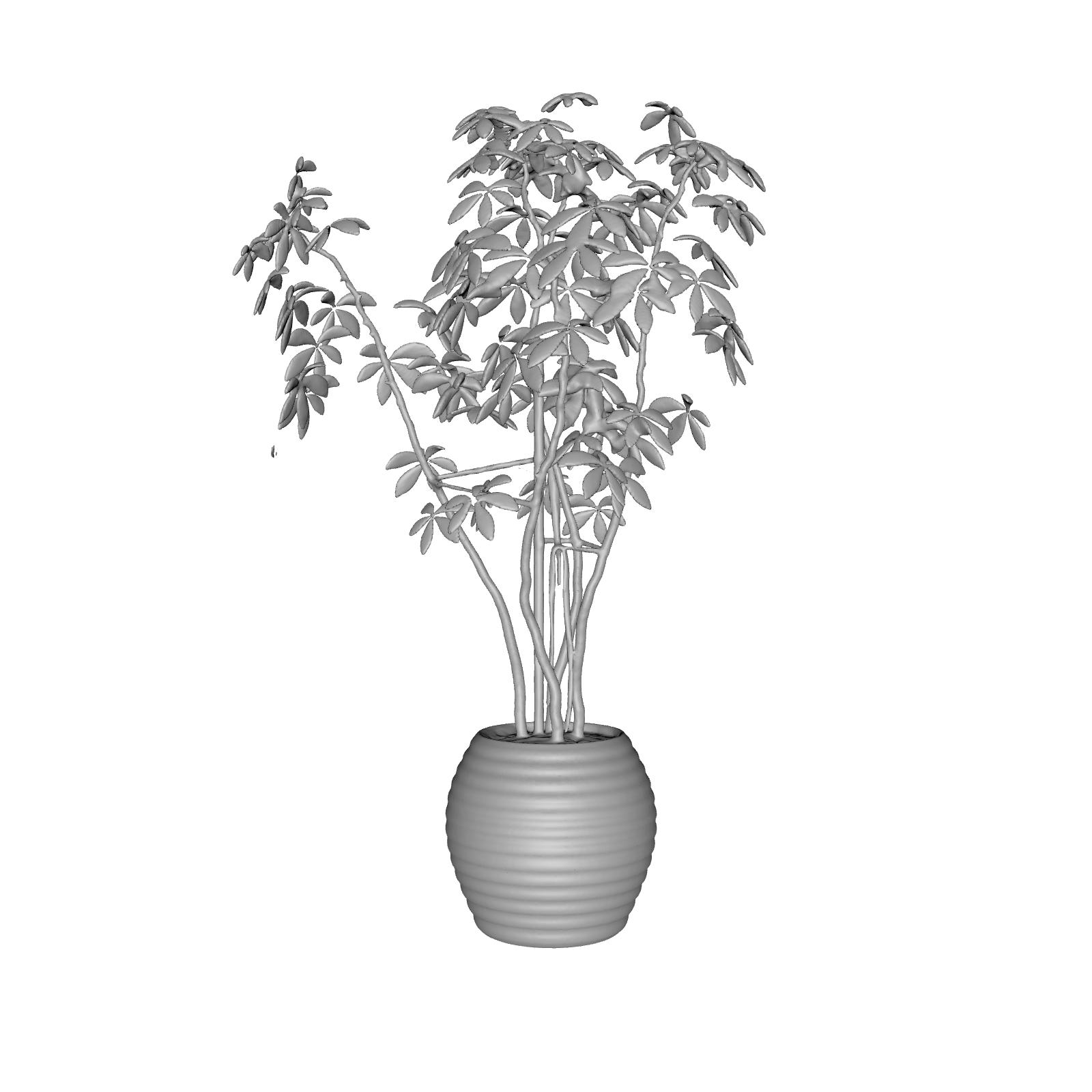} &
        \hspace{\mrg}
        \includegraphics[width=\wid]{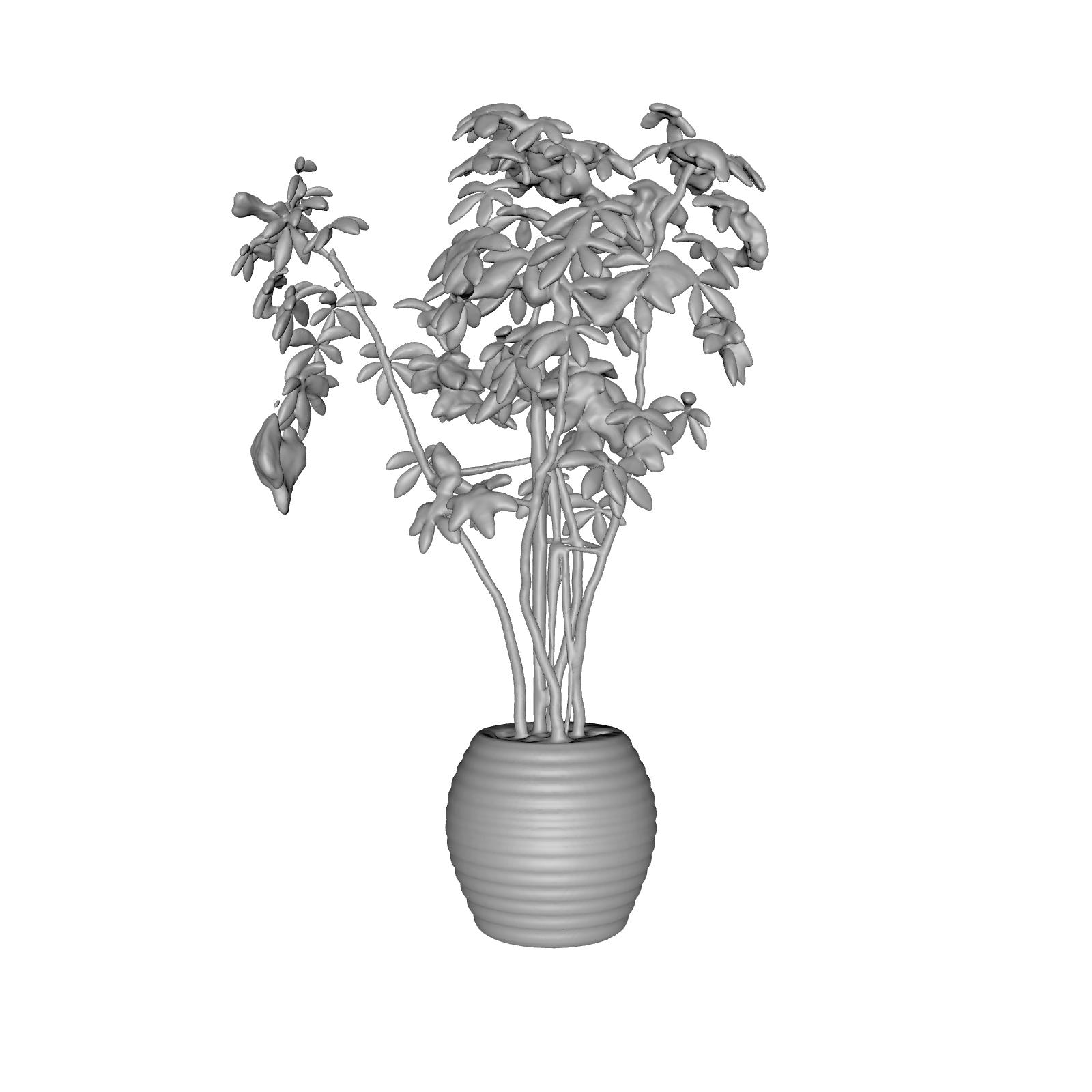} 
        \\
        & \hspace{\mrg} 0.40  & \hspace{\mrg} 0.52 & \hspace{\mrg} 0.87
        \\
        \textbf{Ours} &
        \hspace{\mrg} 
        \includegraphics[width=\wid]{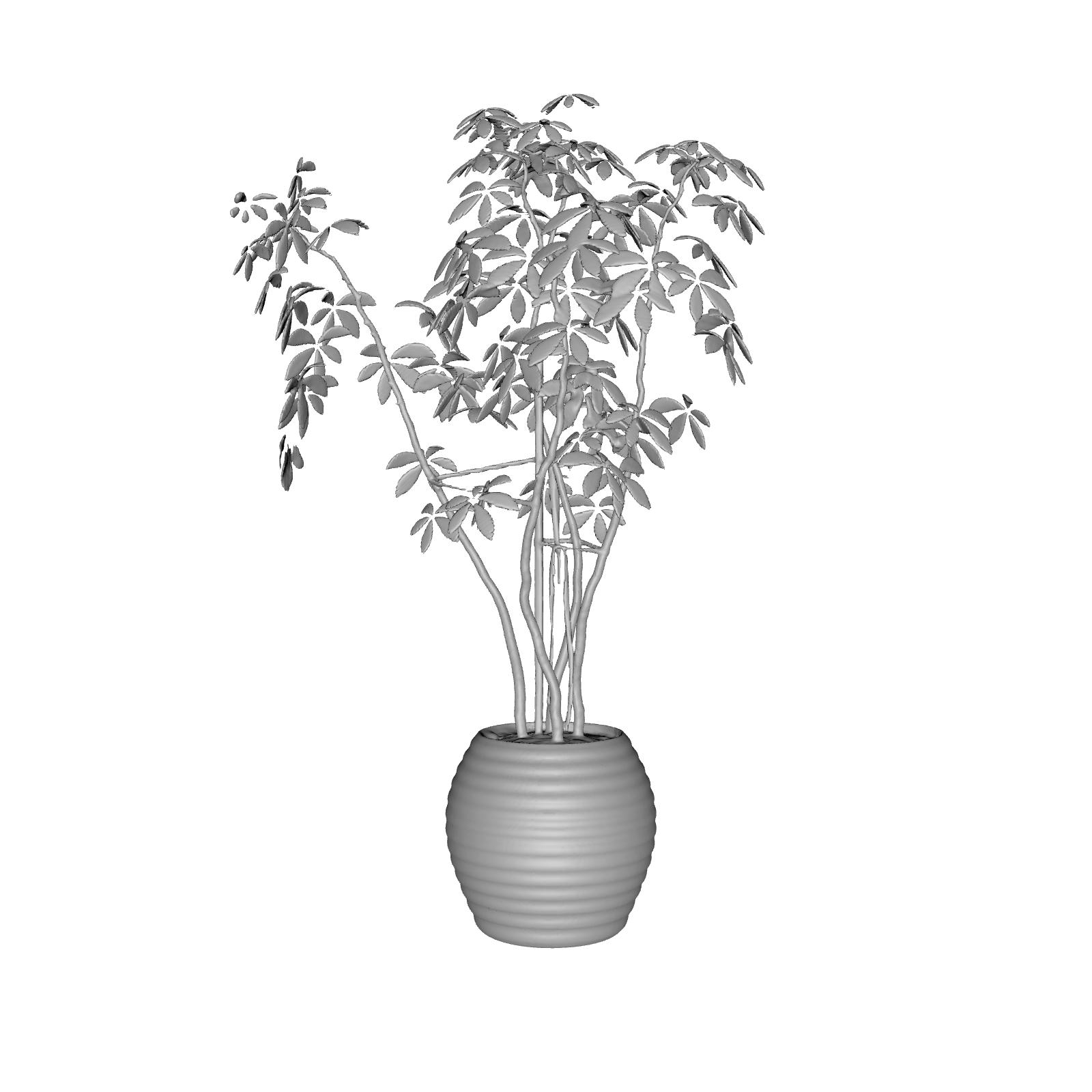} &
        \hspace{\mrg}
        \includegraphics[width=\wid]{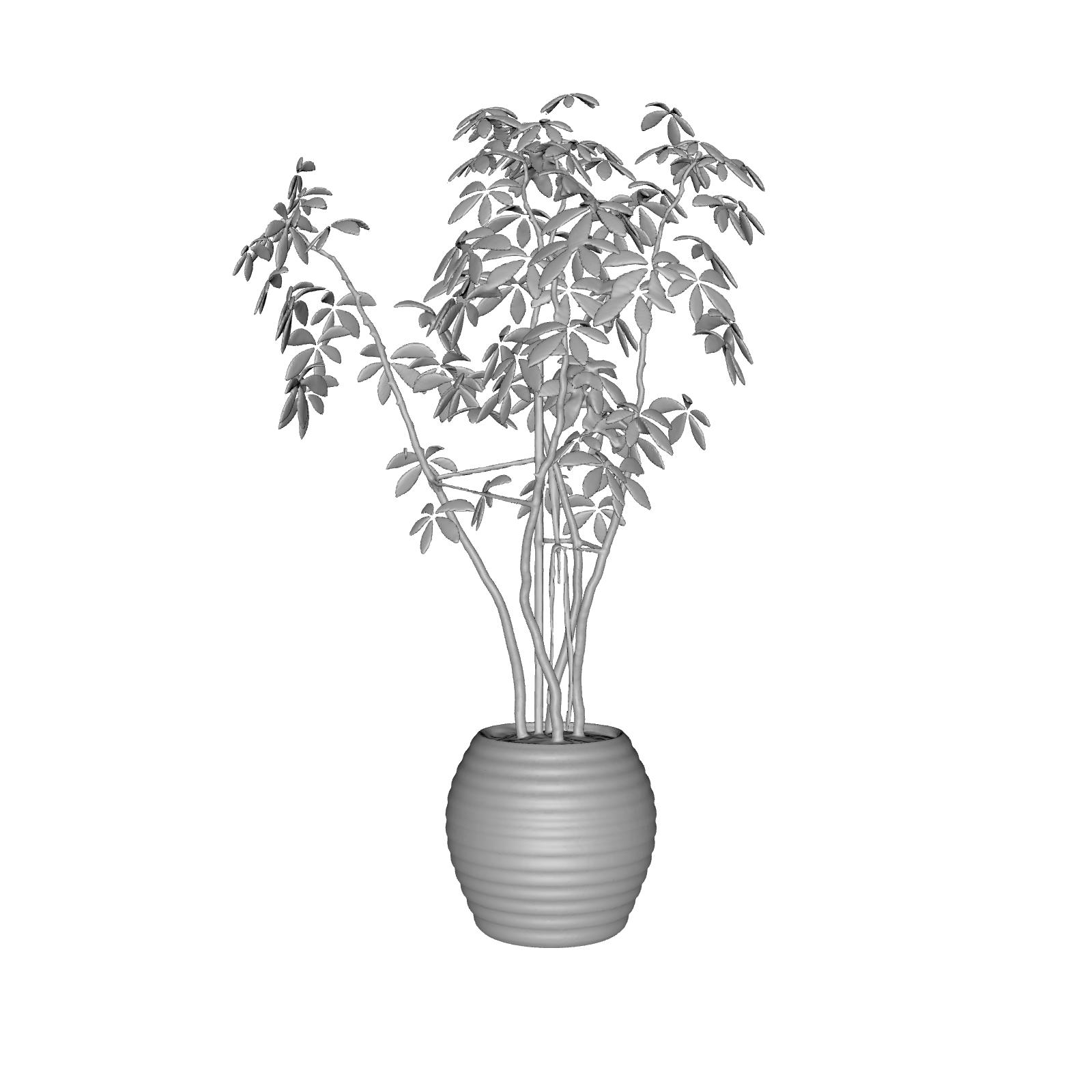} &
        \hspace{\mrg}
        \includegraphics[width=\wid]{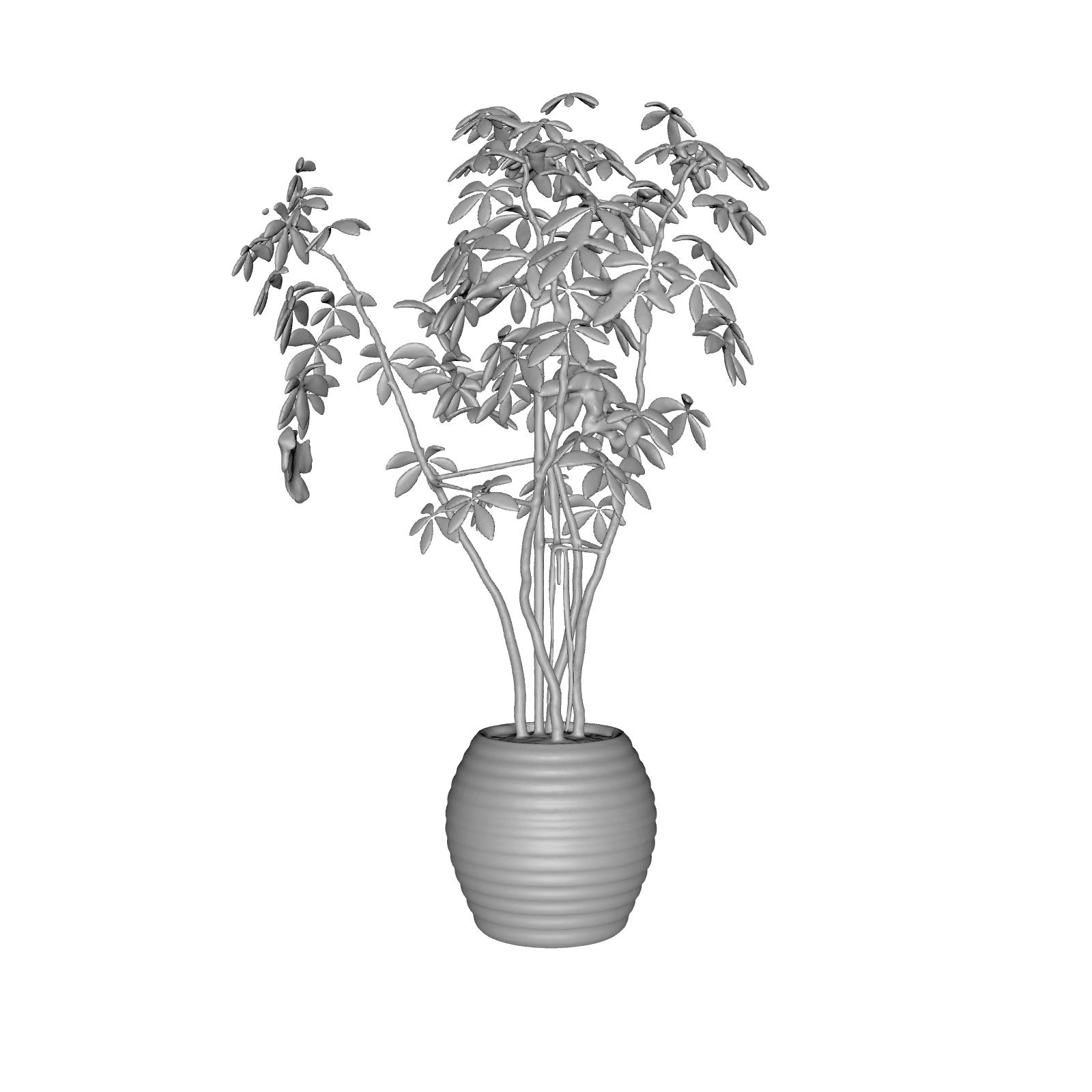} 
        \\
        & \hspace{\mrg} 0.40  & \hspace{\mrg} 0.45 & \hspace{\mrg} 0.55
    \end{tabular}
    \caption{Performance comparison when training with a reduced number of points per ray. For each reconstruction, Chamfer distances are reported under the rendered meshes. Our model achieves larger improvements when the number of points per ray is lower.
    }
    \label{fig:ablation_appendix}
\end{figure*}
\begin{table*}[!ht]
    \centering
    \begin{tabular}{l cccccccc }
        & \multicolumn{8}{c}{Scene name}
        \\
        Method & Chair & Drums & Ficus & Hotdog & Lego & Mats & Mic & Ship
        \\
        \hline
        NeuS & 0.38 & 1.88 & 0.51 & \textbf{0.52} & 0.68 & 0.40 & \textbf{0.60} & 0.60 
        \\
        w/o $\S$-guided ray sampling & \textbf{0.36} & 1.79 & 0.49 & 0.55 & 0.69 & 0.43 & 0.72 & 0.79 \\
        
        w/o $\S$-guided ray marching & 0.38 & \textbf{0.81} & \textbf{0.40} & 0.54 & \textbf{0.61} & \textbf{0.29} & 0.69 & 0.67 \\
        
        NeuS (ours) & 0.39 & 1.20 & \textbf{0.40} & 0.57 & \textbf{0.61} & 0.31 & 0.67 & \textbf{0.54}
        \\
        \hline
        NeuS 64 & \textbf{0.35} & 2.28 & 0.61 & \textbf{0.52} & 0.73 & 0.35 & 0.57 & 0.84
        \\
        w/o S-guided ray marching 64 & 0.38 & 1.64 & 0.52 & 0.55 & 0.65 & \textbf{0.28} & 0.58 & 0.75 
        \\
        NeuS (ours) 64 & 0.40 & \textbf{0.79} & \textbf{0.45} & 0.59 & \textbf{0.60} & \textbf{0.28} & \textbf{0.56} & \textbf{0.69} \\
        \hline
        \vspace{-0.2cm}
    \end{tabular}
    \caption{Ablation study on the importance of sphere-based ray sampling and marching. We evaluate the base setting with 128 points per ray and also using 64 points per ray (as fewer points emphasize the effect of improved ray-marching).
    }\label{tab:ablation_app}
    \vspace{-0.2cm}
\end{table*}
\begin{figure}
    \centering
    \includegraphics[width=0.45\textwidth]{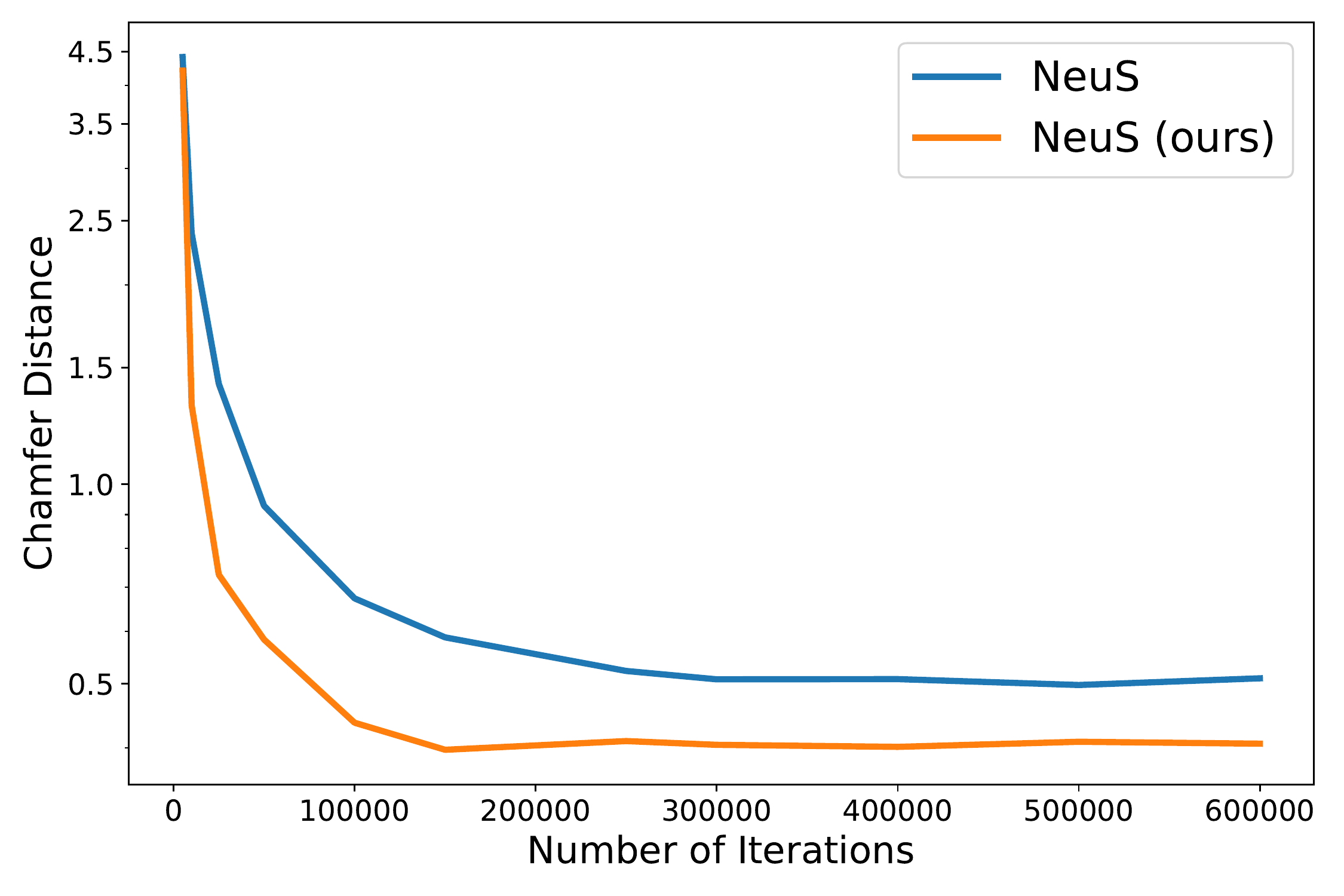}\vspace{-0.3cm}
    \caption{The learning curves for the ficus scene of the Realistic Synthetic 360 dataset showcase that our method converges to a better local optima than the base system, while using the same set of hyperparameters.}
    \label{fig:convergence}
    \vspace{-0.6cm}
\end{figure}

\paragraph{DTU dataset experiments.}

\begin{figure*}
    \centering    
    \setlength{\wid}{0.20\textwidth}
    \setlength{\mrg}{-0.45cm}
    \setlength{\mrgv}{-0.05cm}
    \begin{tabular}{c cccc}
        \vspace{\mrgv}
        \includegraphics[width=\wid]{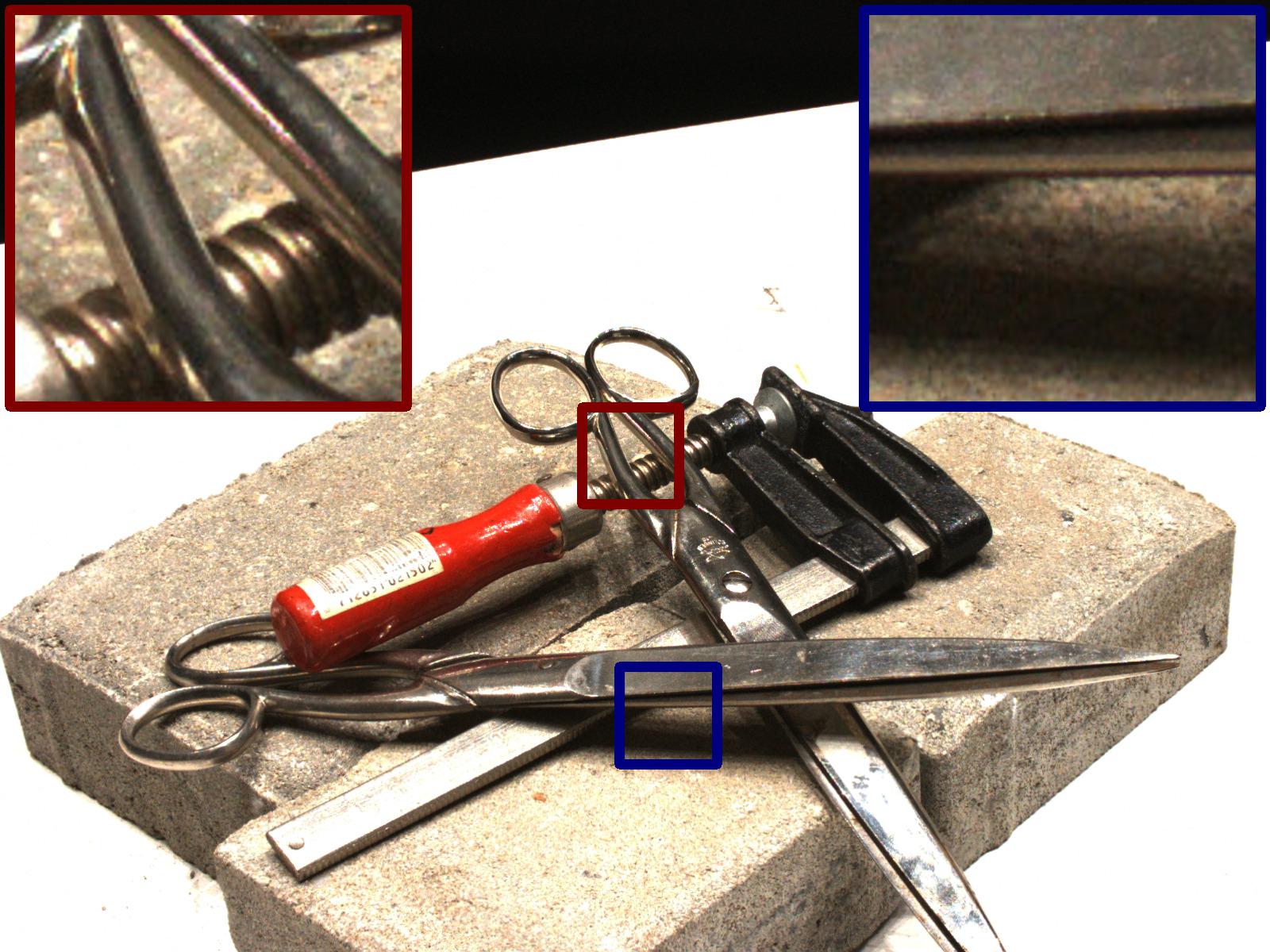} &
        \hspace{\mrg}
        \includegraphics[width=\wid]{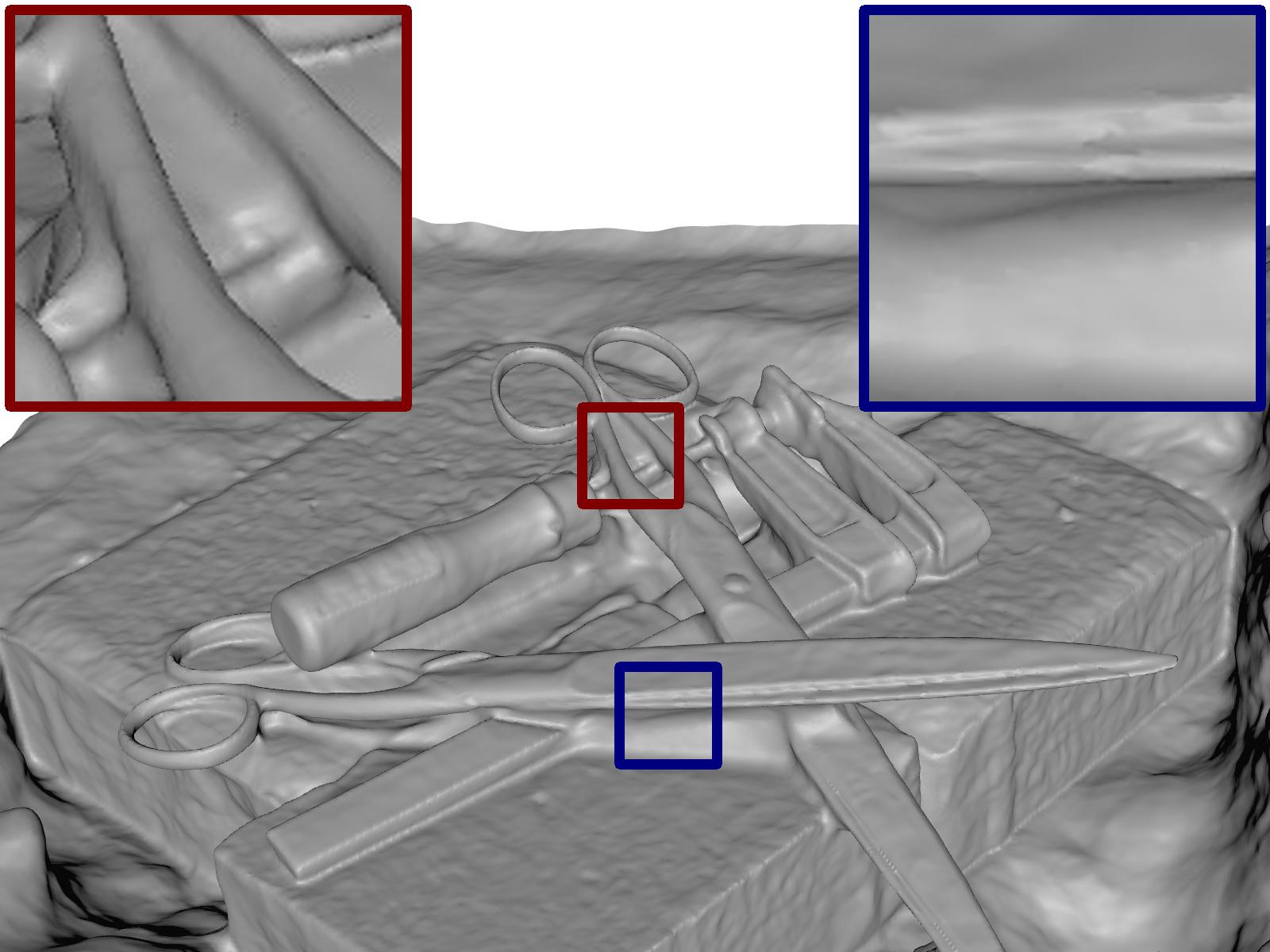} &
        \hspace{\mrg}
        \includegraphics[width=\wid]{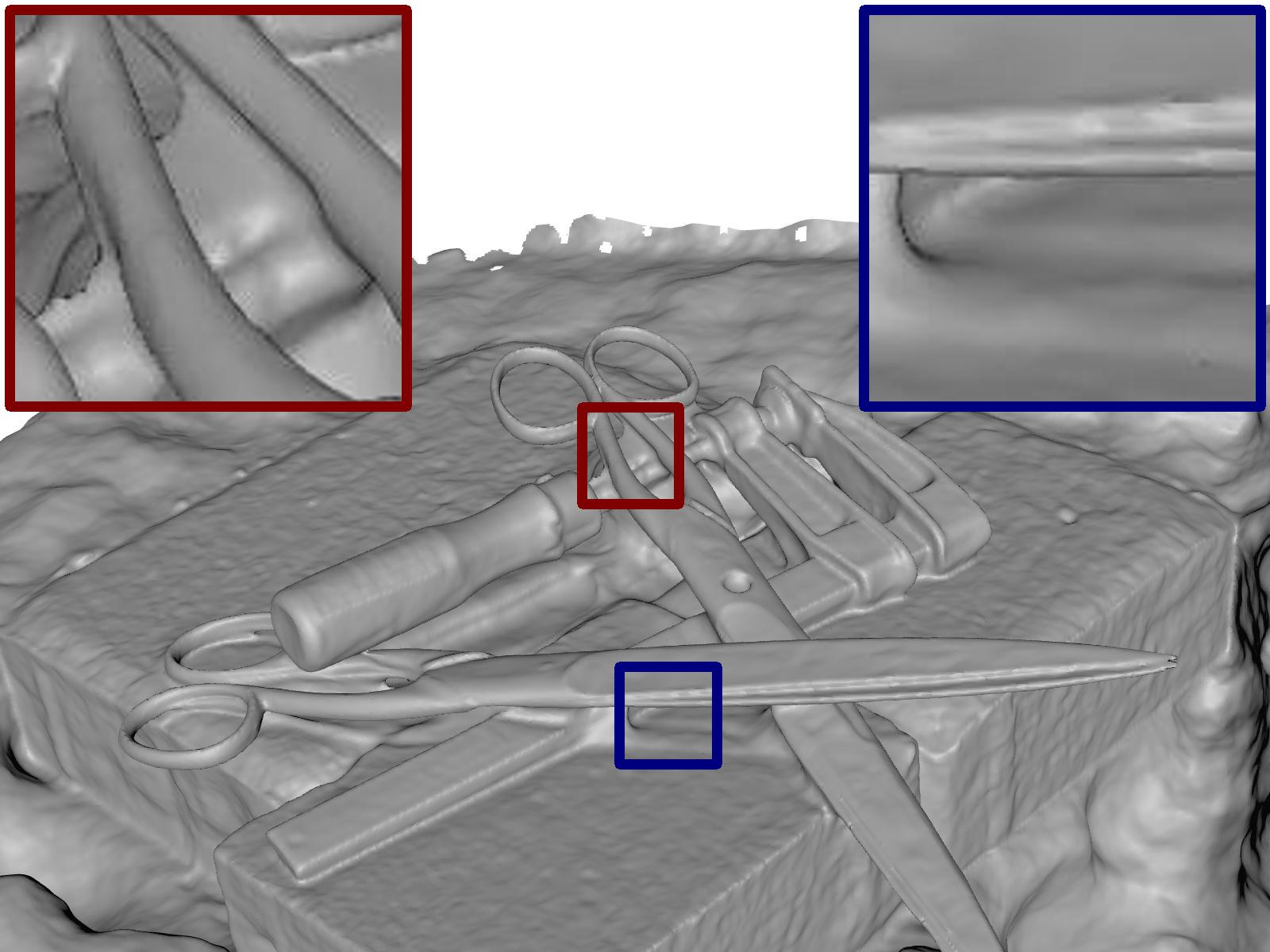}
        \\ 
        \vspace{\mrgv}
        \includegraphics[width=\wid]{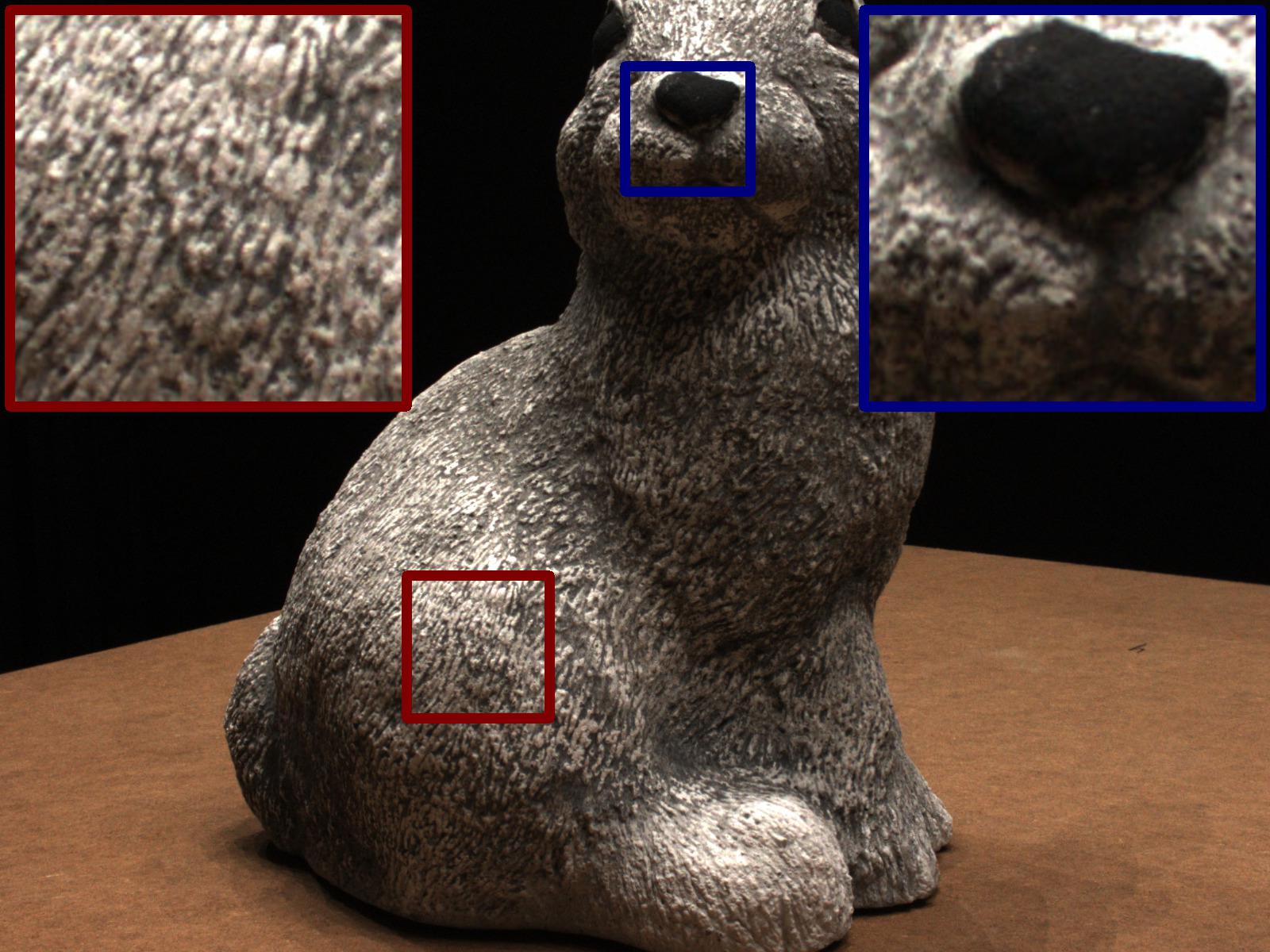} &
        \hspace{\mrg}
        \includegraphics[width=\wid]{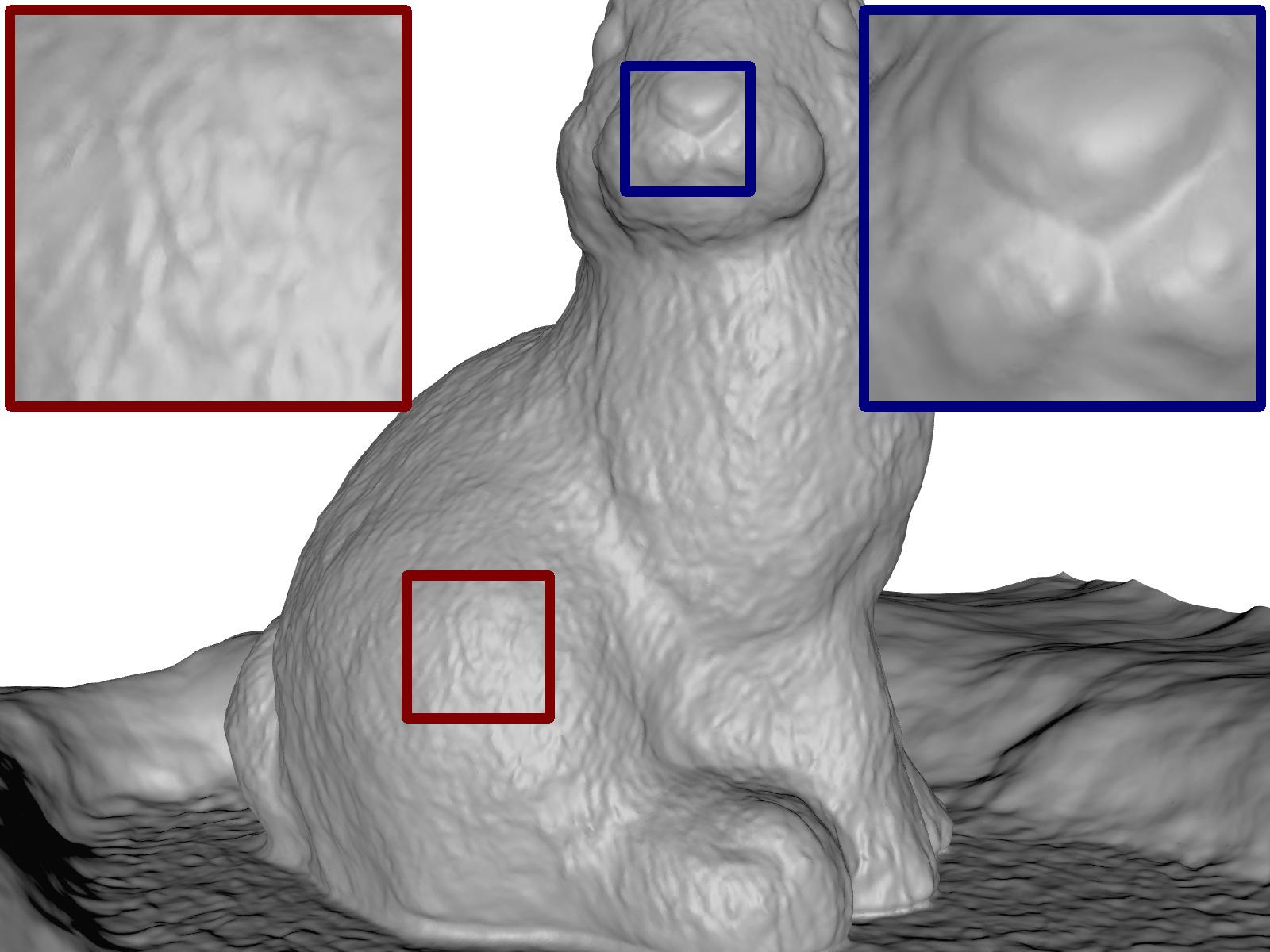} &
        \hspace{\mrg}
        \includegraphics[width=\wid]{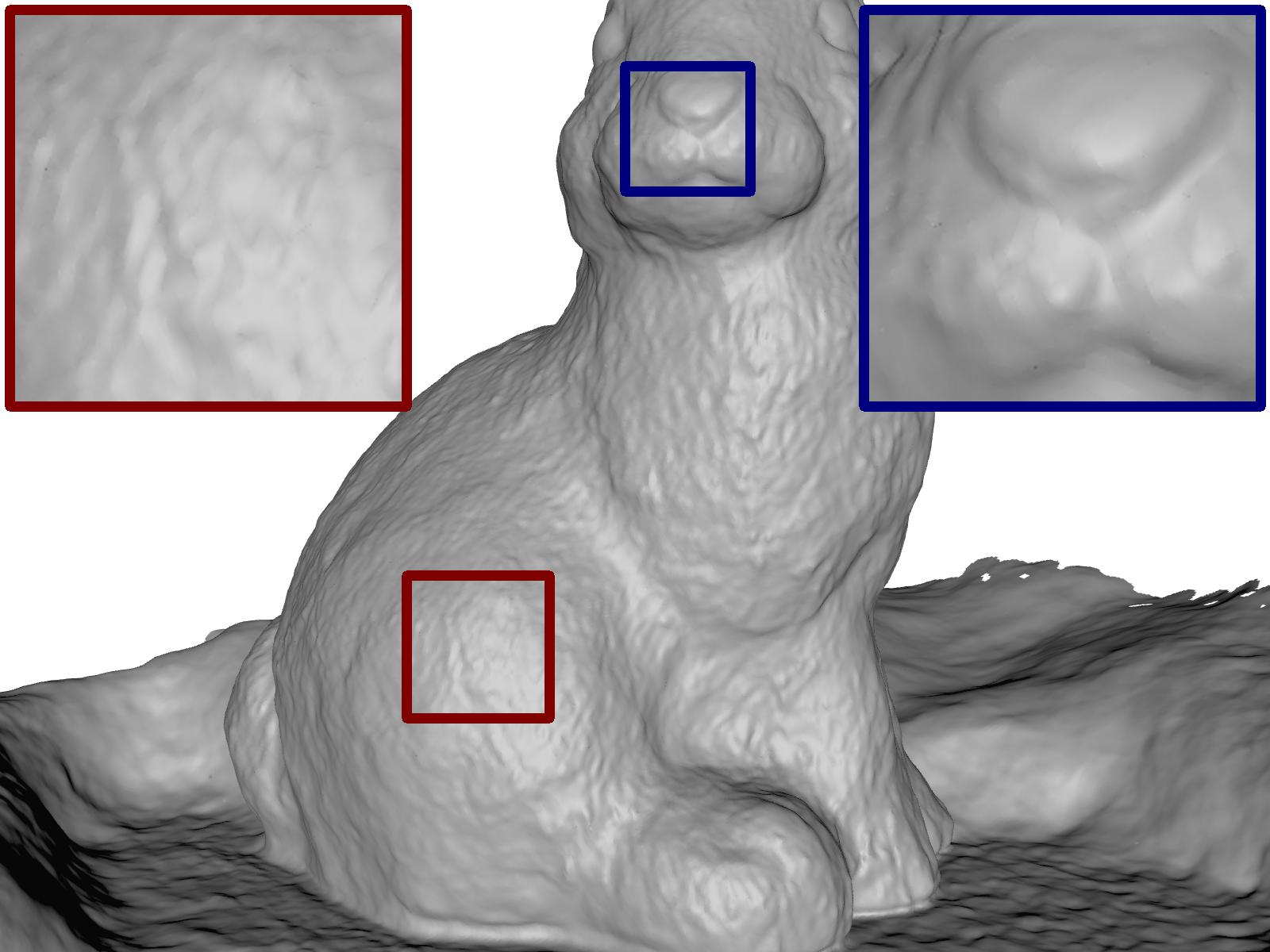}
        \\ 
        \vspace{\mrgv}
        \includegraphics[width=\wid]{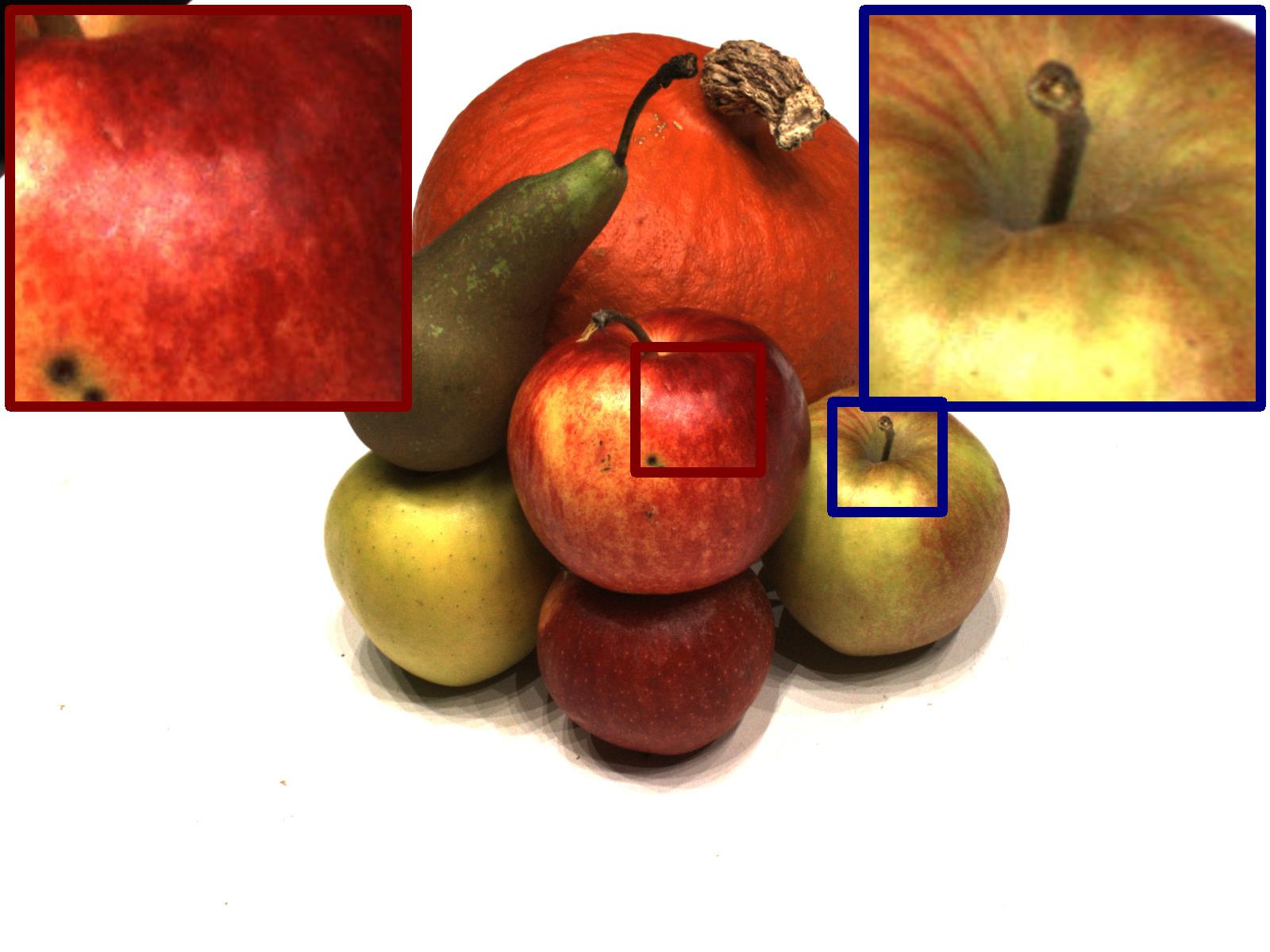} &
        \hspace{\mrg}
        \includegraphics[width=\wid]{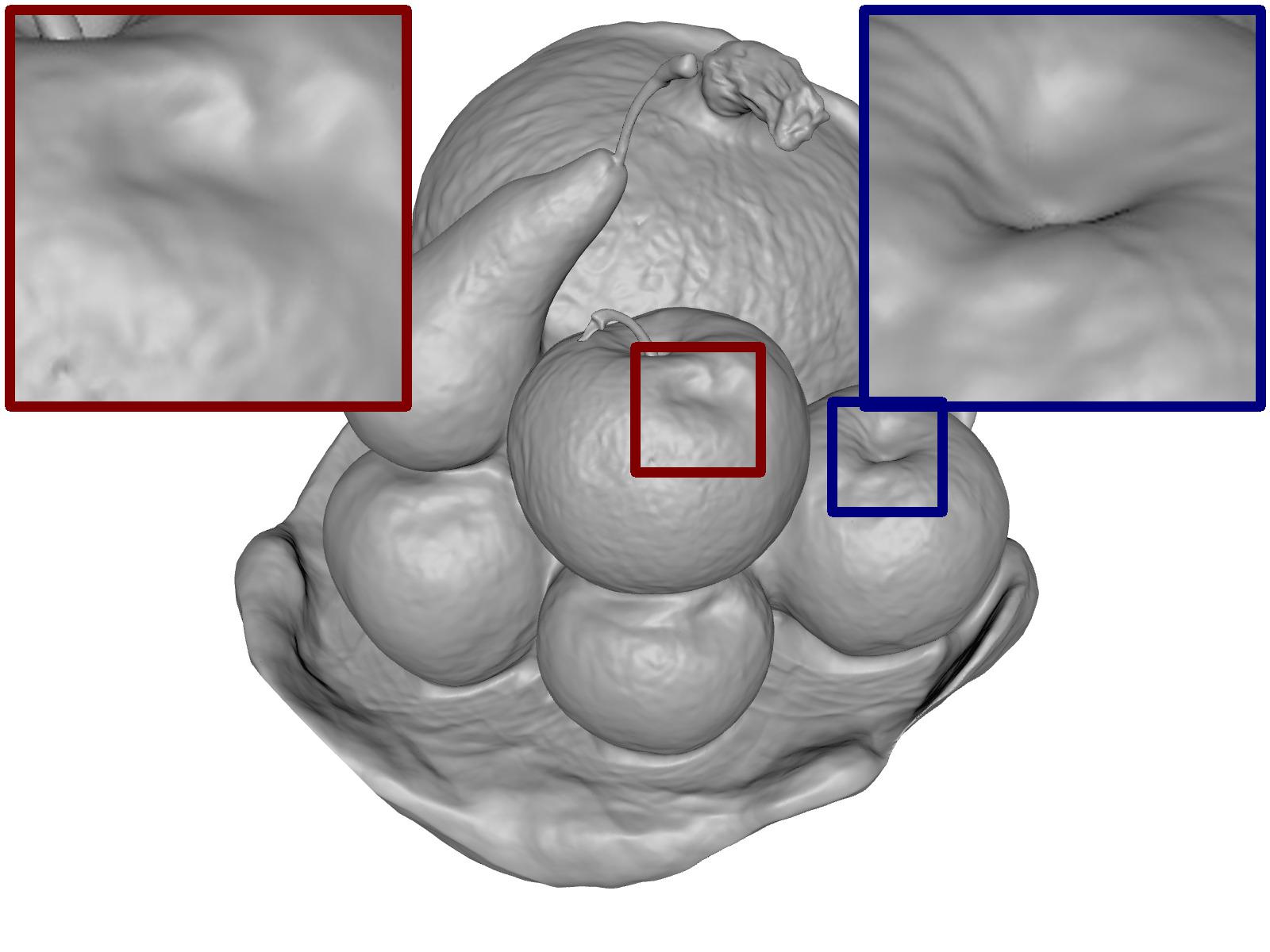} &
        \hspace{\mrg}
        \includegraphics[width=\wid]{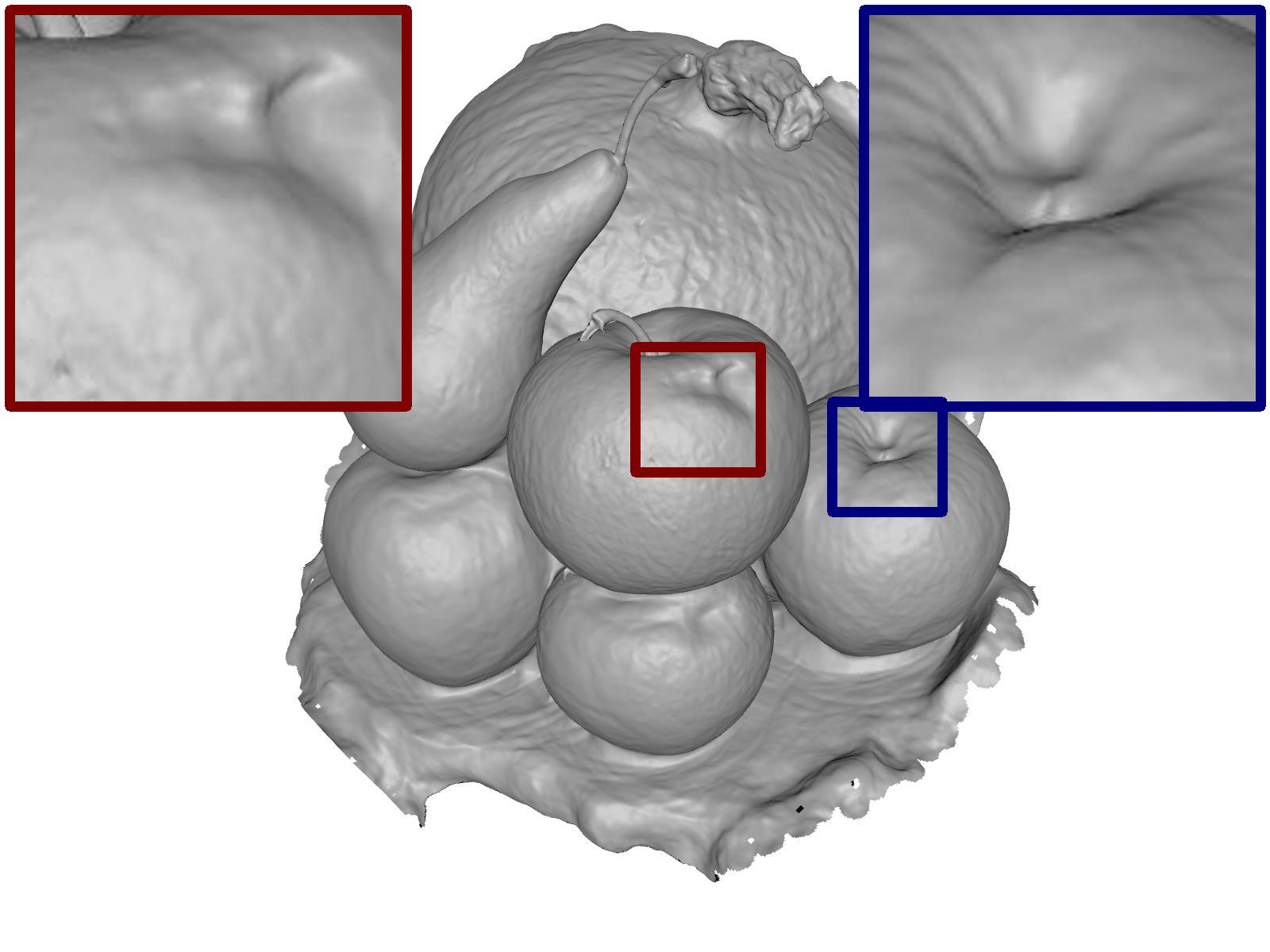}
        \\ 
        \vspace{\mrgv}
        \includegraphics[width=\wid]{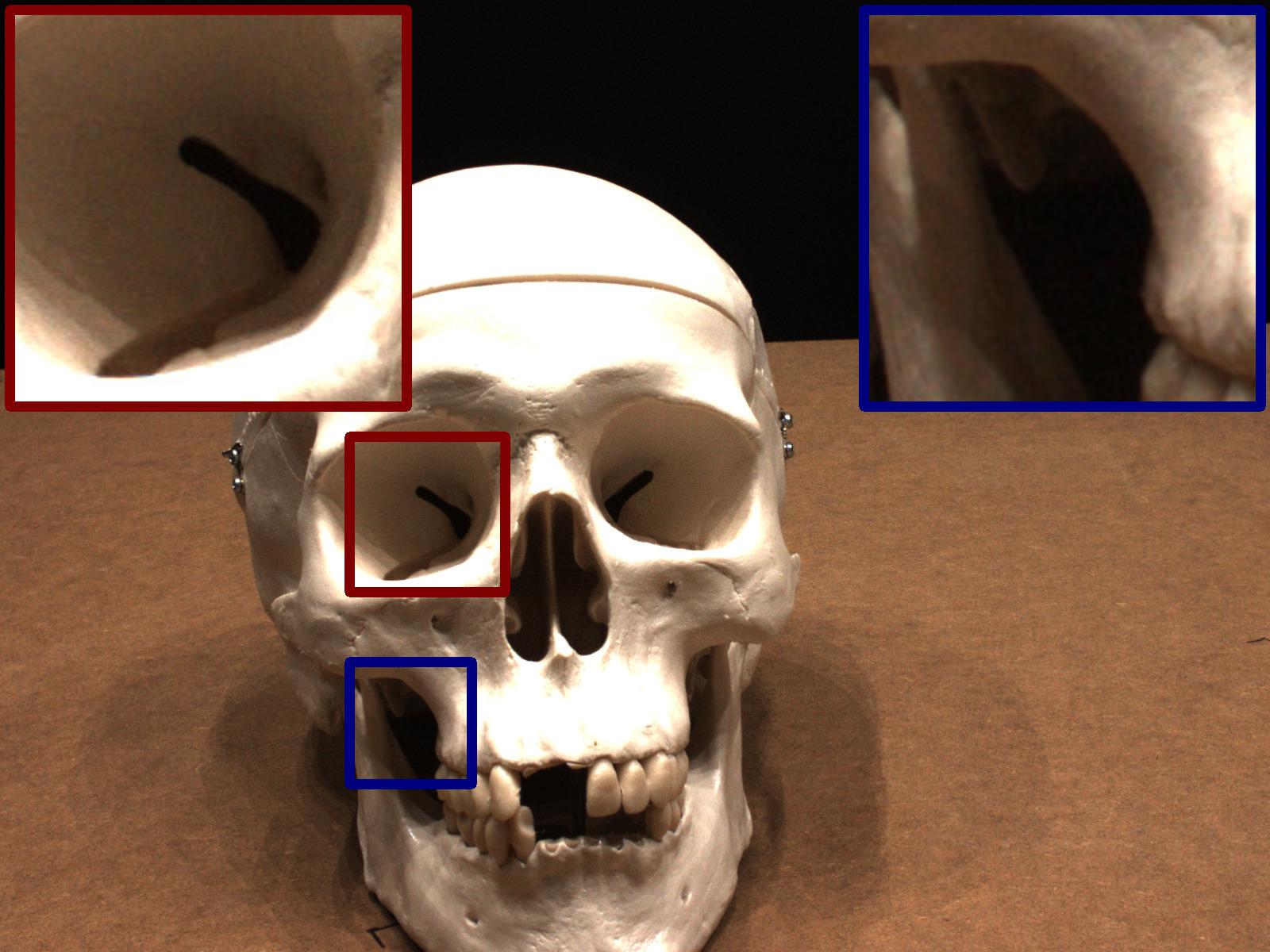} &
        \hspace{\mrg}
        \includegraphics[width=\wid]{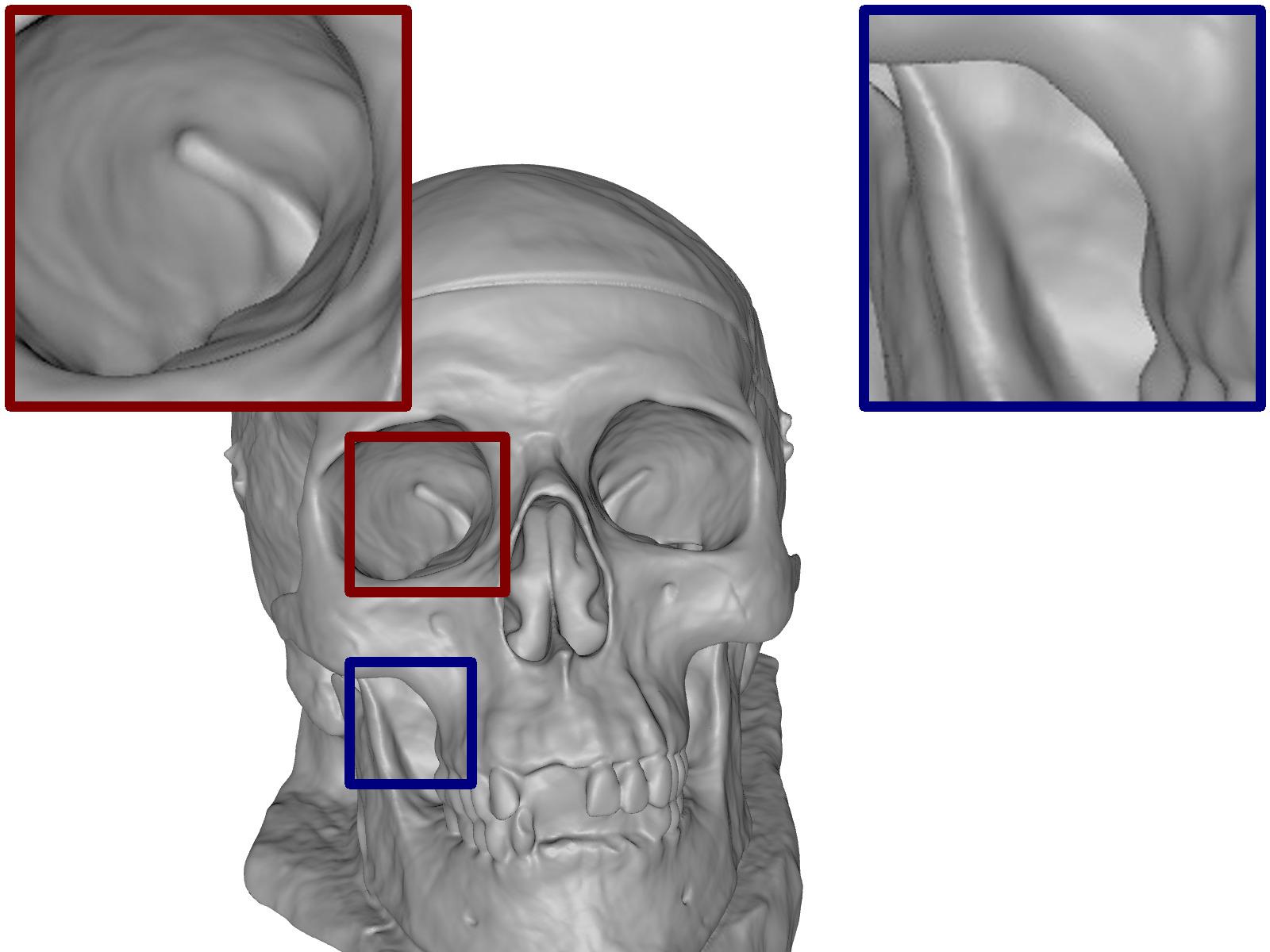} &
        \hspace{\mrg}
        \includegraphics[width=\wid]{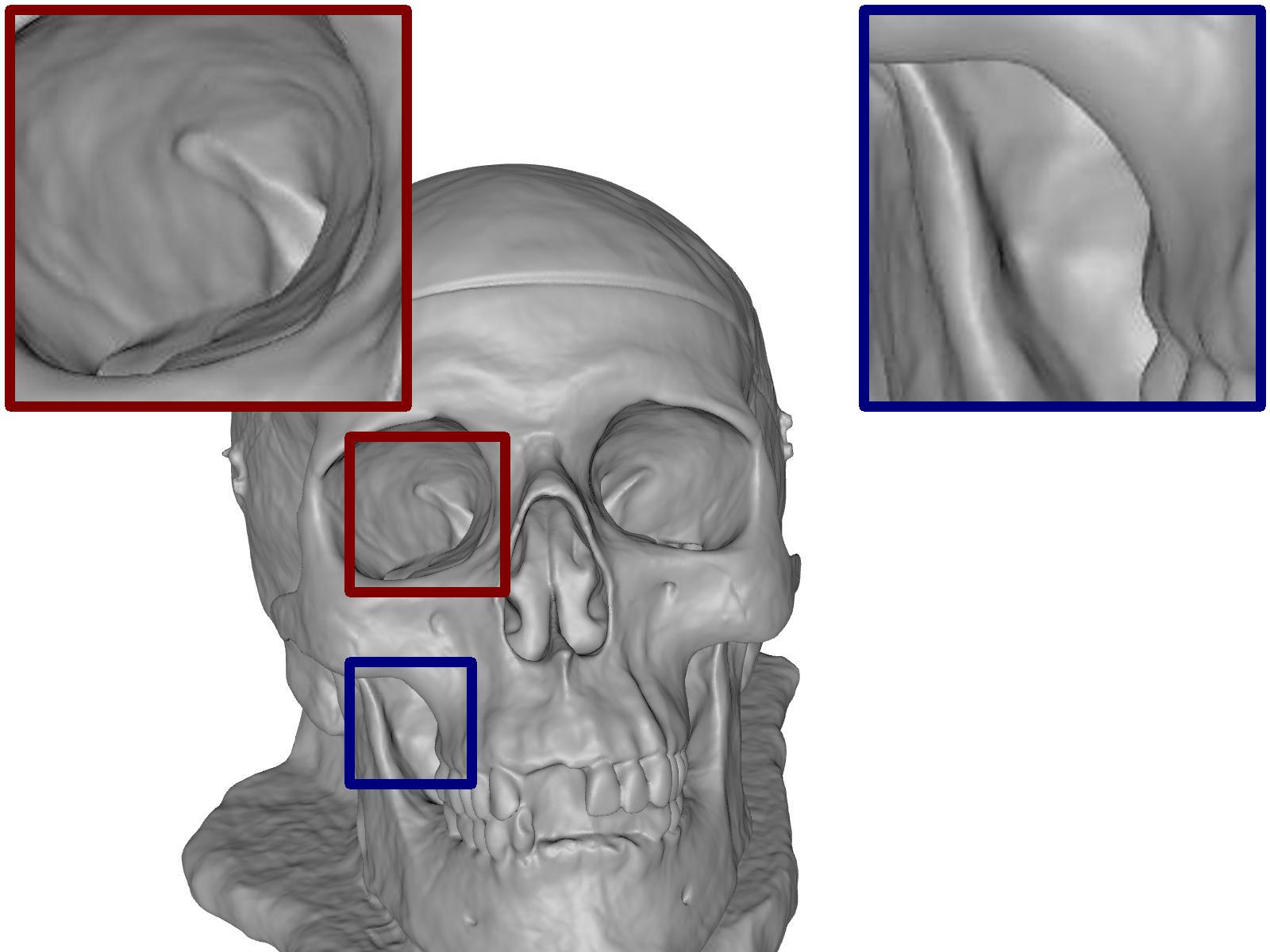}
        \\ 
        \vspace{\mrgv}
        \includegraphics[width=\wid]{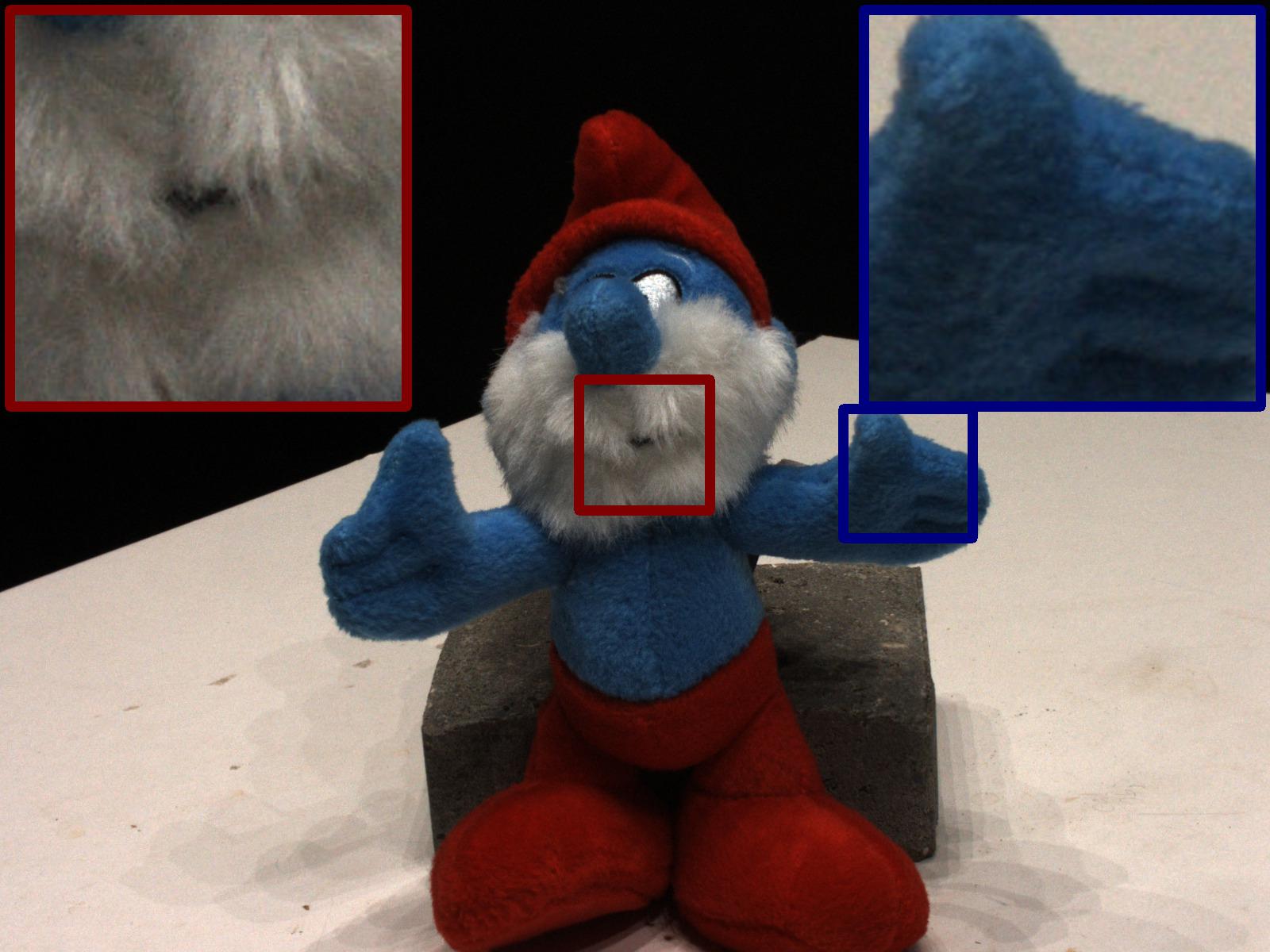} &
        \hspace{\mrg}
        \includegraphics[width=\wid]{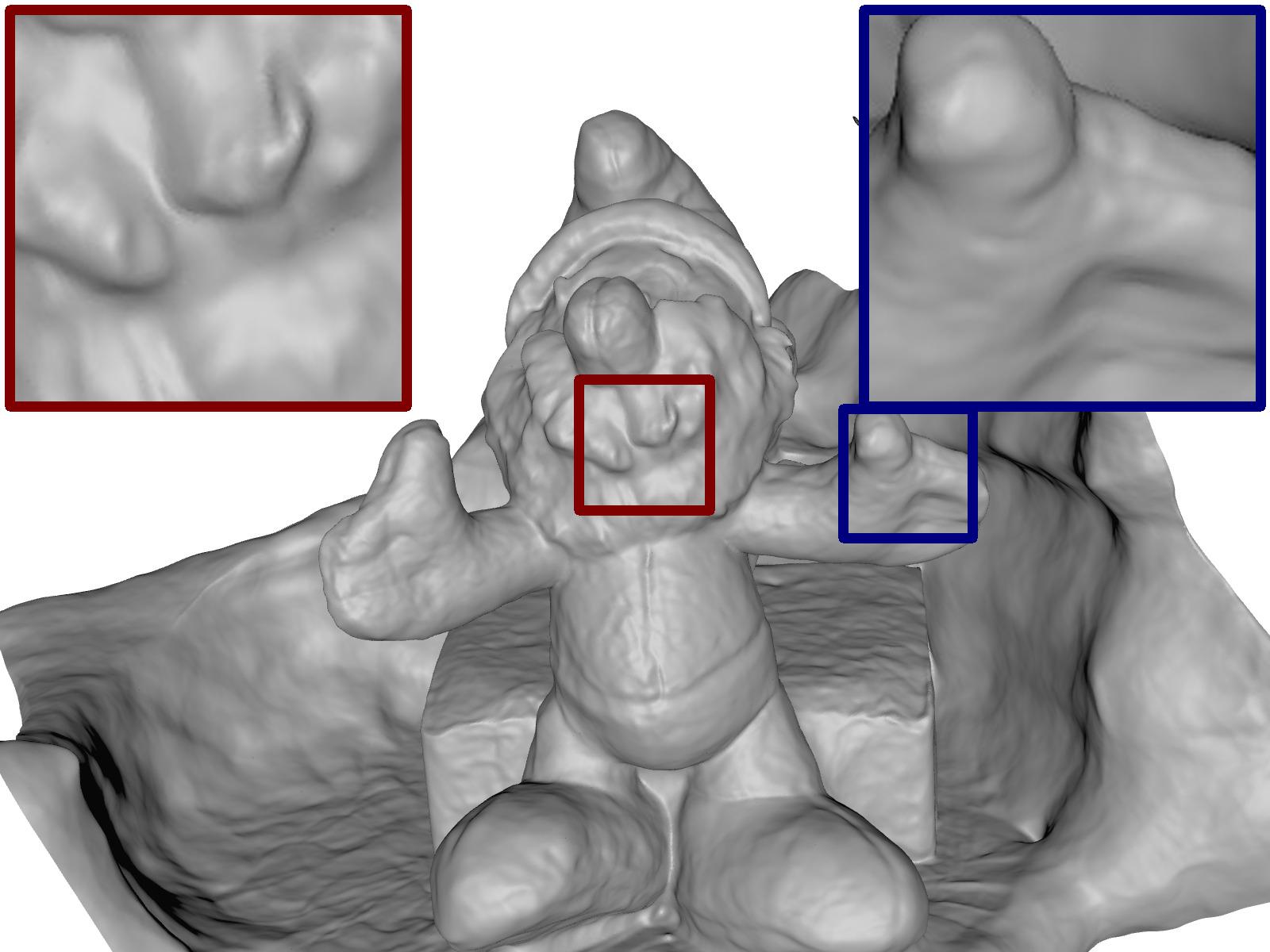} &
        \hspace{\mrg}
        \includegraphics[width=\wid]{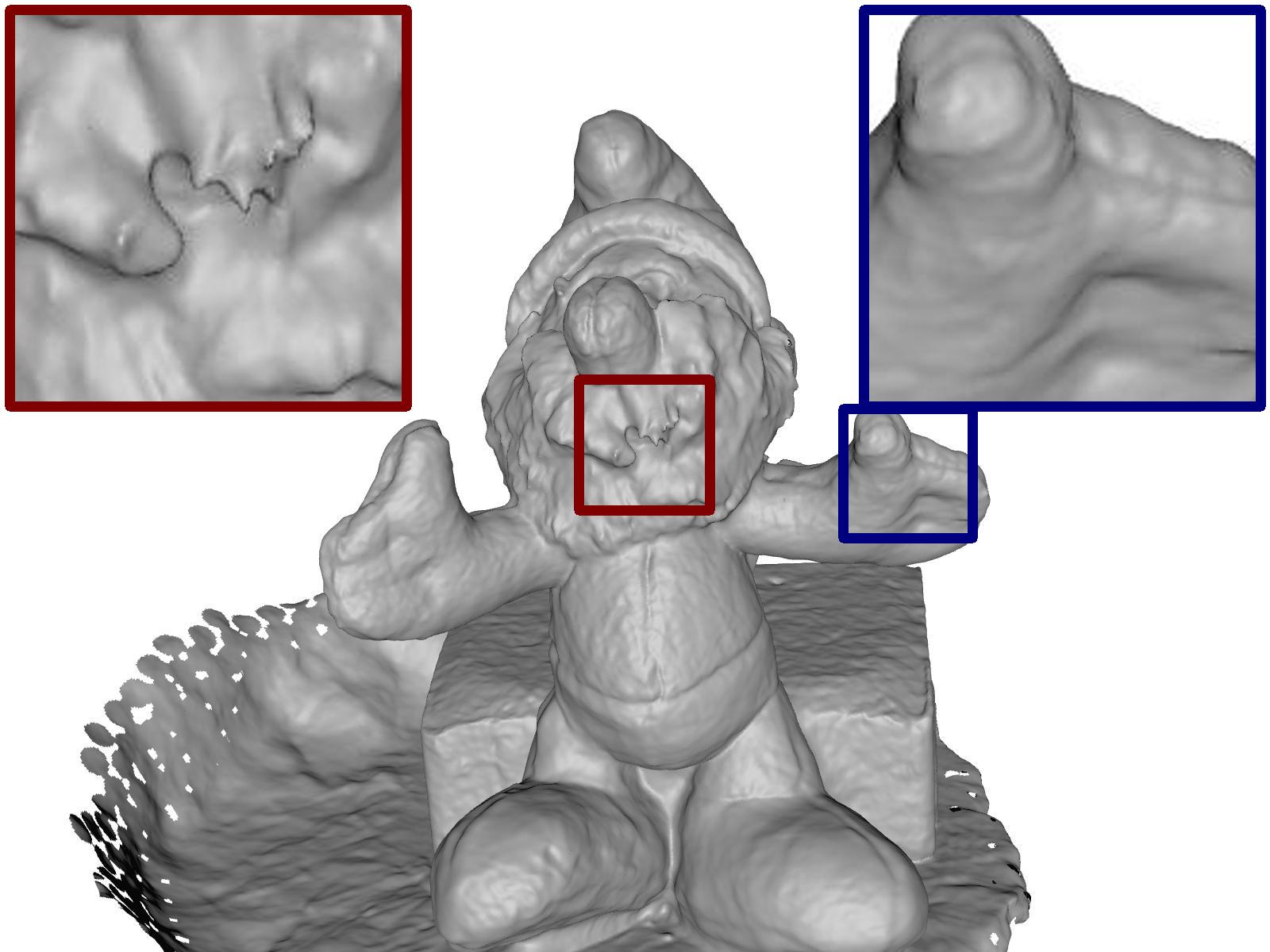}
        \\
        \vspace{\mrgv}
        \includegraphics[width=\wid]{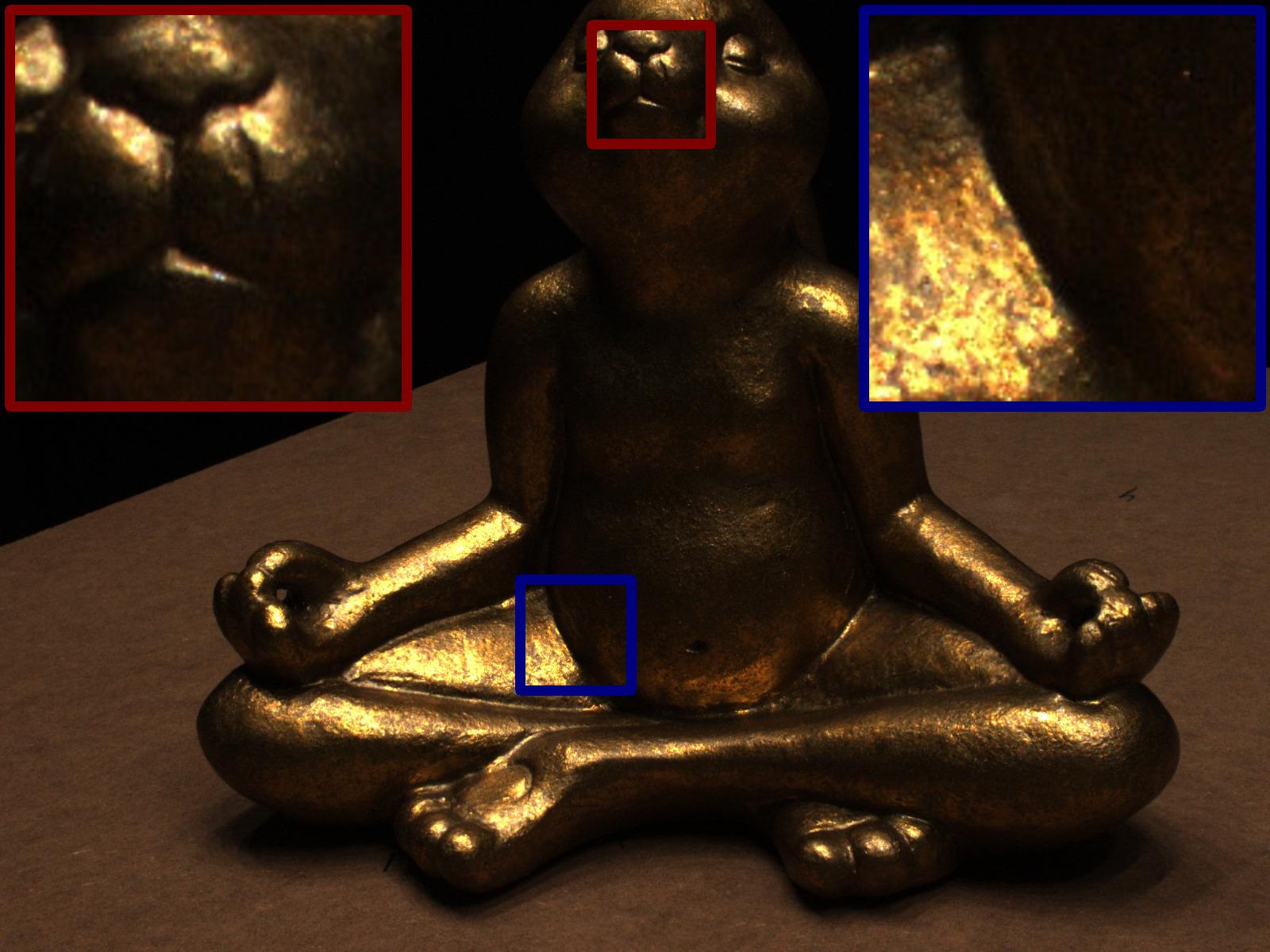} &
        \hspace{\mrg}
        \includegraphics[width=\wid]{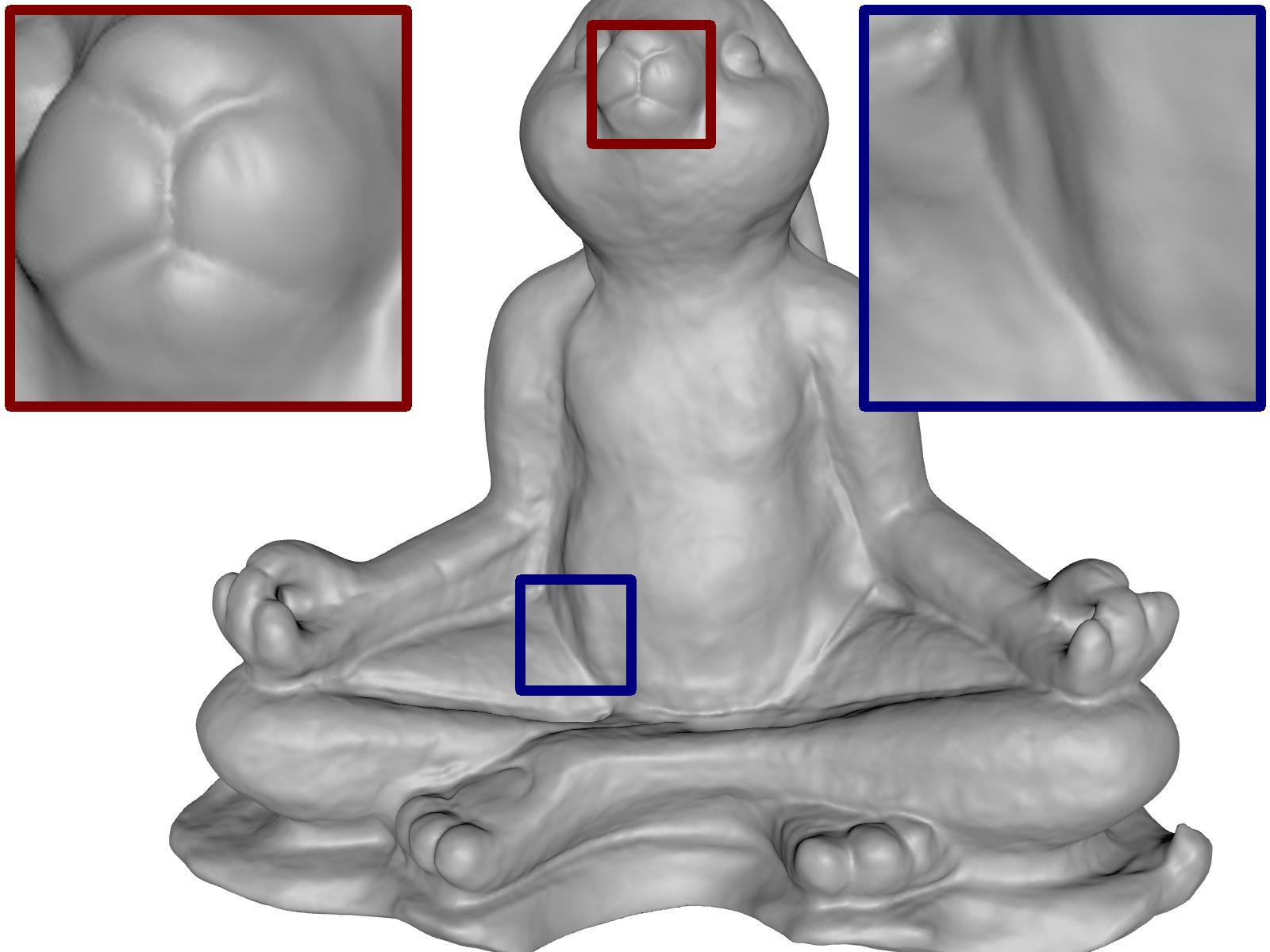} &
        \hspace{\mrg}
        \includegraphics[width=\wid]{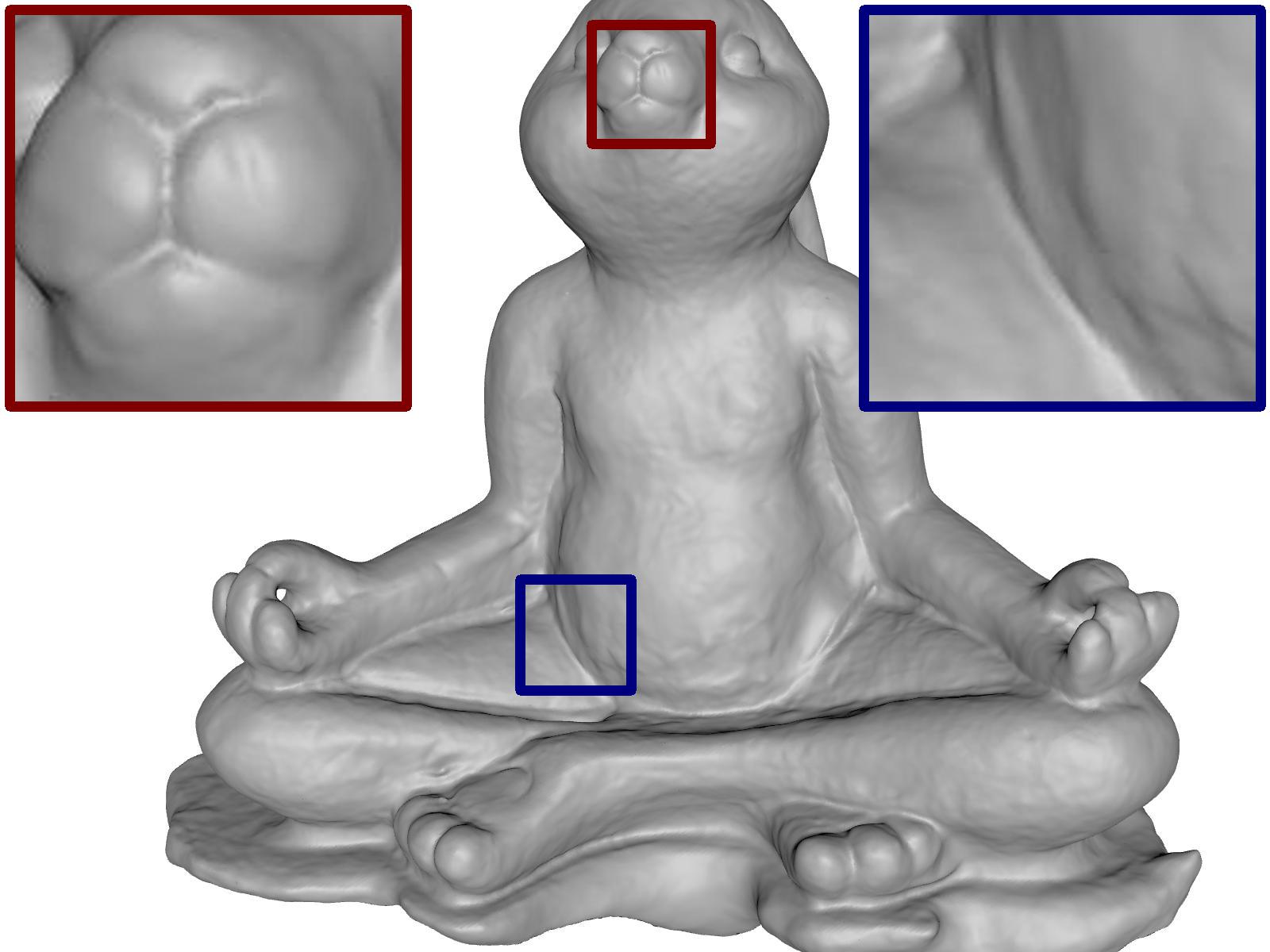}
        \\
        \vspace{\mrgv}
        \includegraphics[width=\wid]{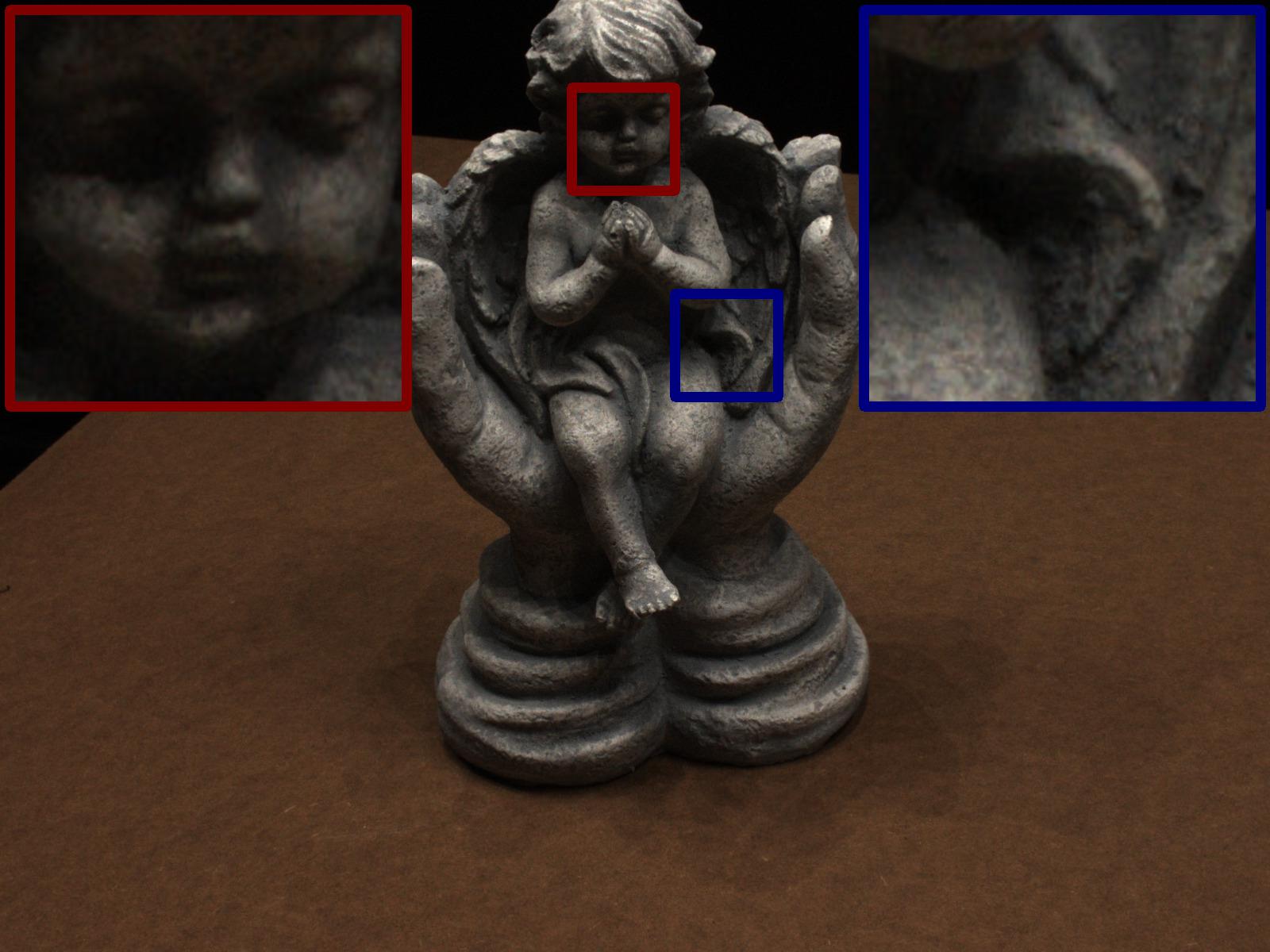} &
        \hspace{\mrg}
        \includegraphics[width=\wid]{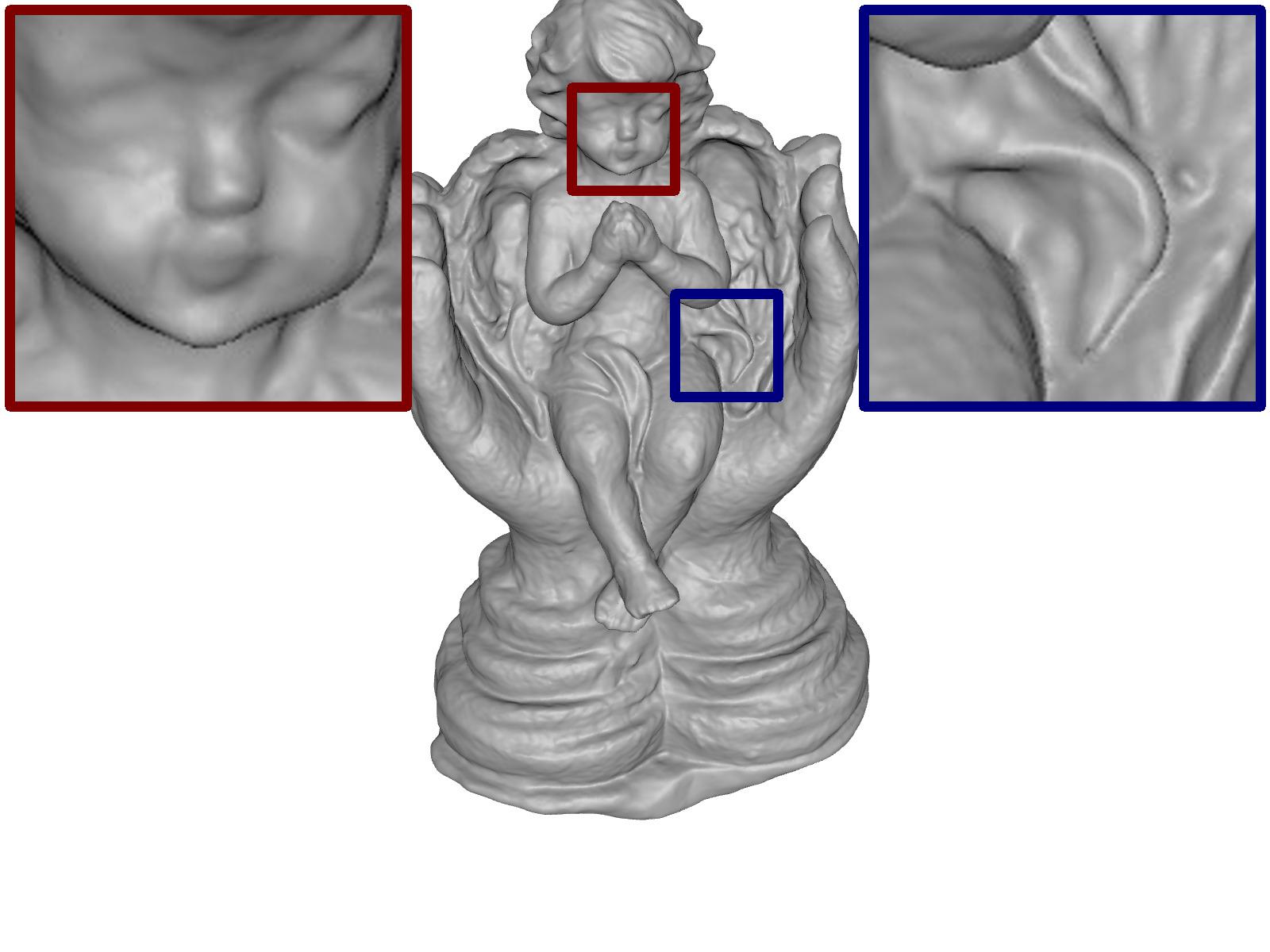} &
        \hspace{\mrg}
        \includegraphics[width=\wid]{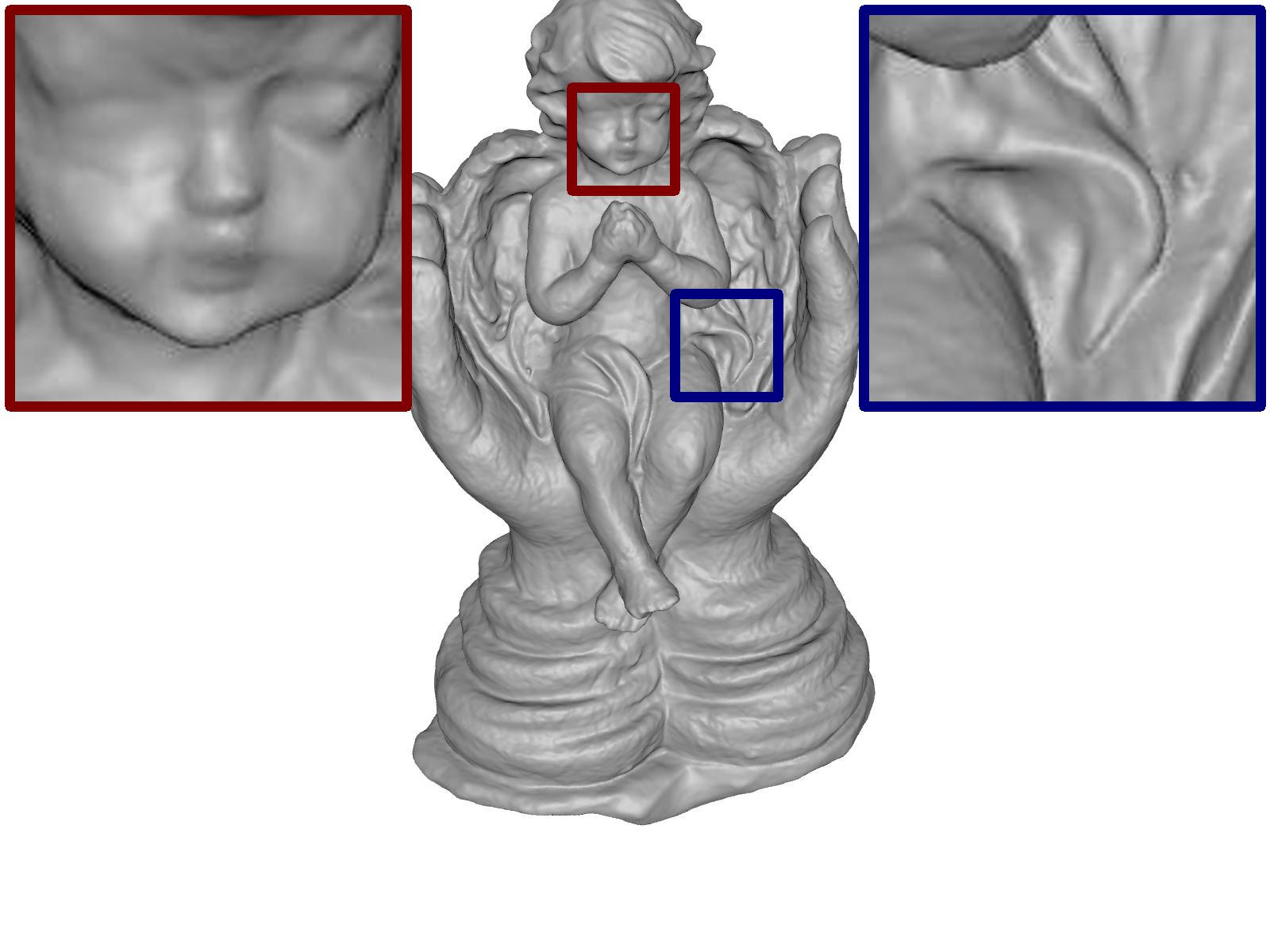}
        \\
        \vspace{\mrgv}
        \includegraphics[width=\wid]{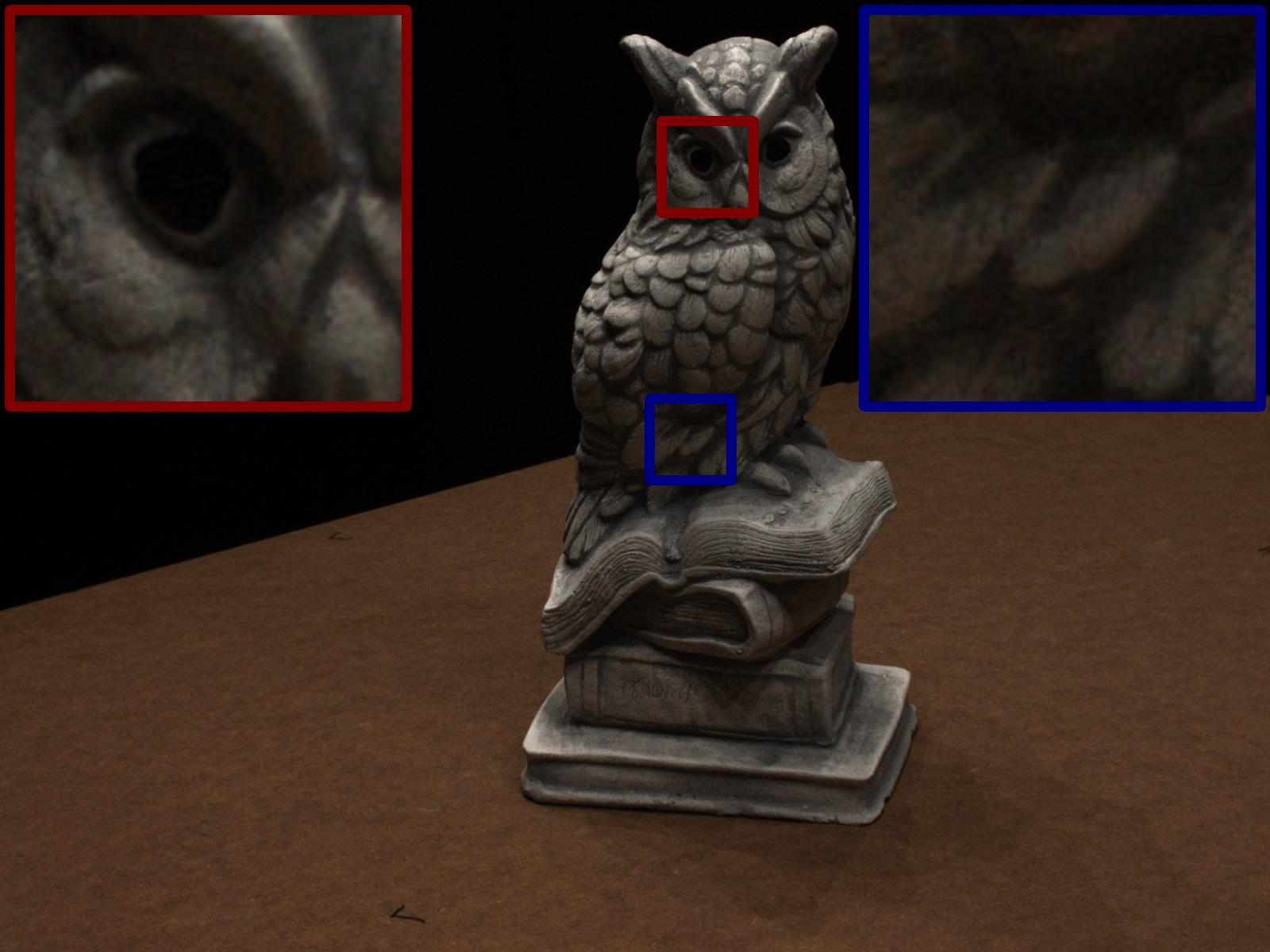} &
        \hspace{\mrg}
        \includegraphics[width=\wid]{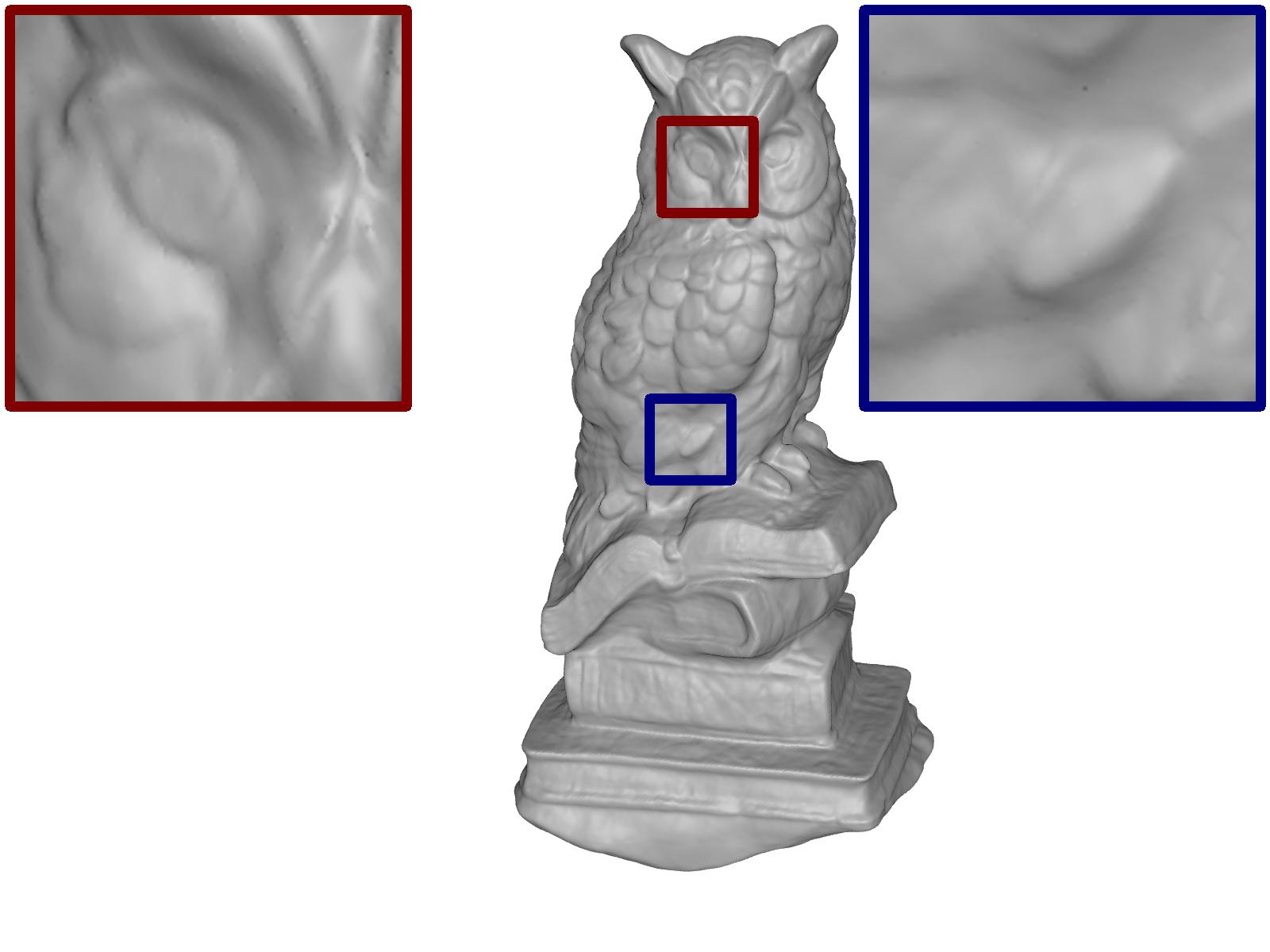} &
        \hspace{\mrg}
        \includegraphics[width=\wid]{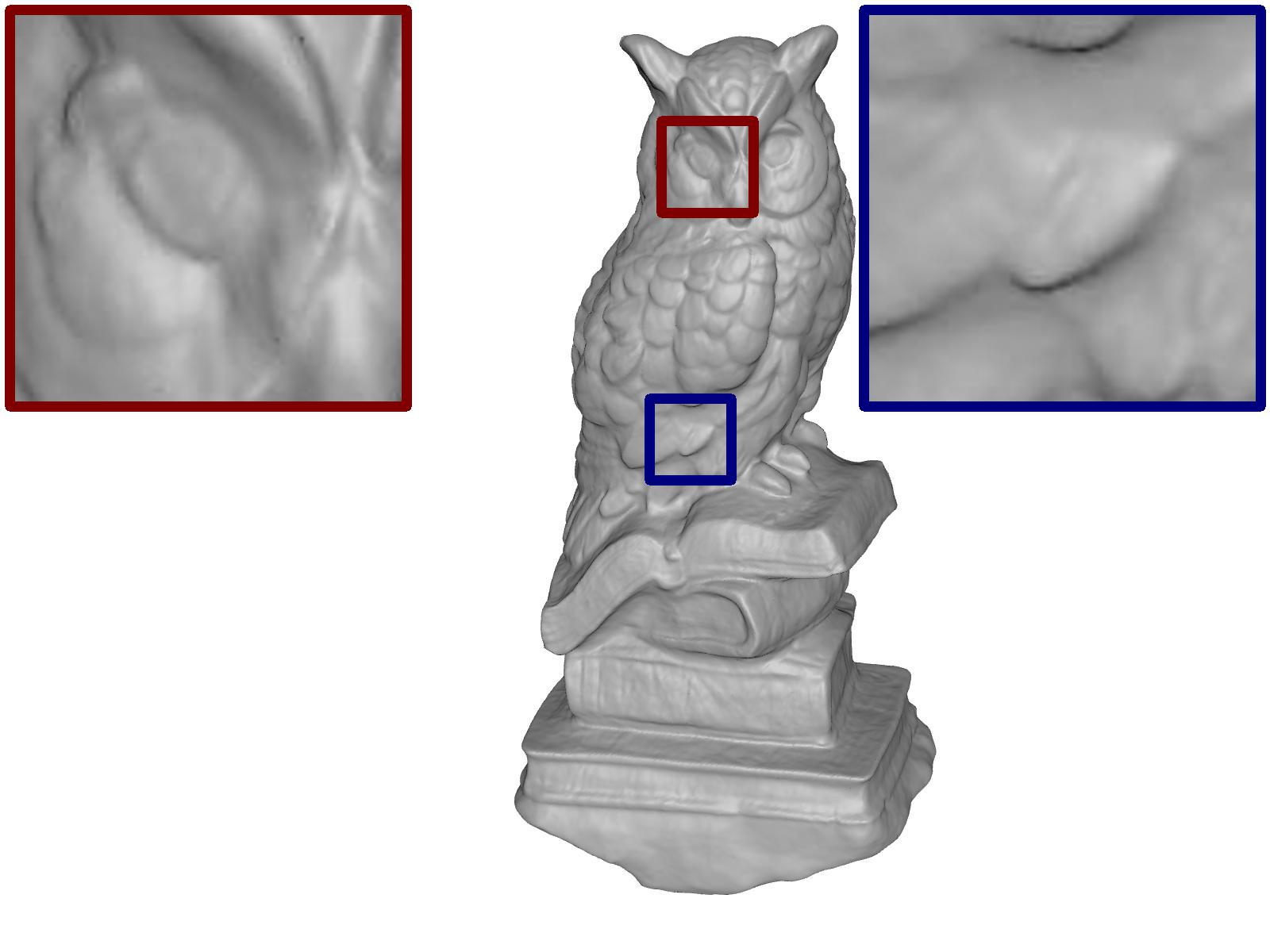}
        \\
        \textbf{Source} & \hspace{\mrg}
        \textbf{NeuS} & \hspace{\mrg}
        \textbf{NeuS (ours)}
    \end{tabular}
    \caption{Additional qualitative results on the DTU~\cite{Jensen2014LargeSM} dataset for NeuS~\cite{Wang2021NeuSLN} method.}
    \label{fig:dtu_qual_appendix_neus}
    \vspace{-0.4cm}
\end{figure*}

\begin{figure*}
    \centering    
    \setlength{\wid}{0.20\textwidth}
    \setlength{\mrg}{-0.45cm}
    \setlength{\mrgv}{-0.05cm}
    \begin{tabular}{c cc}
        \vspace{\mrgv}
        \includegraphics[width=\wid]{figures/dtu_qual/gt/37.jpg} &
        \hspace{\mrg}
        \includegraphics[width=\wid]{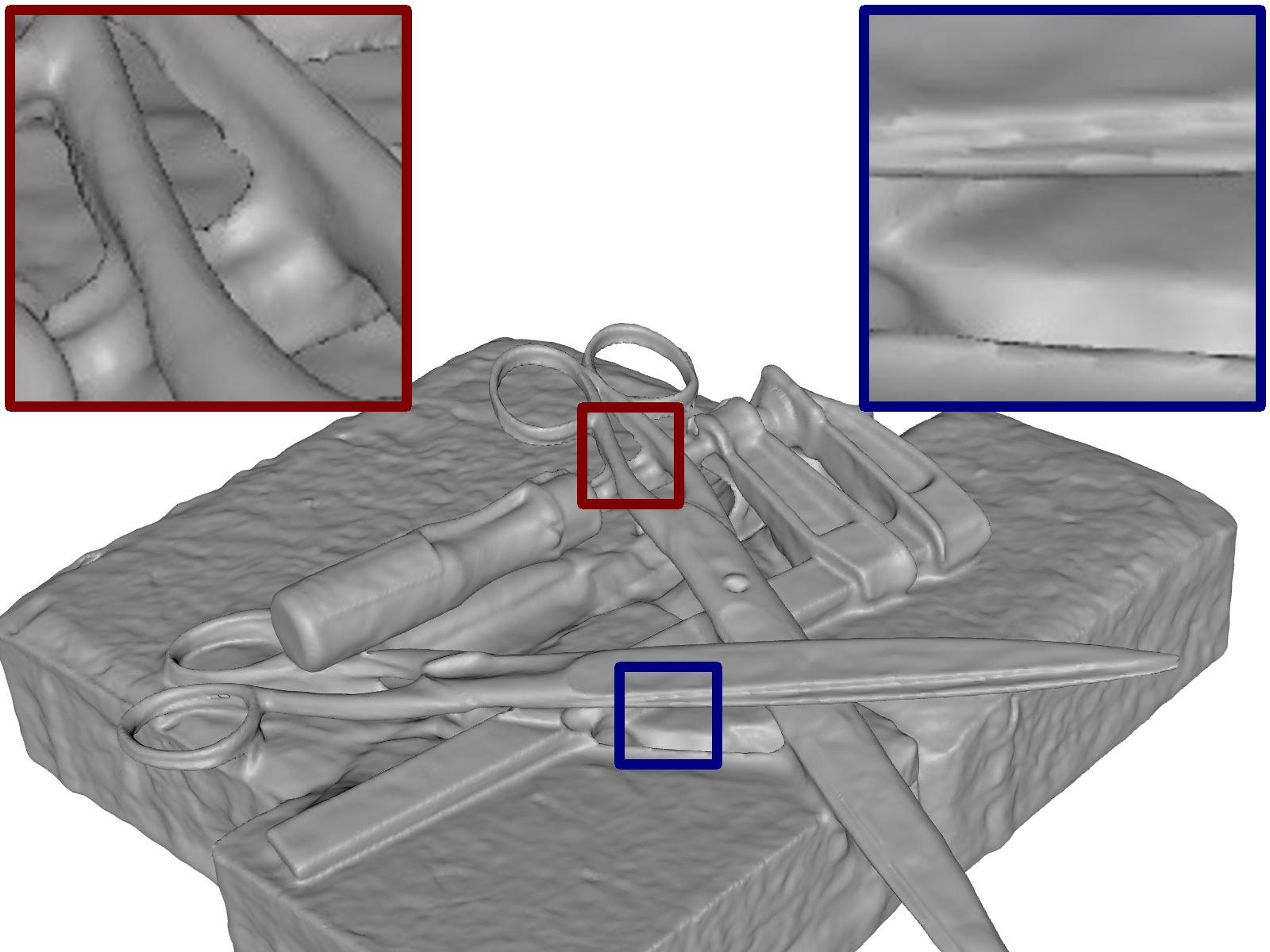} &
        \hspace{\mrg}
        \includegraphics[width=\wid]{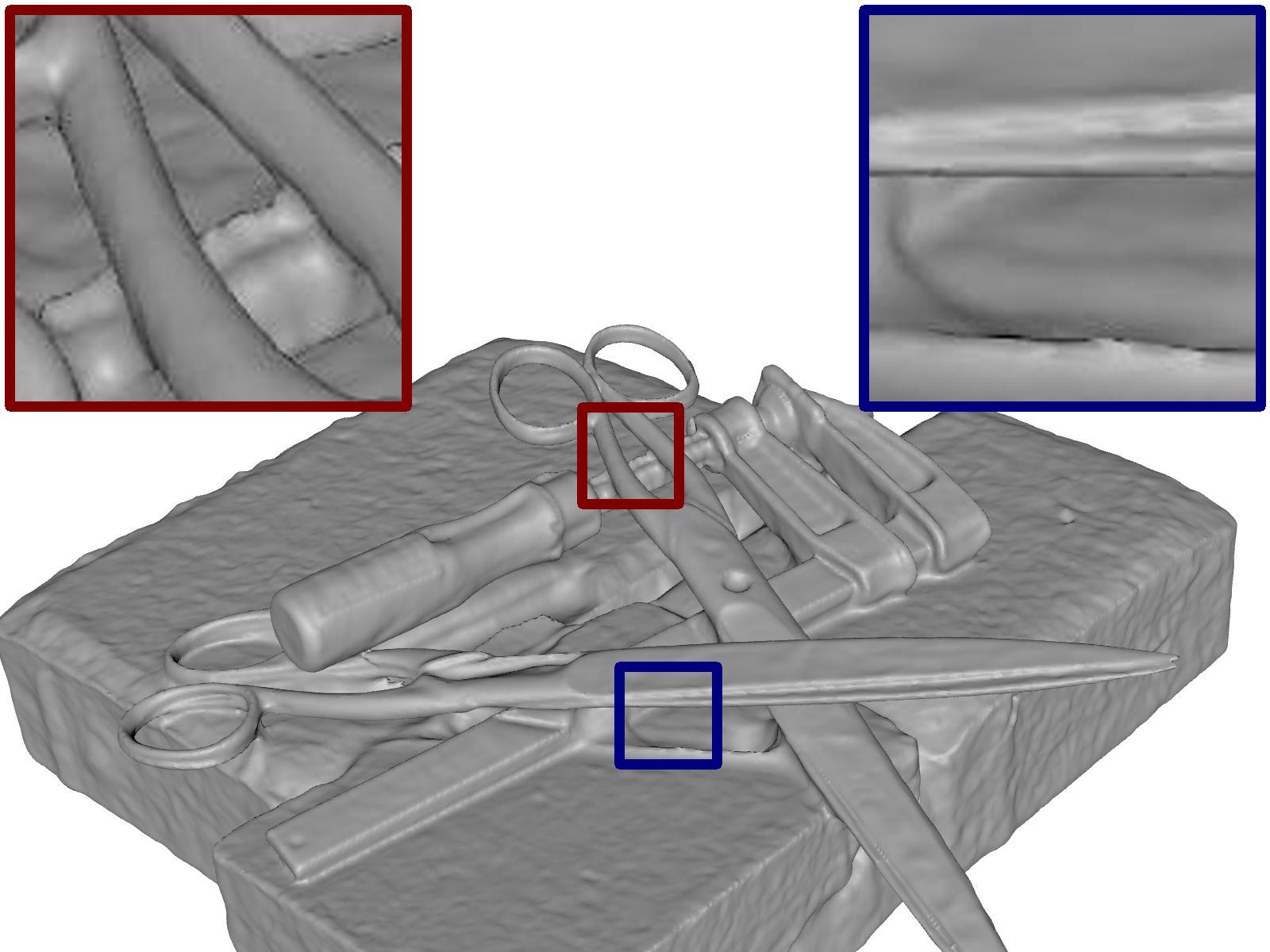}
        \\ 
        \vspace{\mrgv}
        \includegraphics[width=\wid]{figures/dtu_qual/gt/55.jpg} &
        \hspace{\mrg}
        \includegraphics[width=\wid]{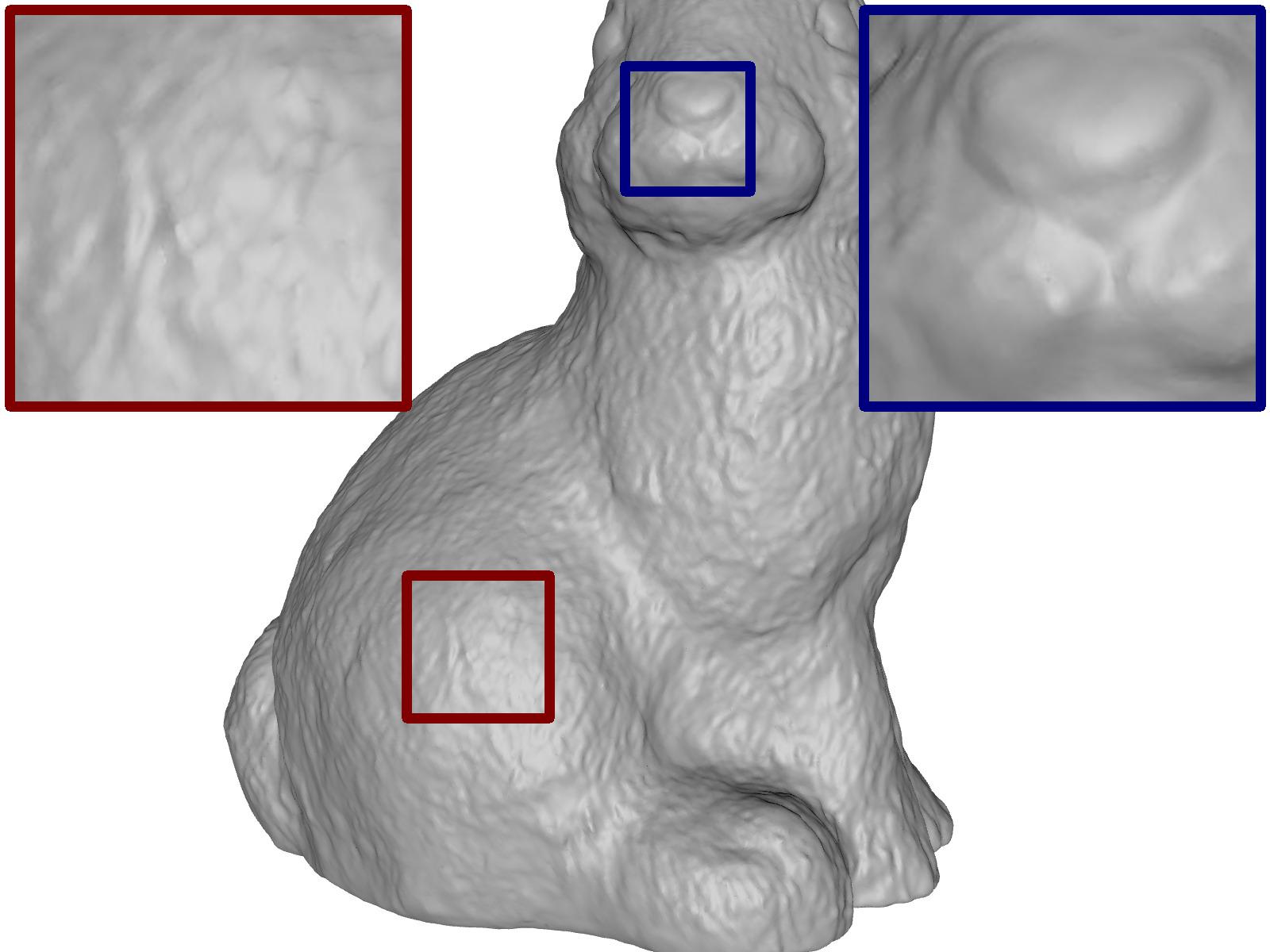} &
        \hspace{\mrg}
        \includegraphics[width=\wid]{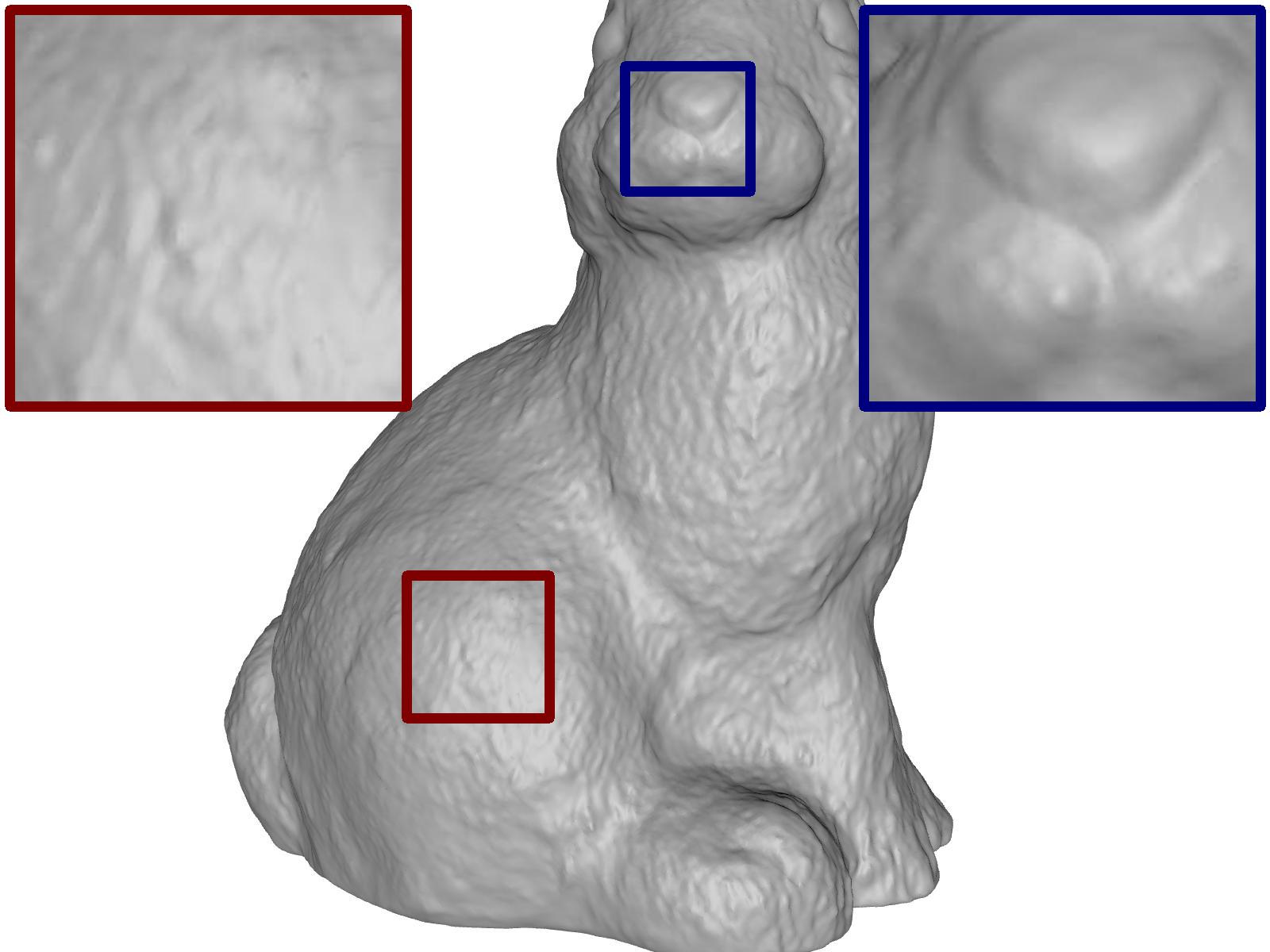}
        \\ 
        \vspace{\mrgv}
        \includegraphics[width=\wid]{figures/dtu_qual/gt/63.jpg} &
        \hspace{\mrg}
        \includegraphics[width=\wid]{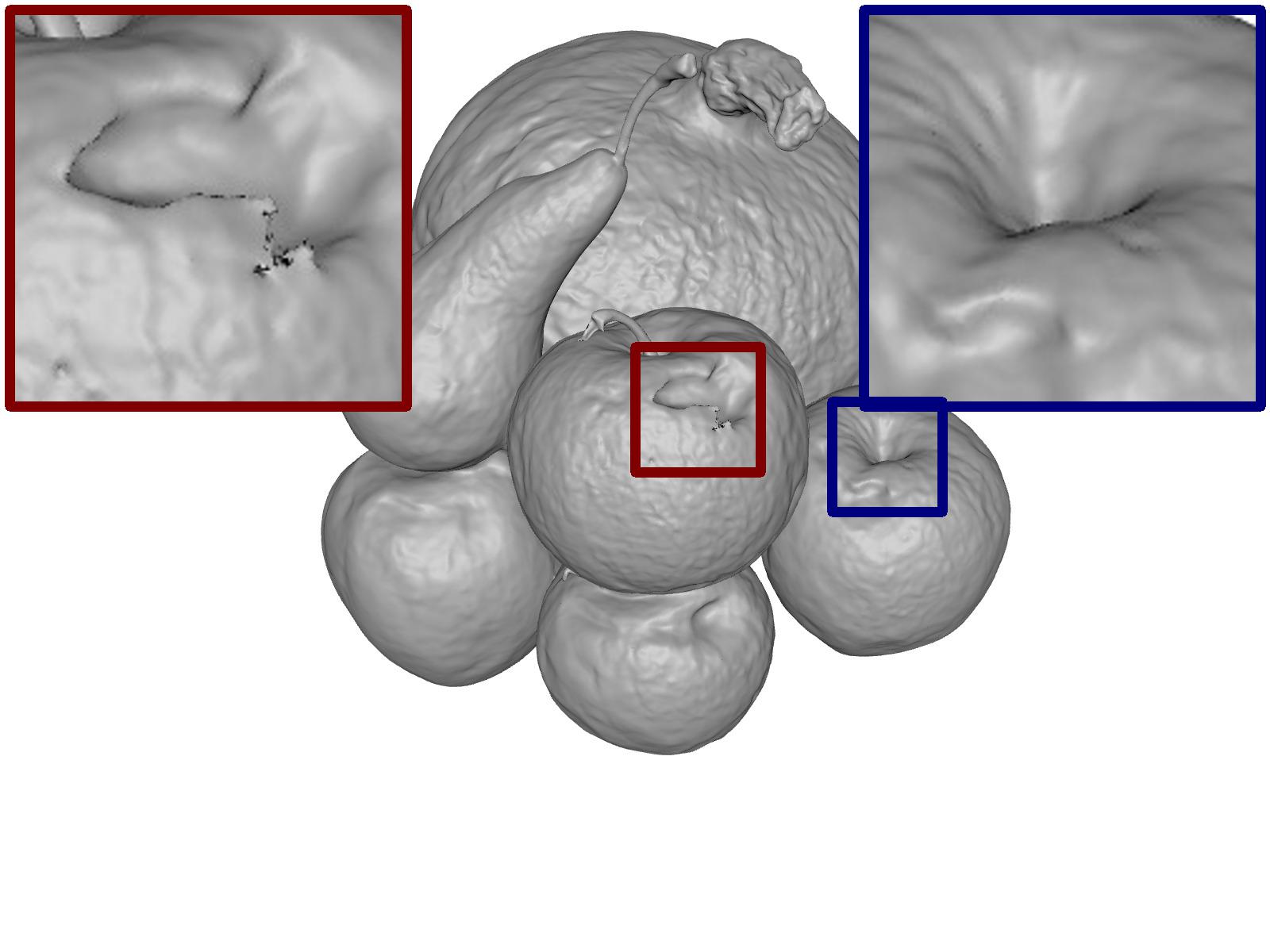} &
        \hspace{\mrg}
        \includegraphics[width=\wid]{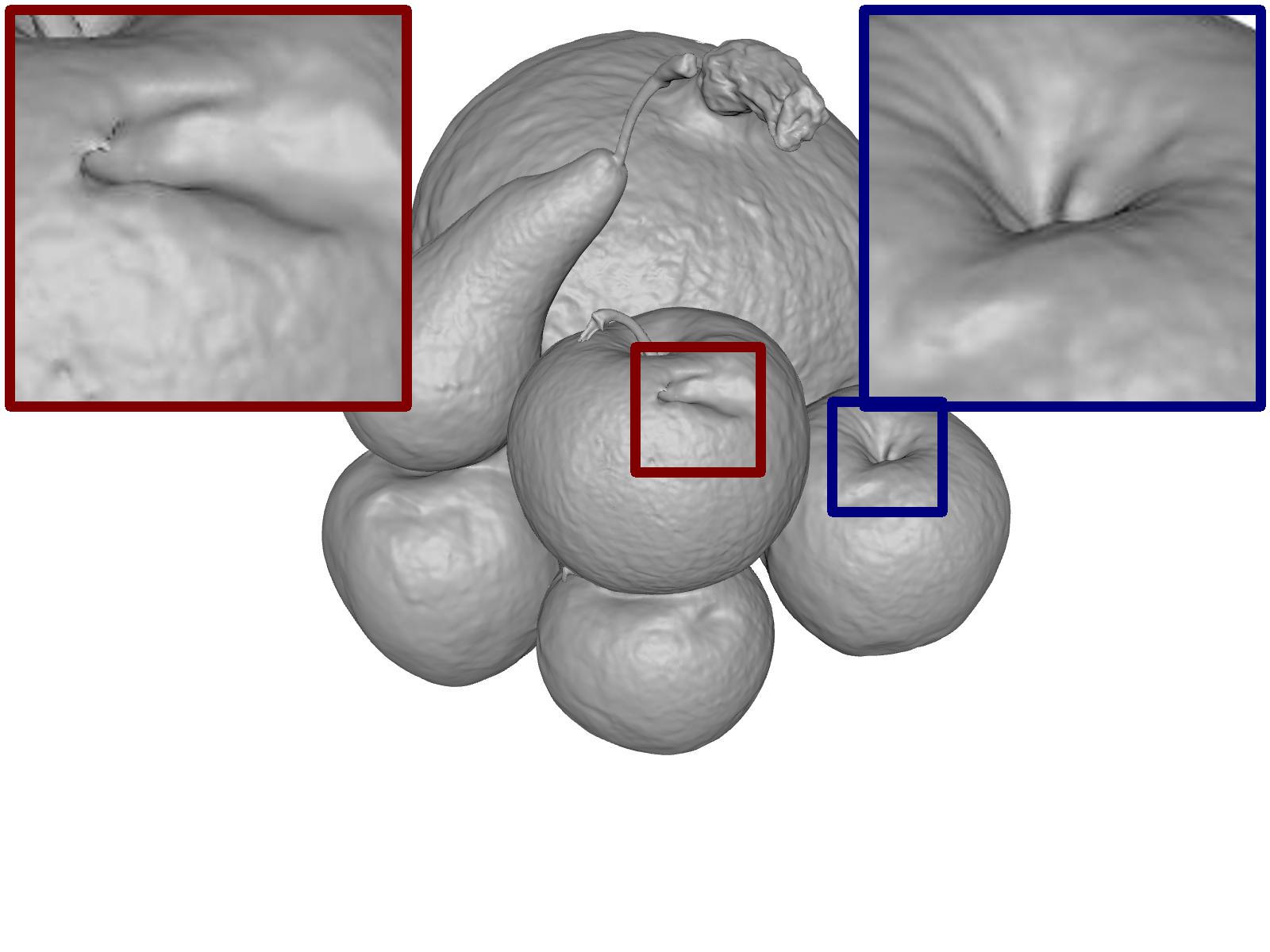}
        \\ 
        \vspace{\mrgv}
        \includegraphics[width=\wid]{figures/dtu_qual/gt/65.jpg} &
        \hspace{\mrg}
        \includegraphics[width=\wid]{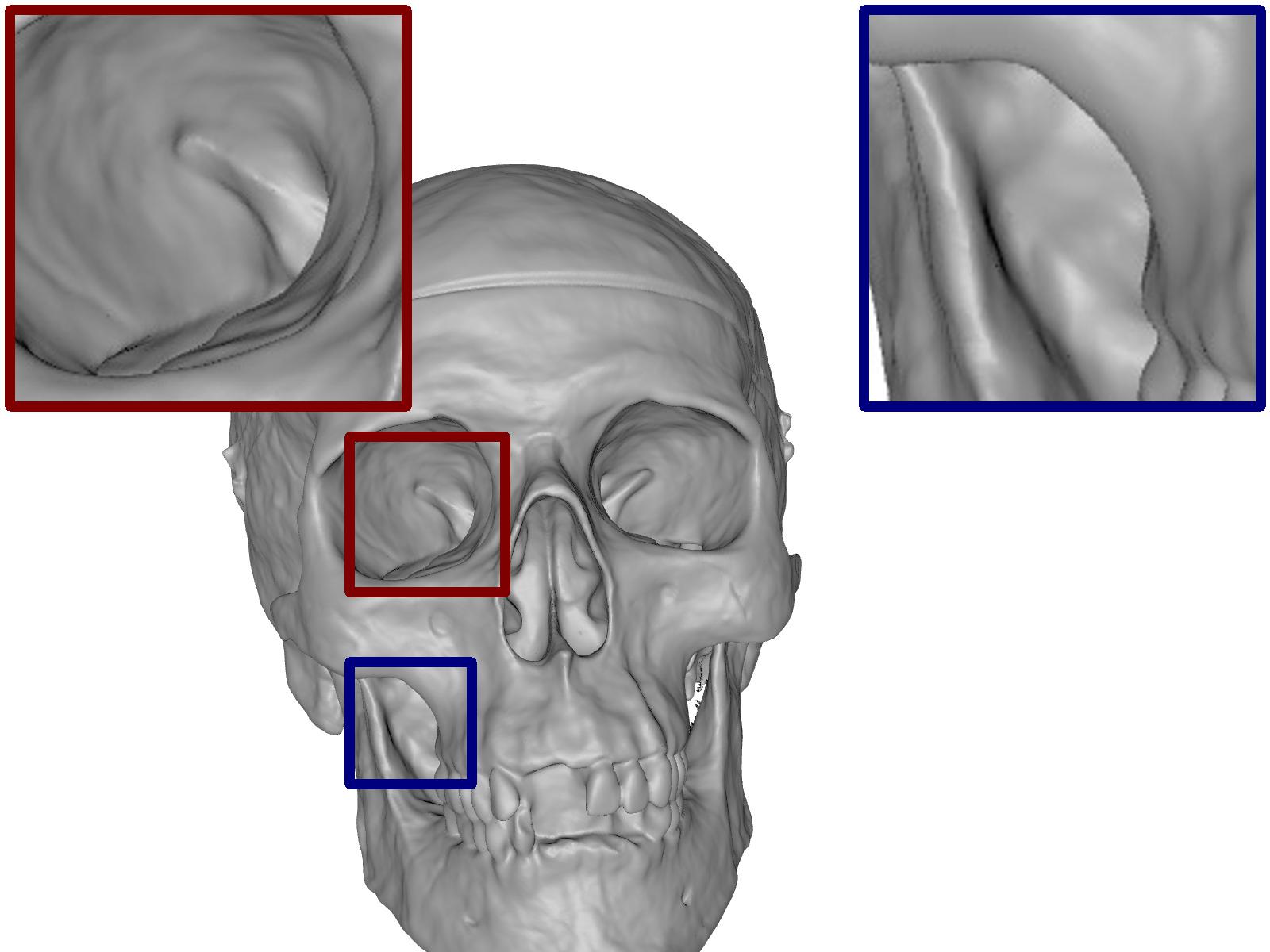} &
        \hspace{\mrg}
        \includegraphics[width=\wid]{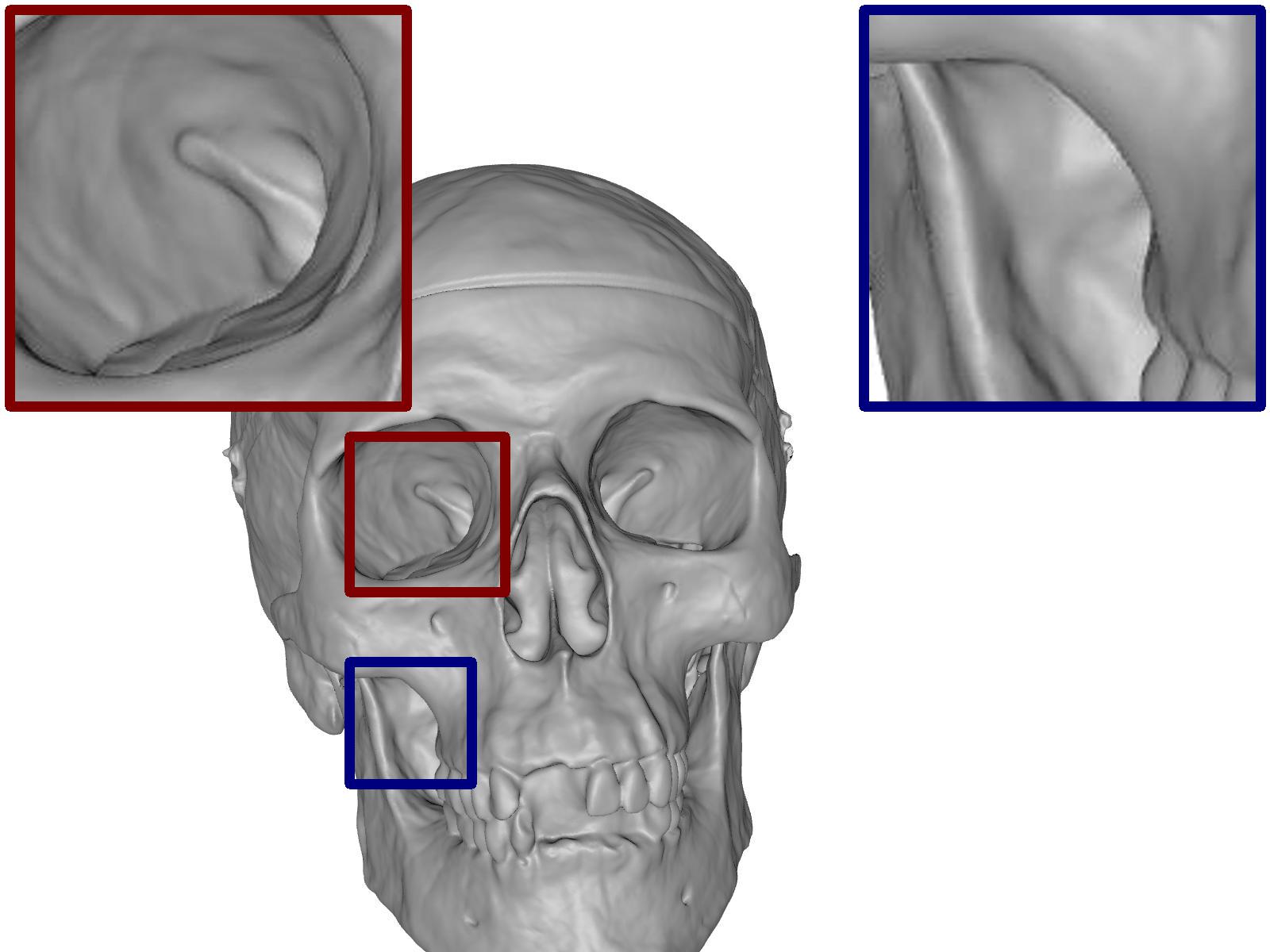}
        \\ 
        \vspace{\mrgv}
        \includegraphics[width=\wid]{figures/dtu_qual/gt/83.jpg} &
        \hspace{\mrg}
        \includegraphics[width=\wid]{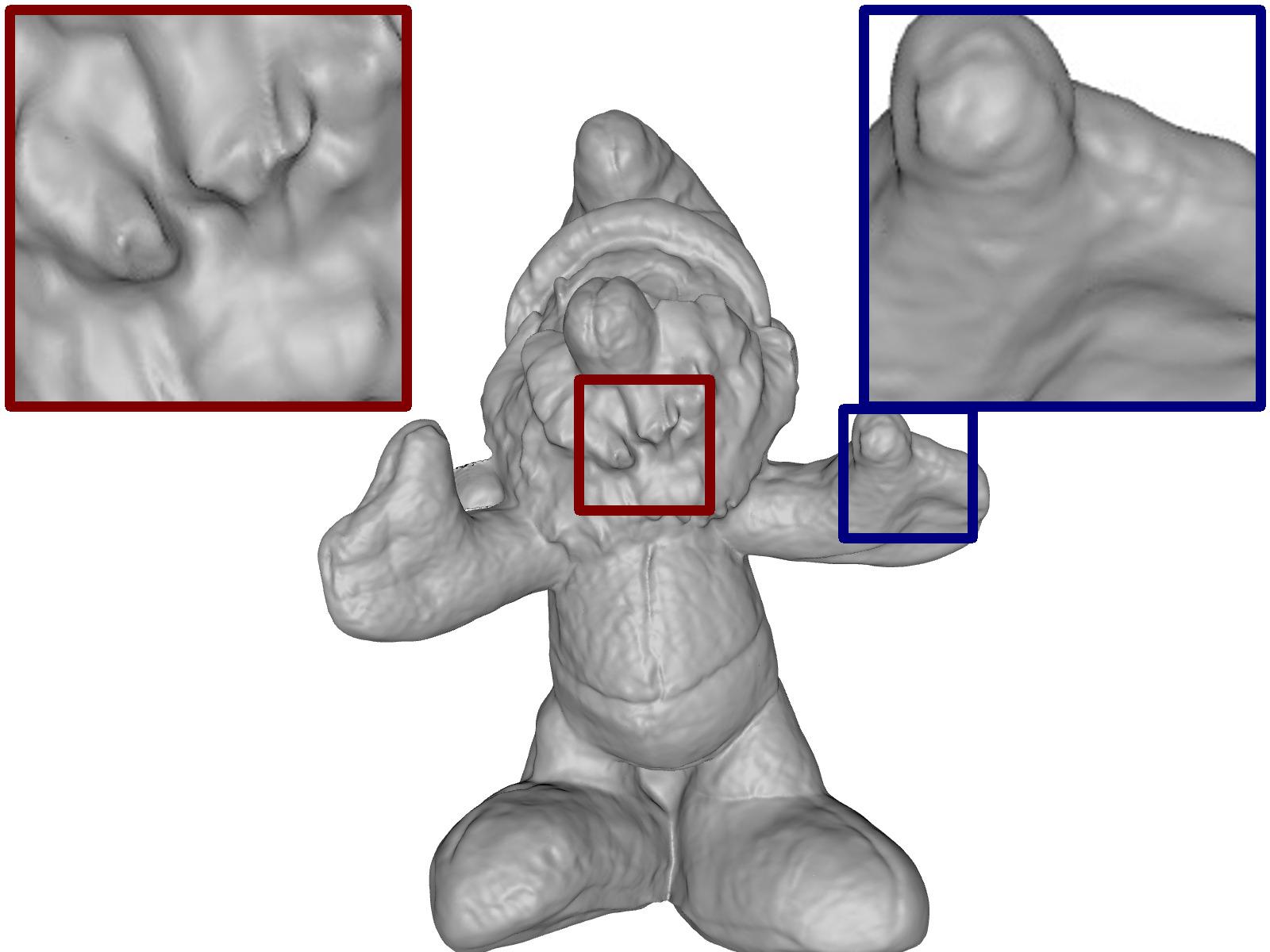} &
        \hspace{\mrg}
        \includegraphics[width=\wid]{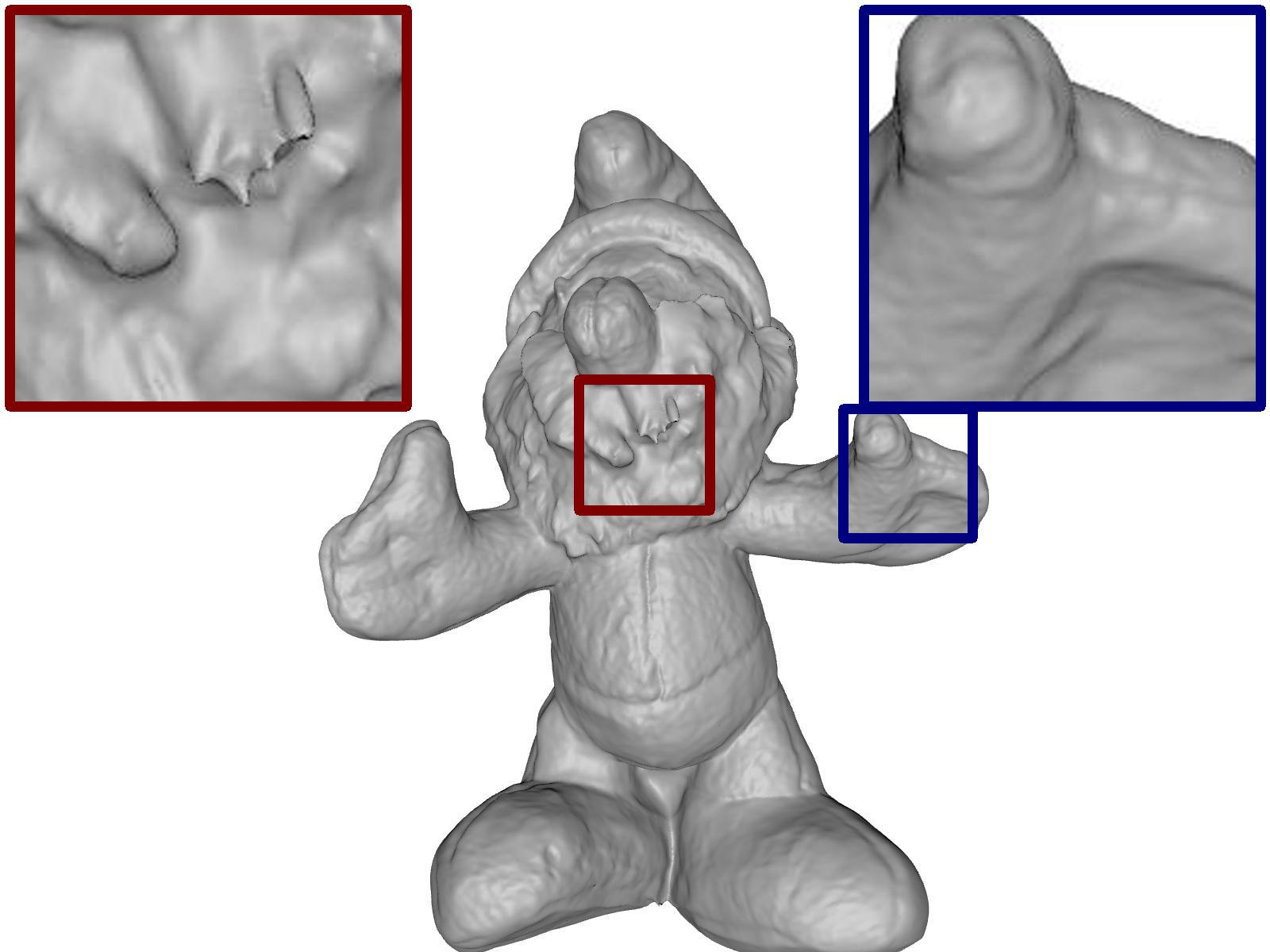}
        \\
        \vspace{\mrgv}
        \includegraphics[width=\wid]{figures/dtu_qual/gt/110.jpg} &
        \hspace{\mrg}
        \includegraphics[width=\wid]{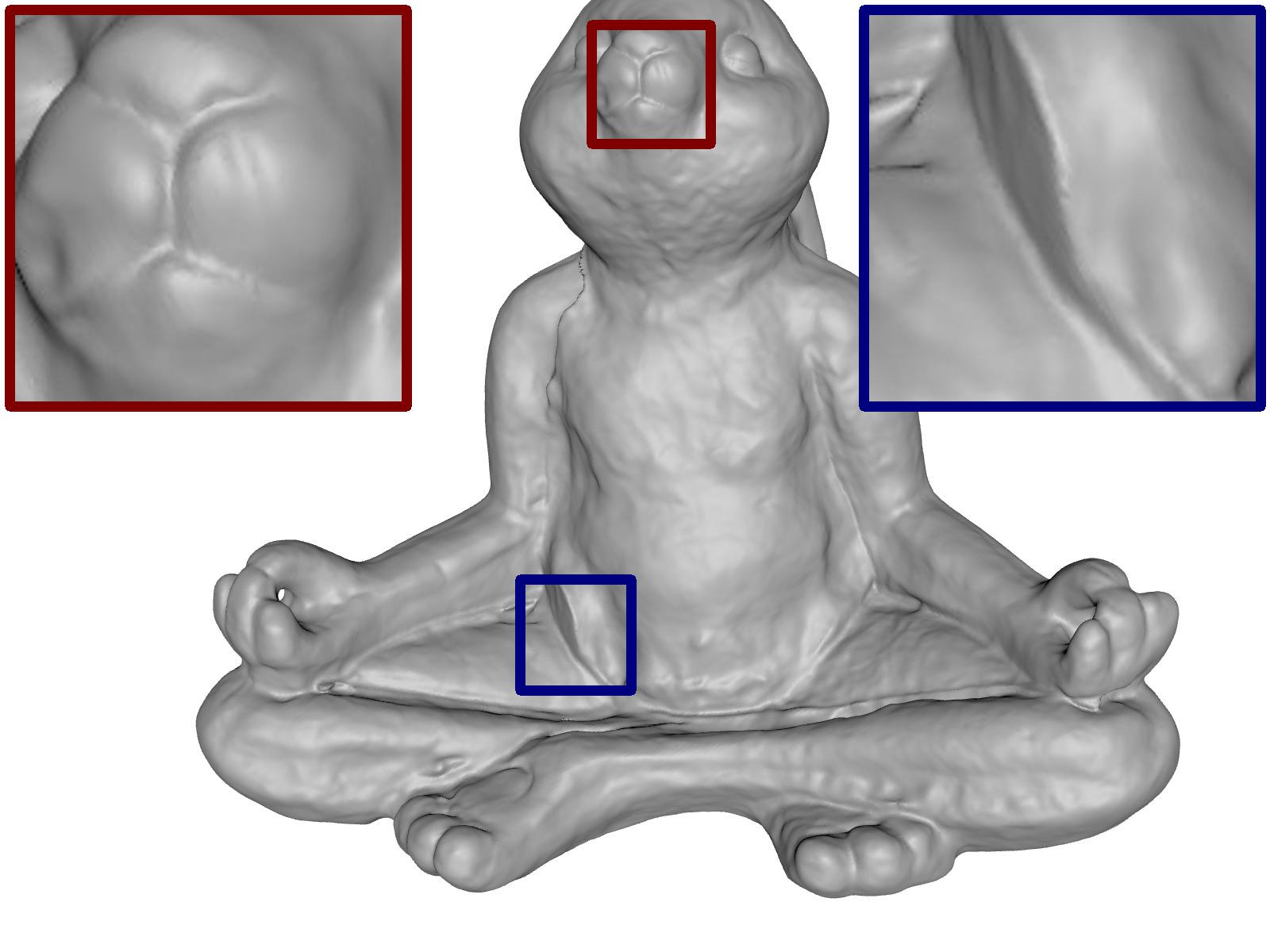} &
        \hspace{\mrg}
        \includegraphics[width=\wid]{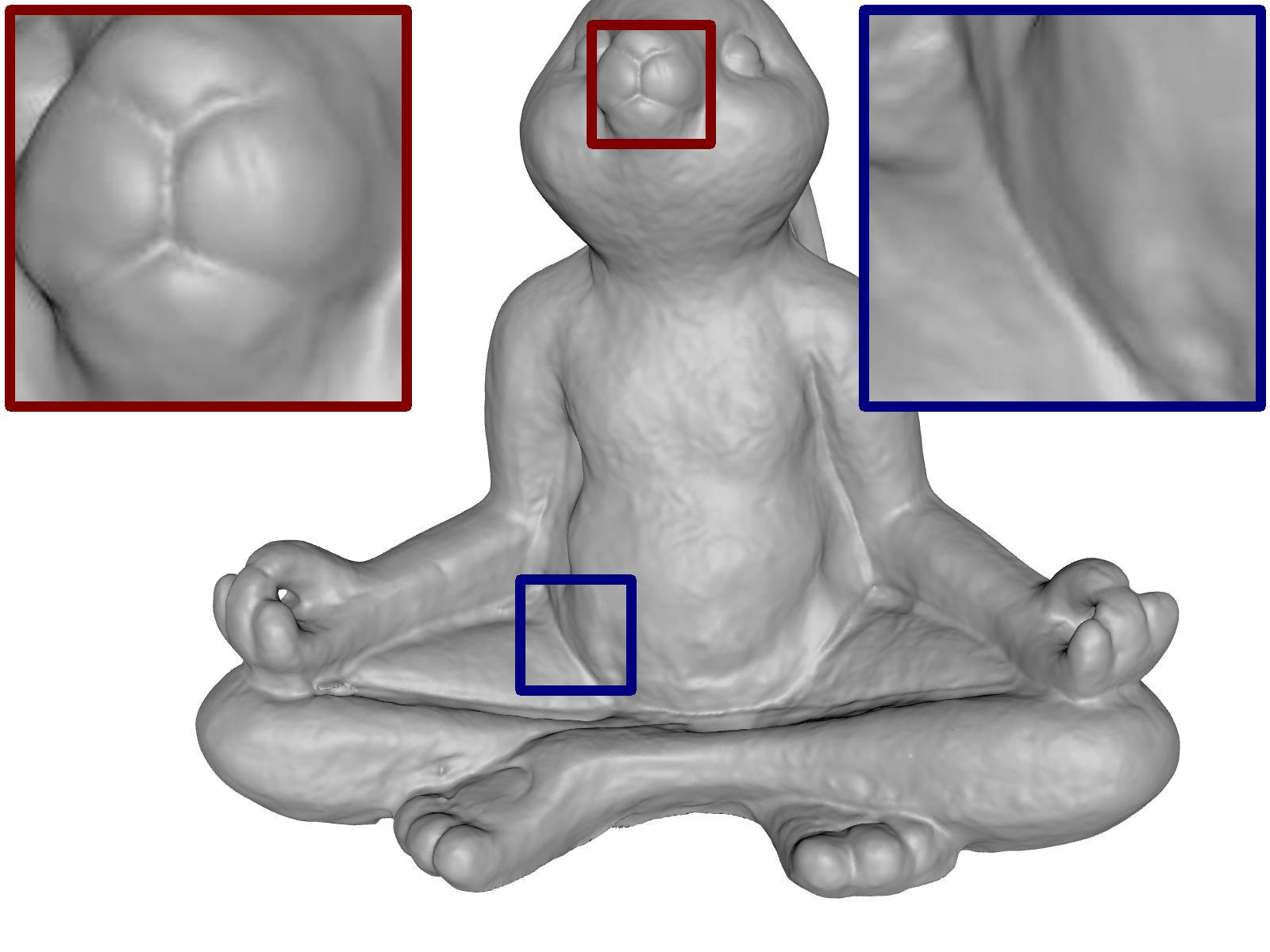}
        \\
        \vspace{\mrgv}
        \includegraphics[width=\wid]{figures/dtu_qual/gt/118.jpg} &
        \hspace{\mrg}
        \includegraphics[width=\wid]{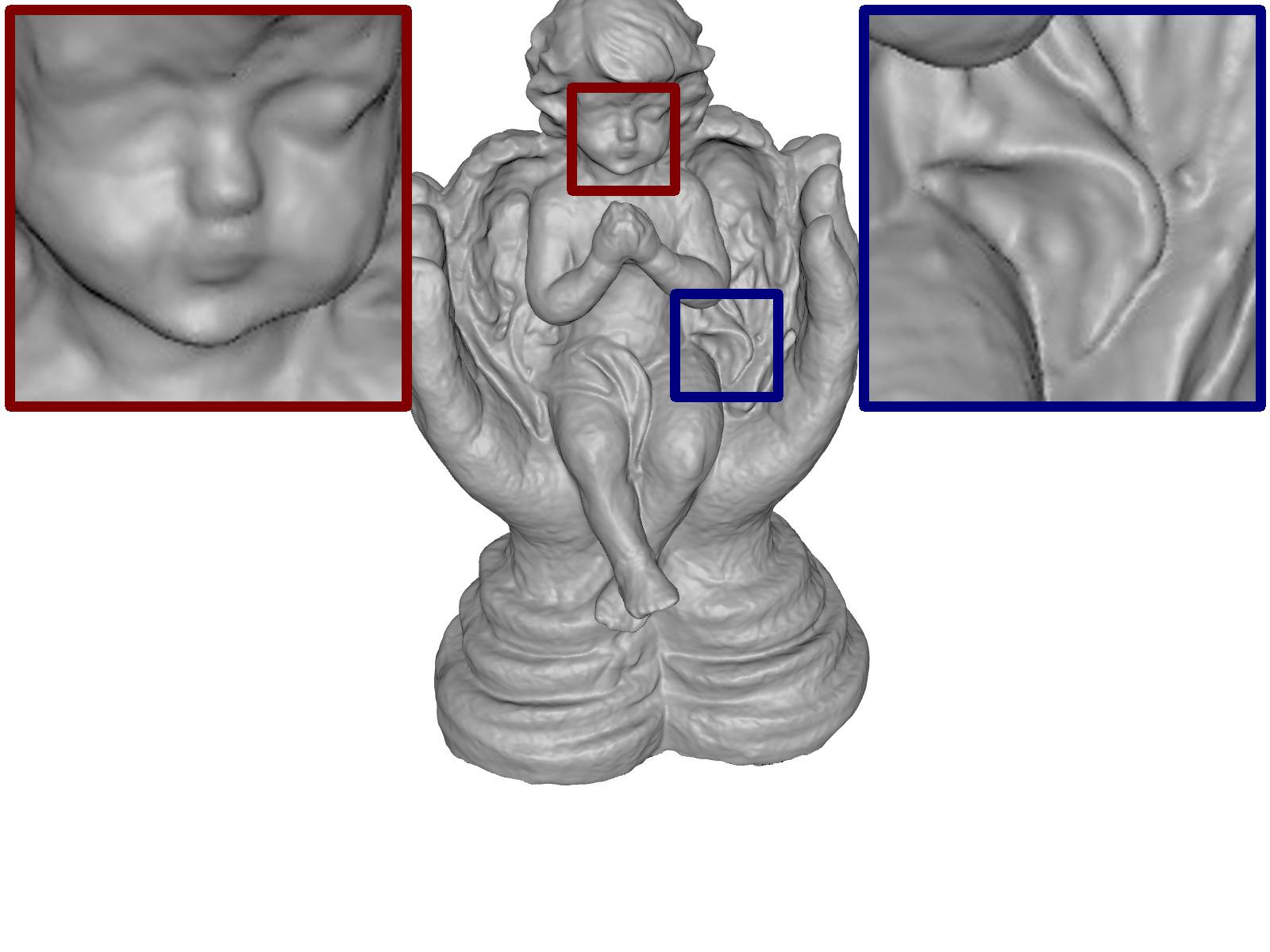} &
        \hspace{\mrg}
        \includegraphics[width=\wid]{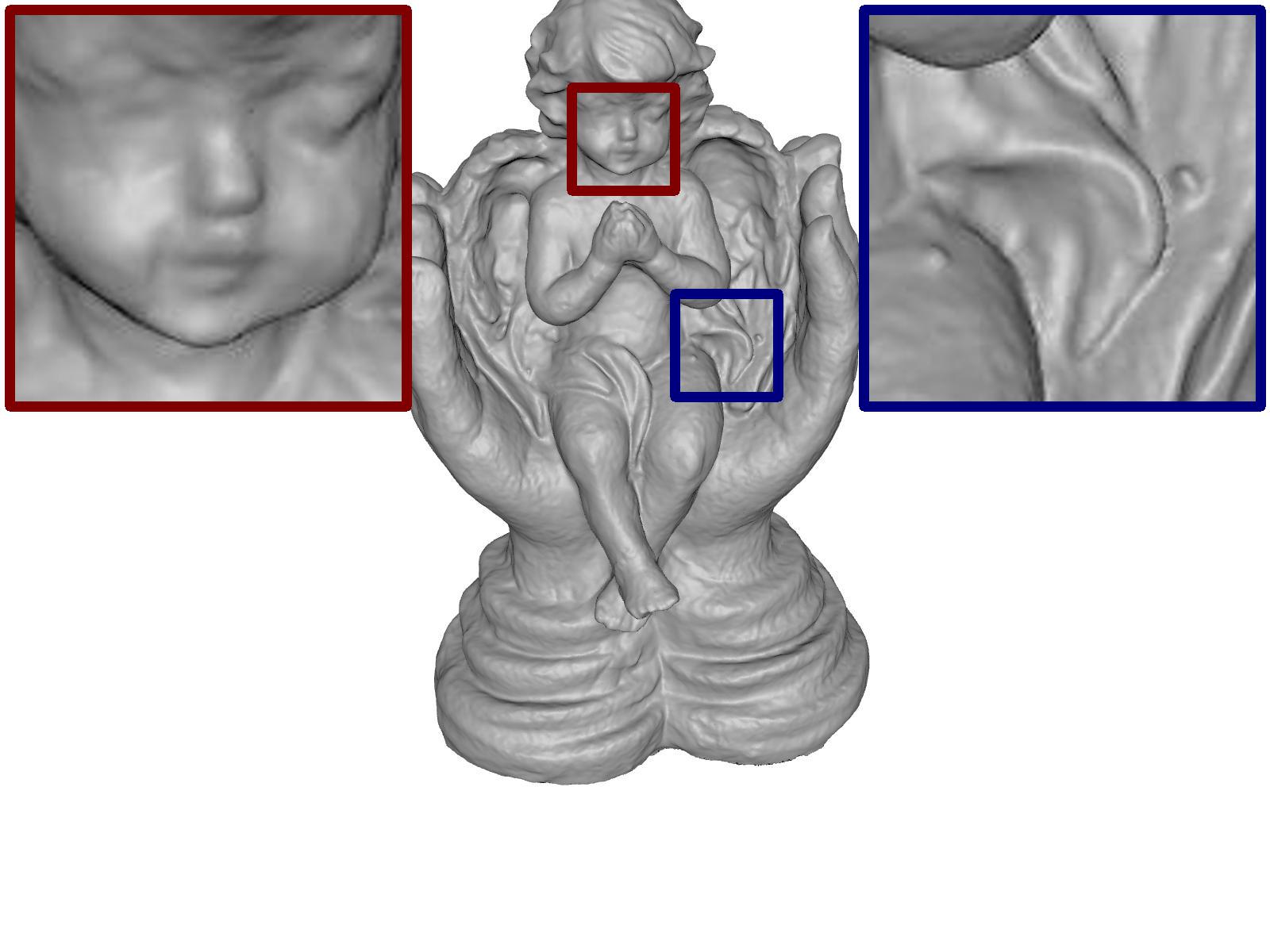}
        \\
        \vspace{\mrgv}
        \includegraphics[width=\wid]{figures/dtu_qual/gt/122.jpg} &
        \hspace{\mrg}
        \includegraphics[width=\wid]{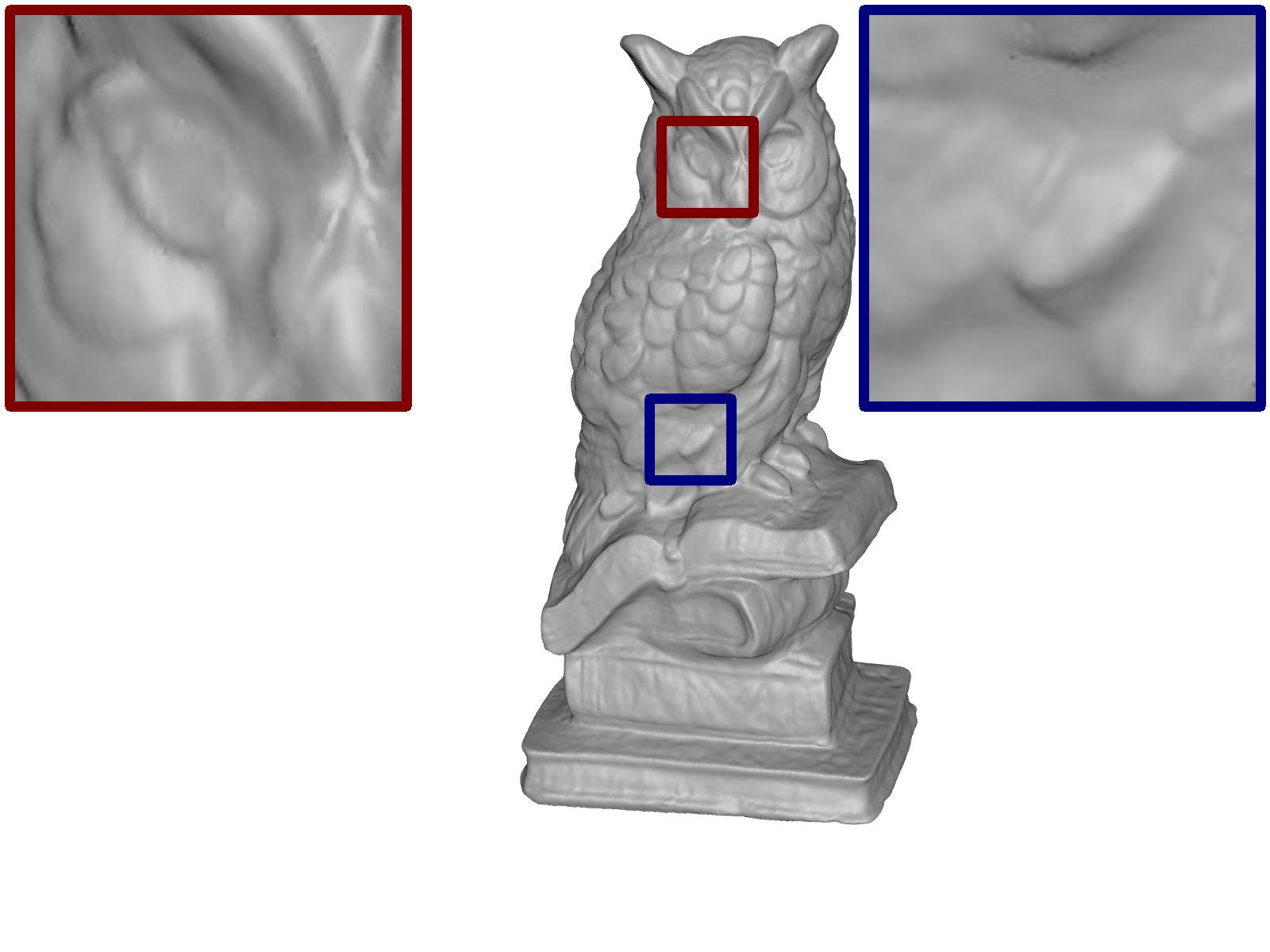} &
        \hspace{\mrg}
        \includegraphics[width=\wid]{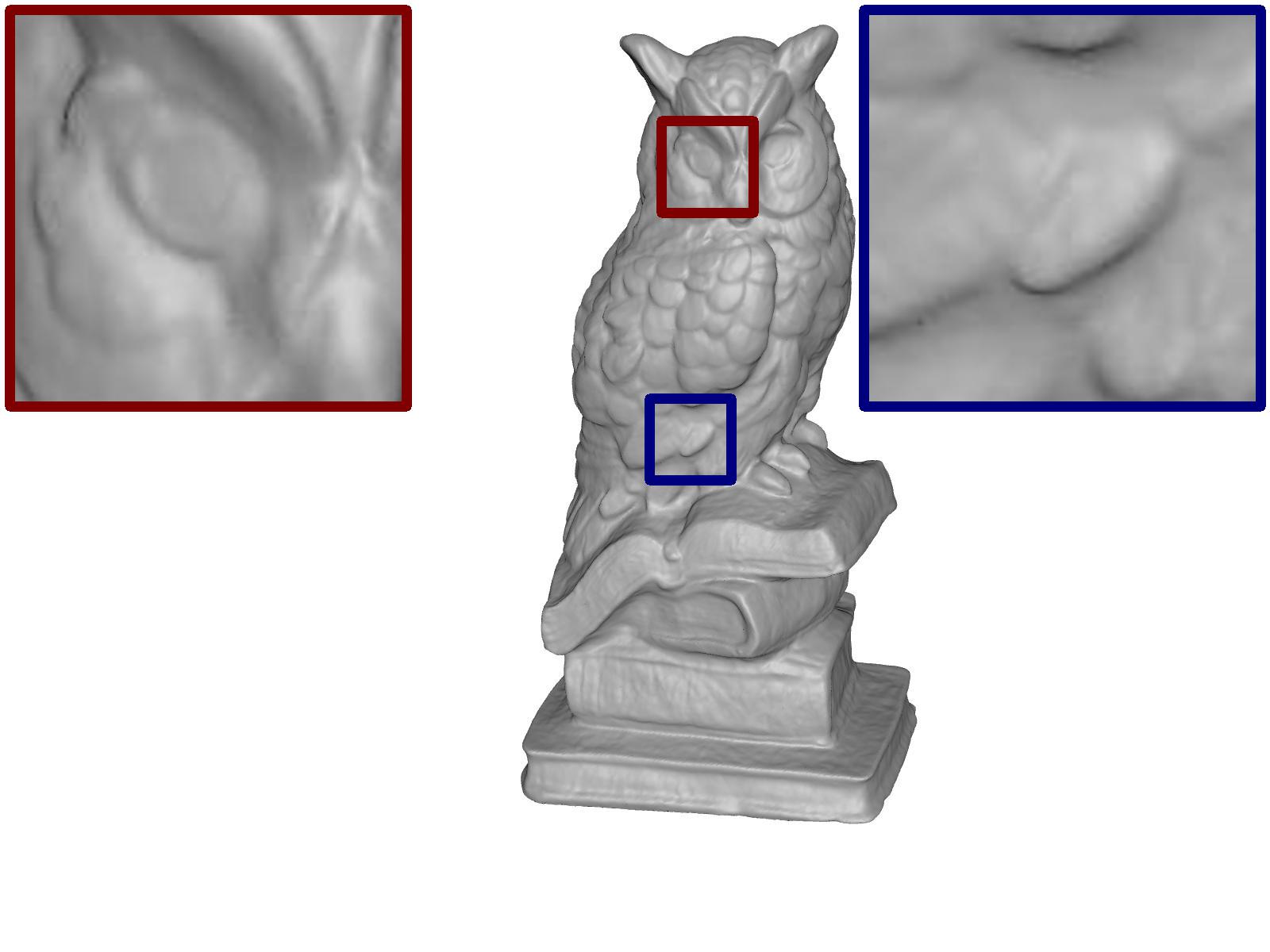}
        \\
        \textbf{Source} & \hspace{\mrg}
        \textbf{NeuS w/ masks} & \hspace{\mrg}
        \textbf{NeuS w/ masks (ours)}
    \end{tabular}
    \caption{Additional qualitative results on the DTU~\cite{Jensen2014LargeSM} dataset for NeuS~\cite{Wang2021NeuSLN} method trained with masks supervision.}
    \label{fig:dtu_qual_appendix_neusm}
    \vspace{-0.4cm}
\end{figure*}

\begin{figure*}
    \centering    
    \setlength{\wid}{0.20\textwidth}
    \setlength{\mrg}{-0.45cm}
    \setlength{\mrgv}{-0.05cm}
    \begin{tabular}{c cc}
        \vspace{\mrgv}
        \includegraphics[width=\wid]{figures/dtu_qual/gt/37.jpg} &
        \hspace{\mrg}
        \includegraphics[width=\wid]{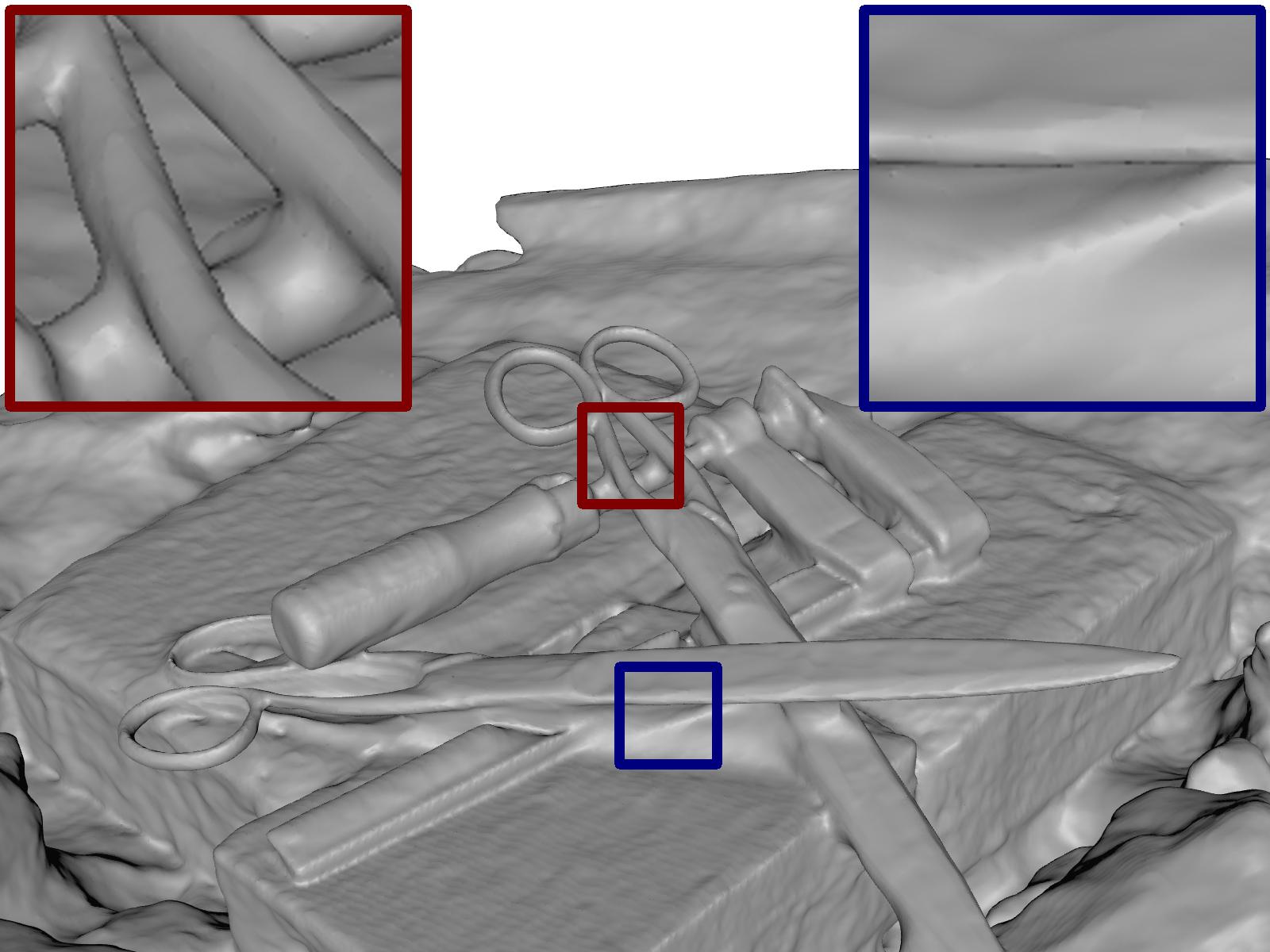} &
        \hspace{\mrg}
        \includegraphics[width=\wid]{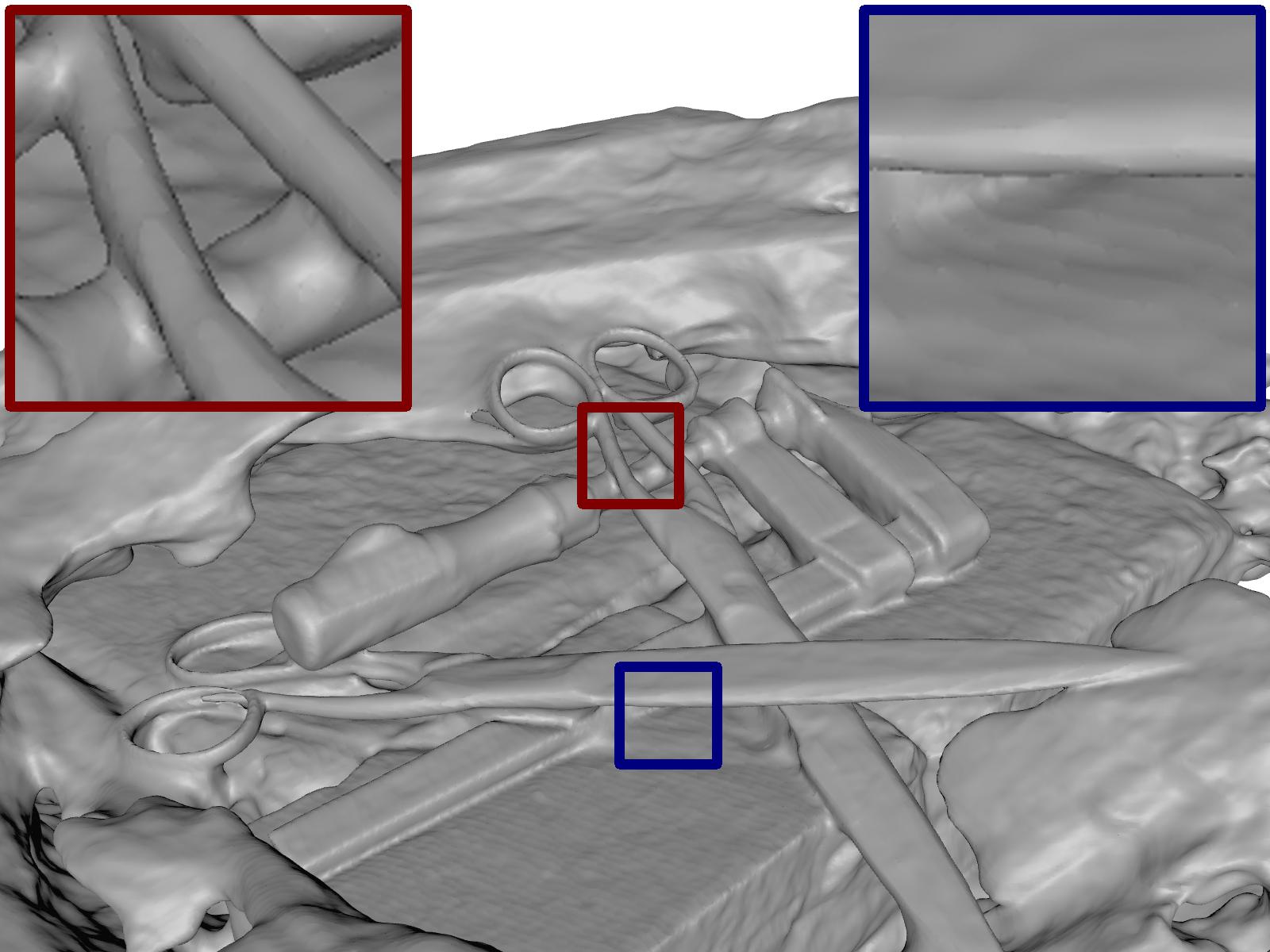}
        \\ 
        \vspace{\mrgv}
        \includegraphics[width=\wid]{figures/dtu_qual/gt/55.jpg} &
        \hspace{\mrg}
        \includegraphics[width=\wid]{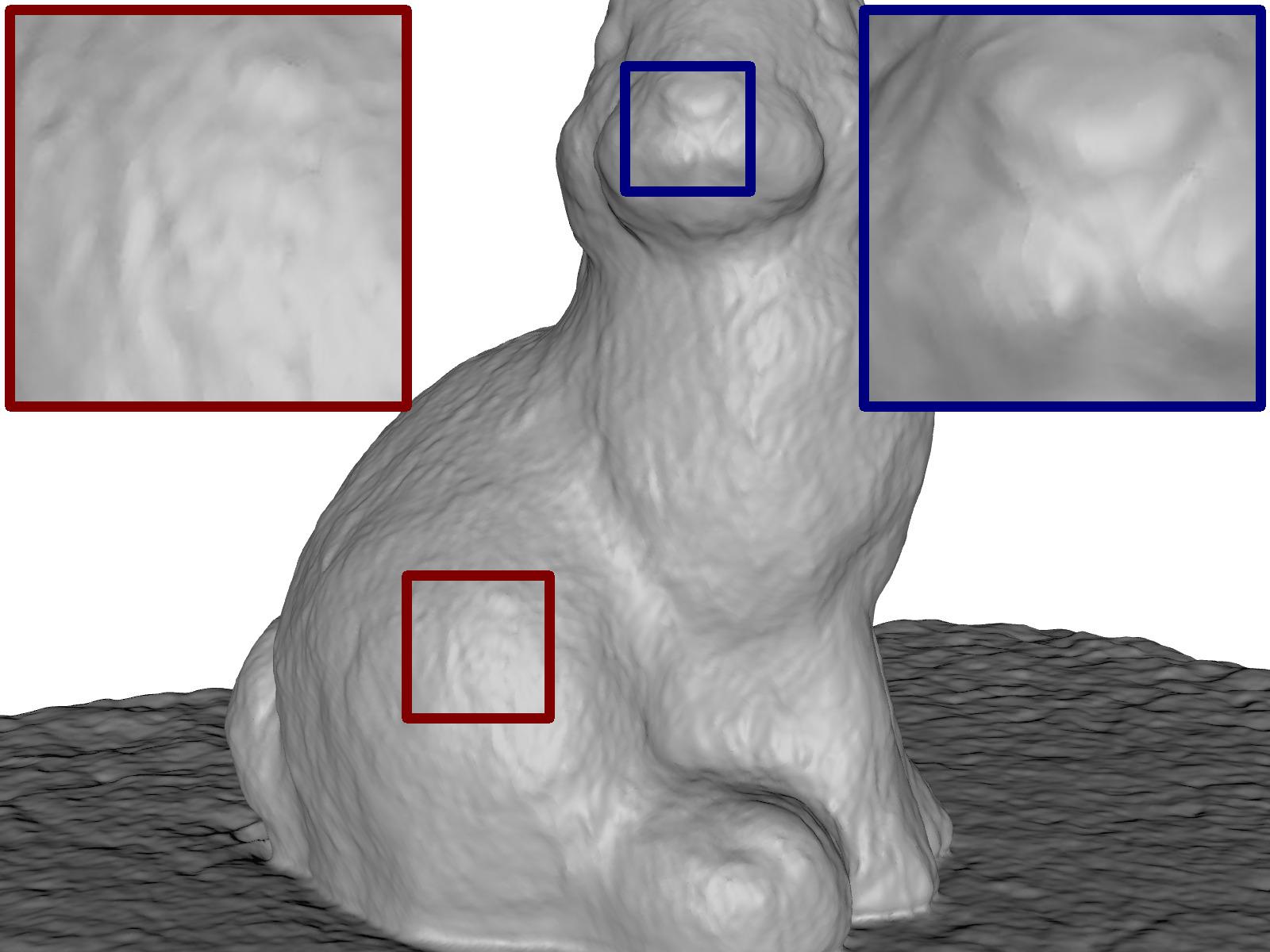} &
        \hspace{\mrg}
        \includegraphics[width=\wid]{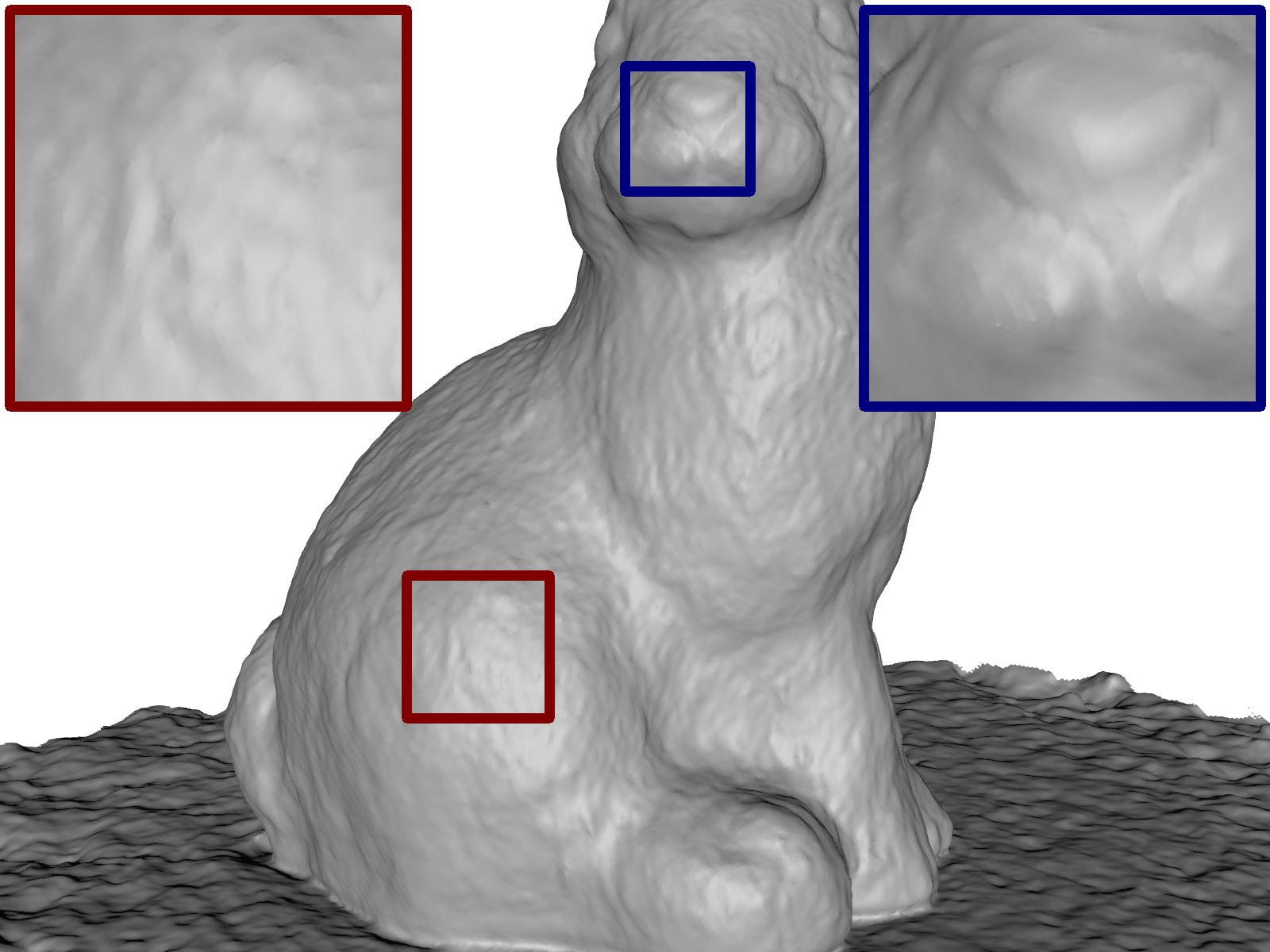}
        \\ 
        \vspace{\mrgv}
        \includegraphics[width=\wid]{figures/dtu_qual/gt/63.jpg} &
        \hspace{\mrg}
        \includegraphics[width=\wid]{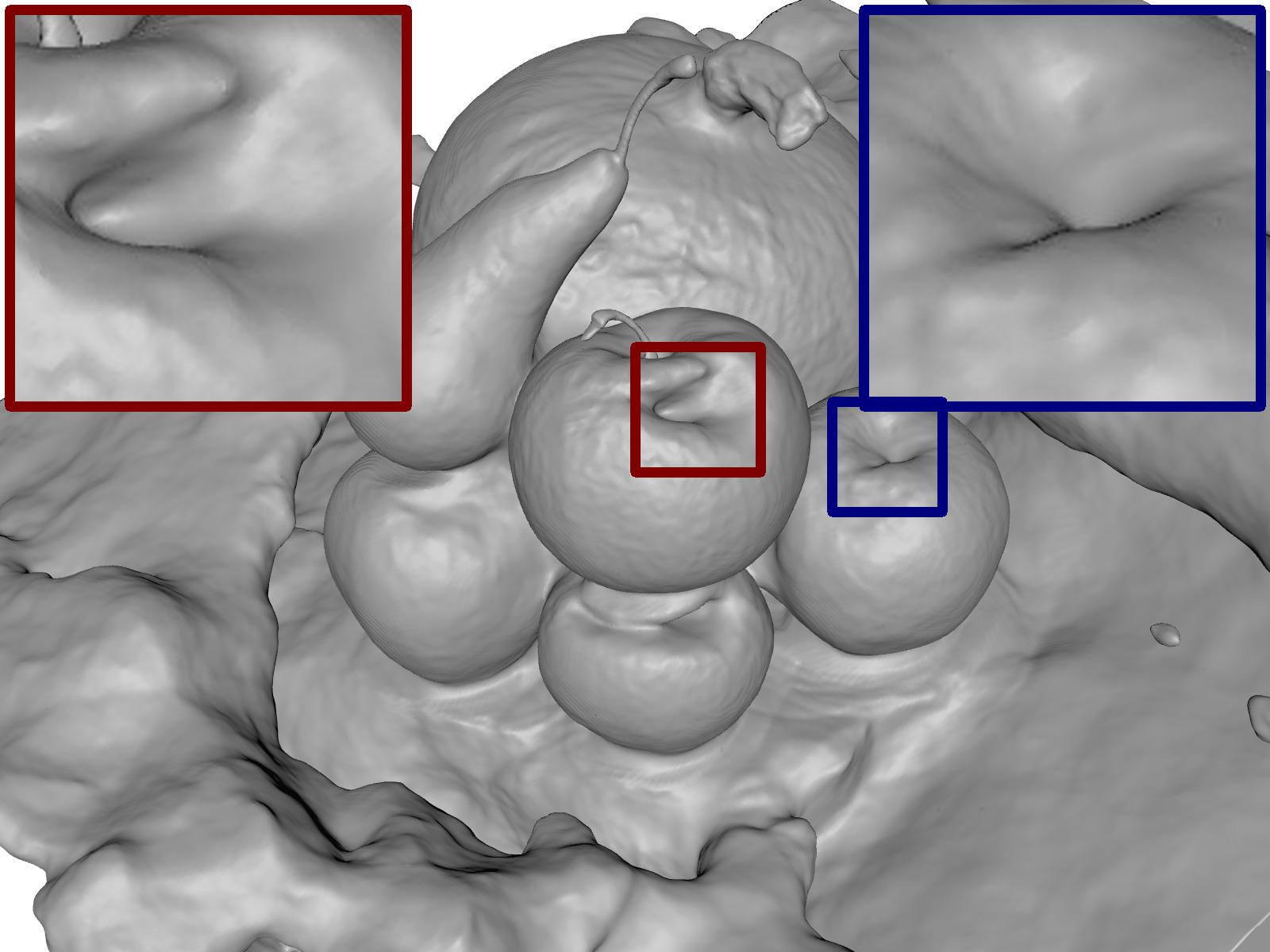} &
        \hspace{\mrg}
        \includegraphics[width=\wid]{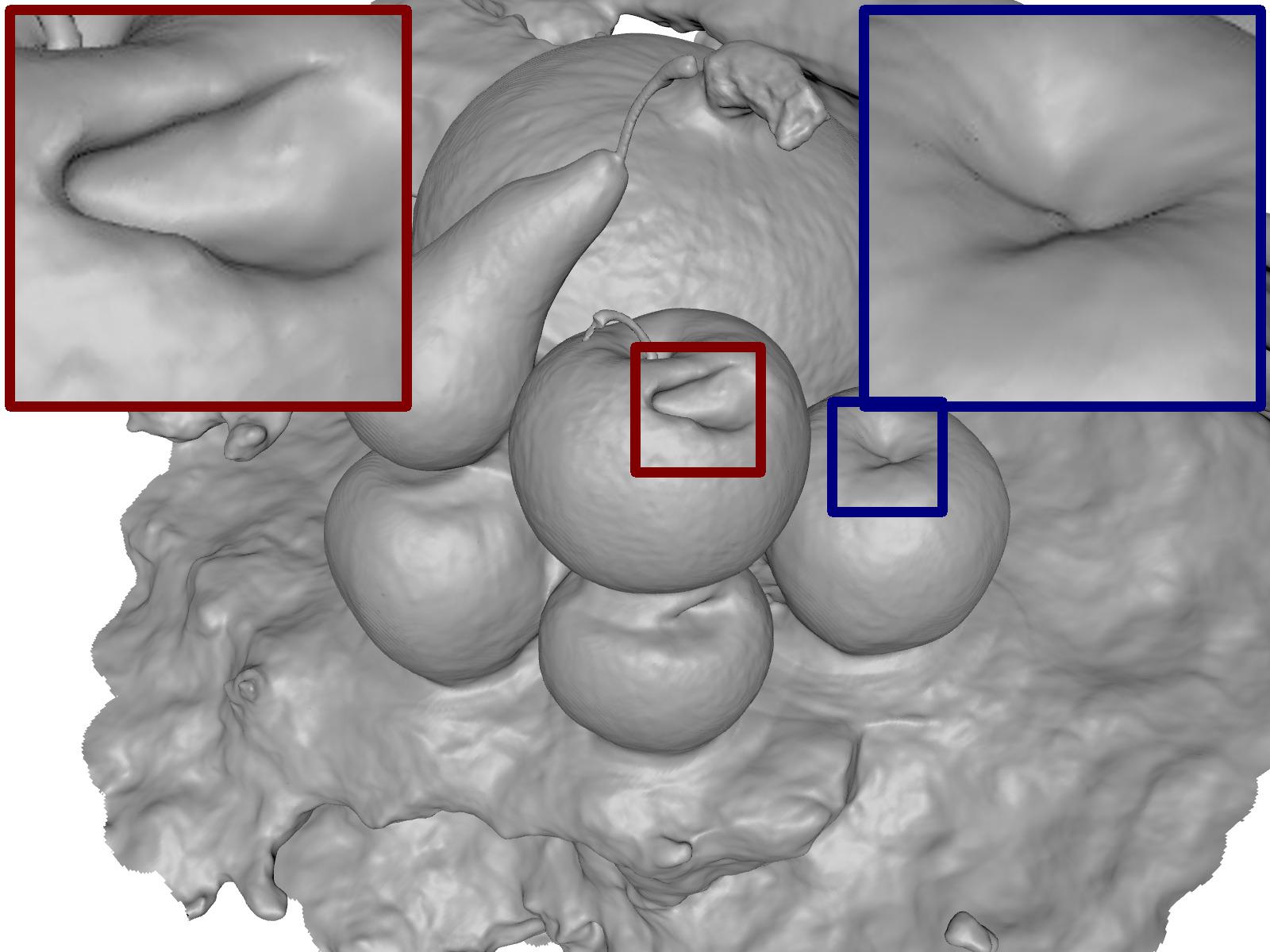}
        \\ 
        \vspace{\mrgv}
        \includegraphics[width=\wid]{figures/dtu_qual/gt/65.jpg} &
        \hspace{\mrg}
        \includegraphics[width=\wid]{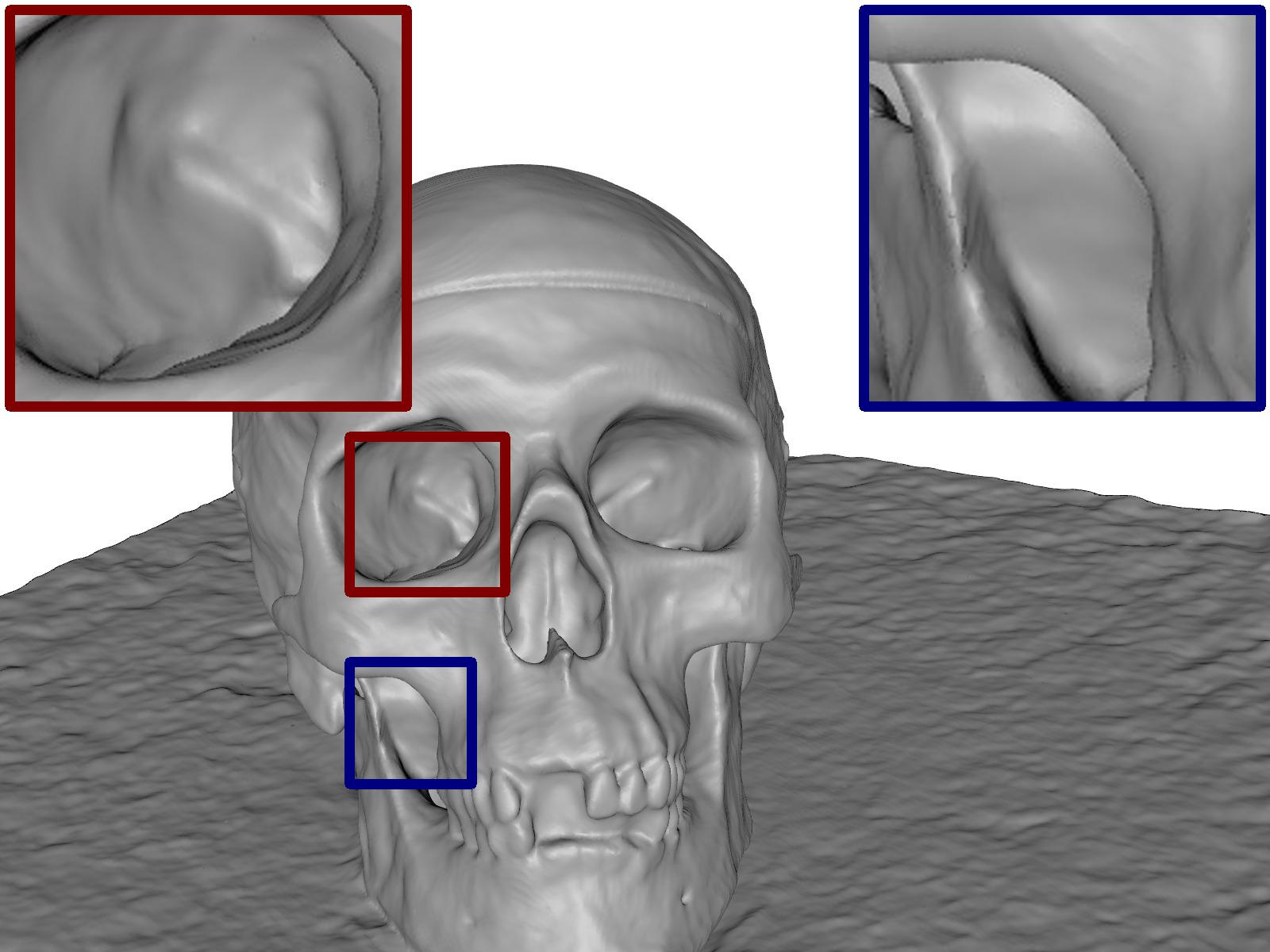} &
        \hspace{\mrg}
        \includegraphics[width=\wid]{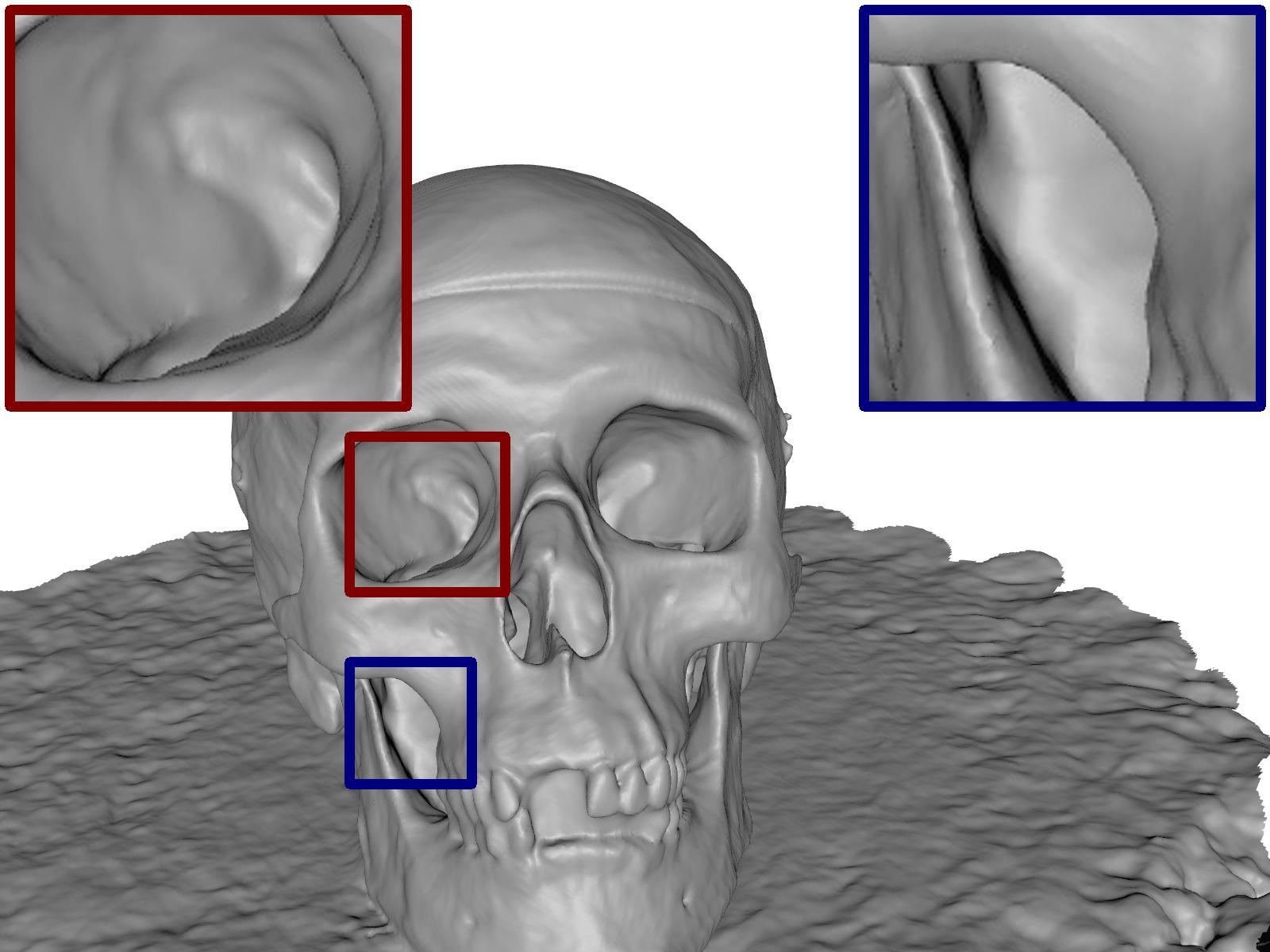}
        \\ 
        \vspace{\mrgv}
        \includegraphics[width=\wid]{figures/dtu_qual/gt/83.jpg} &
        \hspace{\mrg}
        \includegraphics[width=\wid]{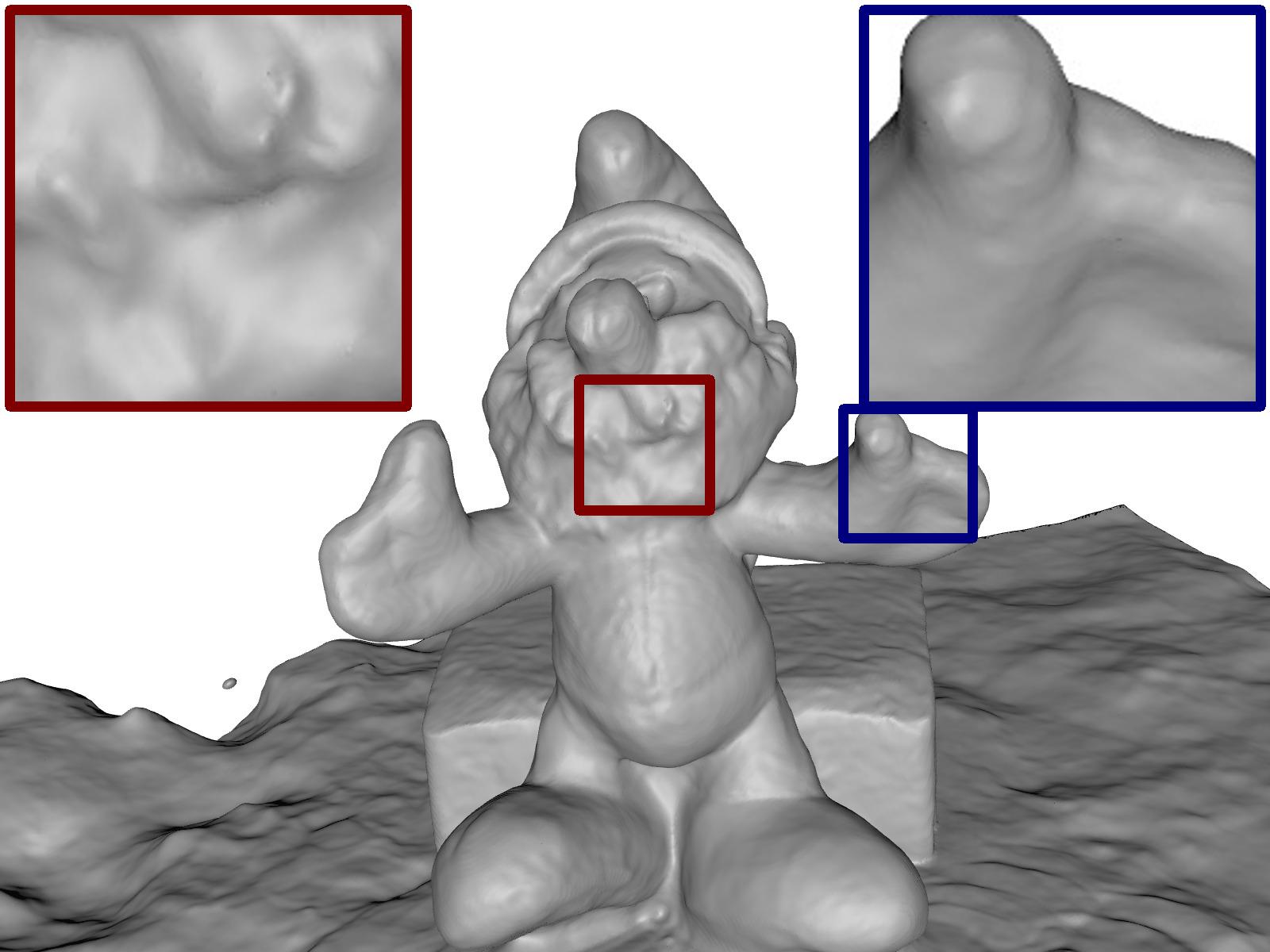} &
        \hspace{\mrg}
        \includegraphics[width=\wid]{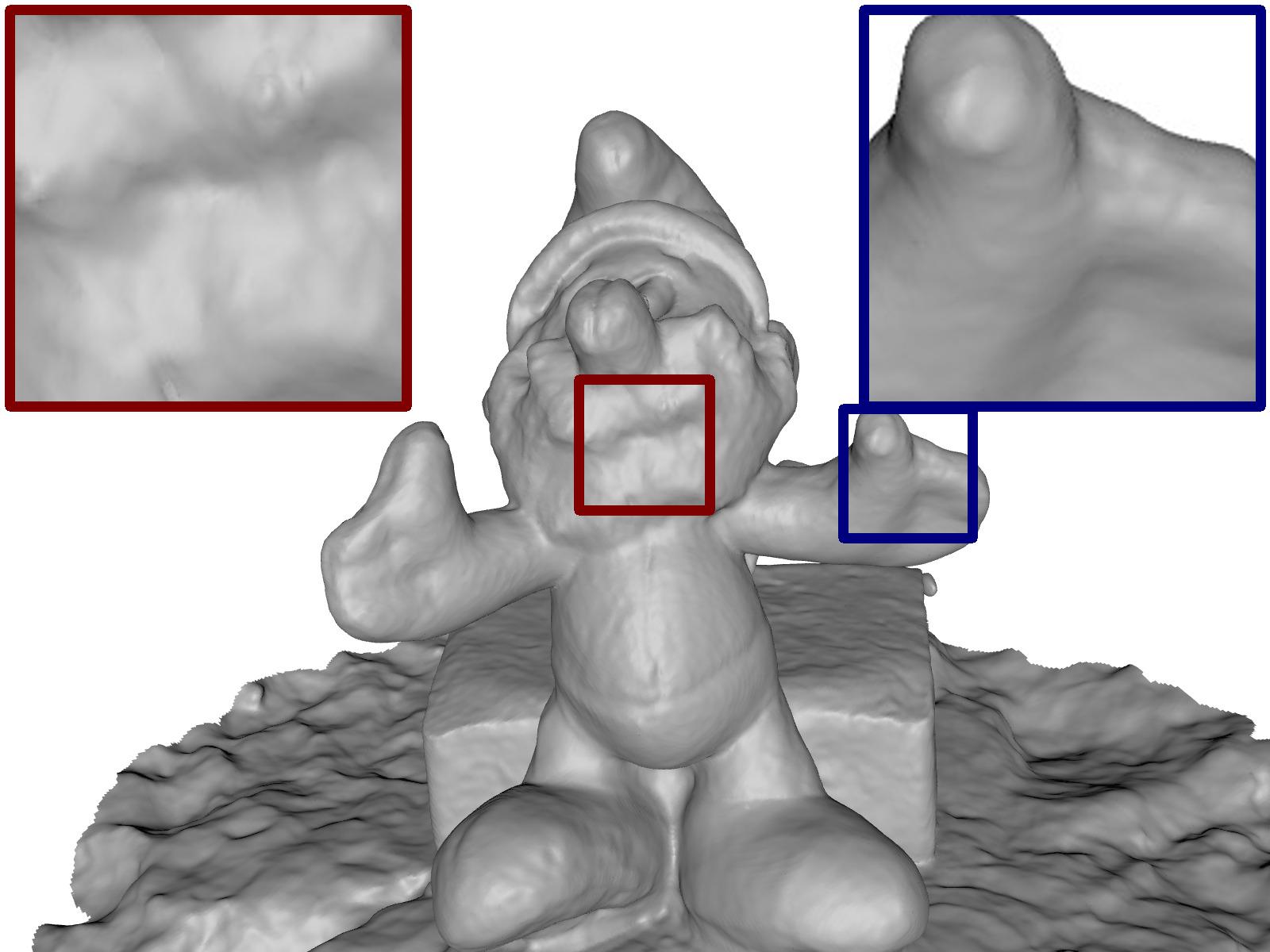}
        \\
        \vspace{\mrgv}
        \includegraphics[width=\wid]{figures/dtu_qual/gt/110.jpg} &
        \hspace{\mrg}
        \includegraphics[width=\wid]{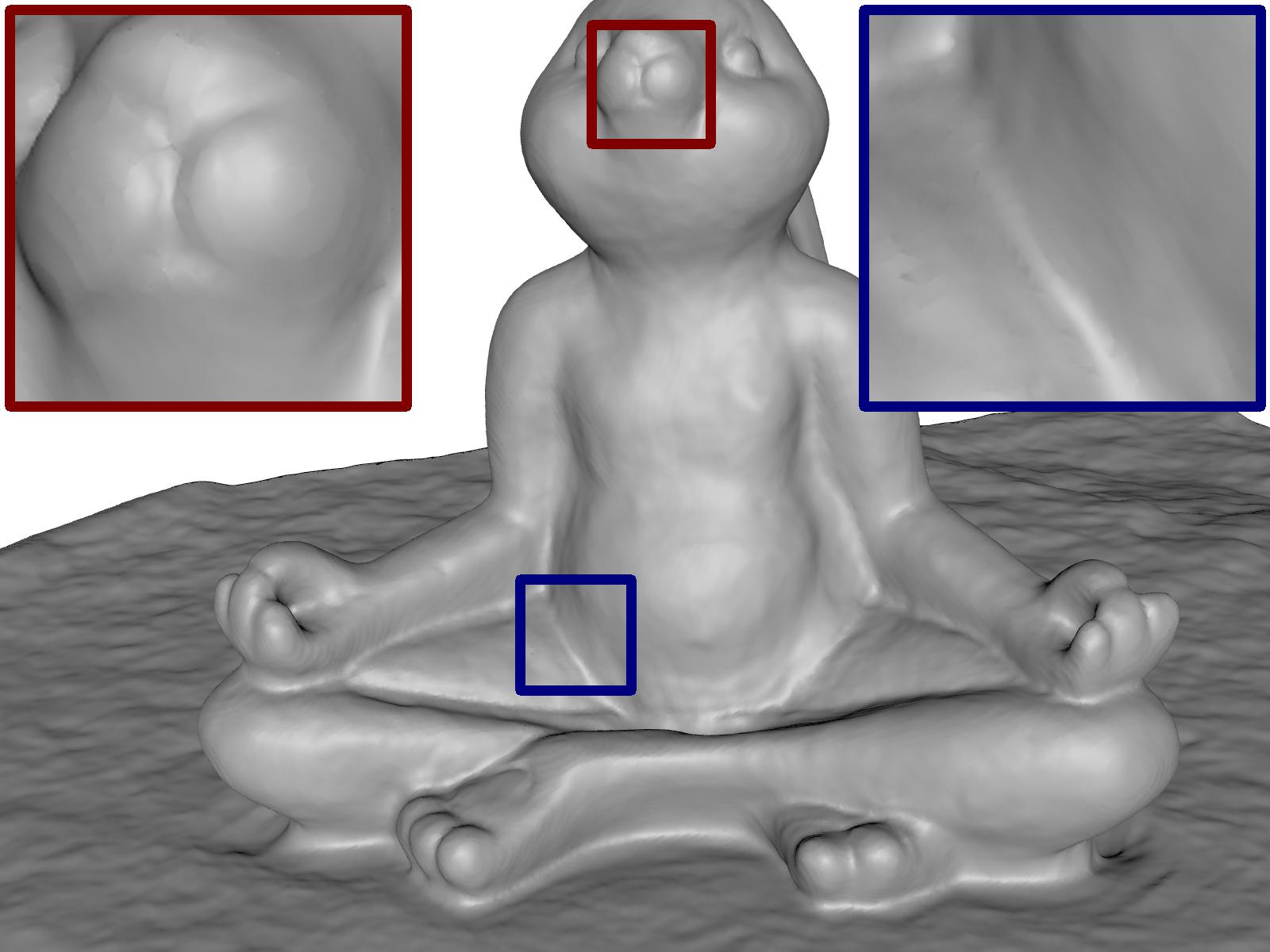} &
        \hspace{\mrg}
        \includegraphics[width=\wid]{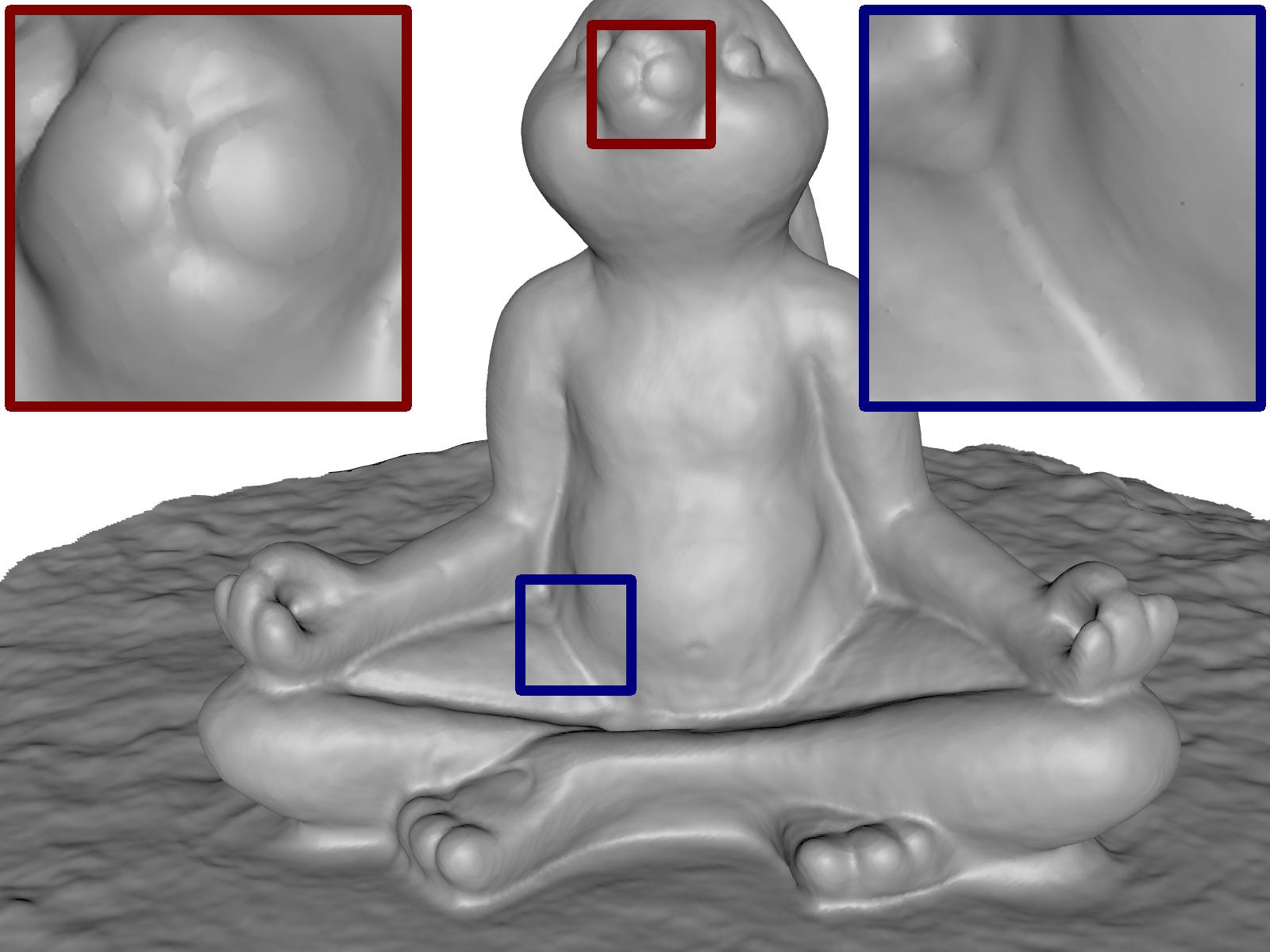}
        \\
        \vspace{\mrgv}
        \includegraphics[width=\wid]{figures/dtu_qual/gt/118.jpg} &
        \hspace{\mrg}
        \includegraphics[width=\wid]{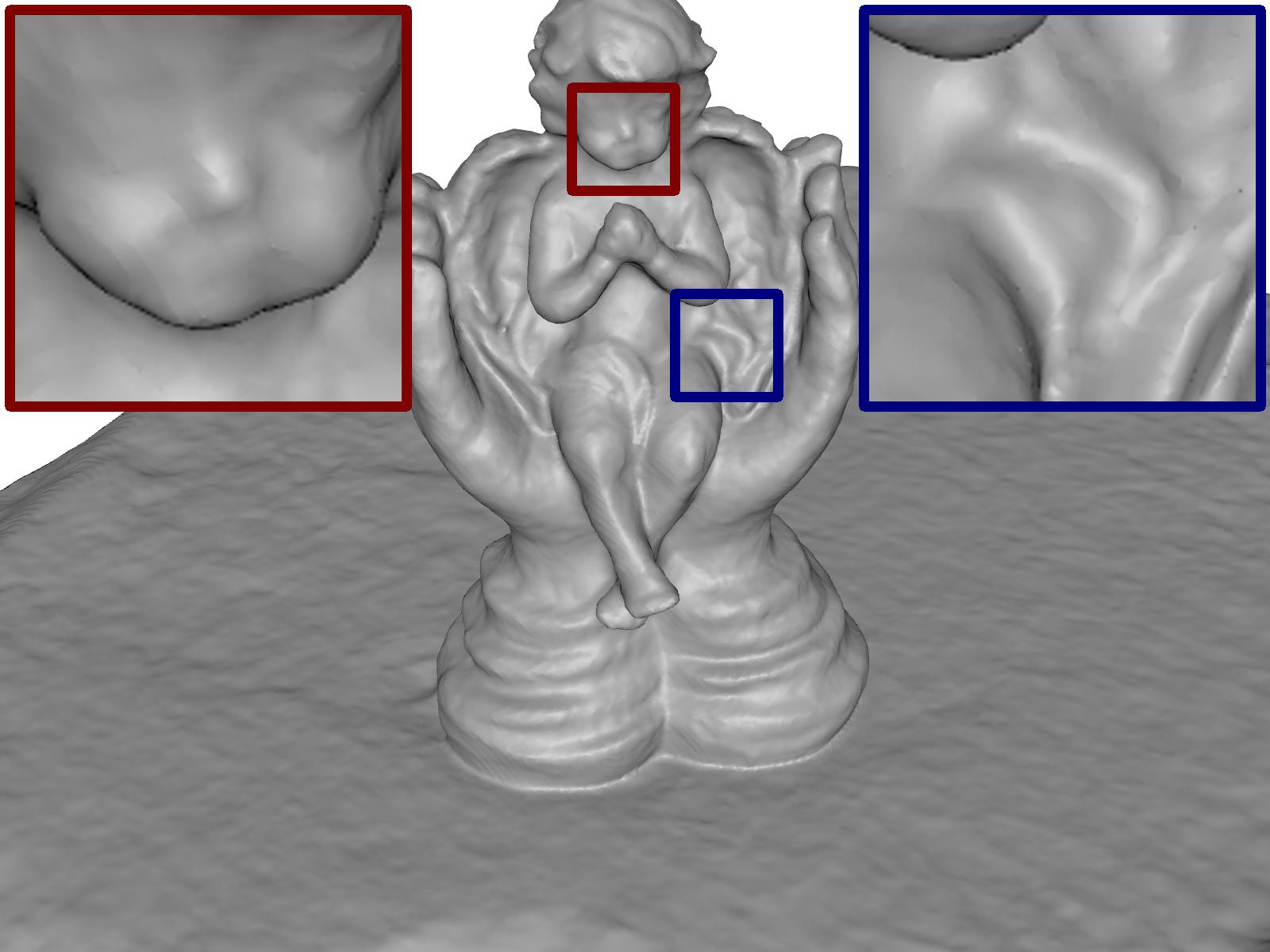} &
        \hspace{\mrg}
        \includegraphics[width=\wid]{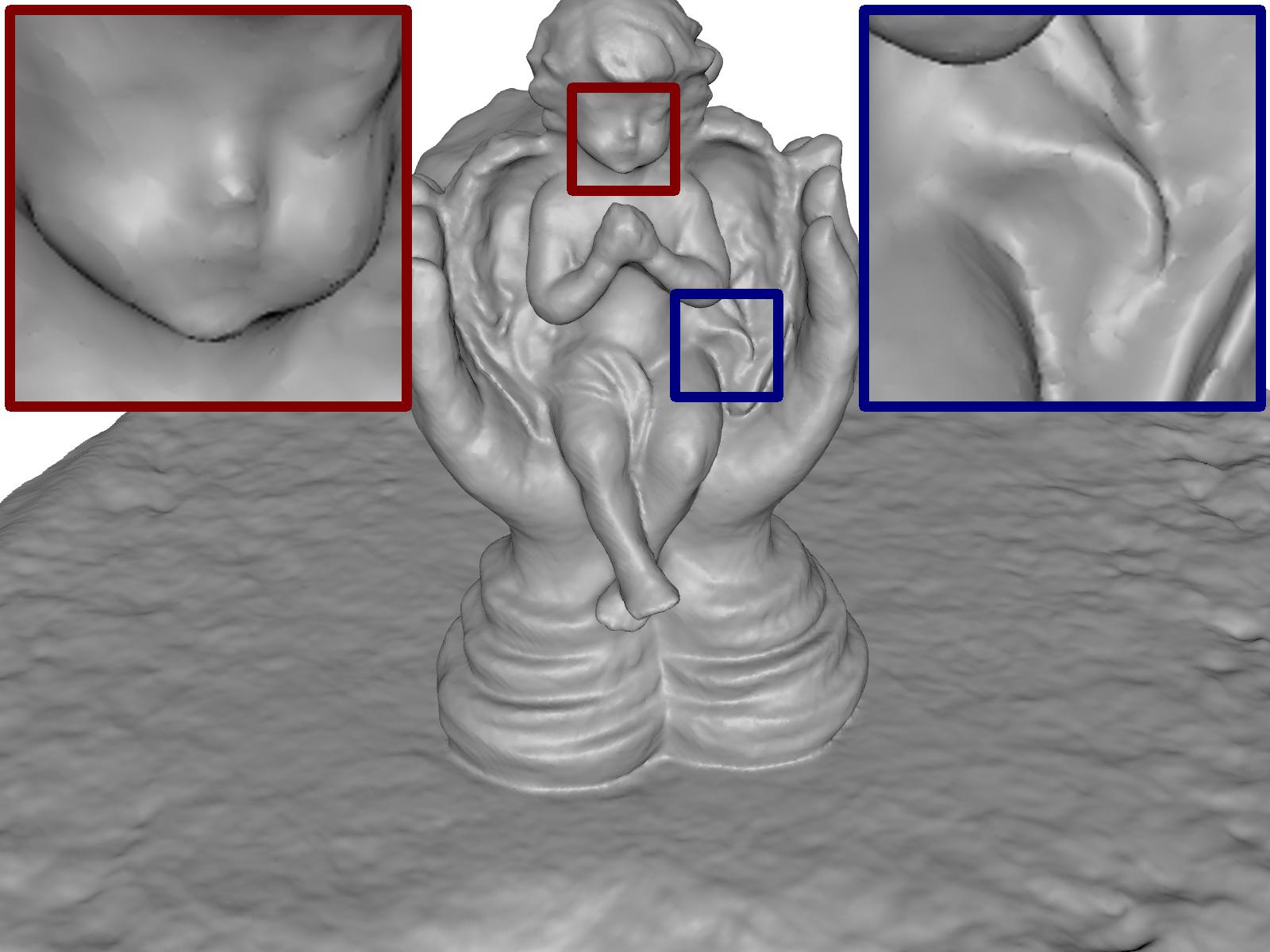}
        \\
        \vspace{\mrgv}
        \includegraphics[width=\wid]{figures/dtu_qual/gt/122.jpg} &
        \hspace{\mrg}
        \includegraphics[width=\wid]{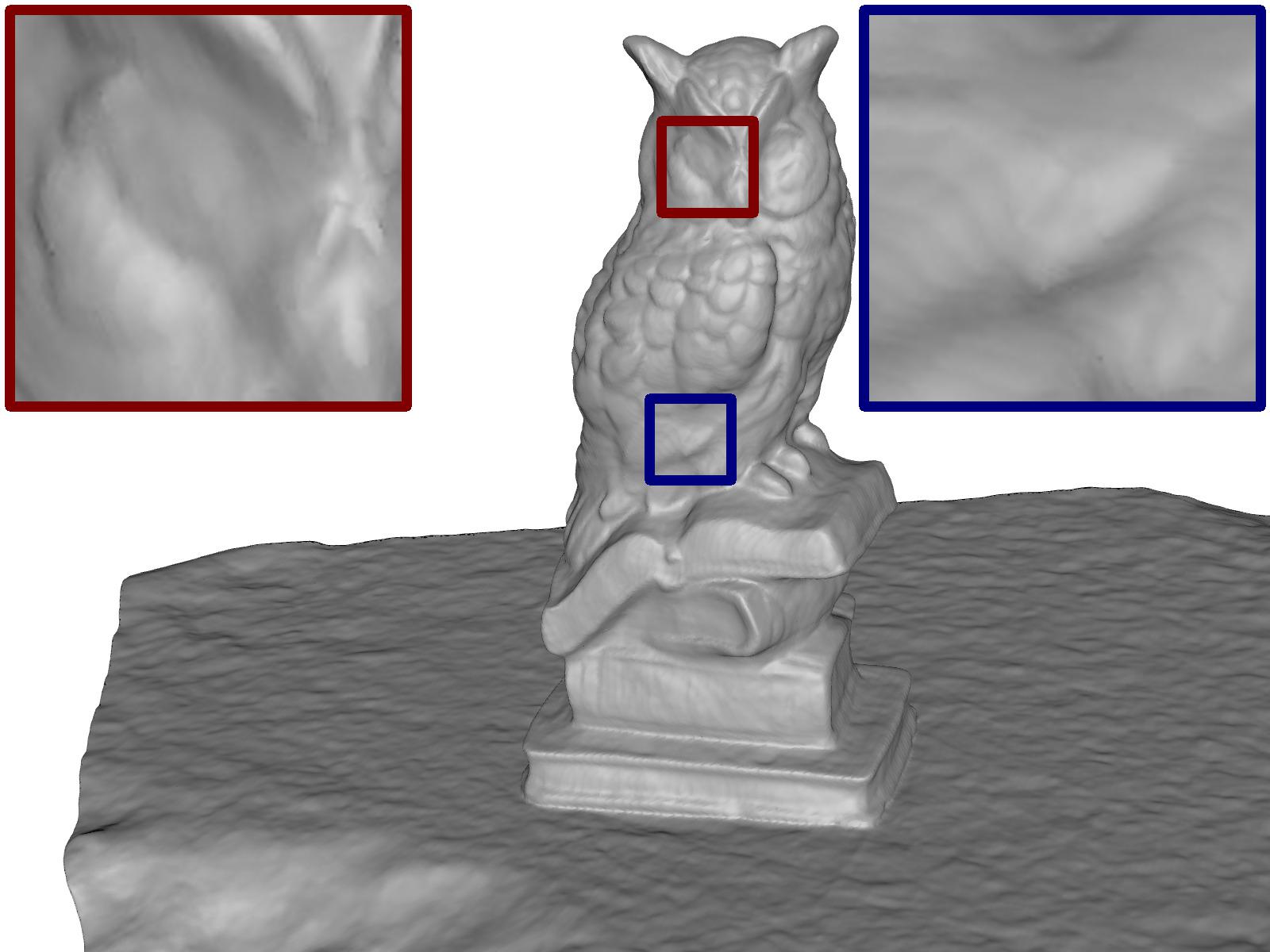} &
        \hspace{\mrg}
        \includegraphics[width=\wid]{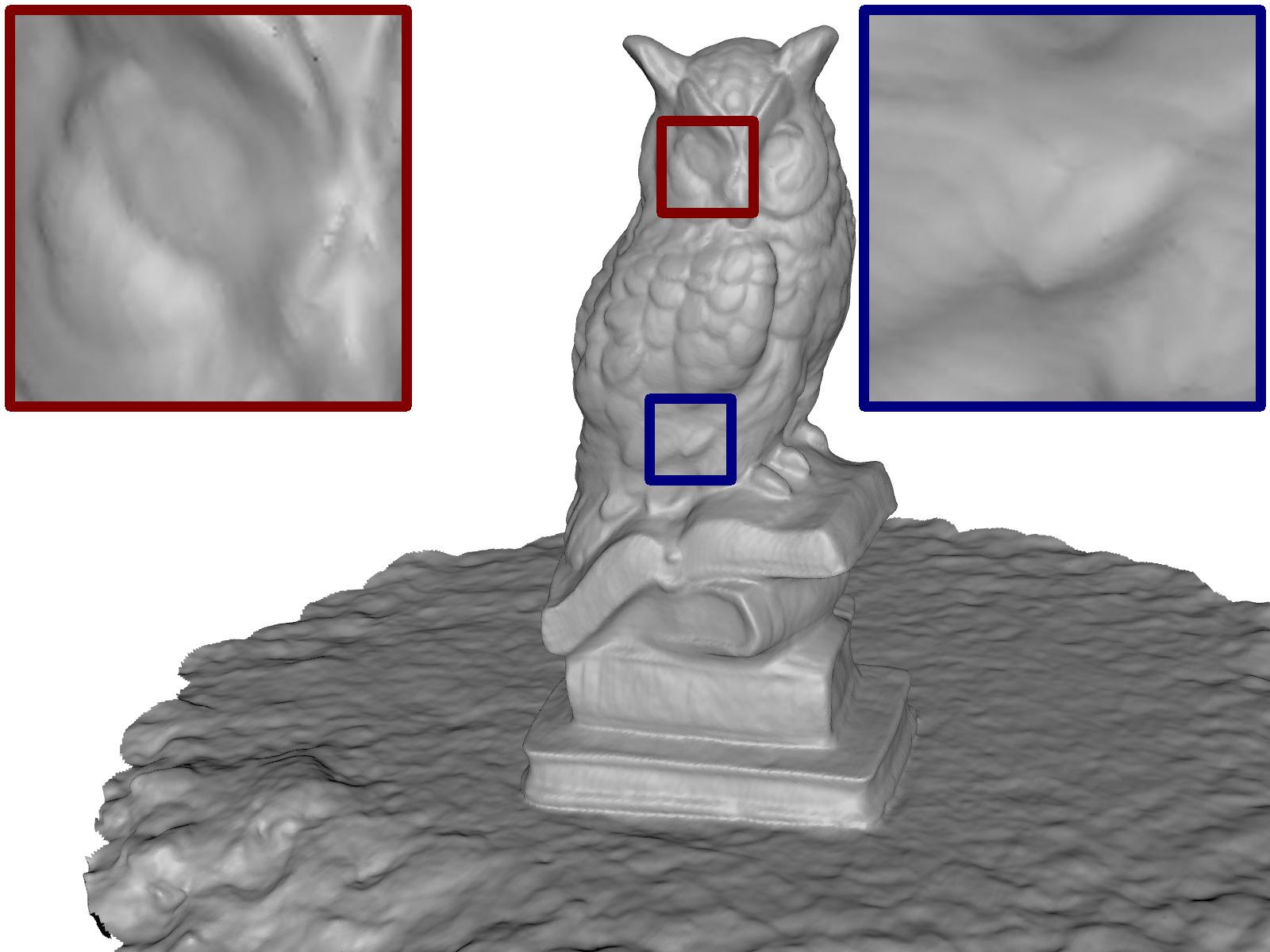}
        \\
        \textbf{Source} & \hspace{\mrg}
        \textbf{UNISURF} & \hspace{\mrg}
        \textbf{UNISURF (ours)}
    \end{tabular}
    \caption{Additional qualitative results on the DTU~\cite{Jensen2014LargeSM} dataset for UNISURF~\cite{Oechsle2021UNISURFUN} method.}
    \label{fig:dtu_qual_appendix_unisurf}
\end{figure*}

\begin{figure*}
    \centering    
    \setlength{\wid}{0.20\textwidth}
    \setlength{\mrg}{-0.45cm}
    \setlength{\mrgv}{-0.05cm}
    \begin{tabular}{c cc}
        \vspace{\mrgv}
        \includegraphics[width=\wid]{figures/dtu_qual/gt/37.jpg} &
        \hspace{\mrg}
        \includegraphics[width=\wid]{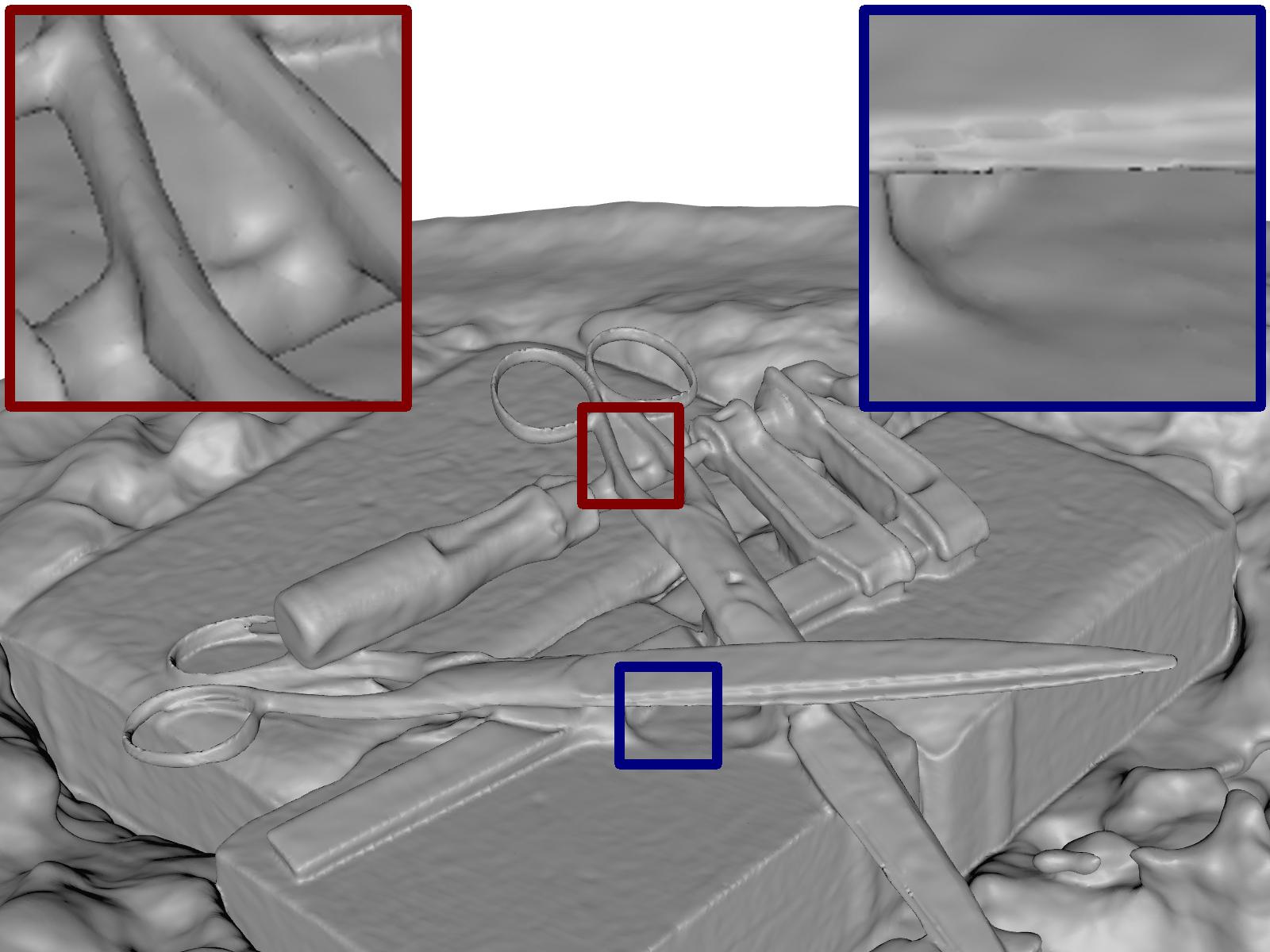} &
        \hspace{\mrg}
        \includegraphics[width=\wid]{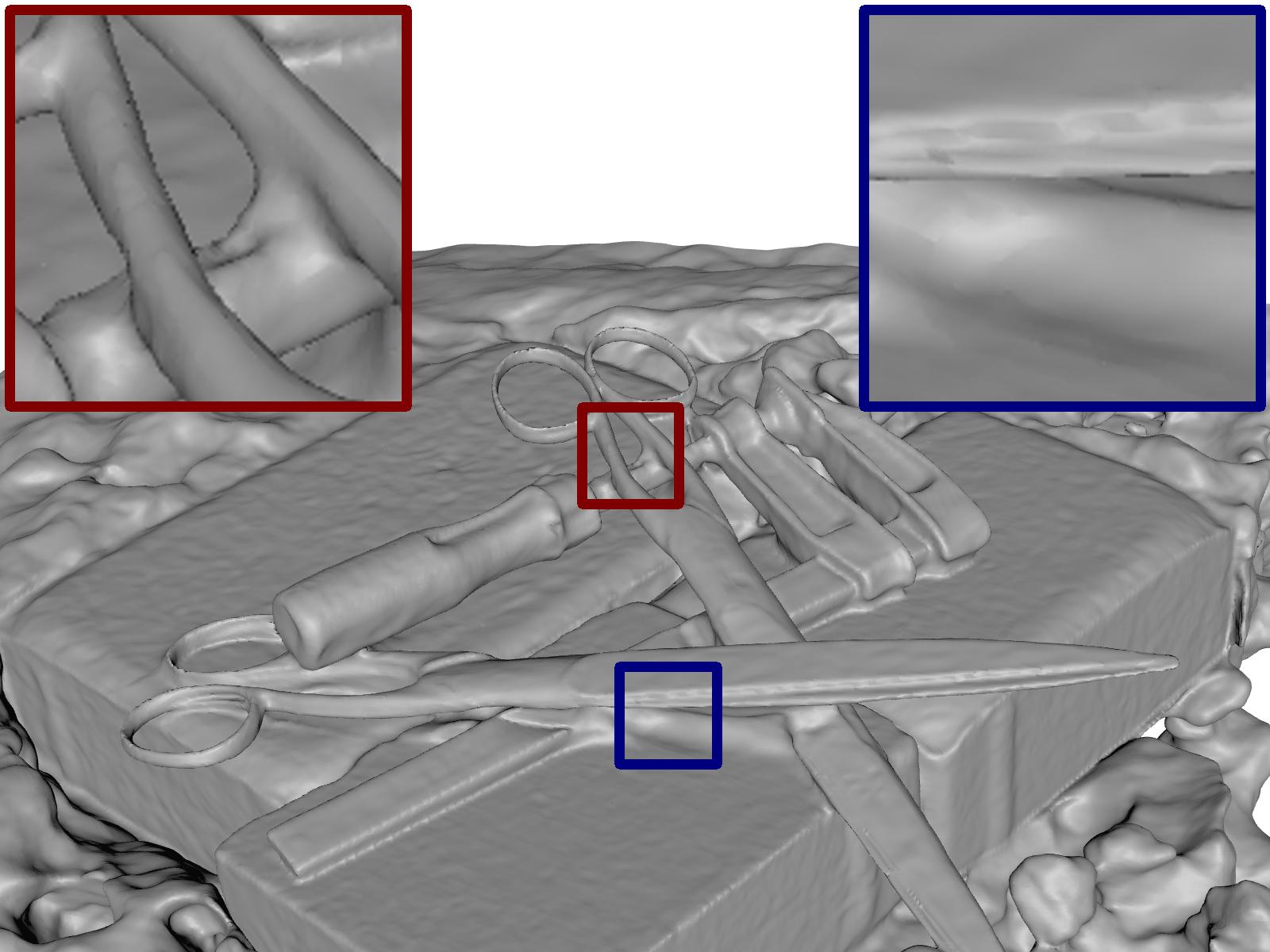}
        \\ 
        \vspace{\mrgv}
        \includegraphics[width=\wid]{figures/dtu_qual/gt/55.jpg} &
        \hspace{\mrg}
        \includegraphics[width=\wid]{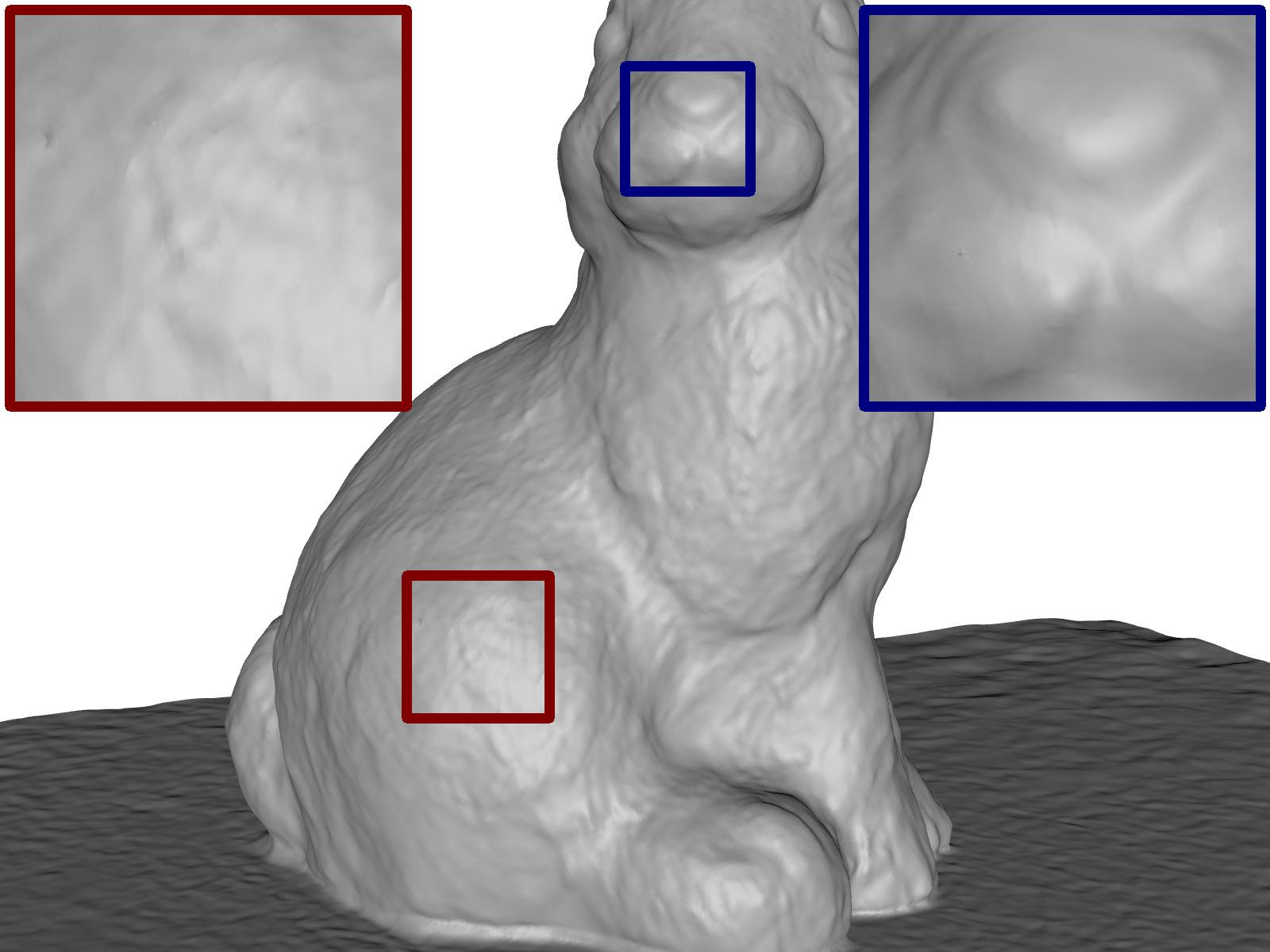} &
        \hspace{\mrg}
        \includegraphics[width=\wid]{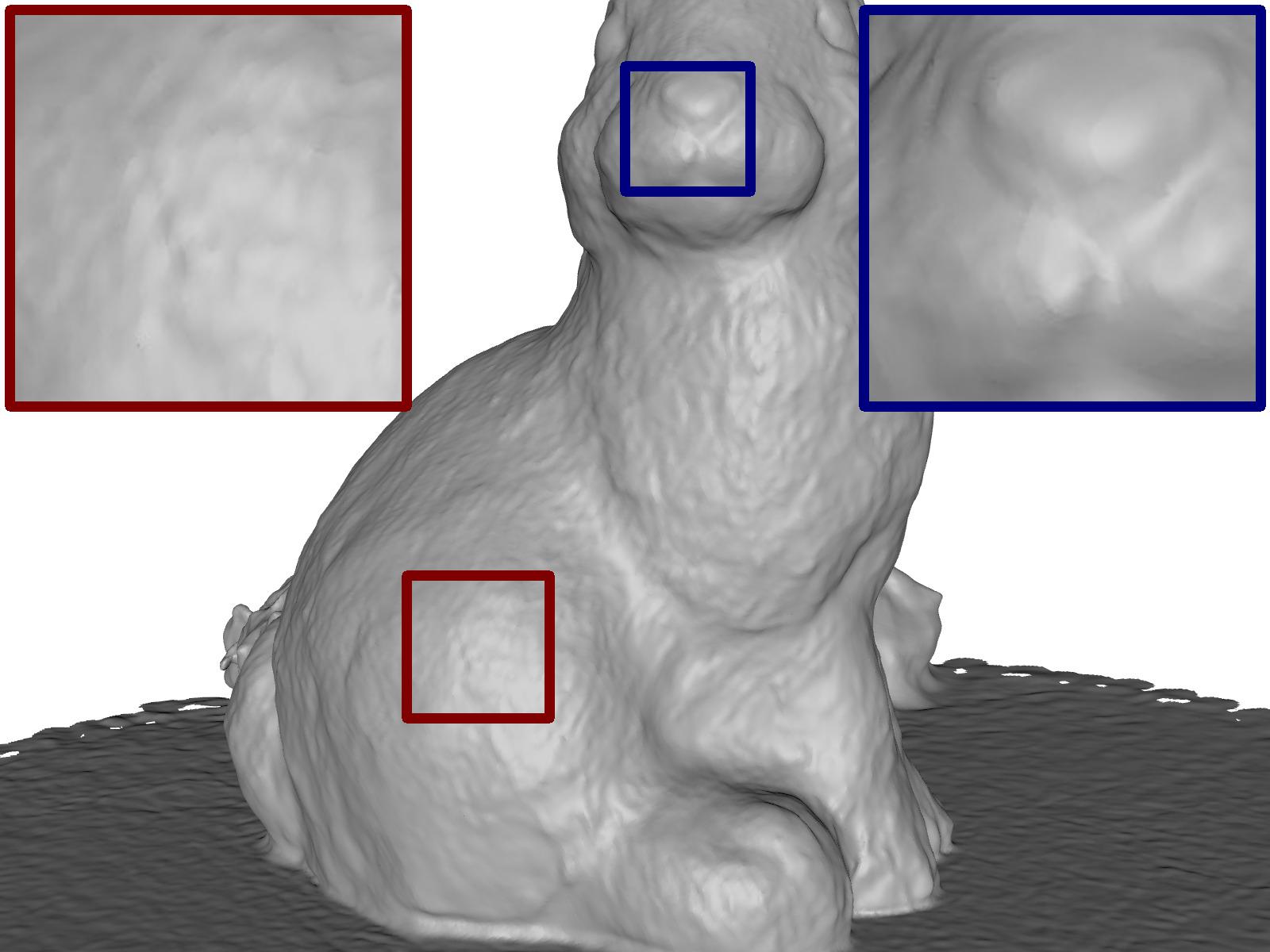}
        \\ 
        \vspace{\mrgv}
        \includegraphics[width=\wid]{figures/dtu_qual/gt/63.jpg} &
        \hspace{\mrg}
        \includegraphics[width=\wid]{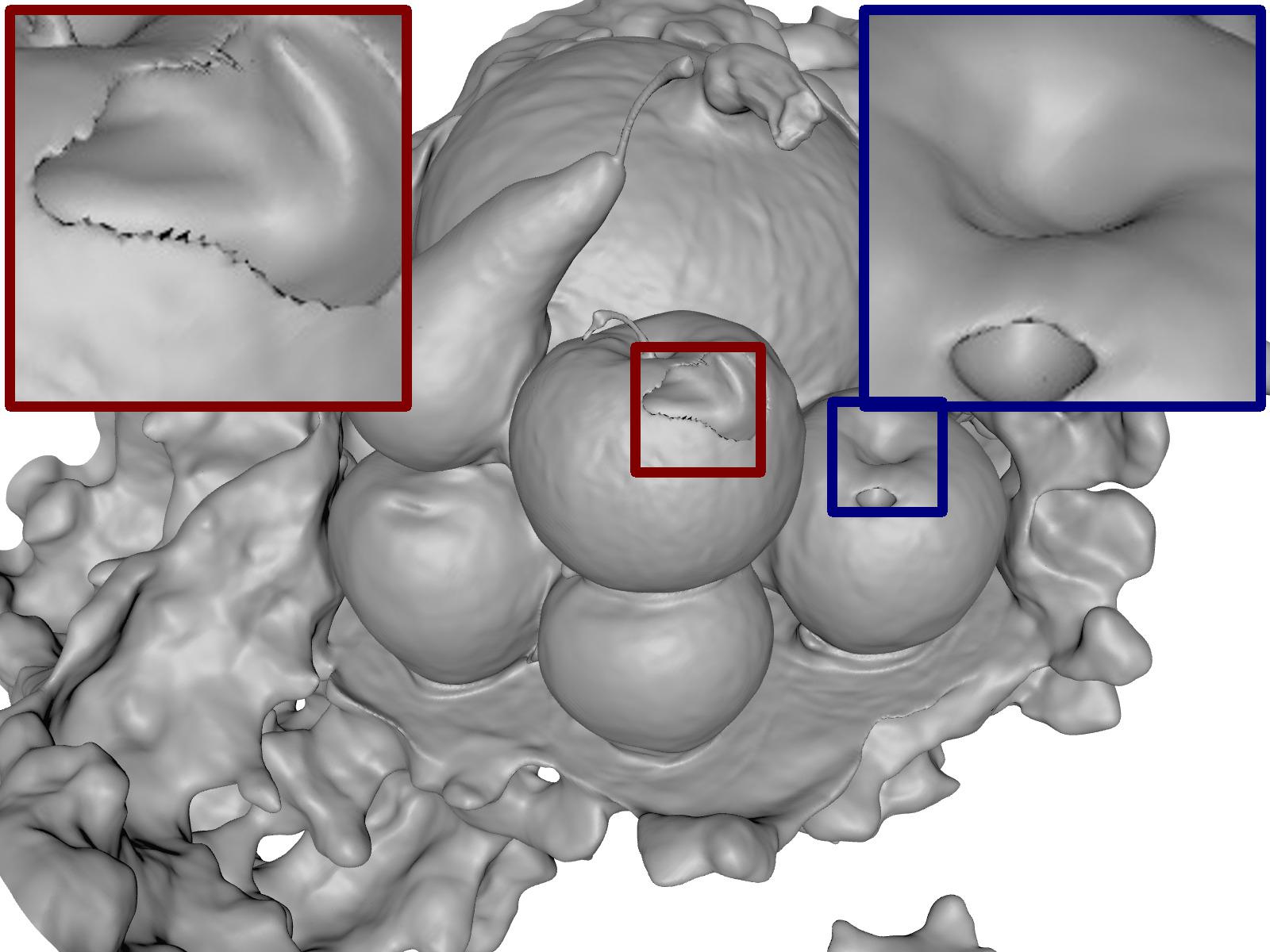} &
        \hspace{\mrg}
        \includegraphics[width=\wid]{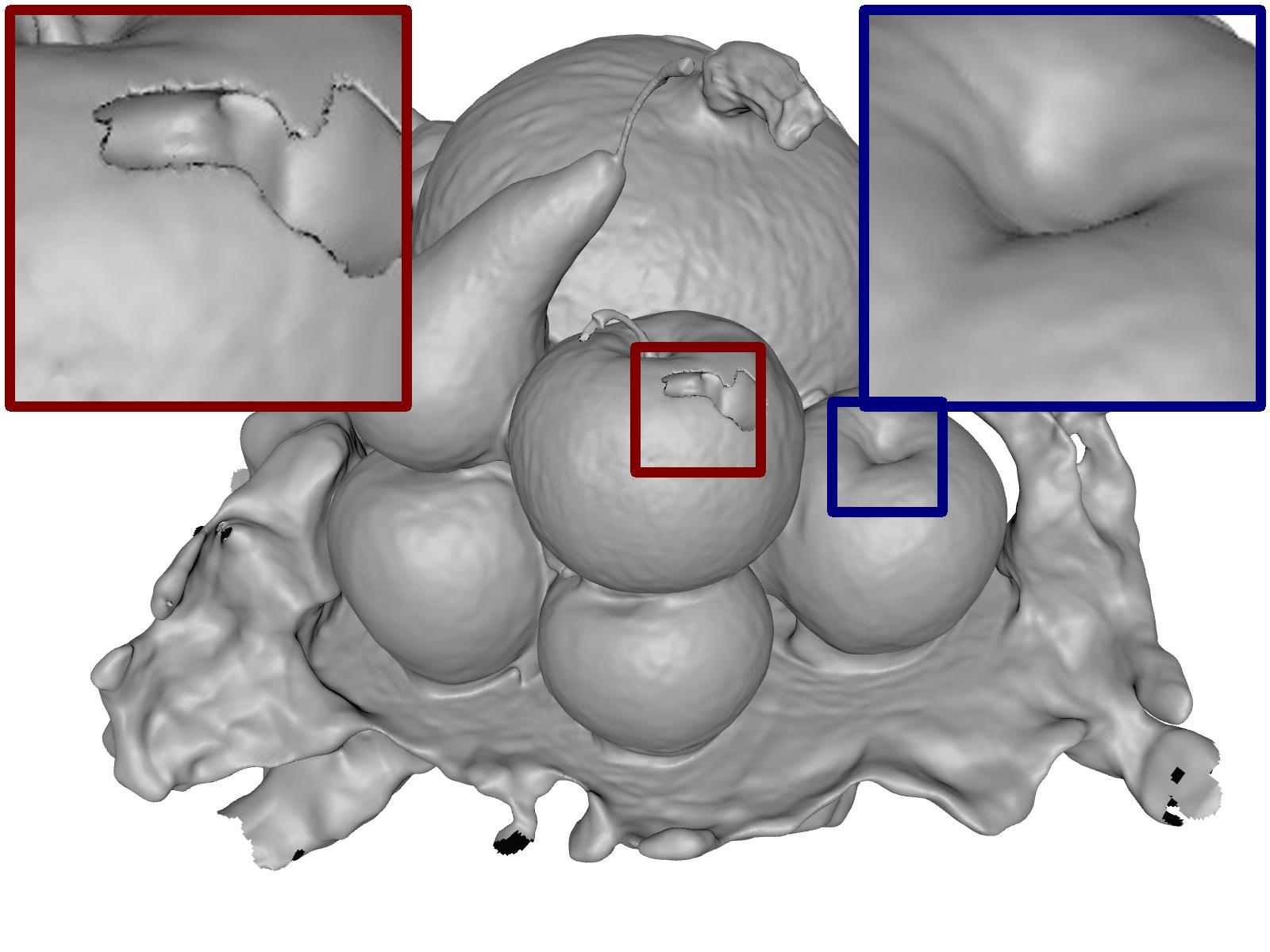}
        \\ 
        \vspace{\mrgv}
        \includegraphics[width=\wid]{figures/dtu_qual/gt/65.jpg} &
        \hspace{\mrg}
        \includegraphics[width=\wid]{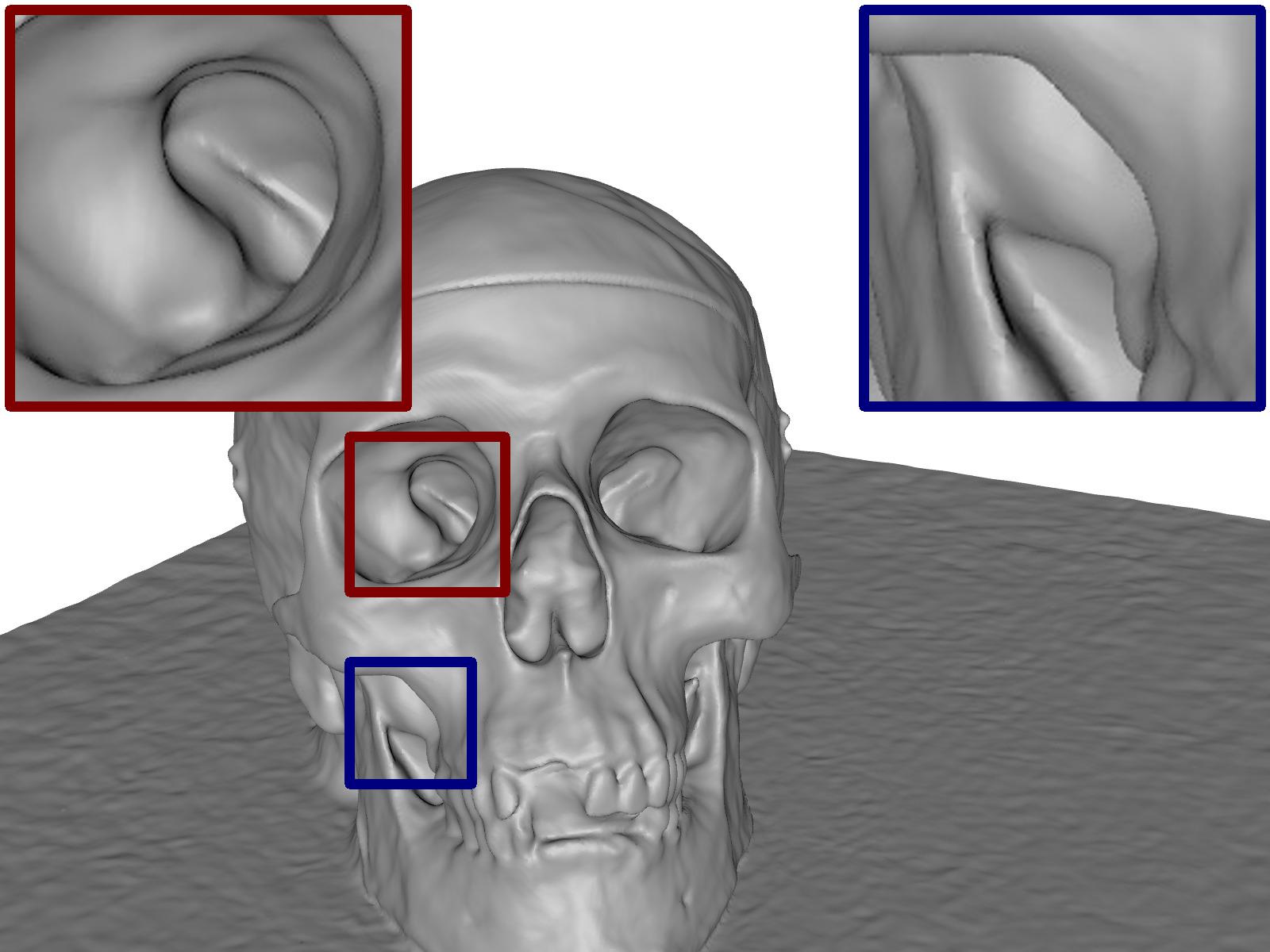} &
        \hspace{\mrg}
        \includegraphics[width=\wid]{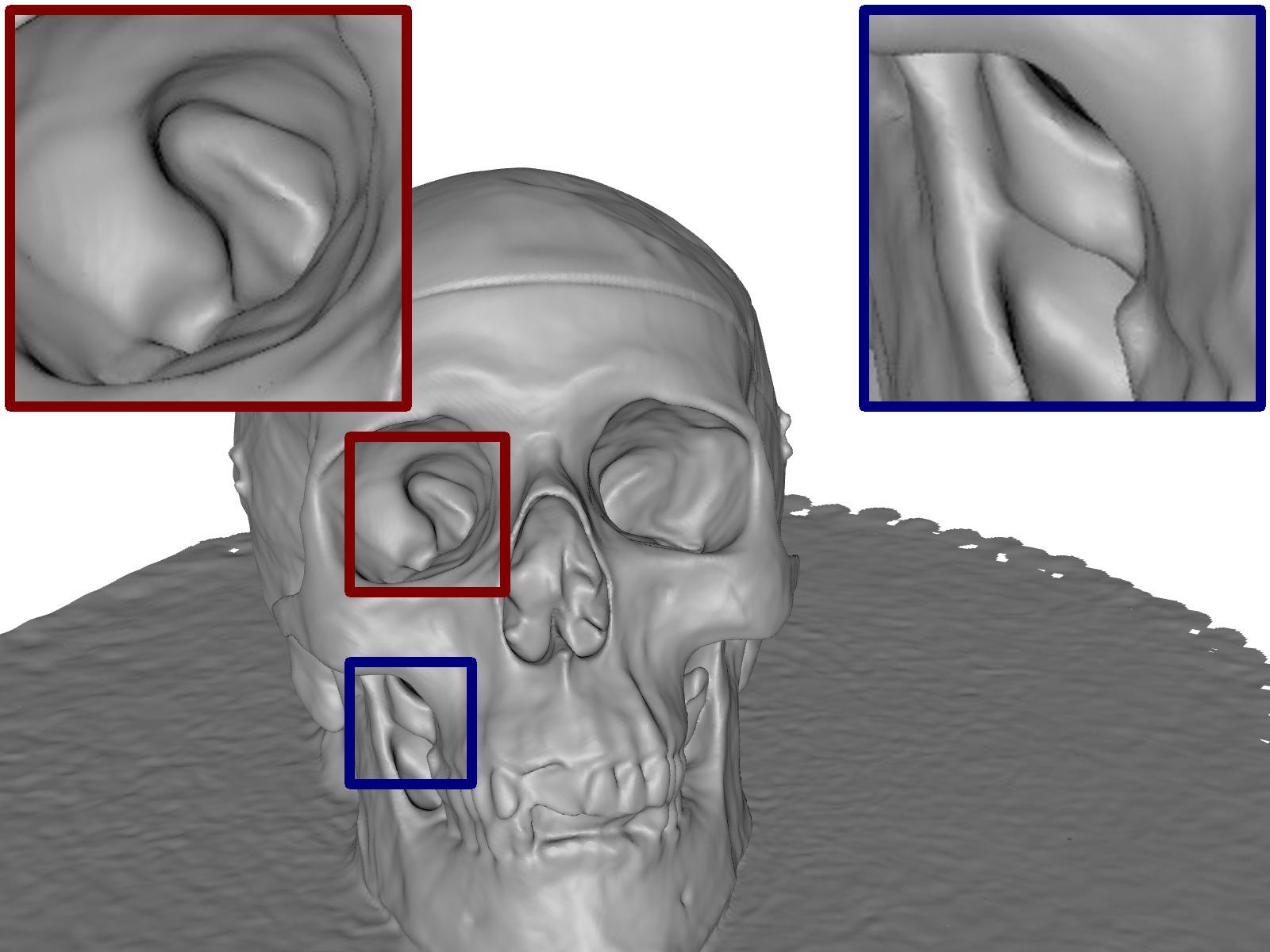}
        \\ 
        \vspace{\mrgv}
        \includegraphics[width=\wid]{figures/dtu_qual/gt/83.jpg} &
        \hspace{\mrg}
        \includegraphics[width=\wid]{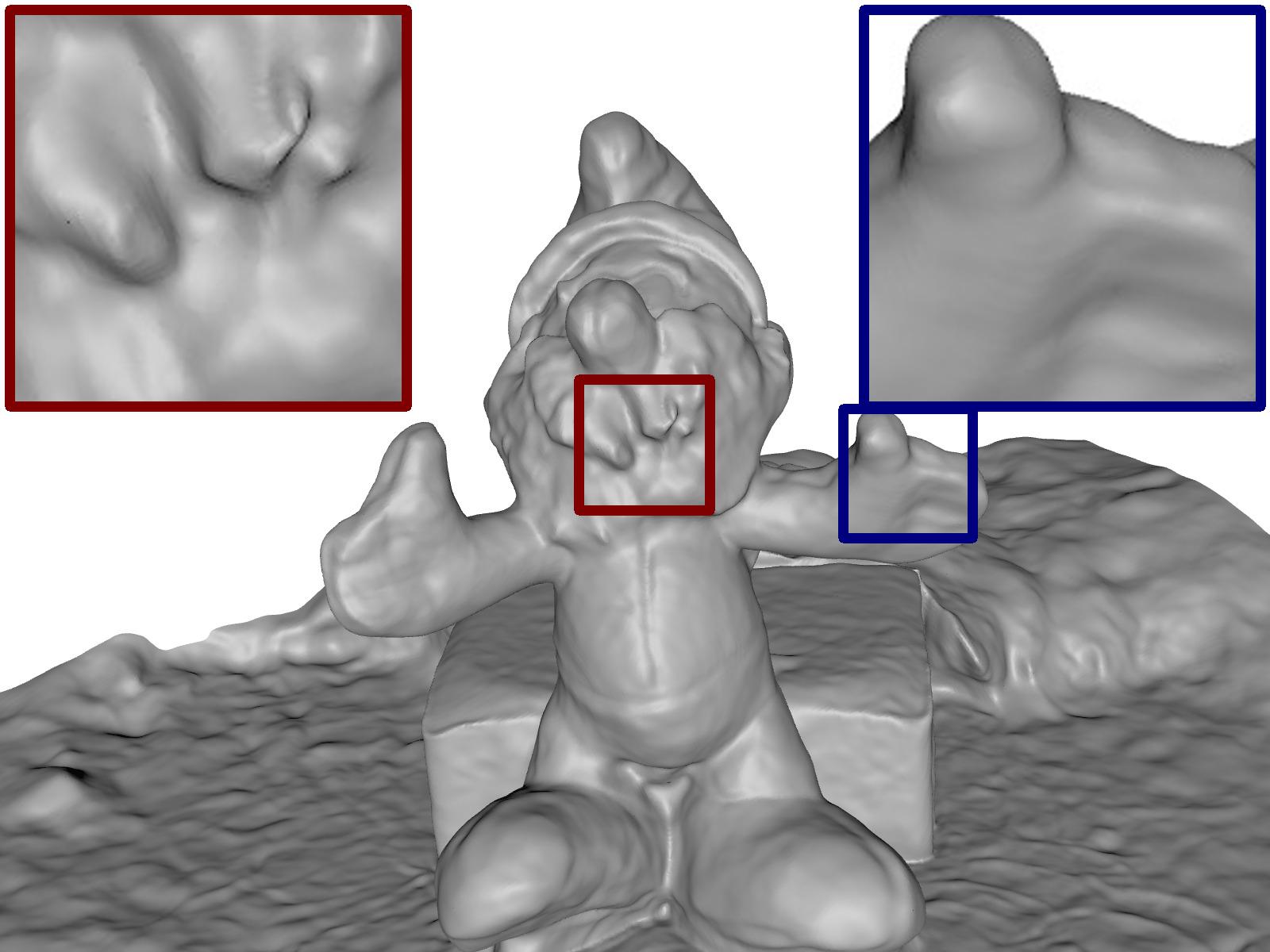} &
        \hspace{\mrg}
        \includegraphics[width=\wid]{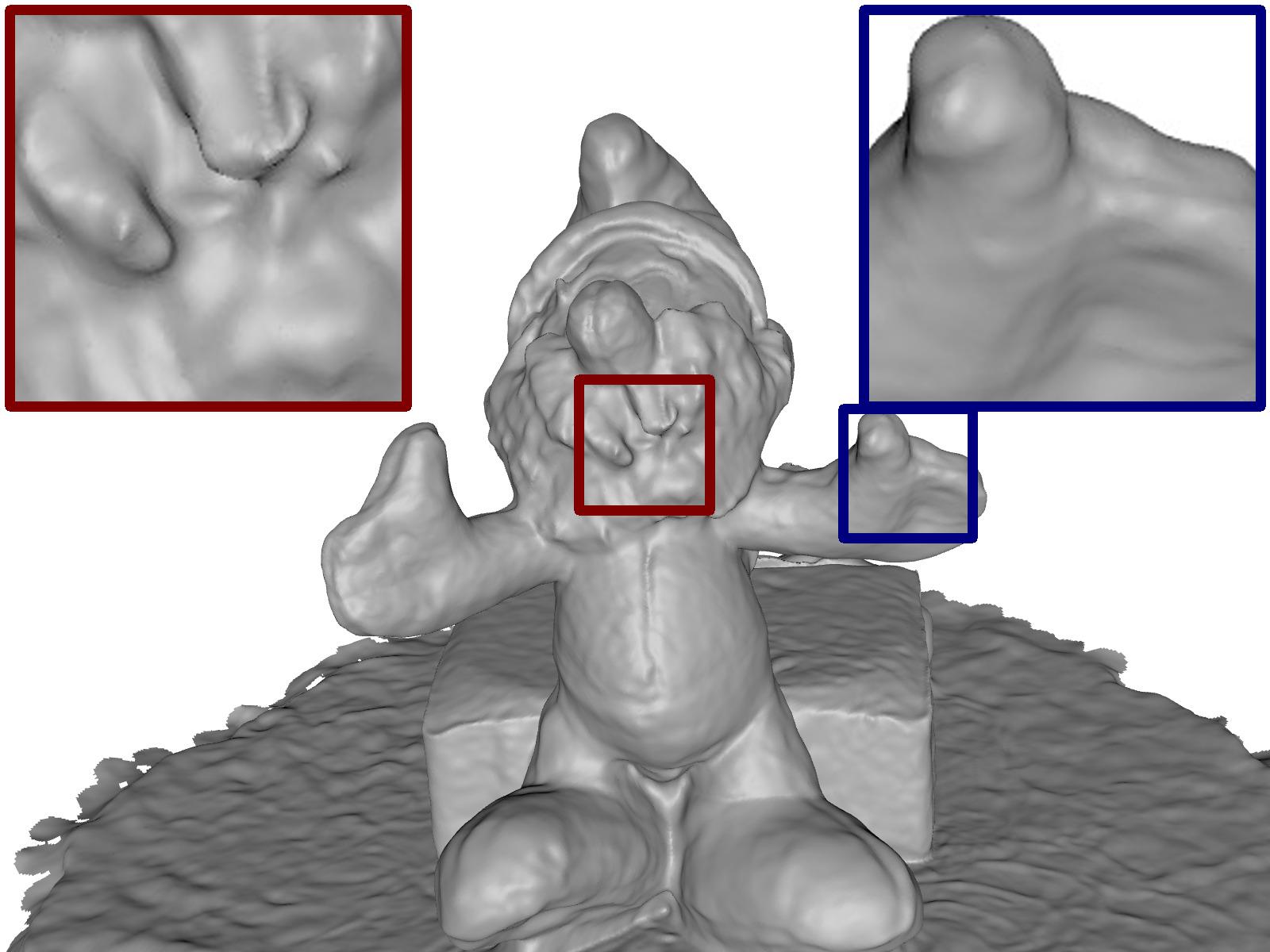}
        \\
        \vspace{\mrgv}
        \includegraphics[width=\wid]{figures/dtu_qual/gt/110.jpg} &
        \hspace{\mrg}
        \includegraphics[width=\wid]{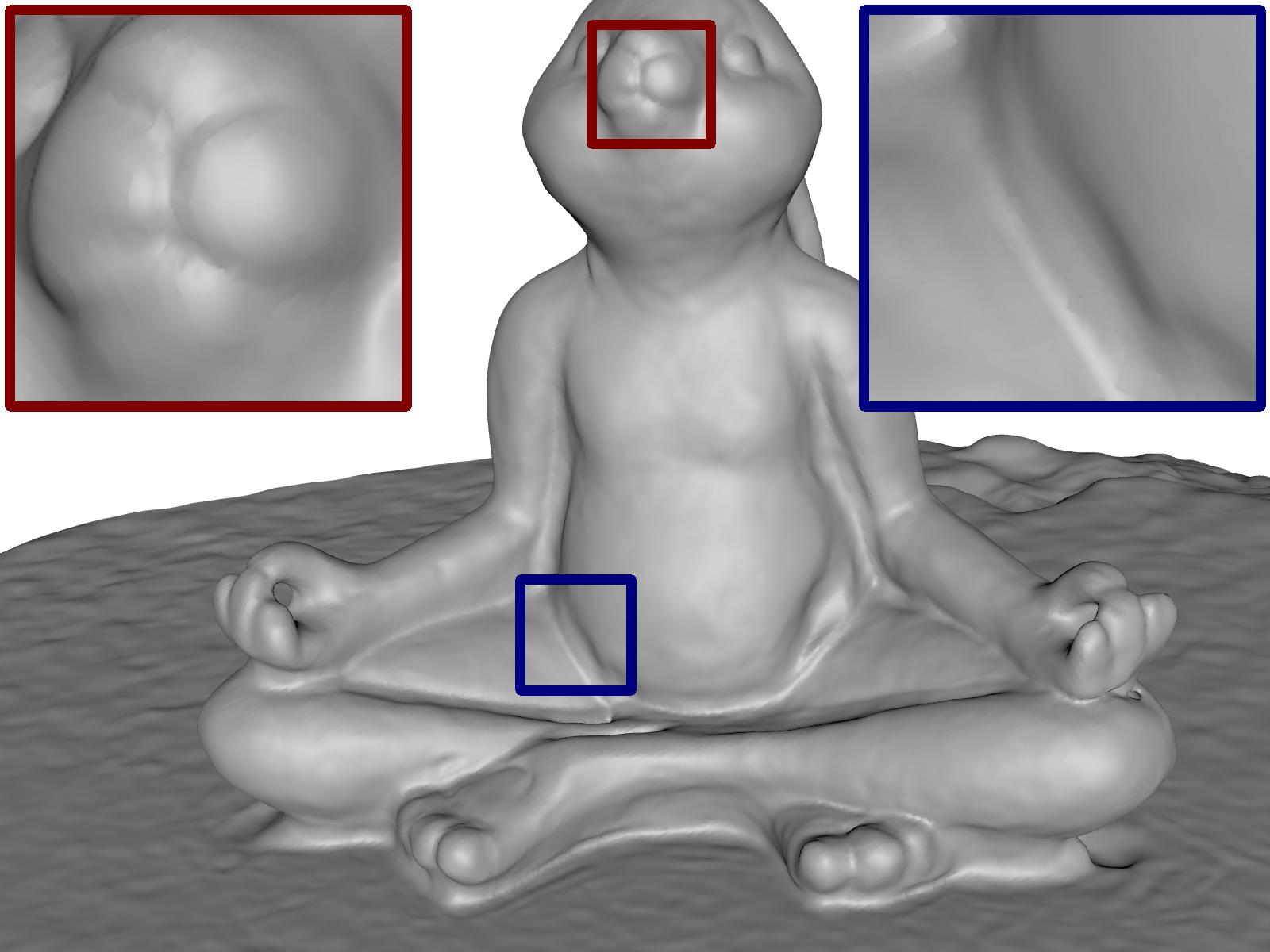} &
        \hspace{\mrg}
        \includegraphics[width=\wid]{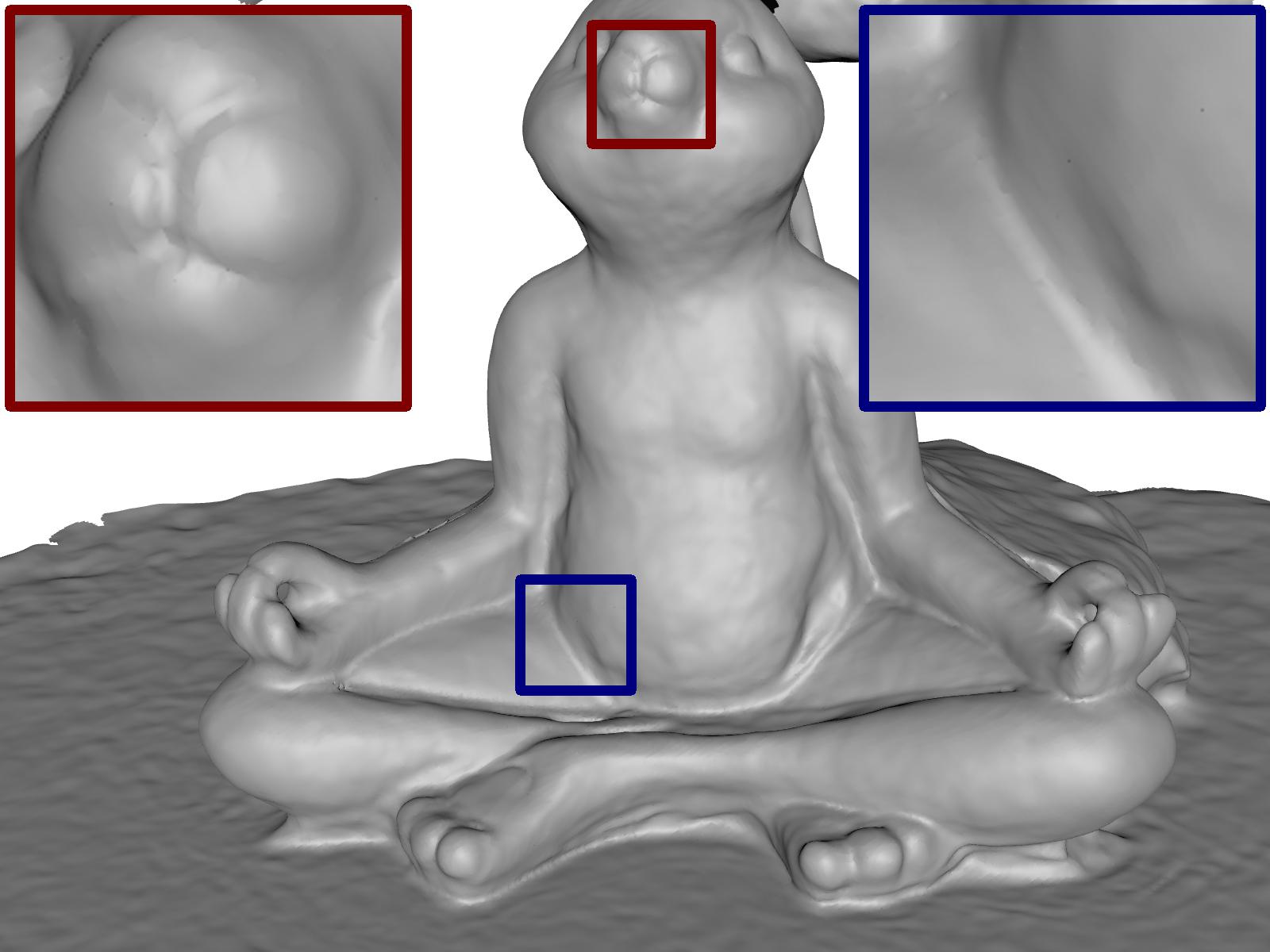}
        \\
        \vspace{\mrgv}
        \includegraphics[width=\wid]{figures/dtu_qual/gt/118.jpg} &
        \hspace{\mrg}
        \includegraphics[width=\wid]{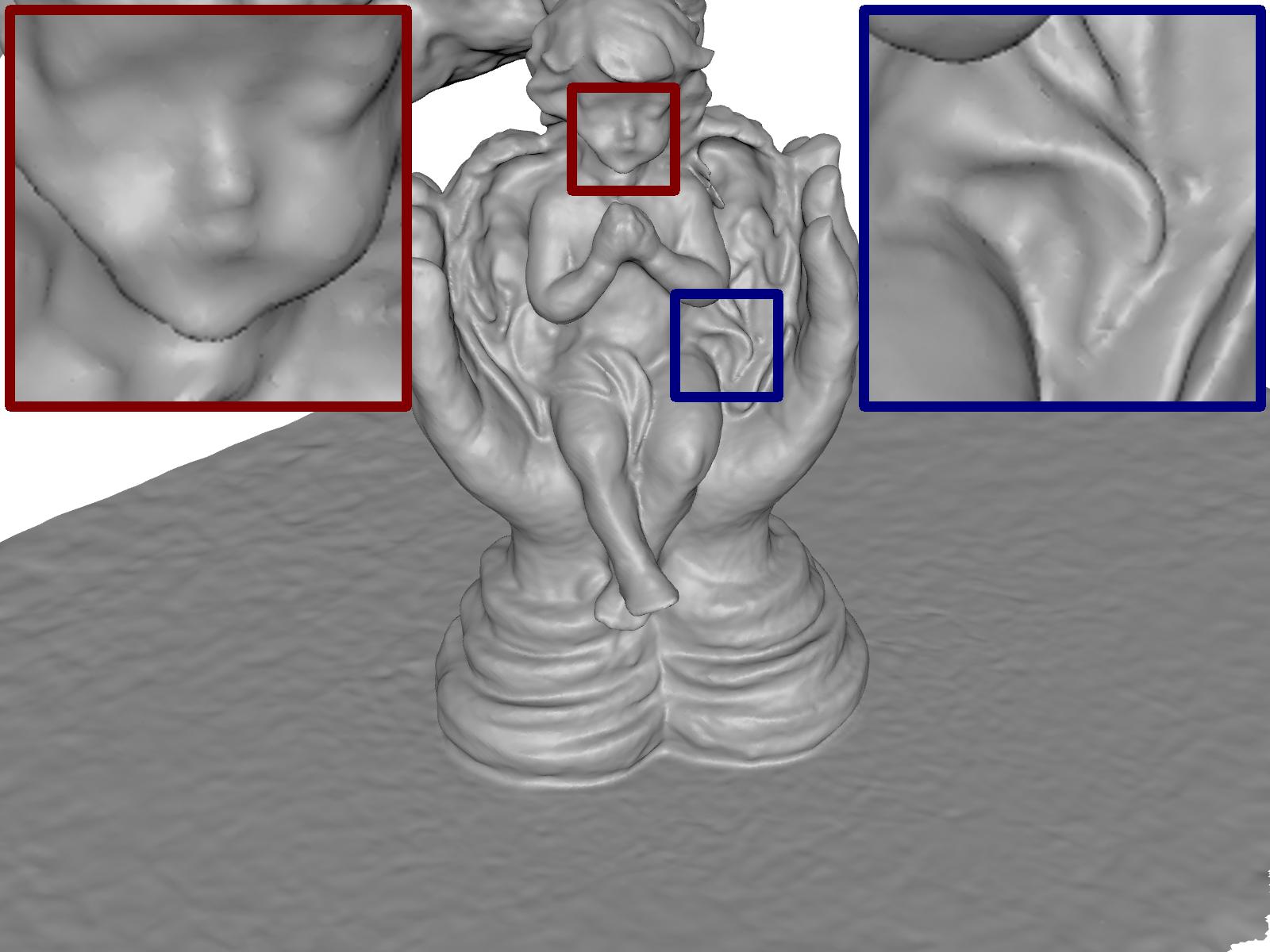} &
        \hspace{\mrg}
        \includegraphics[width=\wid]{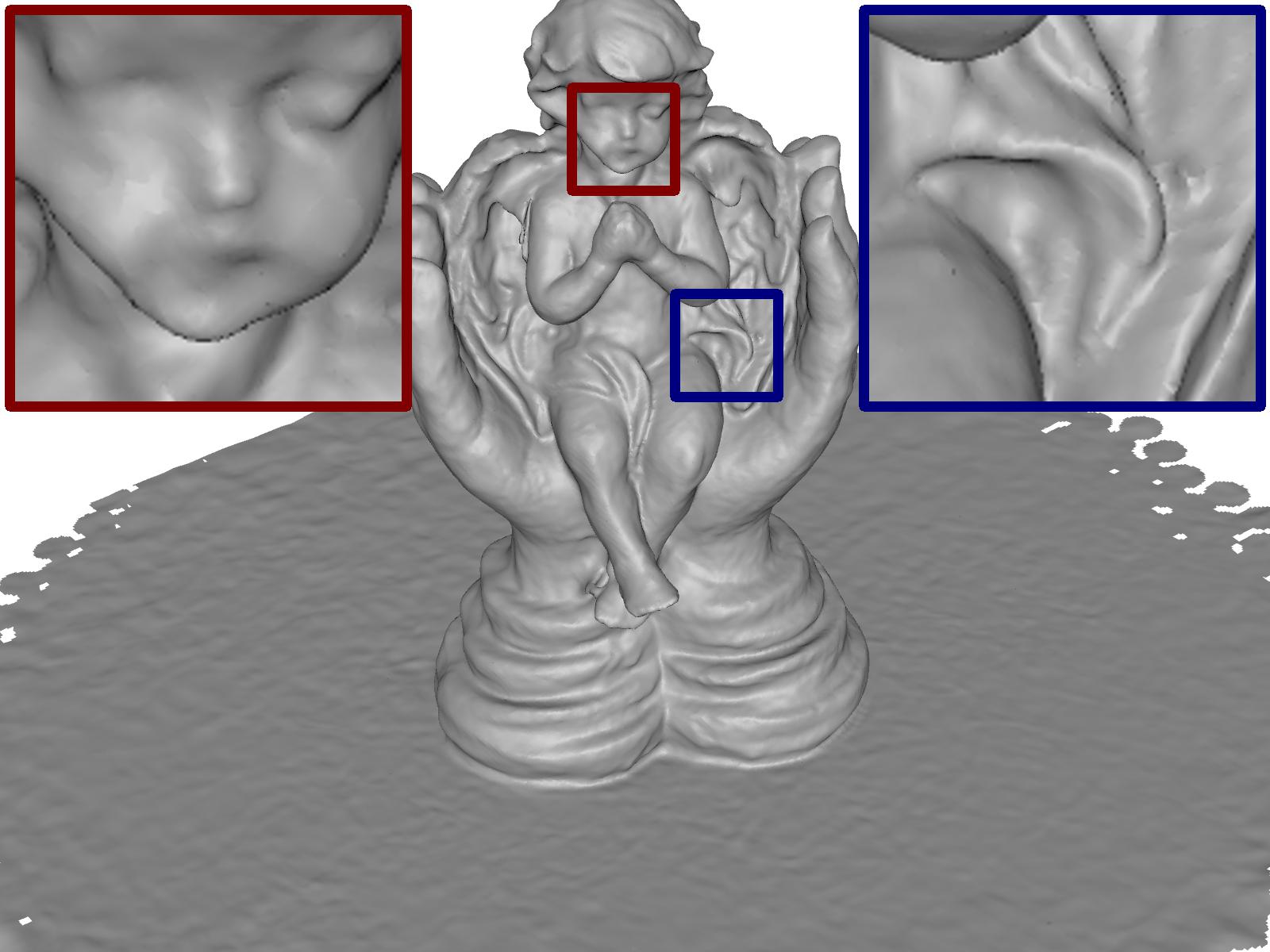}
        \\
        \vspace{\mrgv}
        \includegraphics[width=\wid]{figures/dtu_qual/gt/122.jpg} &
        \hspace{\mrg}
        \includegraphics[width=\wid]{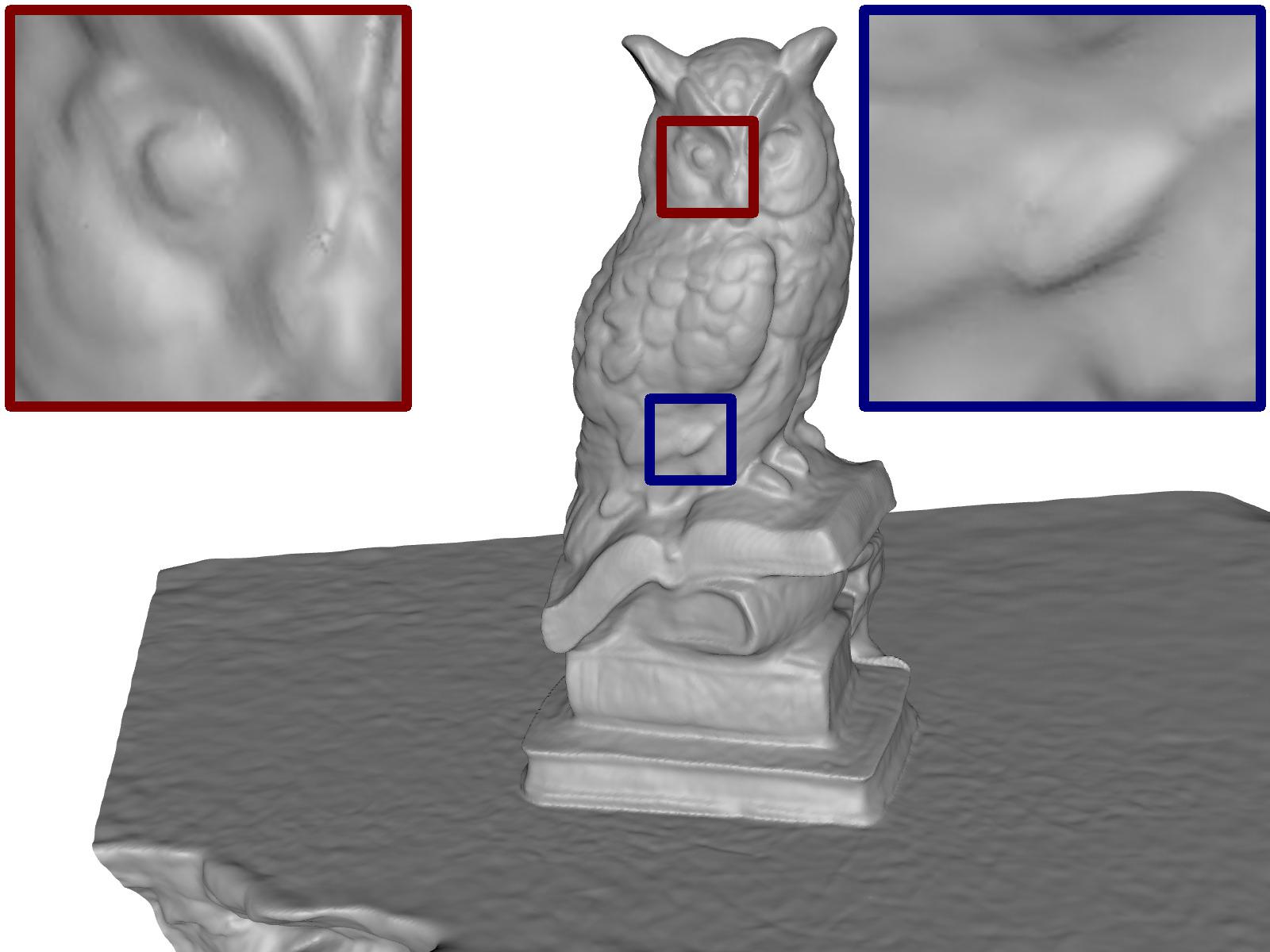} &
        \hspace{\mrg}
        \includegraphics[width=\wid]{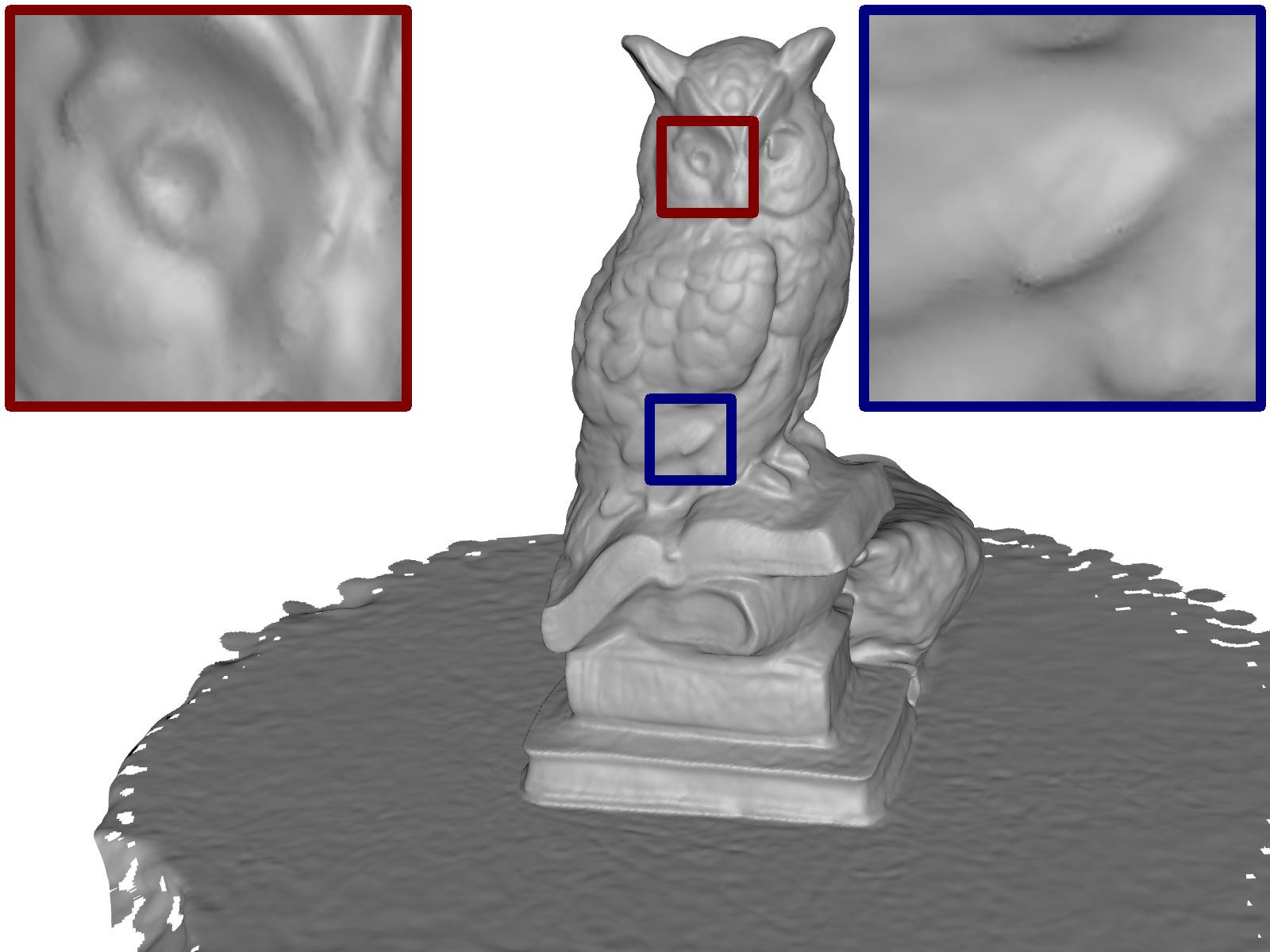}
        \\
        \textbf{Source} & \hspace{\mrg}
        \textbf{NeuralWarp} & \hspace{\mrg}
        \textbf{NeuralWarp (ours)}
    \end{tabular}
    \caption{Additional qualitative results on the DTU~\cite{Jensen2014LargeSM} dataset for NeuralWarp~\cite{darmon2022improving} method.}
    \label{fig:dtu_qual_appendix_nwarp}
\end{figure*}

We show additional quantitative (Figure~\ref{fig:dtu_quant_appendix}) and qualitative (Figures~\ref{fig:dtu_qual_appendix_neus}-\ref{fig:dtu_qual_appendix_nwarp}) evaluation of our method on the DTU~\cite{Jensen2014LargeSM} dataset. In Figure~\ref{fig:dtu_quant_appendix}, we investigate more in-depth the comparative performance between our method and NeuralWarp. To this end, we randomly picked five scenes and trained four models with different random seeds. We add these data points to the ones reported in the main paper and present them in this figure. We conclude that, on average, our method performs better than raw NeuralWarp, when the stochasticity of the base method is accounted for. Our method on average achieves a mean Chamfer distance of $0.73$ against the NeuralWarp's $0.76$. We also show additional qualitative results for eight more scenes of DTU dataset in Figures~\ref{fig:dtu_qual_appendix_neus}-\ref{fig:dtu_qual_appendix_nwarp}.

\paragraph{Realistic Synthetic 360 dataset~\cite{Mildenhall2020NeRFRS} experiments.}
This dataset has more complex geometries (with multiple objects per scene and fine details) compared with the DTU dataset, making it more challenging for the networks to accurately reconstruct them. We found that the default Eikonal regularization weight of $0.1$ discourages the NeuS-based models without mask supervision from reconstructing disconnected objects (such as the chair in the drums scene). Therefore, the weight was divided by 10 for all experiments on this dataset, including the results in the main paper and the ablations.
Before evaluating the Chamfer distance of the reconstructed meshes we filter the geometries using the ground truth segmentation masks dilated with a radius of 12 as in the DTU evaluation.

\begin{figure*}
    \centering    
    \setlength{\wid}{0.23\textwidth}
    \setlength{\mrg}{-0.45cm}
    \setlength{\mrgv}{-0.05cm}
    \begin{tabular}{c cc}
        \vspace{\mrgv}
        \includegraphics[width=\wid]{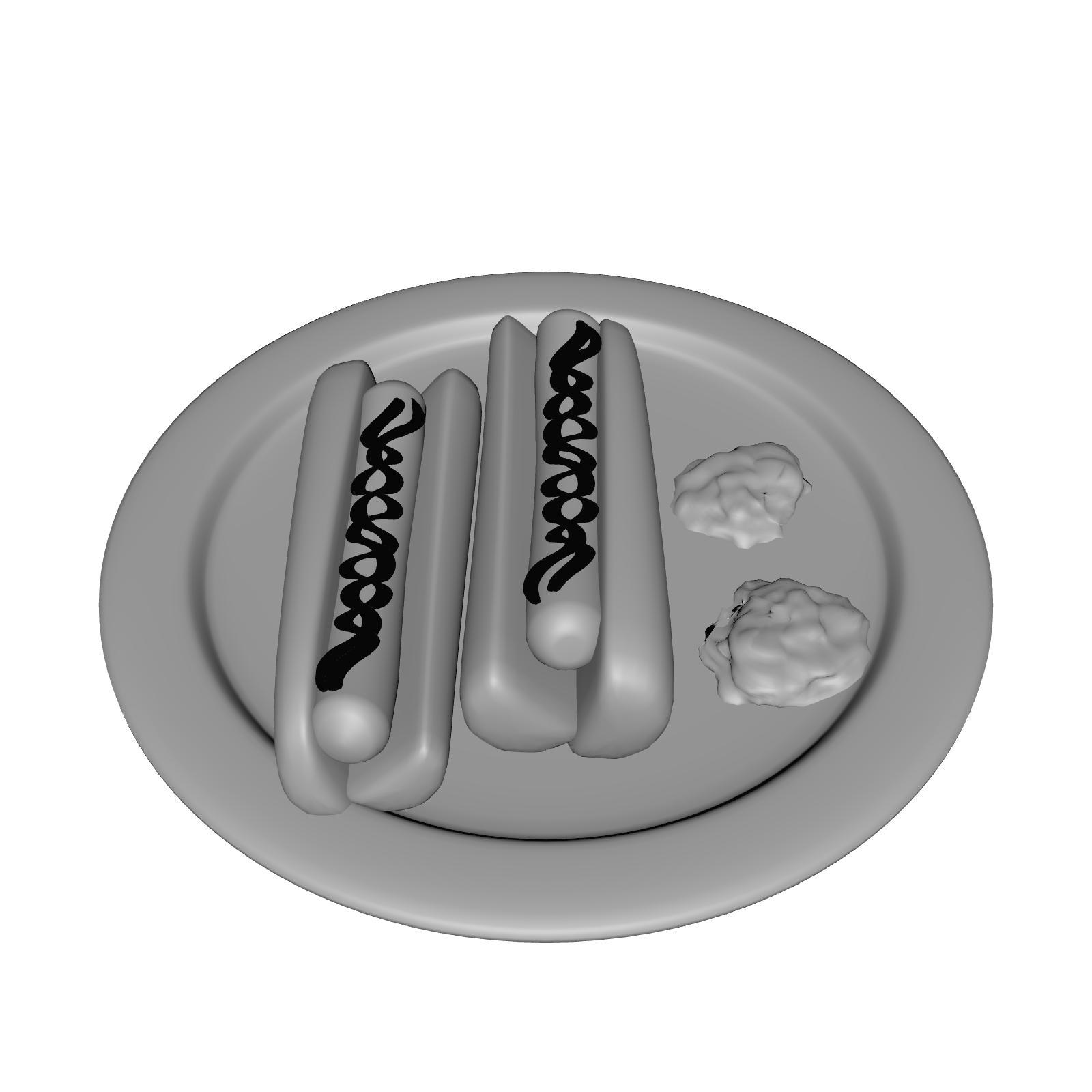} &
        \hspace{\mrg}
        \includegraphics[width=\wid]{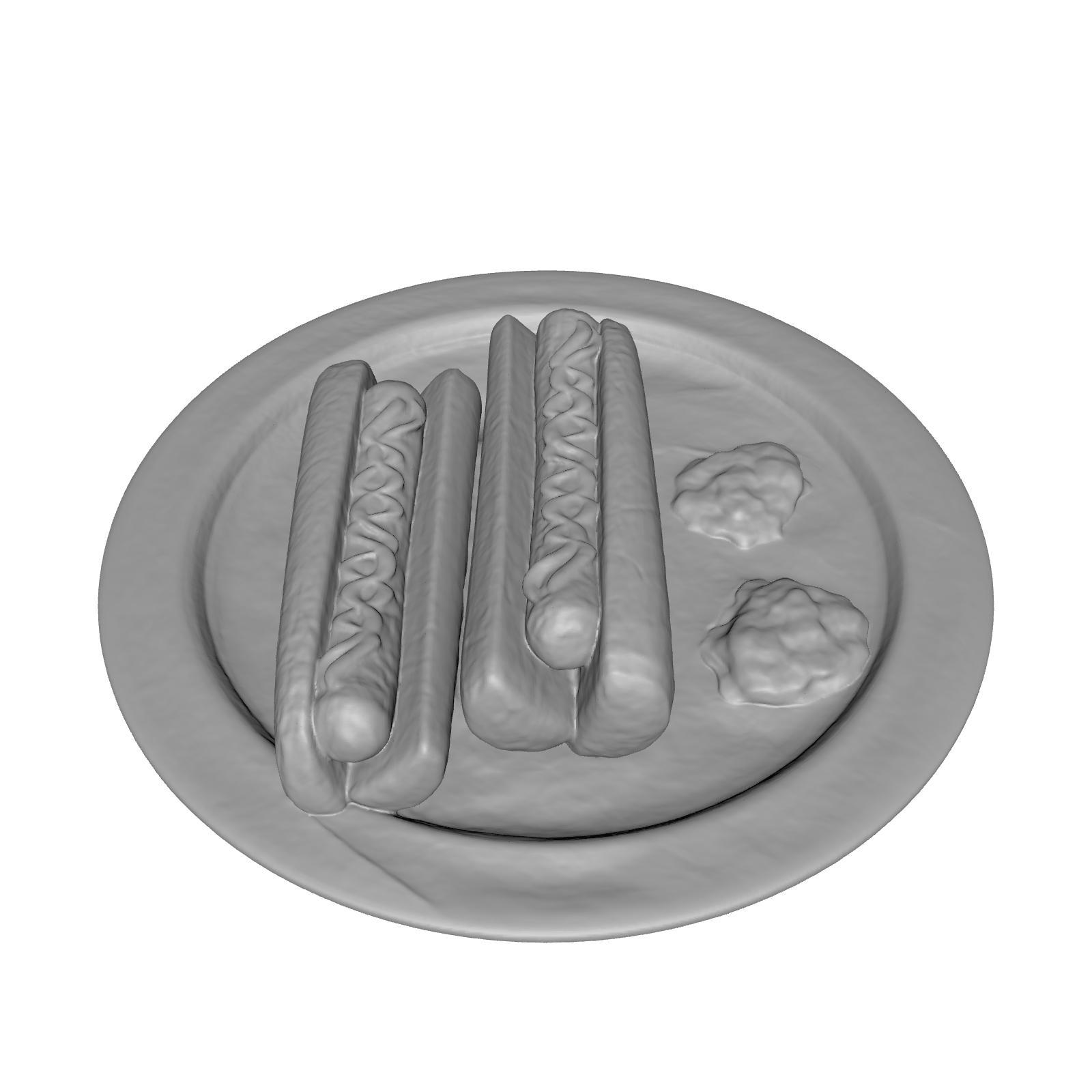} &
        \hspace{\mrg}
        \includegraphics[width=\wid]{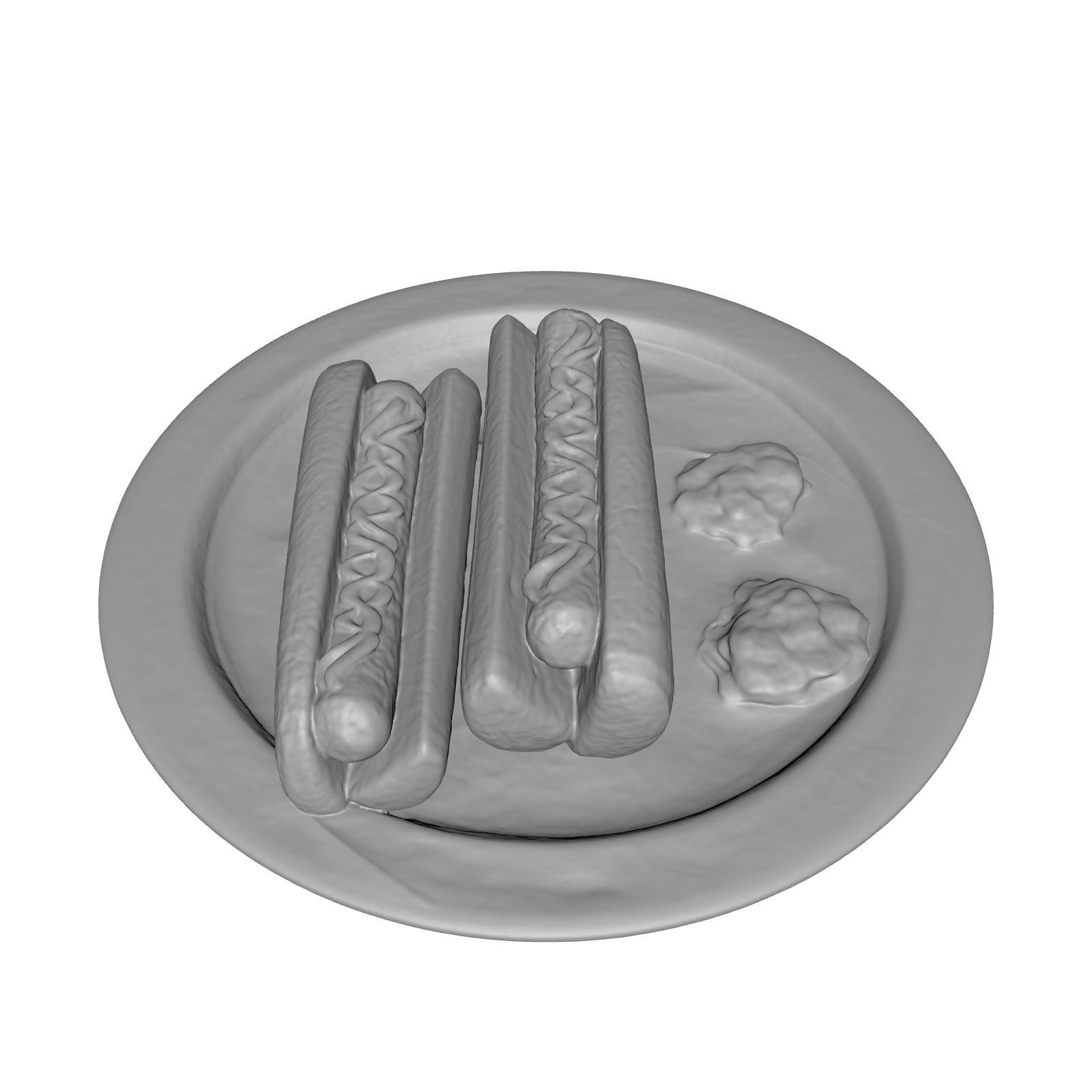}
        \\ 
        \vspace{\mrgv}
        \includegraphics[width=\wid]{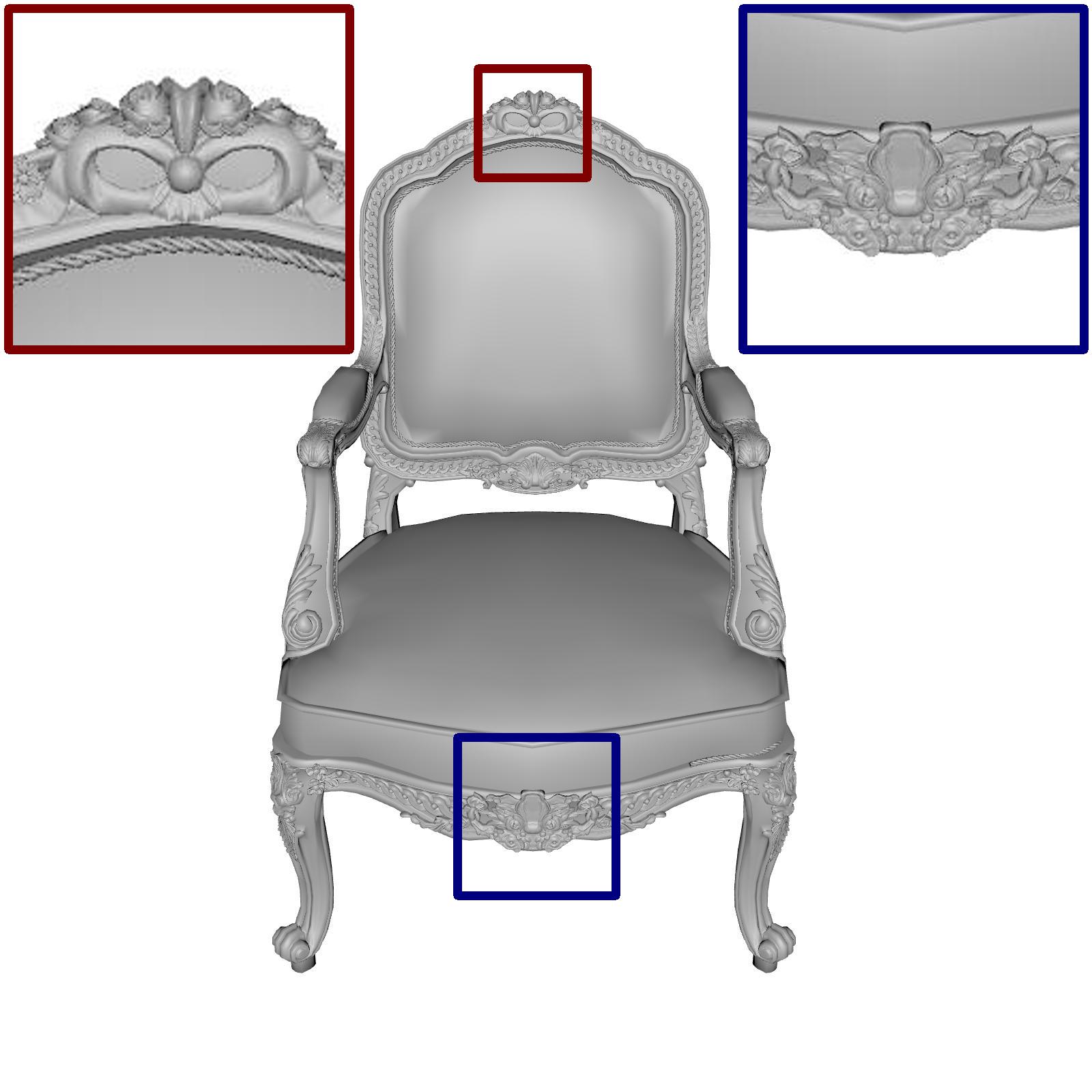} &
        \hspace{\mrg}
        \includegraphics[width=\wid]{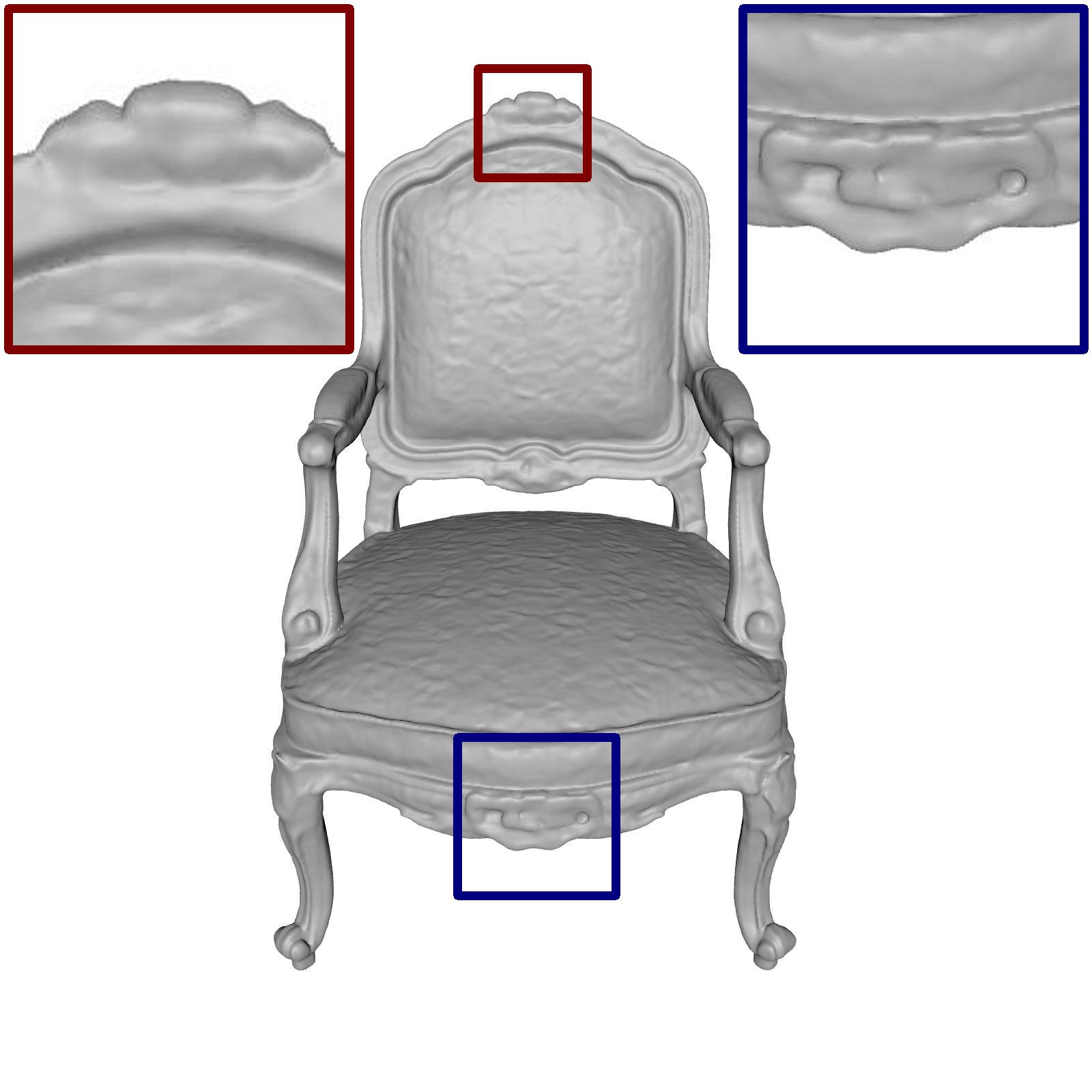} &
        \hspace{\mrg}
        \includegraphics[width=\wid]{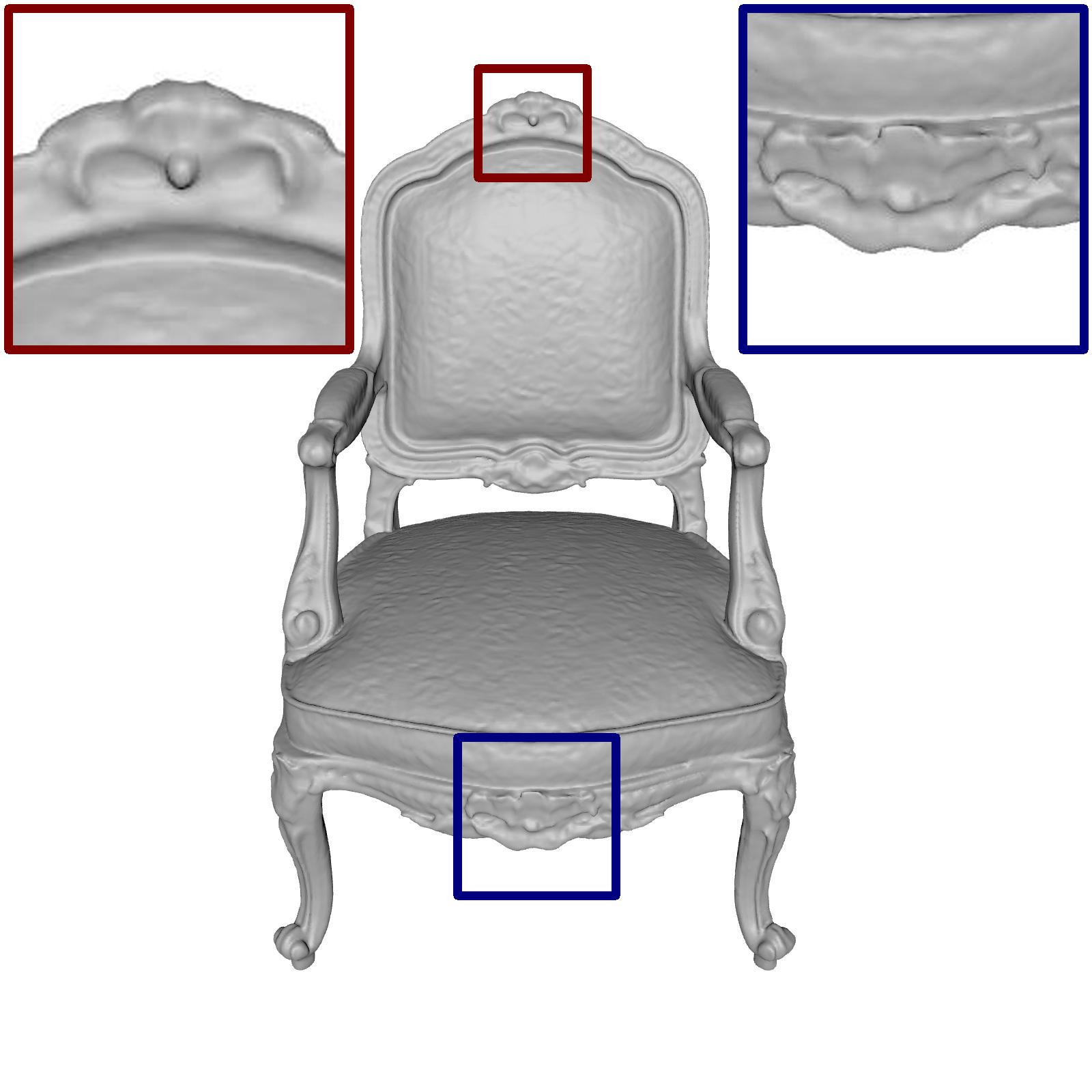}
        \\ 
        \vspace{\mrgv}
        \includegraphics[width=\wid]{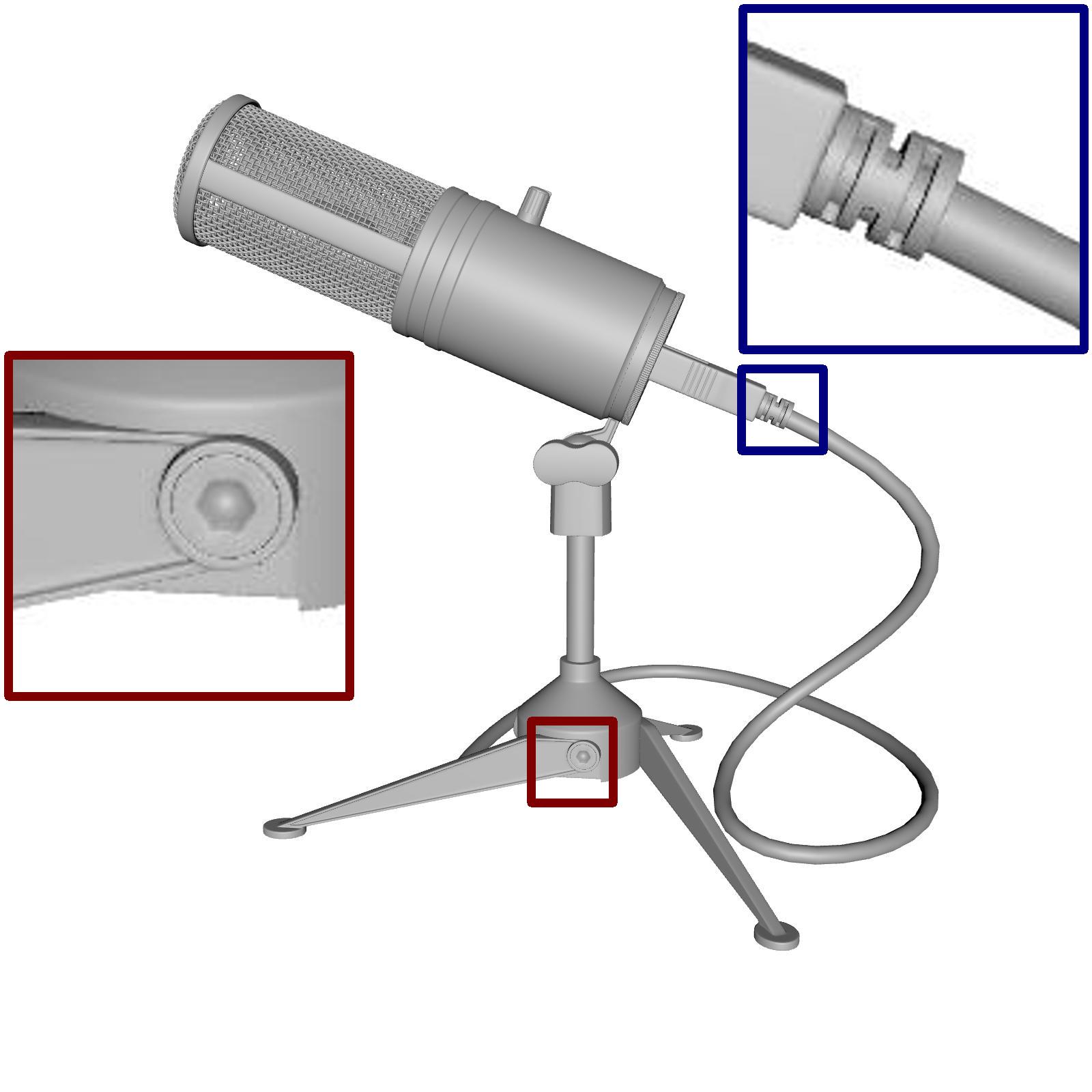} &
        \hspace{\mrg}
        \includegraphics[width=\wid]{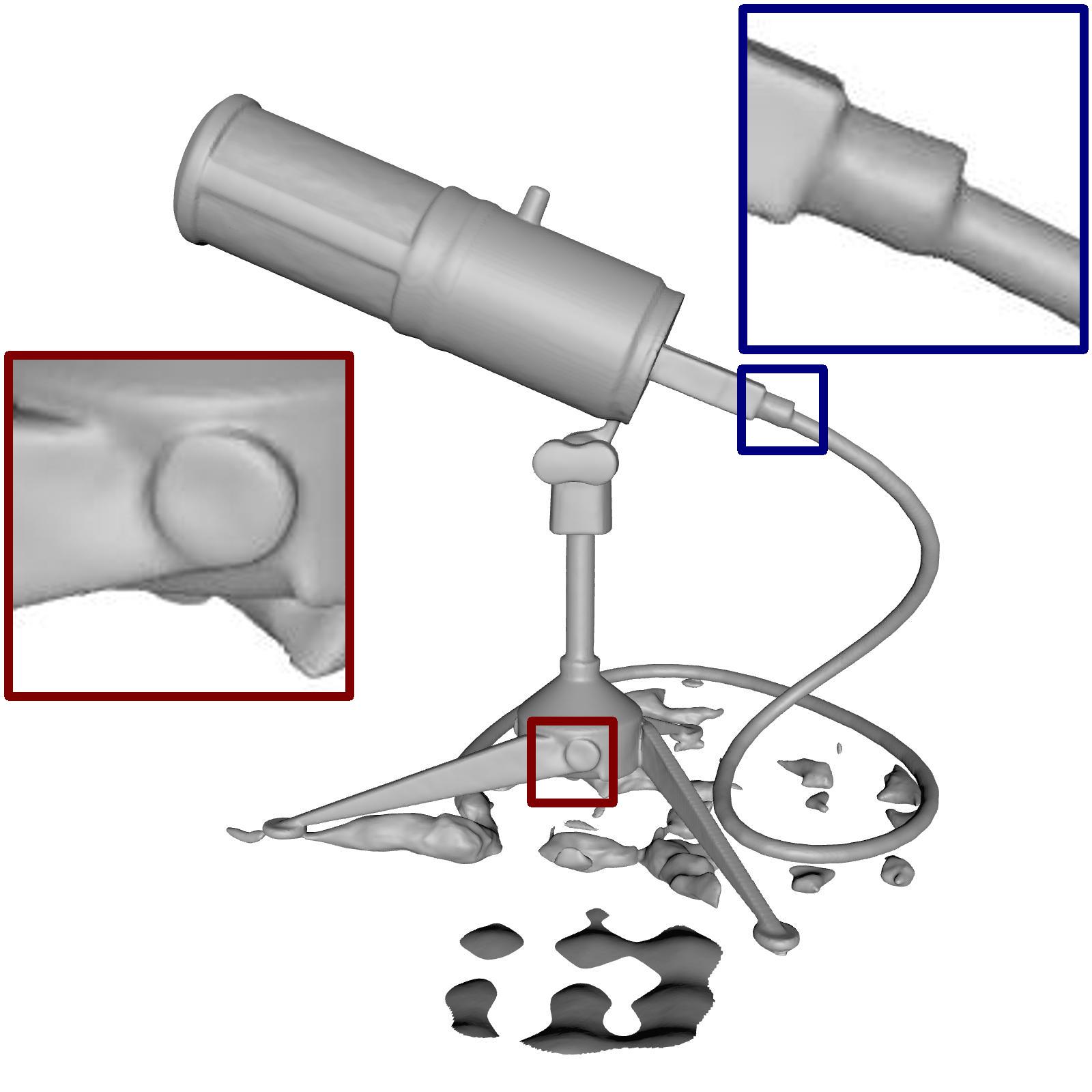} &
        \hspace{\mrg}
        \includegraphics[width=\wid]{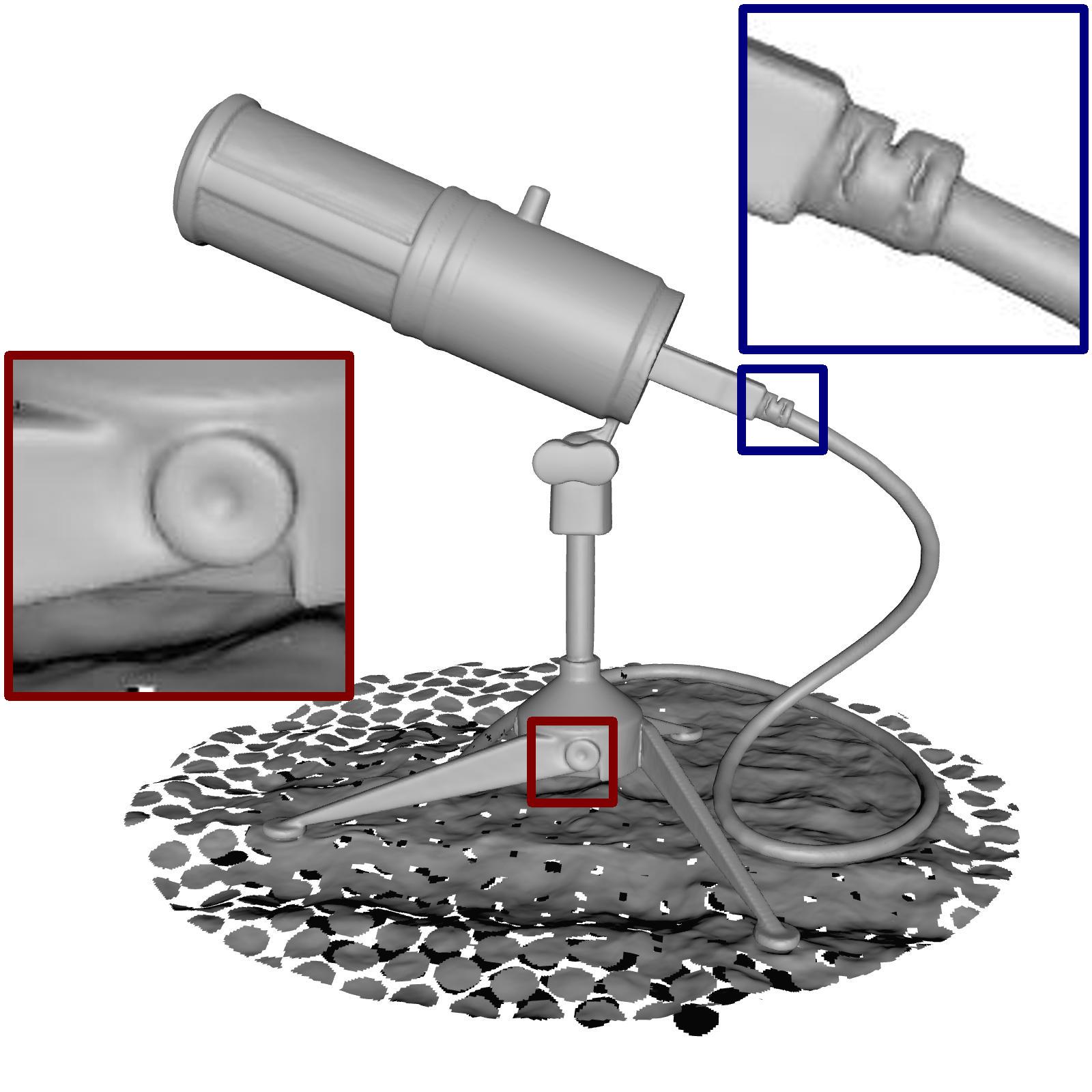}
        \\ 
        \vspace{\mrgv}
        \includegraphics[width=\wid]{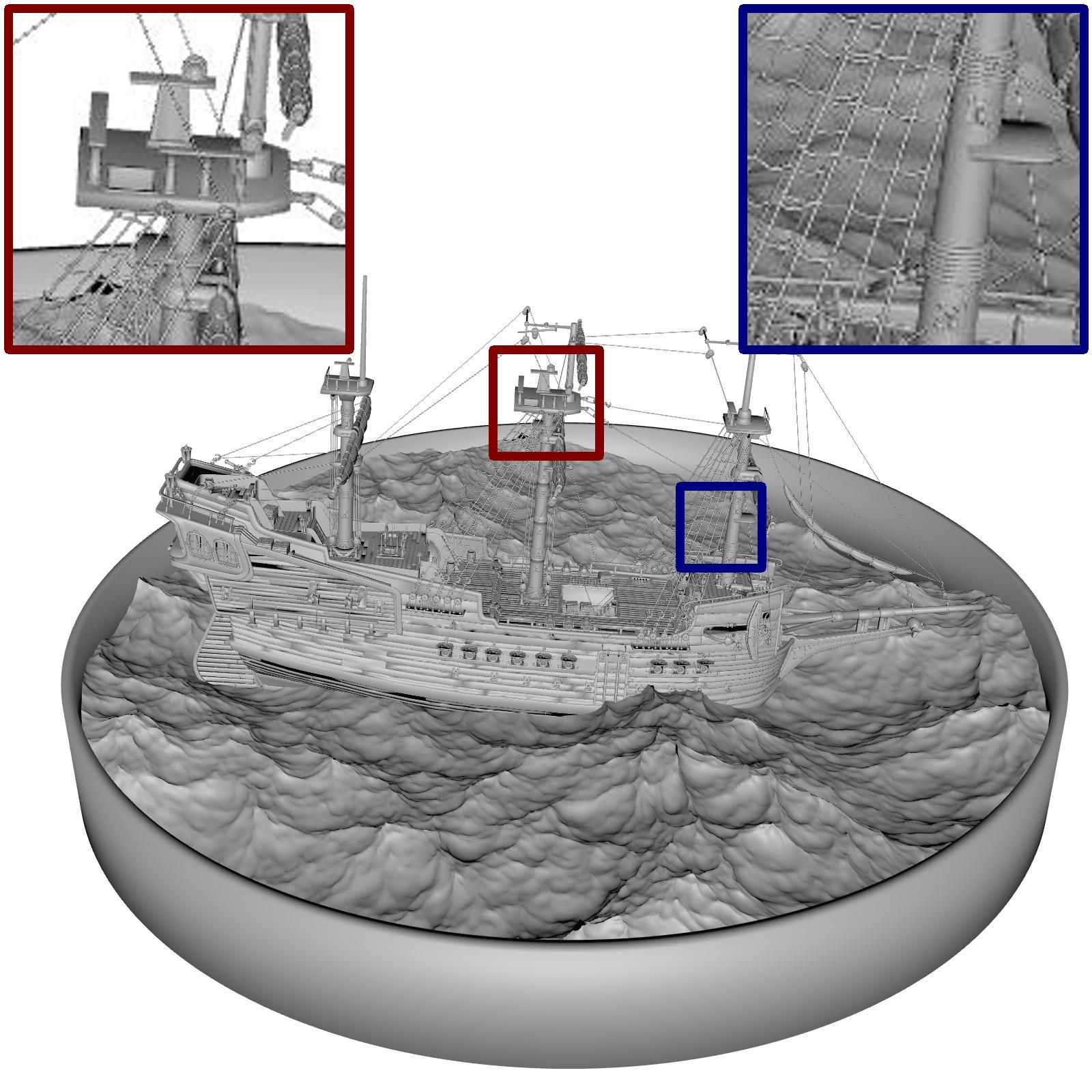} &
        \hspace{\mrg}
        \includegraphics[width=\wid]{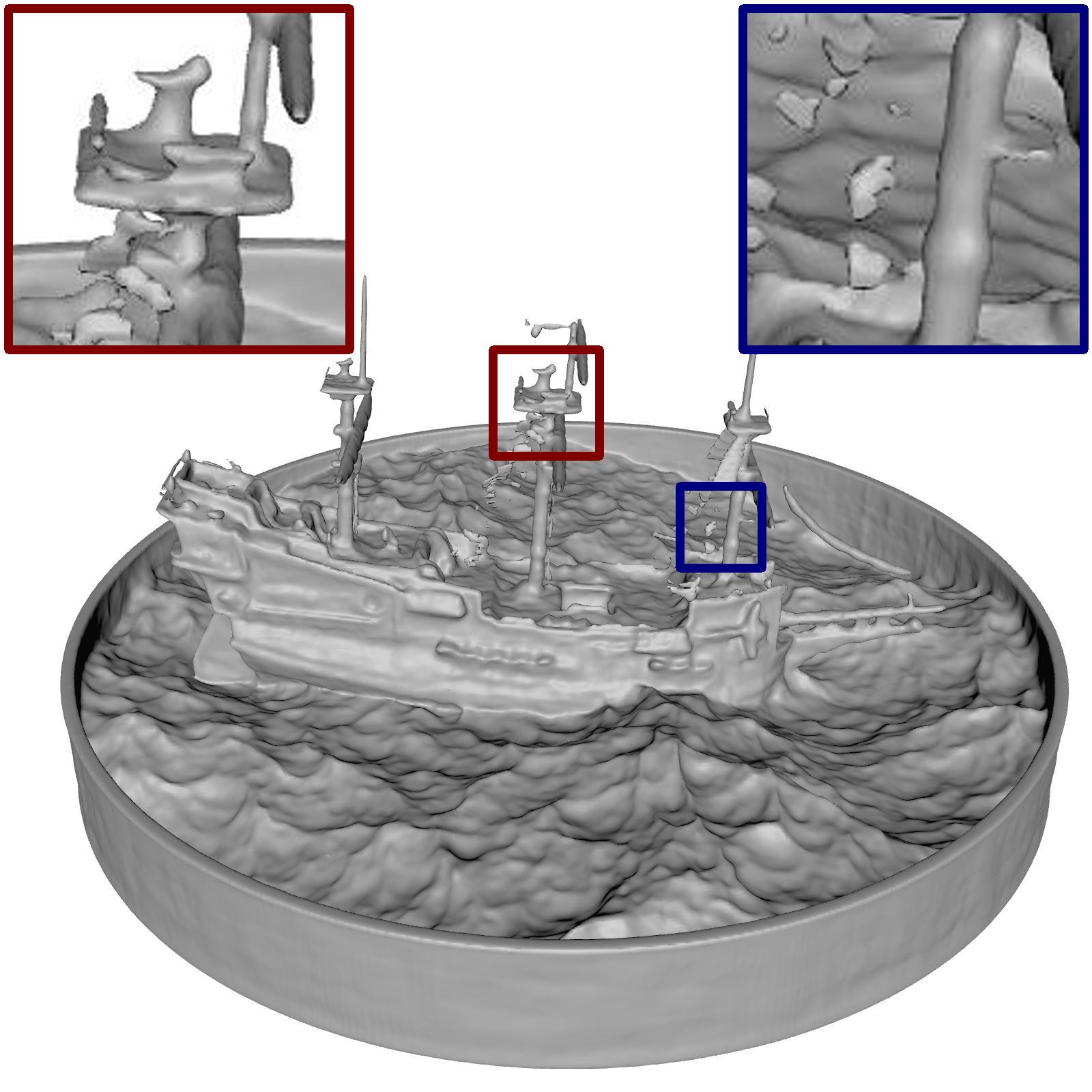} &
        \hspace{\mrg}
        \includegraphics[width=\wid]{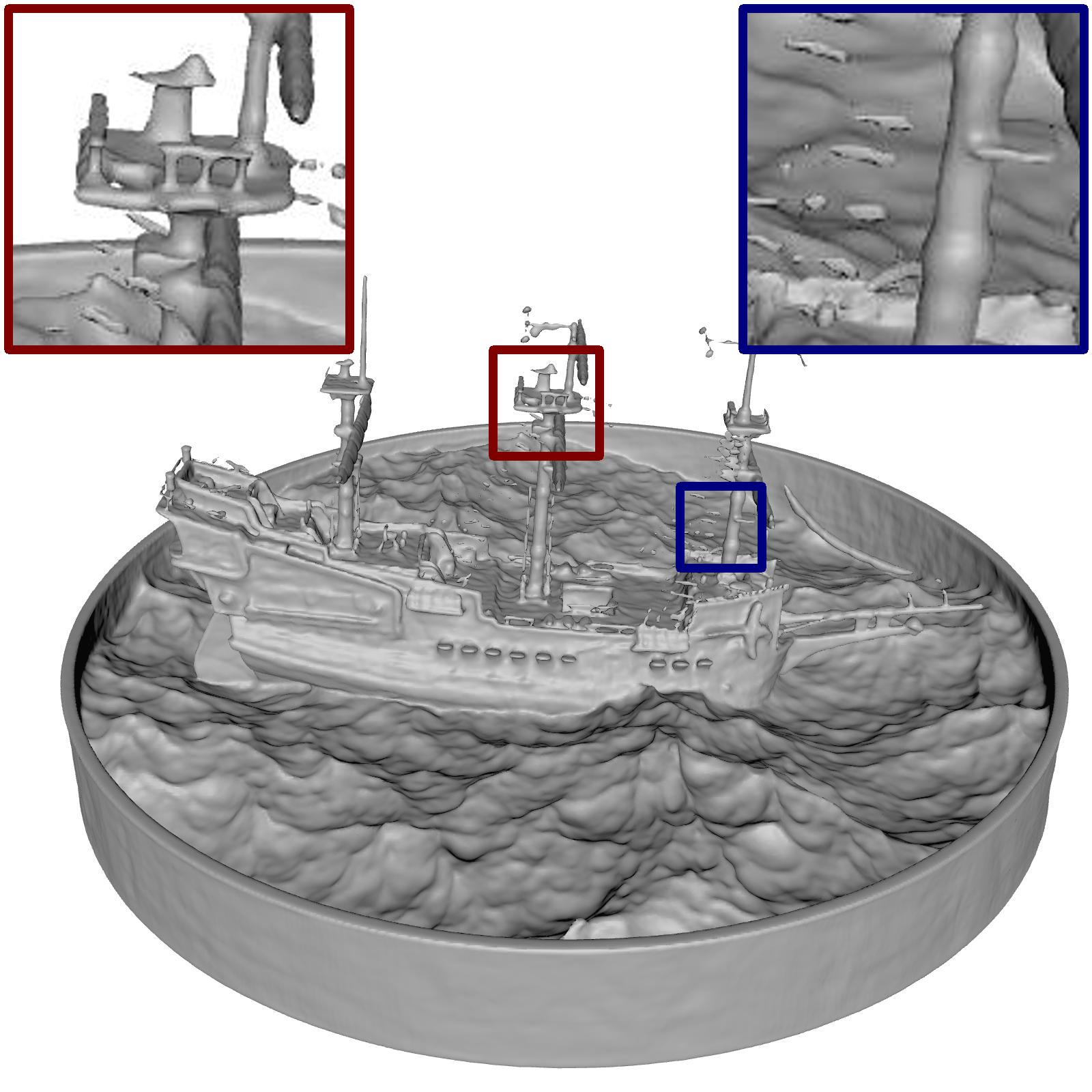}
        \\
        \vspace{\mrgv}
        \includegraphics[width=\wid]{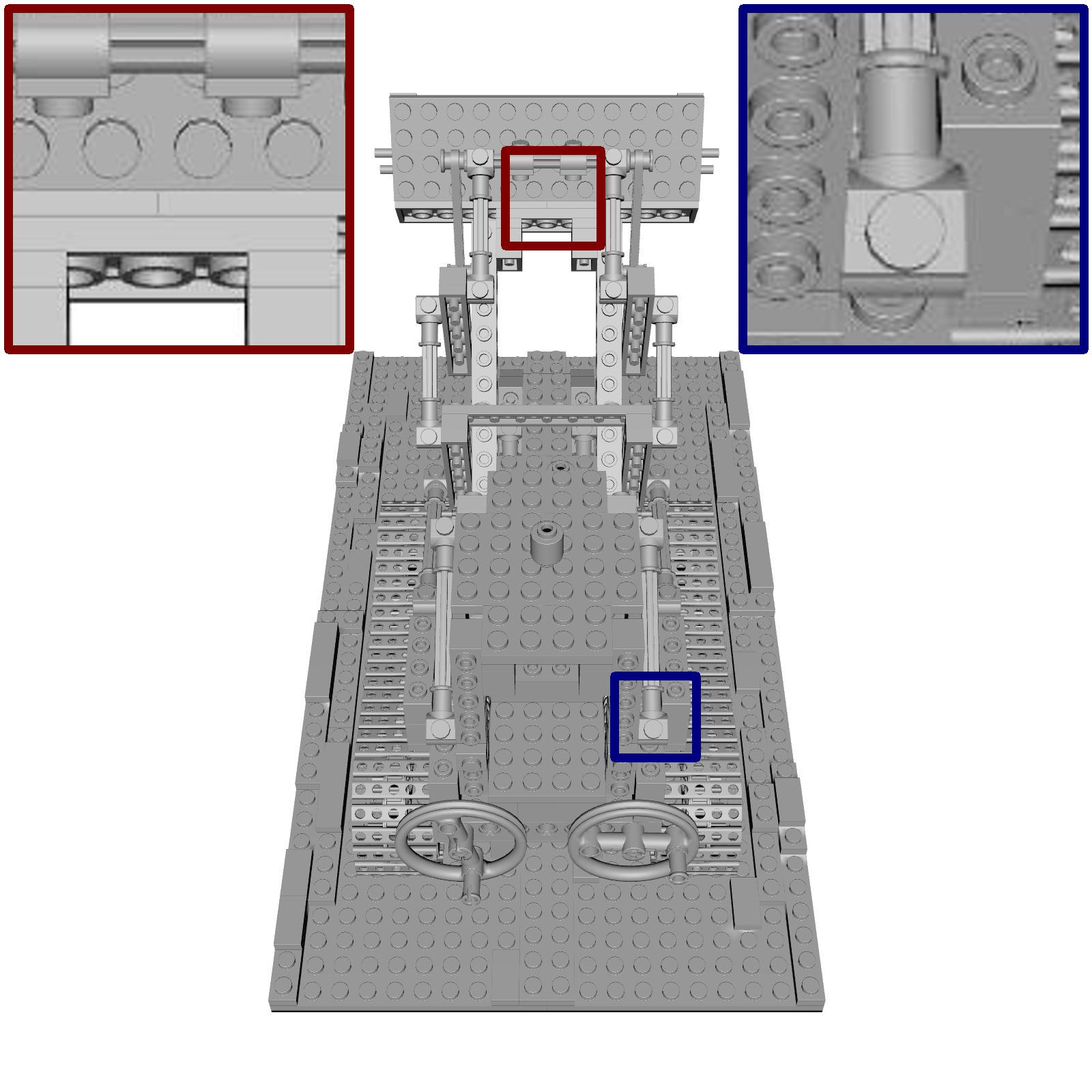} &
        \hspace{\mrg}
        \includegraphics[width=\wid]{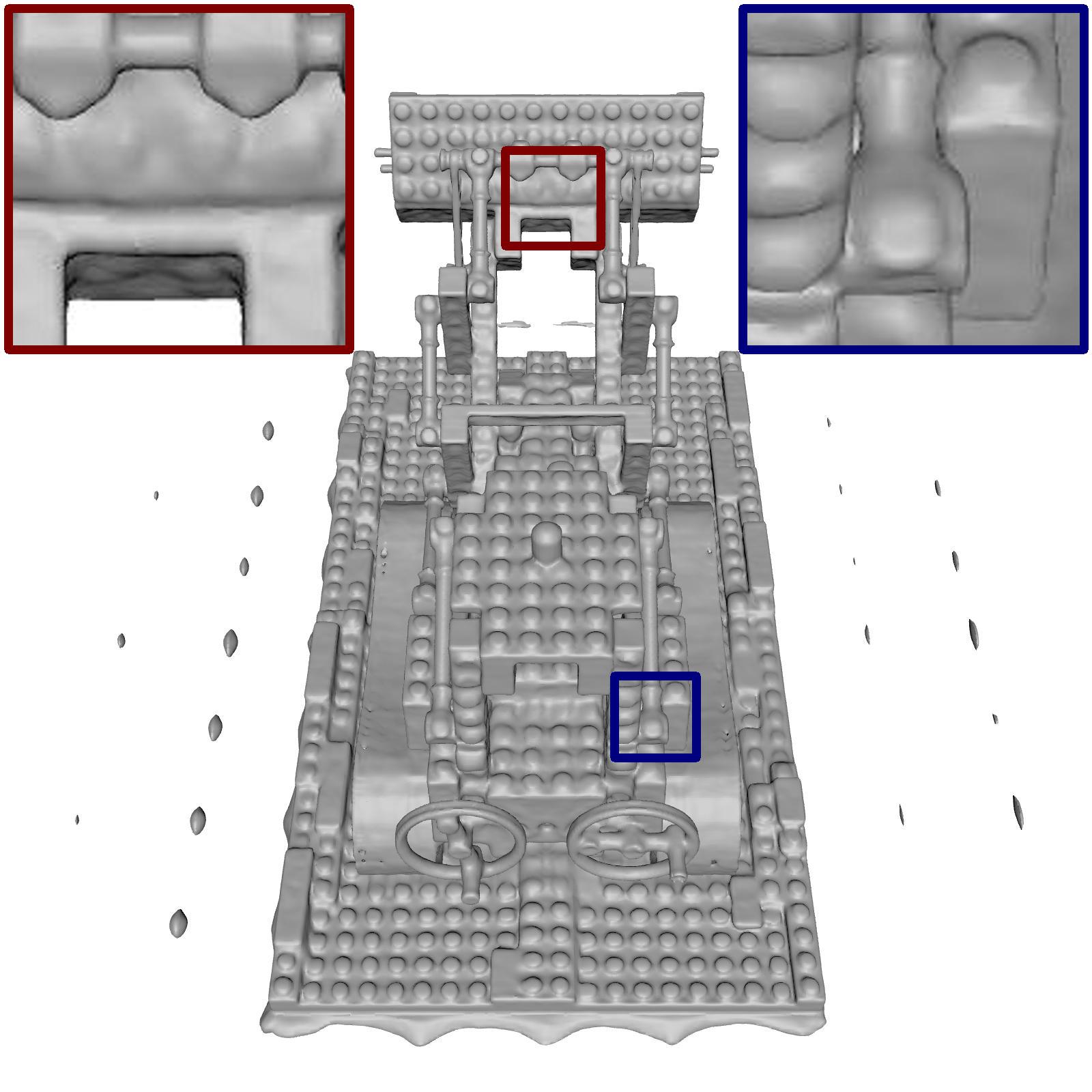} &
        \hspace{\mrg}
        \includegraphics[width=\wid]{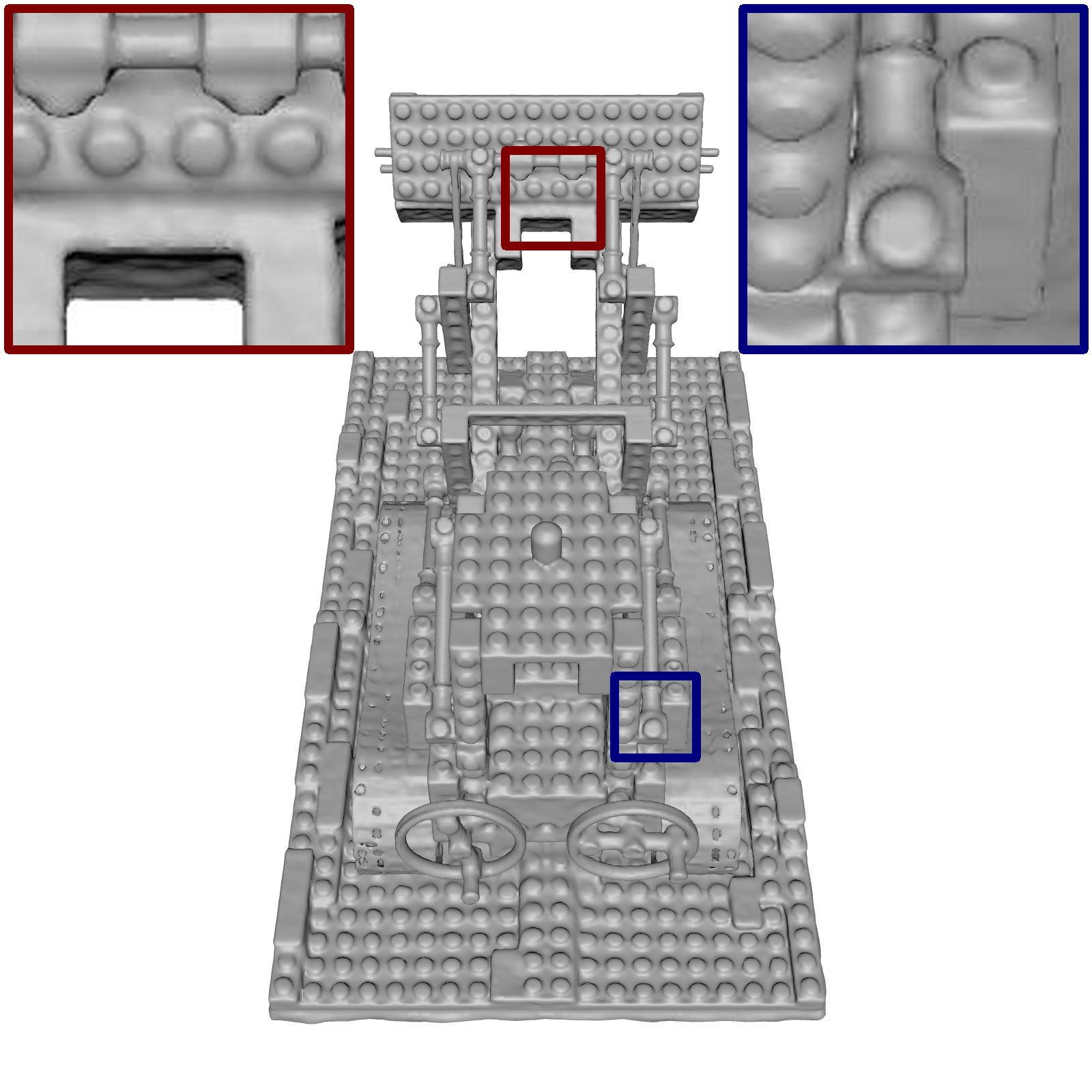}
        \\
        \textbf{Ground truth} & \hspace{\mrg}
        \textbf{NeuS} & \hspace{\mrg}
        \textbf{NeuS (ours)}
    \end{tabular}
    \caption{Additional qualitative results on the Realistic Synthetic 360 dataset~\cite{Mildenhall2020NeRFRS} for NeuS~\cite{Wang2021NeuSLN} method.}
    \label{fig:nerf_qual_appendix_neus}
\end{figure*}

\begin{figure*}
    \centering    
    \setlength{\wid}{0.23\textwidth}
    \setlength{\mrg}{-0.45cm}
    \setlength{\mrgv}{-0.05cm}
    \begin{tabular}{c cc}
        \vspace{\mrgv}
        \includegraphics[width=\wid]{figures/nerf_qual_ap/gt/hotdog.jpg} &
        \hspace{\mrg}
        \includegraphics[width=\wid]{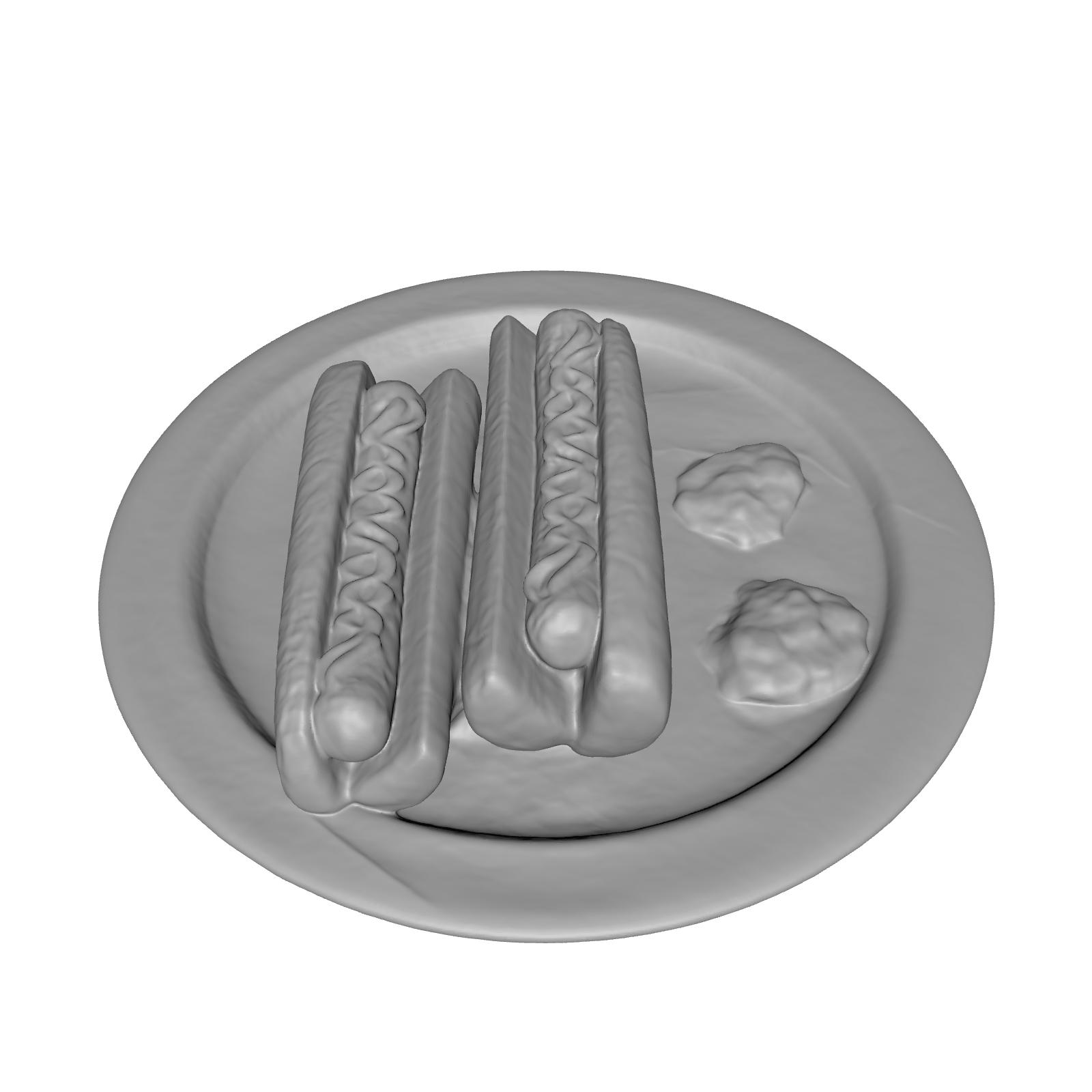} &
        \hspace{\mrg}
        \includegraphics[width=\wid]{figures/nerf_qual_ap/volsdf/hotdog.jpg}
        \\ 
        \vspace{\mrgv}
        \includegraphics[width=\wid]{figures/nerf_qual_ap/gt/chair.jpg} &
        \hspace{\mrg}
        \includegraphics[width=\wid]{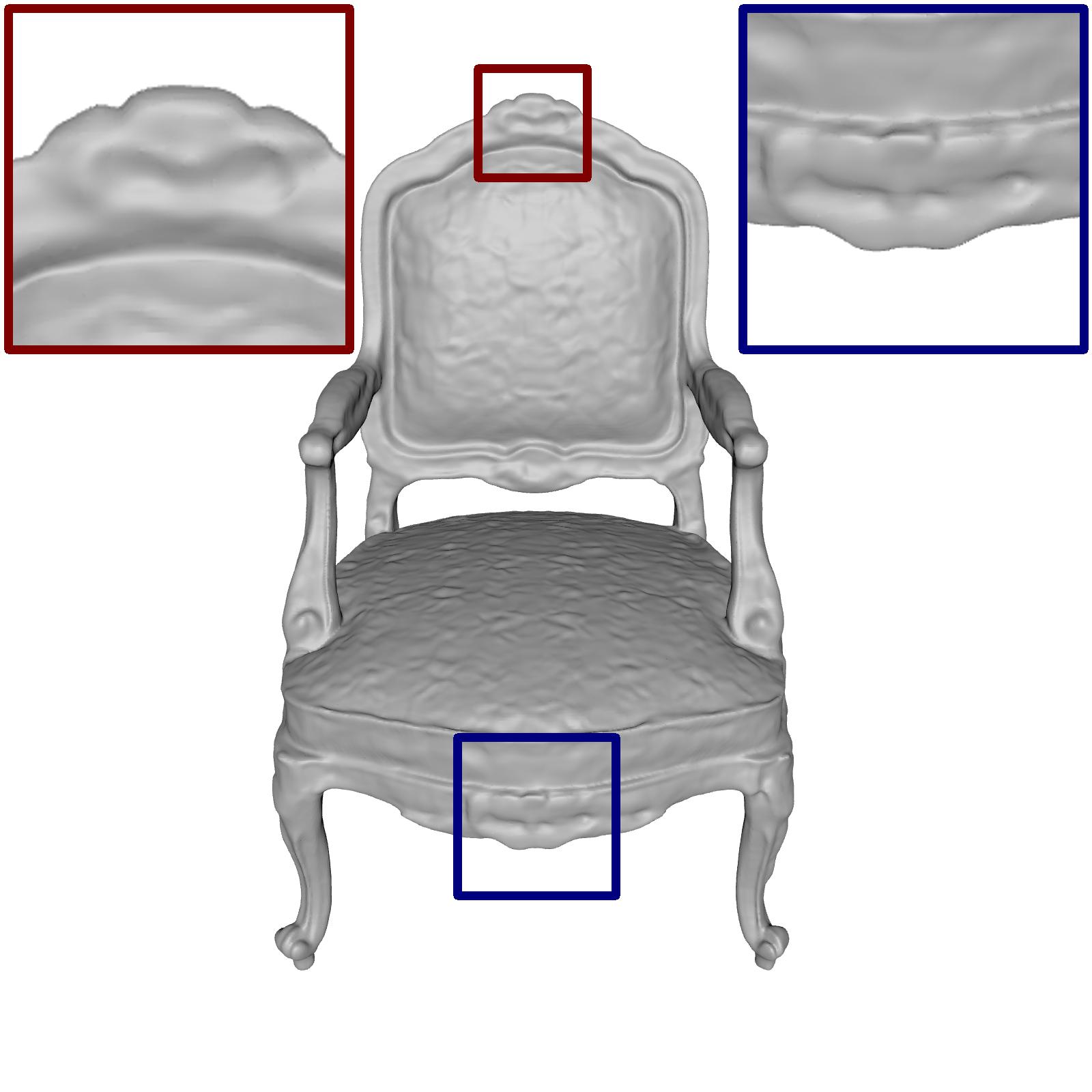} &
        \hspace{\mrg}
        \includegraphics[width=\wid]{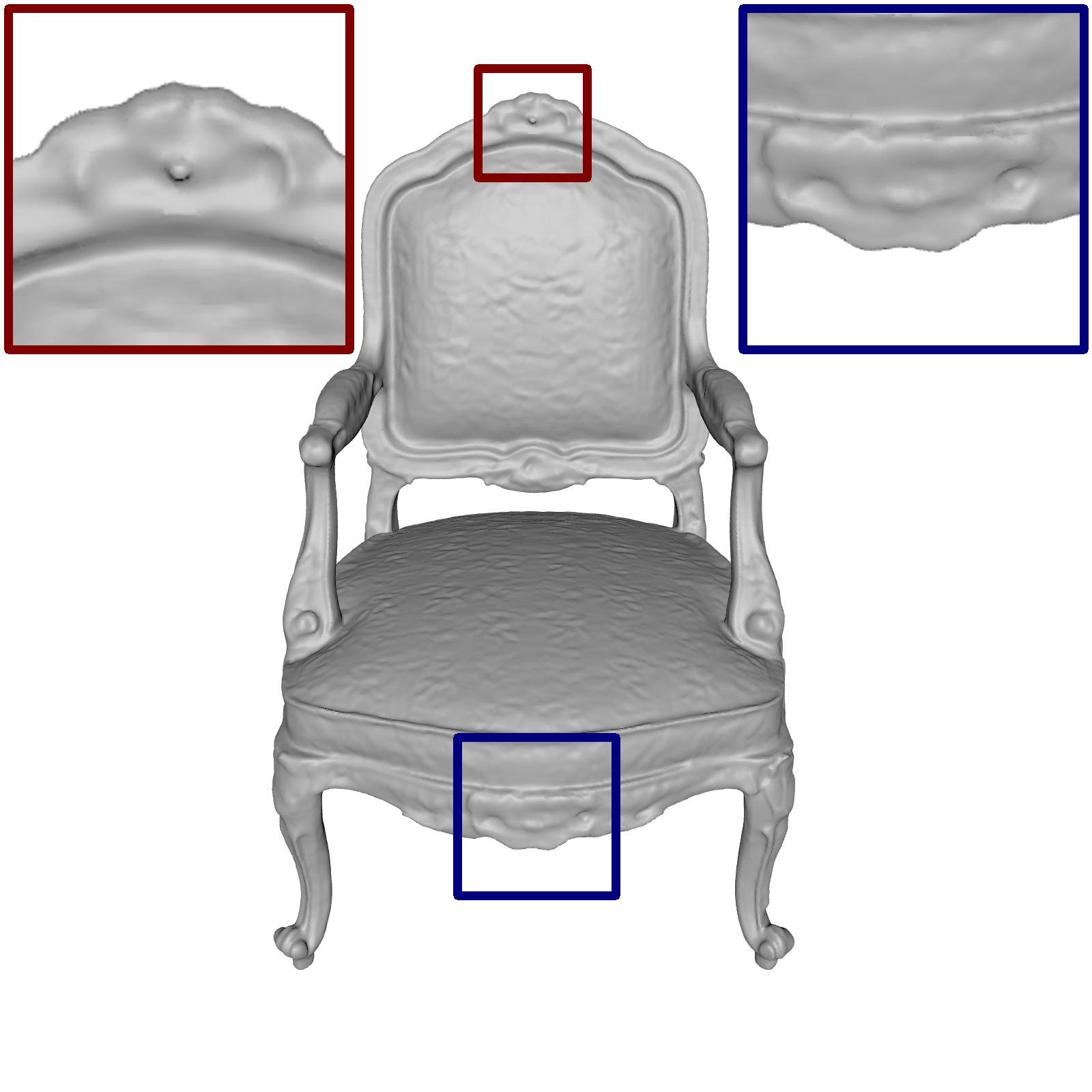}
        \\ 
        \vspace{\mrgv}
        \includegraphics[width=\wid]{figures/nerf_qual_ap/gt/mic.jpg} &
        \hspace{\mrg}
        \includegraphics[width=\wid]{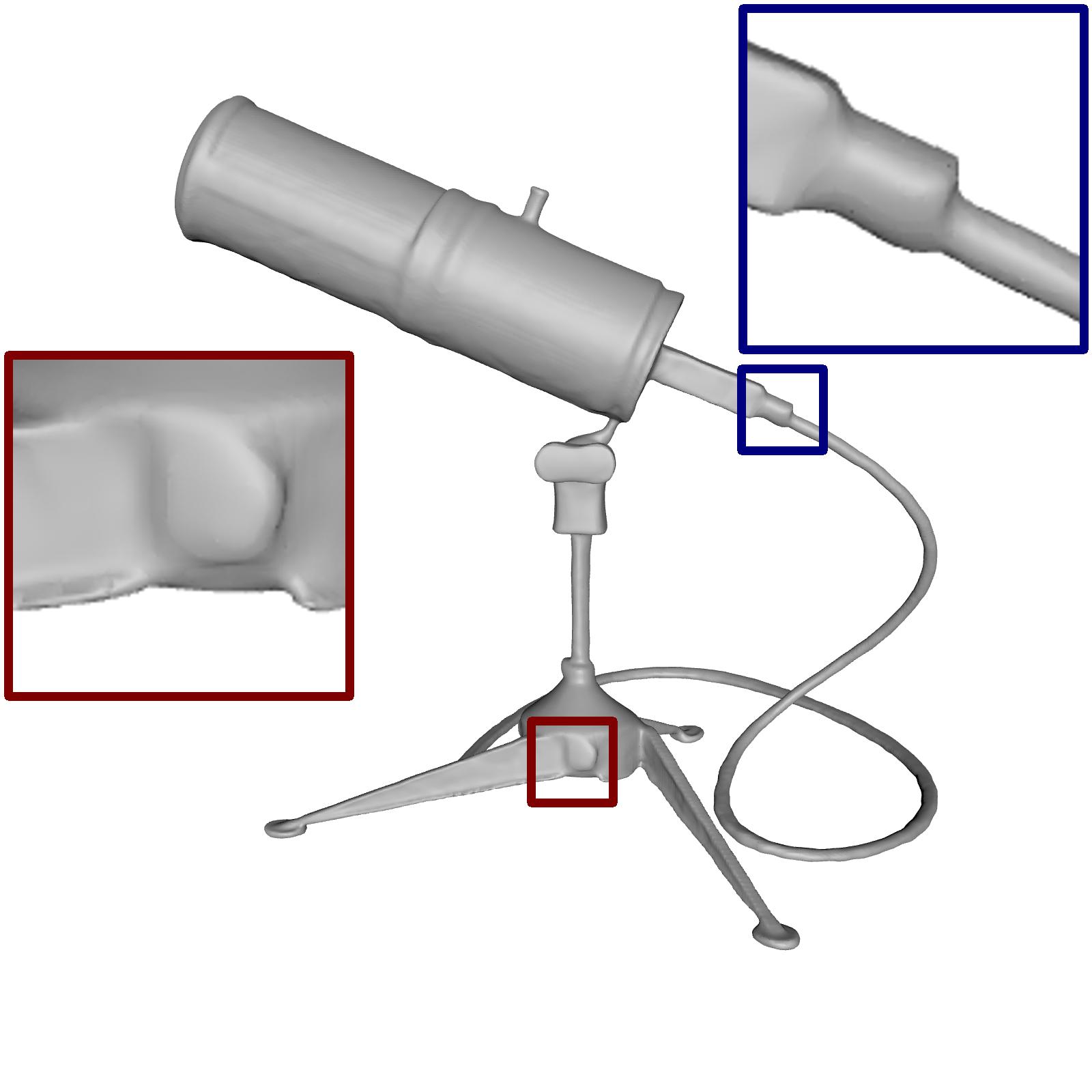} &
        \hspace{\mrg}
        \includegraphics[width=\wid]{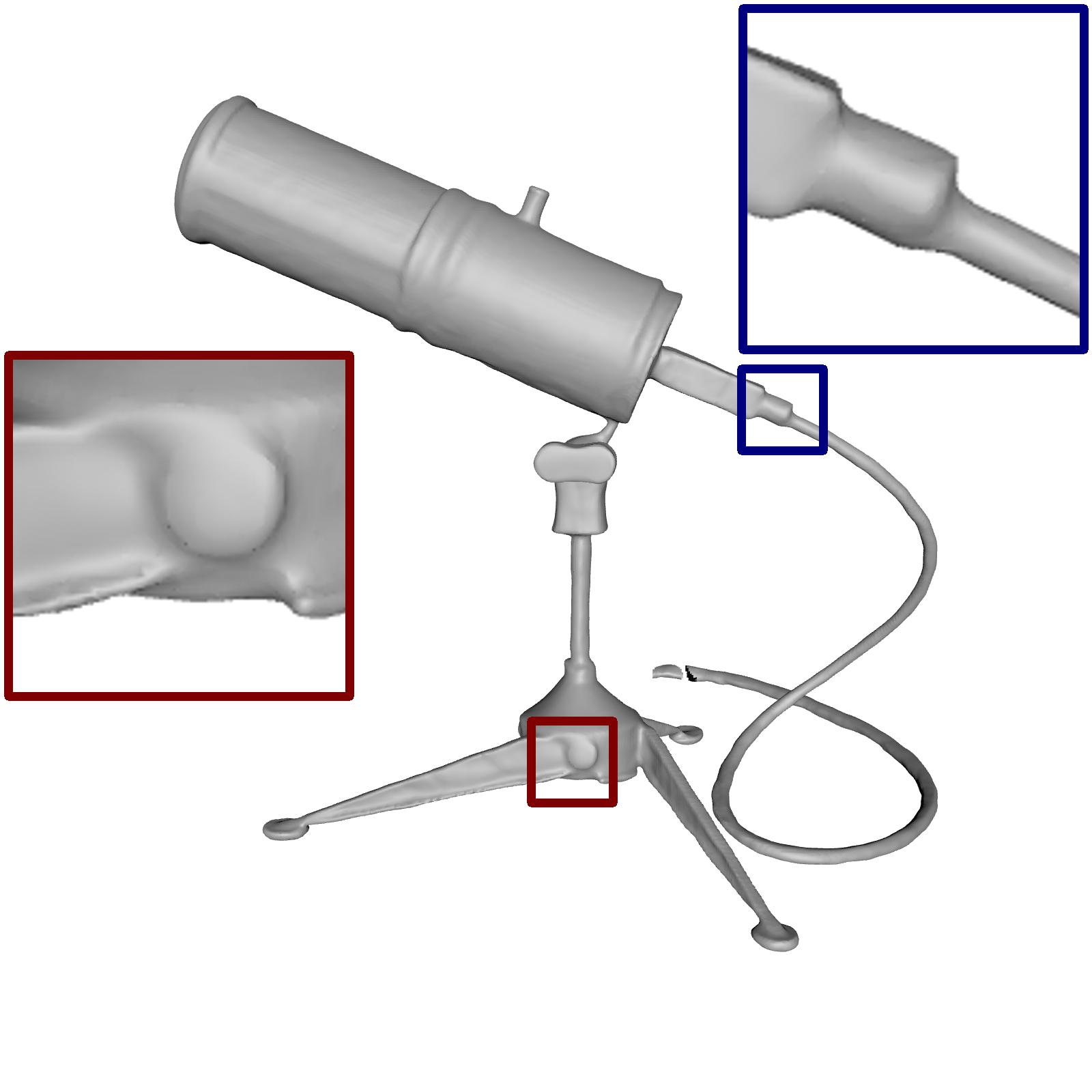}
        \\ 
        \vspace{\mrgv}
        \includegraphics[width=\wid]{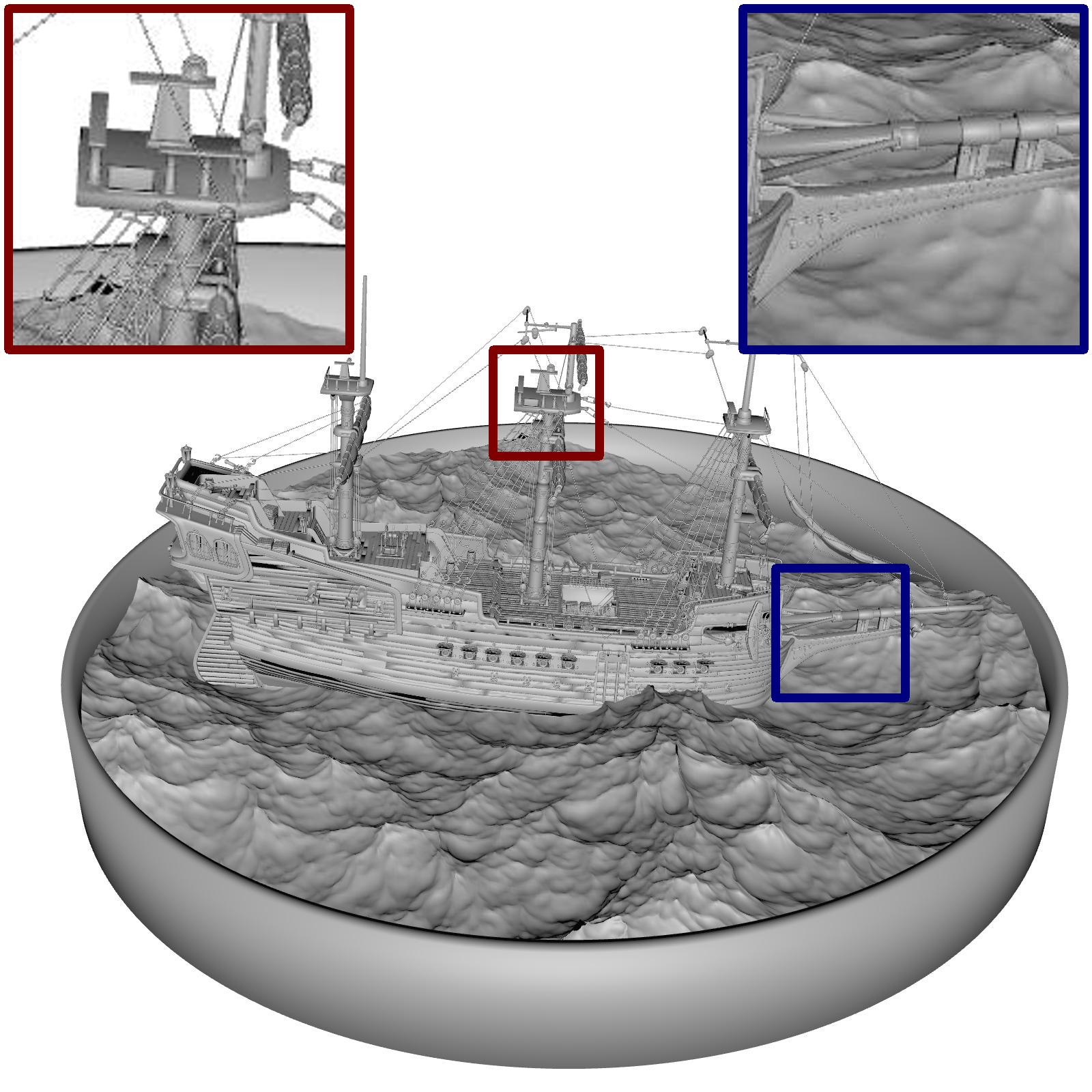} &
        \hspace{\mrg}
        \includegraphics[width=\wid]{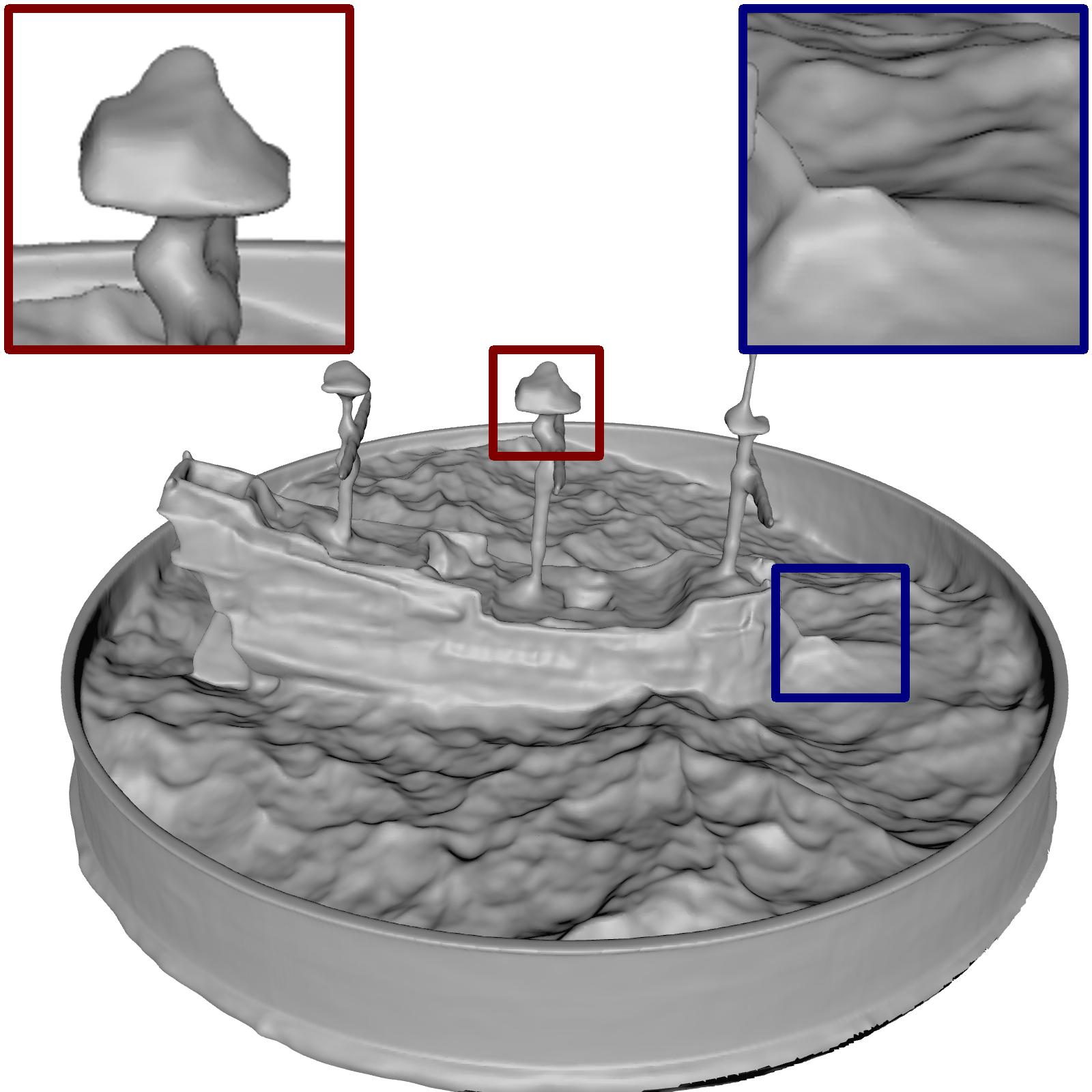} &
        \hspace{\mrg}
        \includegraphics[width=\wid]{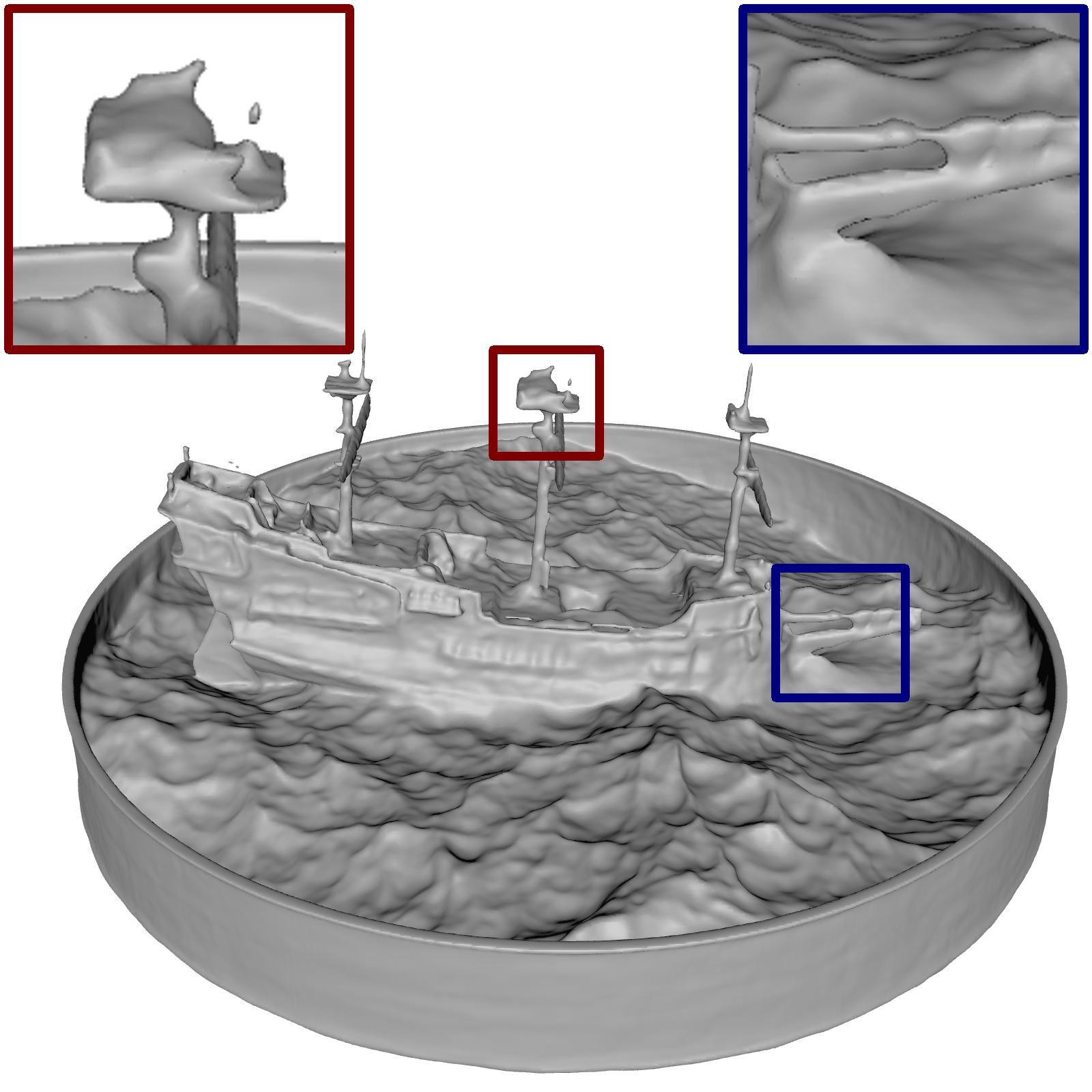}
        \\
        \vspace{\mrgv}
        \includegraphics[width=\wid]{figures/nerf_qual_ap/gt/lego.jpg} &
        \hspace{\mrg}
        \includegraphics[width=\wid]{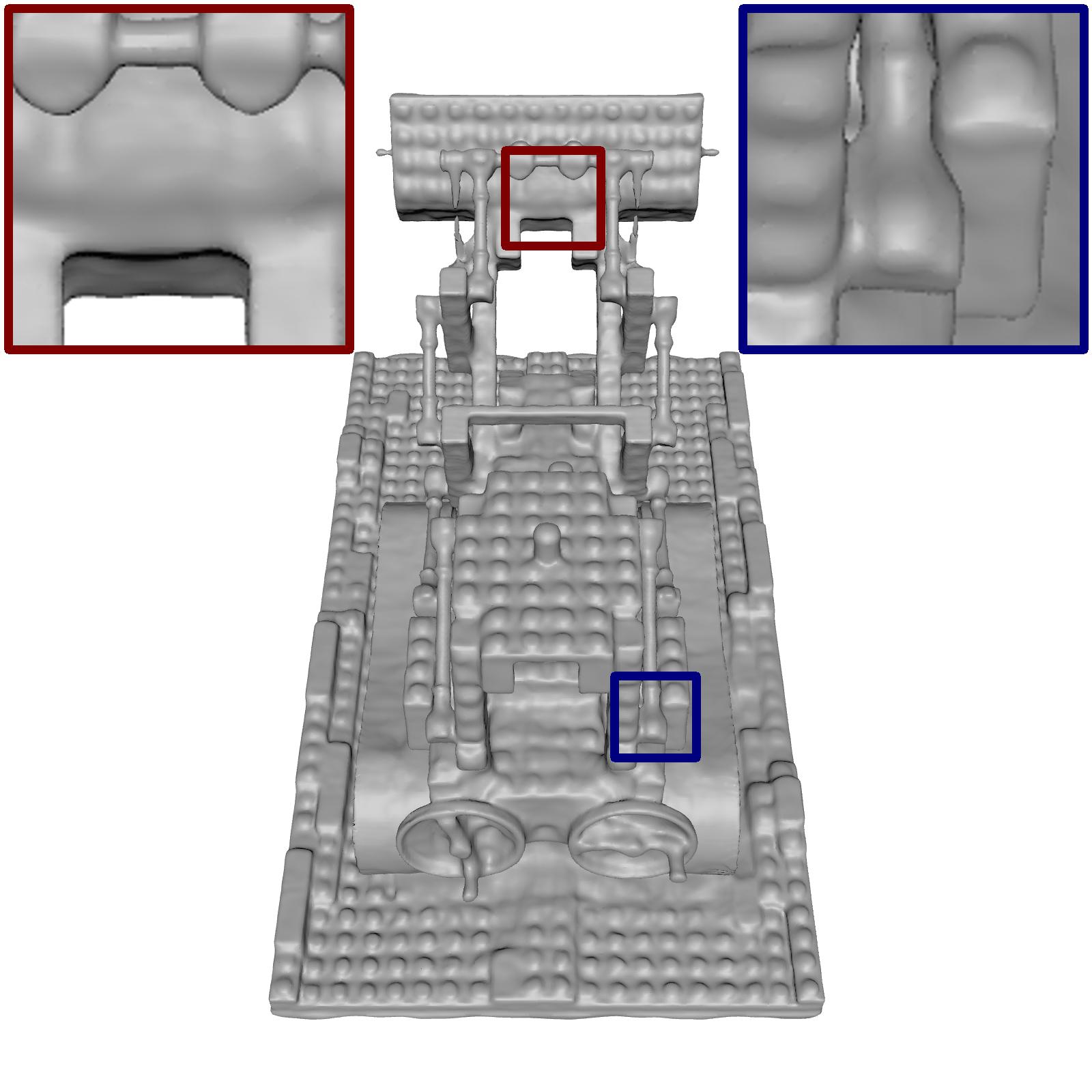} &
        \hspace{\mrg}
        \includegraphics[width=\wid]{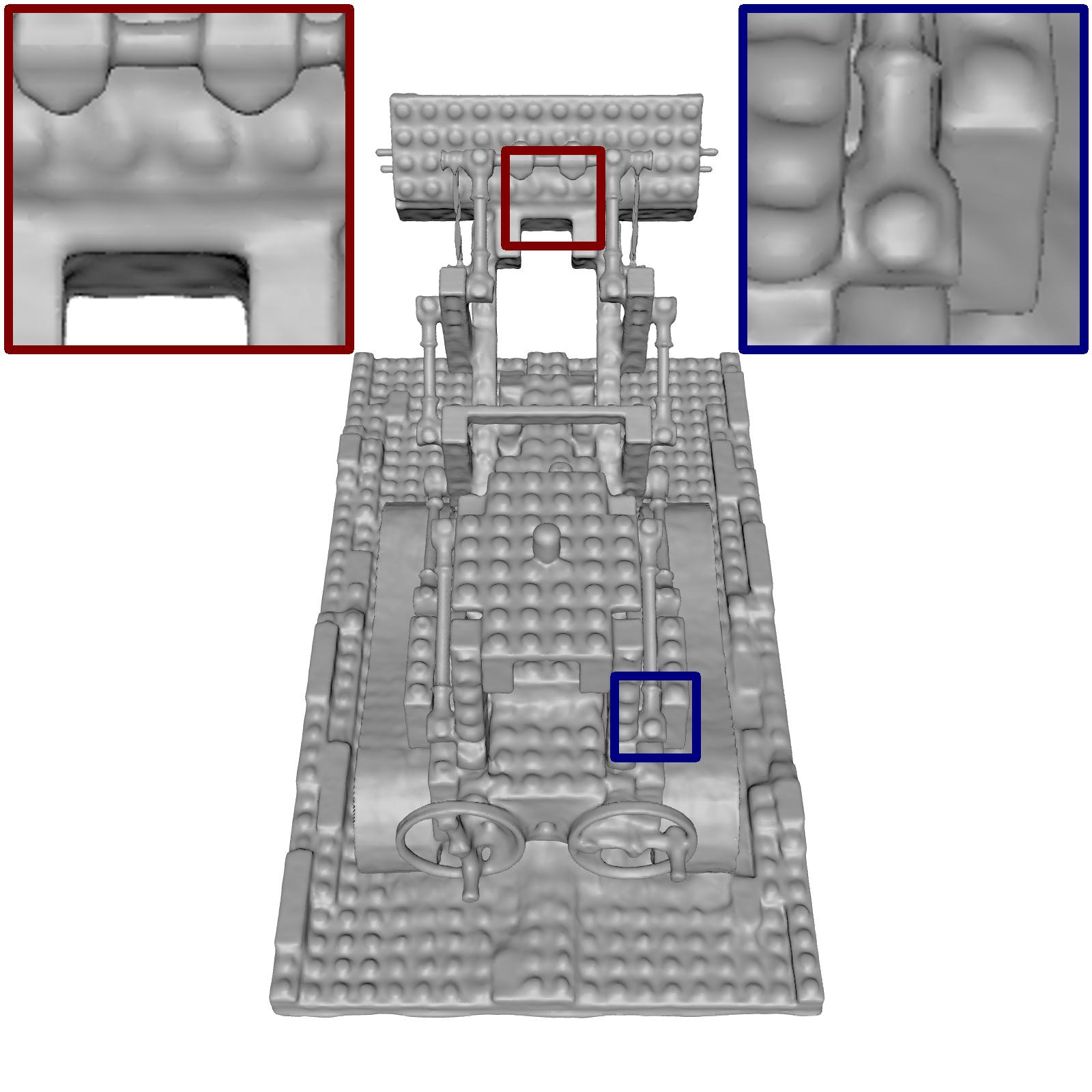}
        \\
        \textbf{Ground truth} & \hspace{\mrg}
        \textbf{VolSDF} & \hspace{\mrg}
        \textbf{VolSDF (ours)}
    \end{tabular}
    \caption{Qualitative results on the Realistic Synthetic 360 dataset~\cite{Mildenhall2020NeRFRS} for the unofficial implementation of VolSDF~\cite{Wang2021NeuSLN} method.}
    \label{fig:nerf_qual_appendix_volsdf}
\end{figure*}

\begin{figure*}
    \centering    
    \setlength{\wid}{0.23\textwidth}
    \setlength{\mrg}{-0.45cm}
    \setlength{\mrgv}{-0.05cm}
    \begin{tabular}{c cc}
        \vspace{\mrgv}
        \includegraphics[width=\wid]{figures/nerf_qual_ap/gt/hotdog.jpg} &
        \hspace{\mrg}
        \includegraphics[width=\wid]{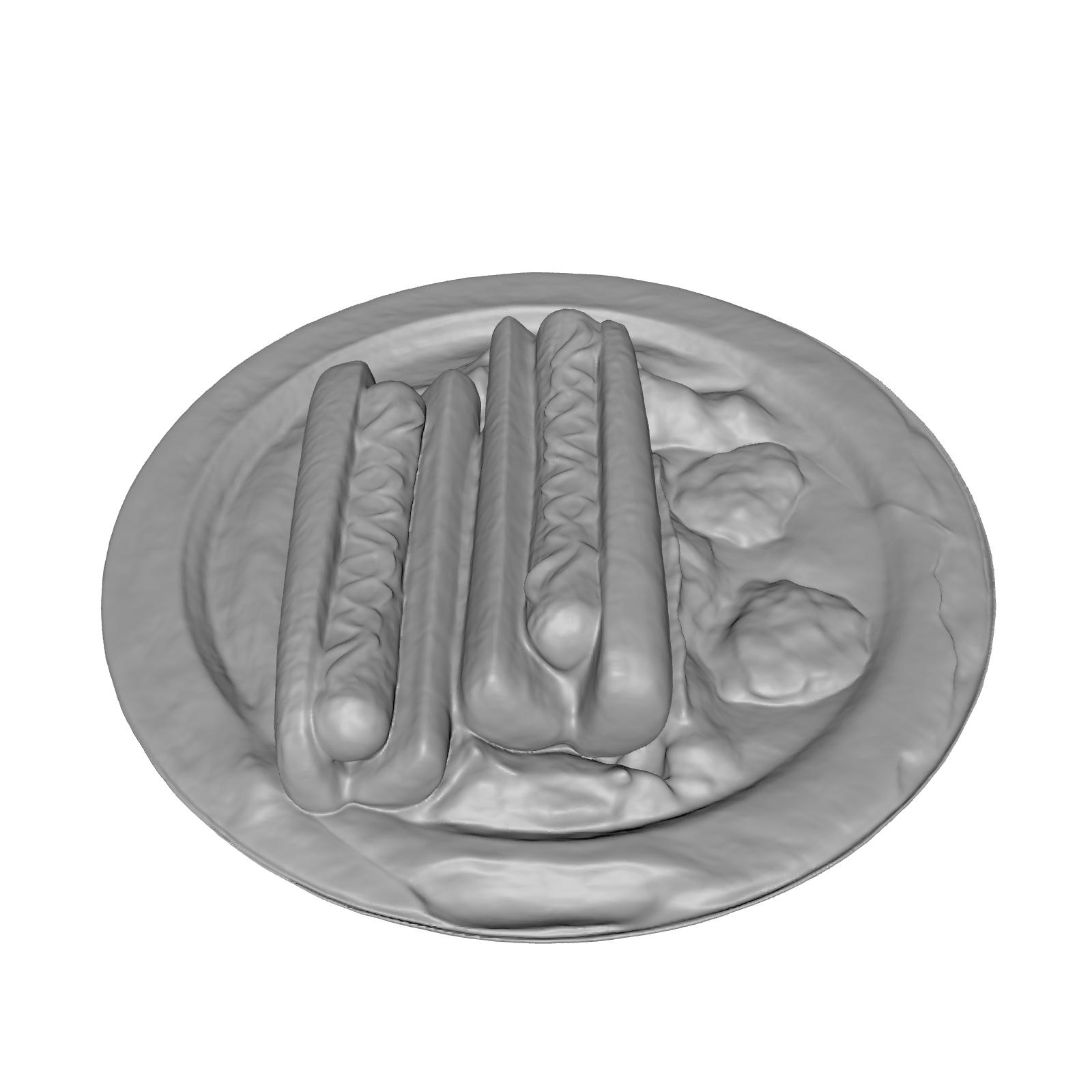} &
        \hspace{\mrg}
        \includegraphics[width=\wid]{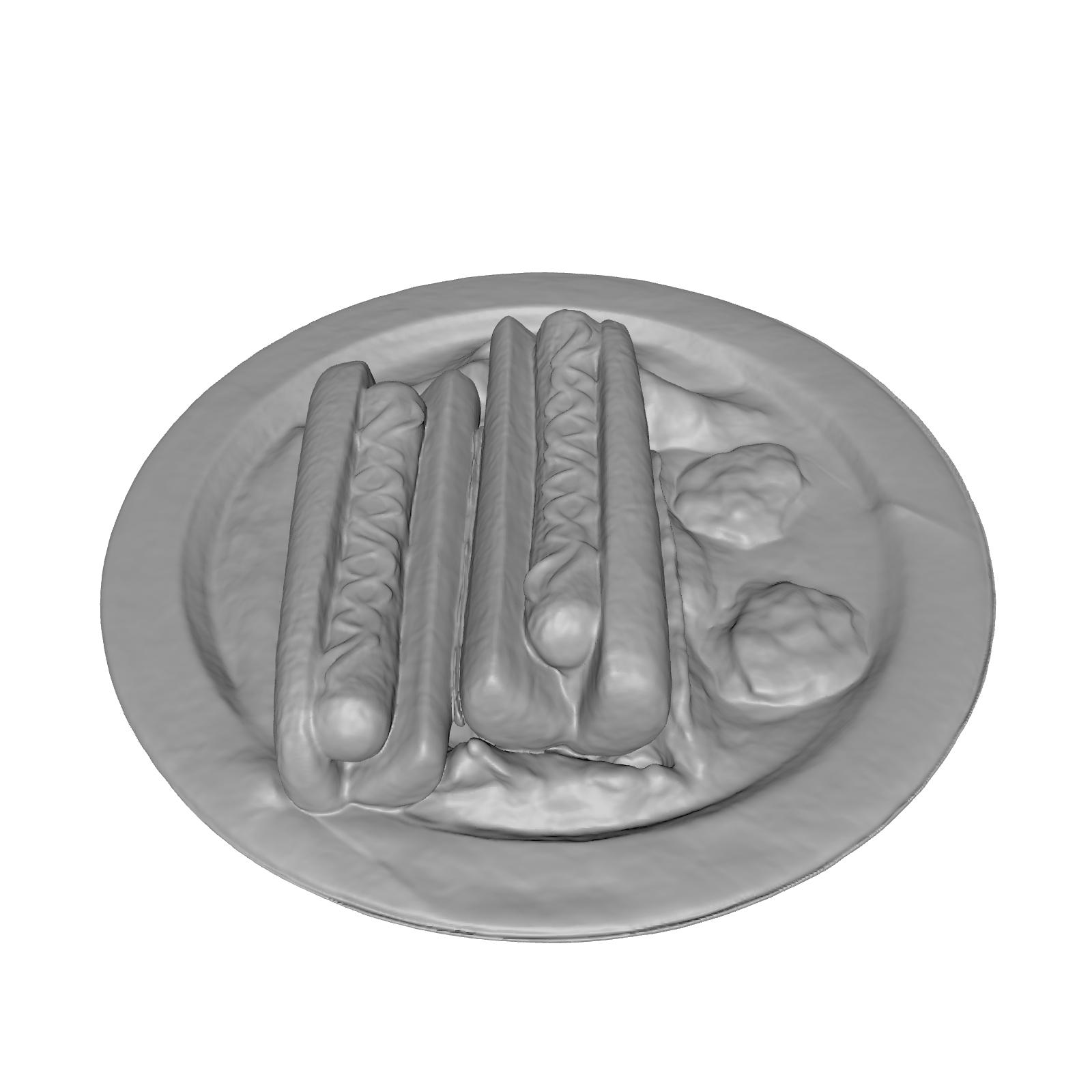}
        \\ 
        \vspace{\mrgv}
        \includegraphics[width=\wid]{figures/nerf_qual_ap/gt/chair.jpg} &
        \hspace{\mrg}
        \includegraphics[width=\wid]{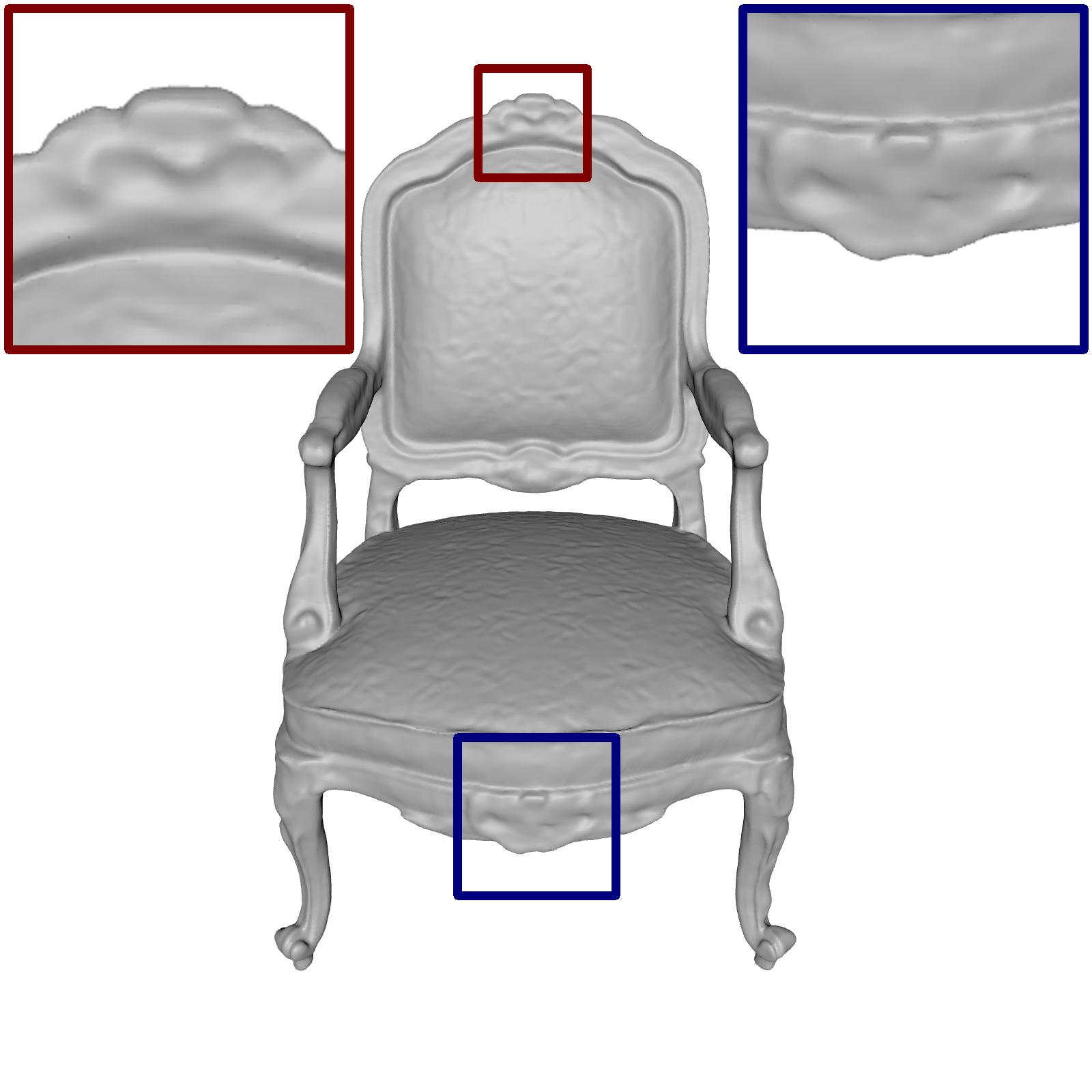} &
        \hspace{\mrg}
        \includegraphics[width=\wid]{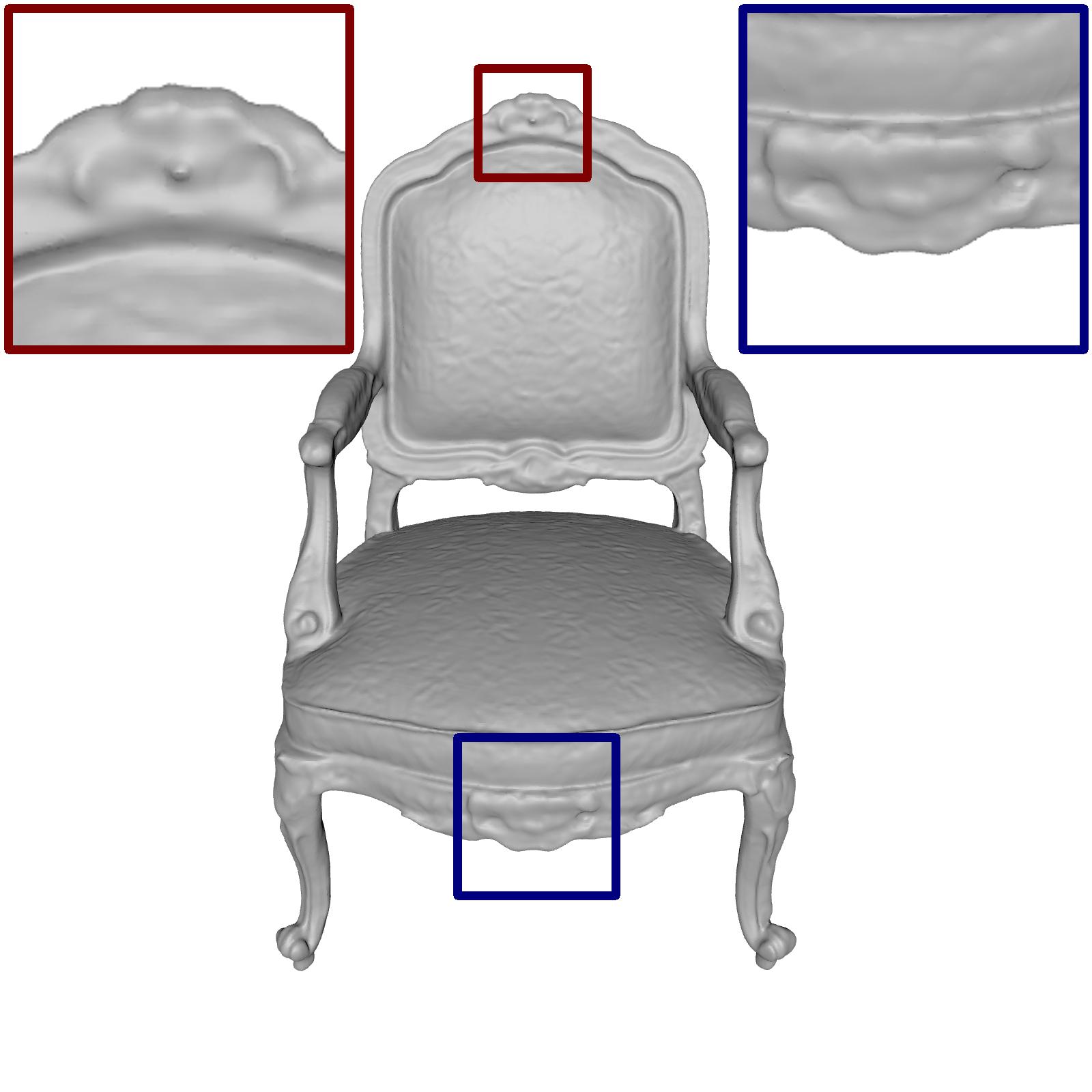}
        \\ 
        \vspace{\mrgv}
        \includegraphics[width=\wid]{figures/nerf_qual_ap/gt/mic.jpg} &
        \hspace{\mrg}
        \includegraphics[width=\wid]{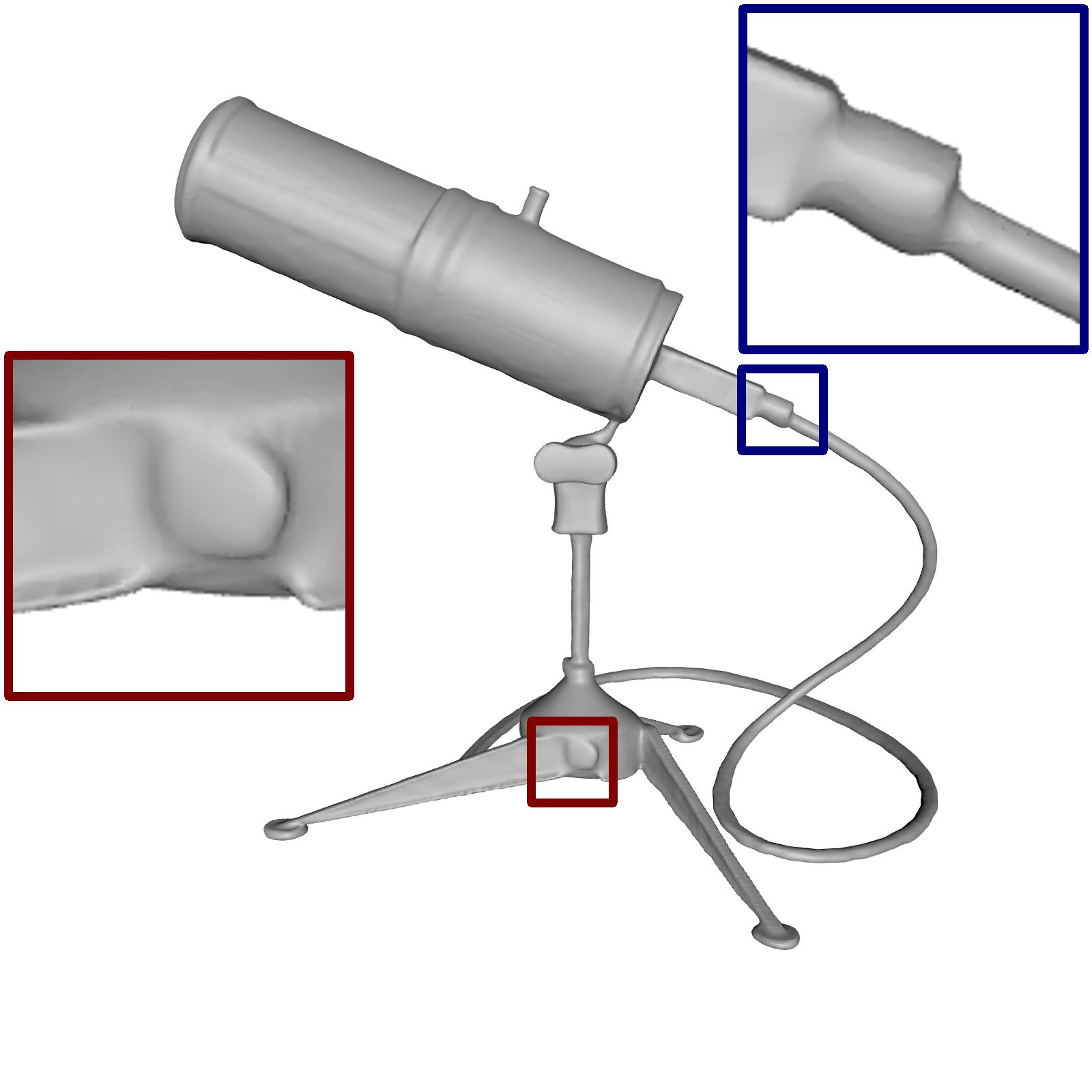} &
        \hspace{\mrg}
        \includegraphics[width=\wid]{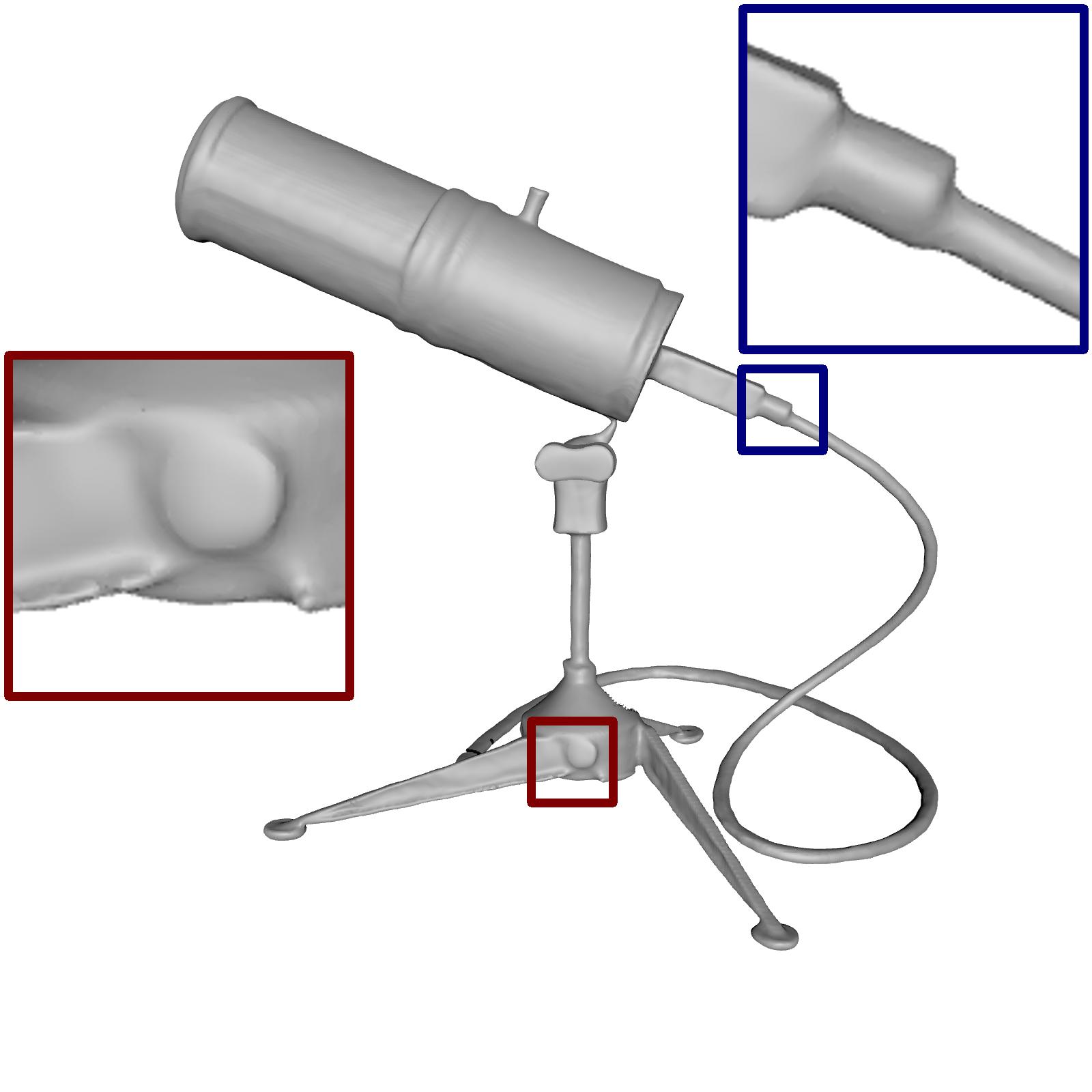}
        \\ 
        \vspace{\mrgv}
        \includegraphics[width=\wid]{figures/nerf_qual_ap/gt/ship_v.jpg} &
        \hspace{\mrg}
        \includegraphics[width=\wid]{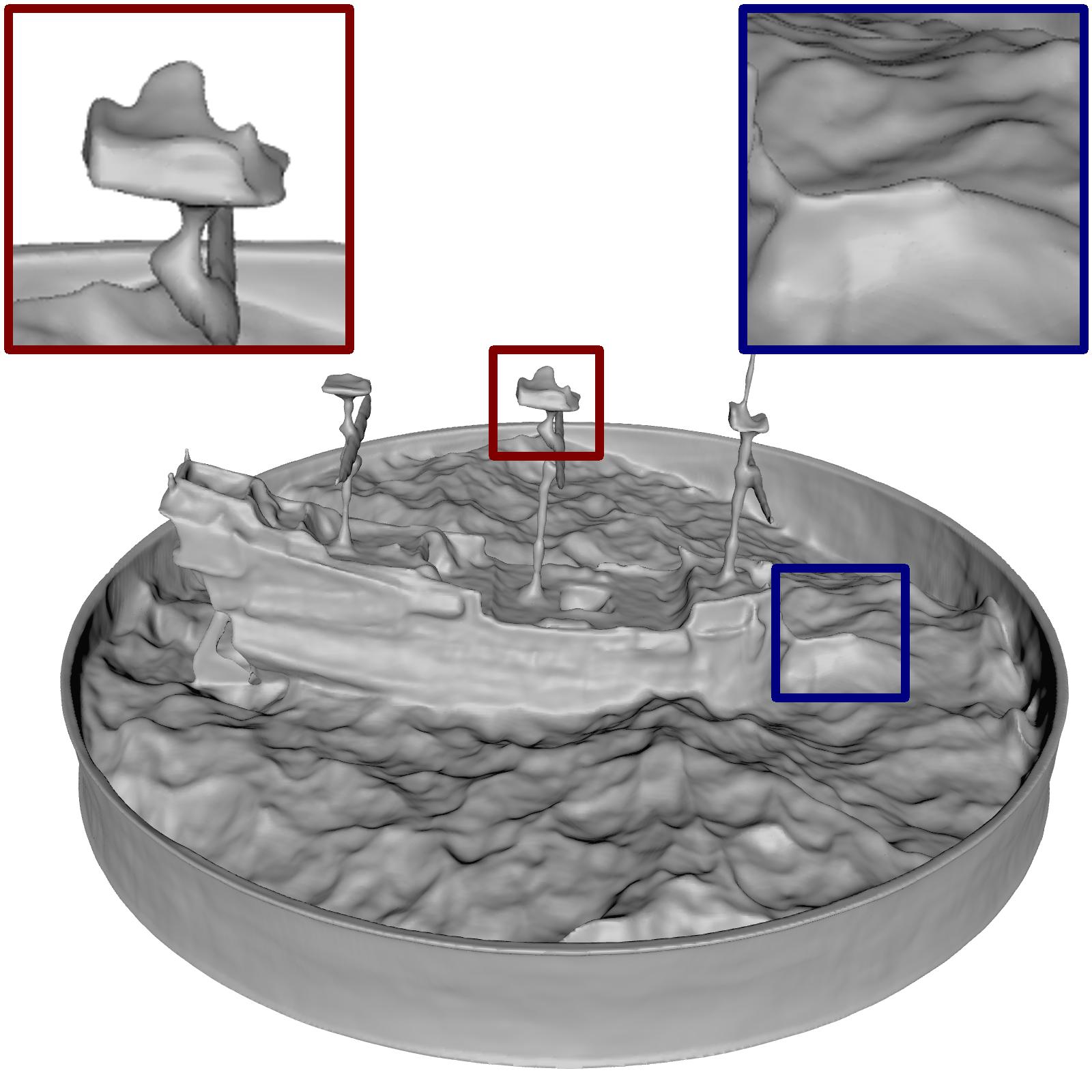} &
        \hspace{\mrg}
        \includegraphics[width=\wid]{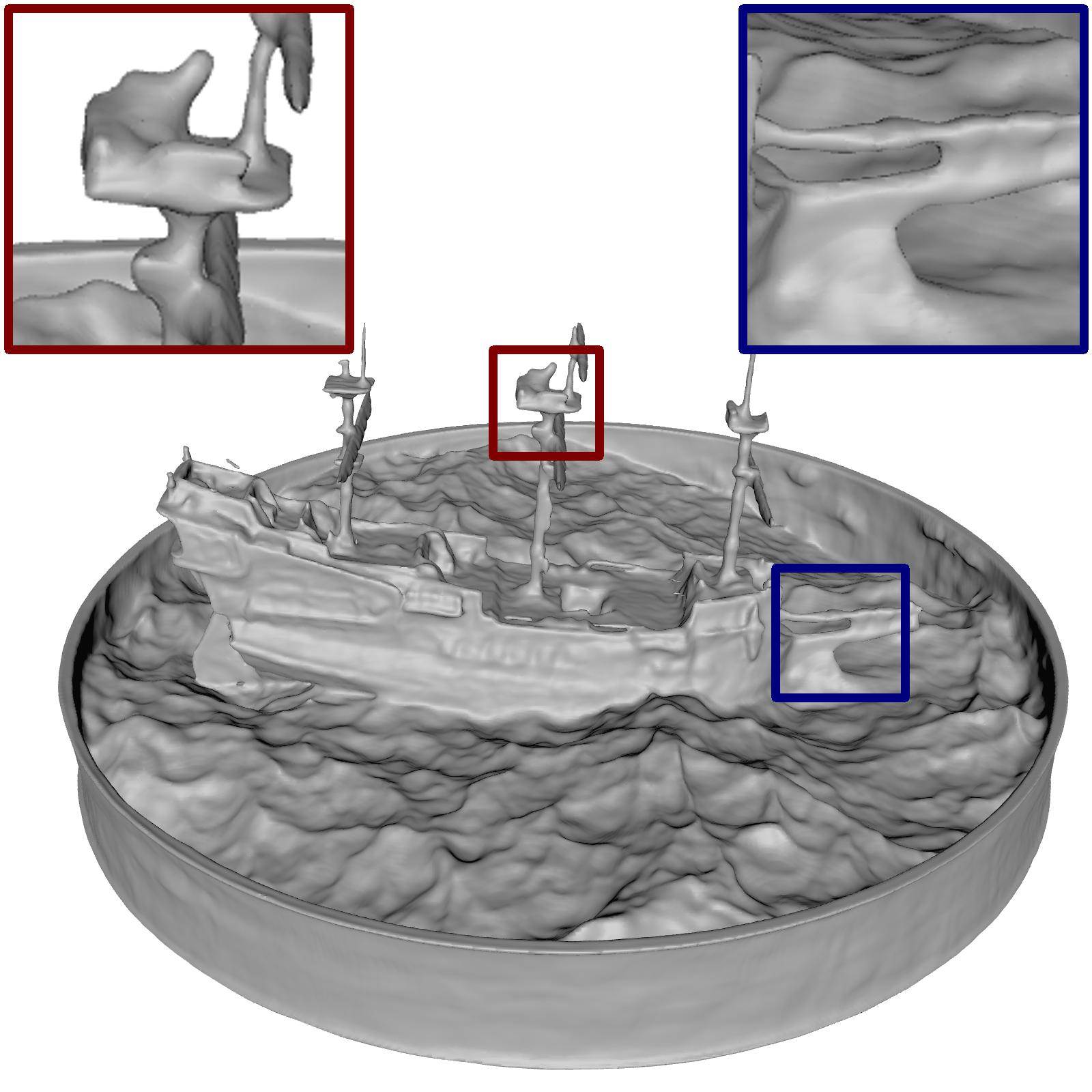}
        \\
        \vspace{\mrgv}
        \includegraphics[width=\wid]{figures/nerf_qual_ap/gt/lego.jpg} &
        \hspace{\mrg}
        \includegraphics[width=\wid]{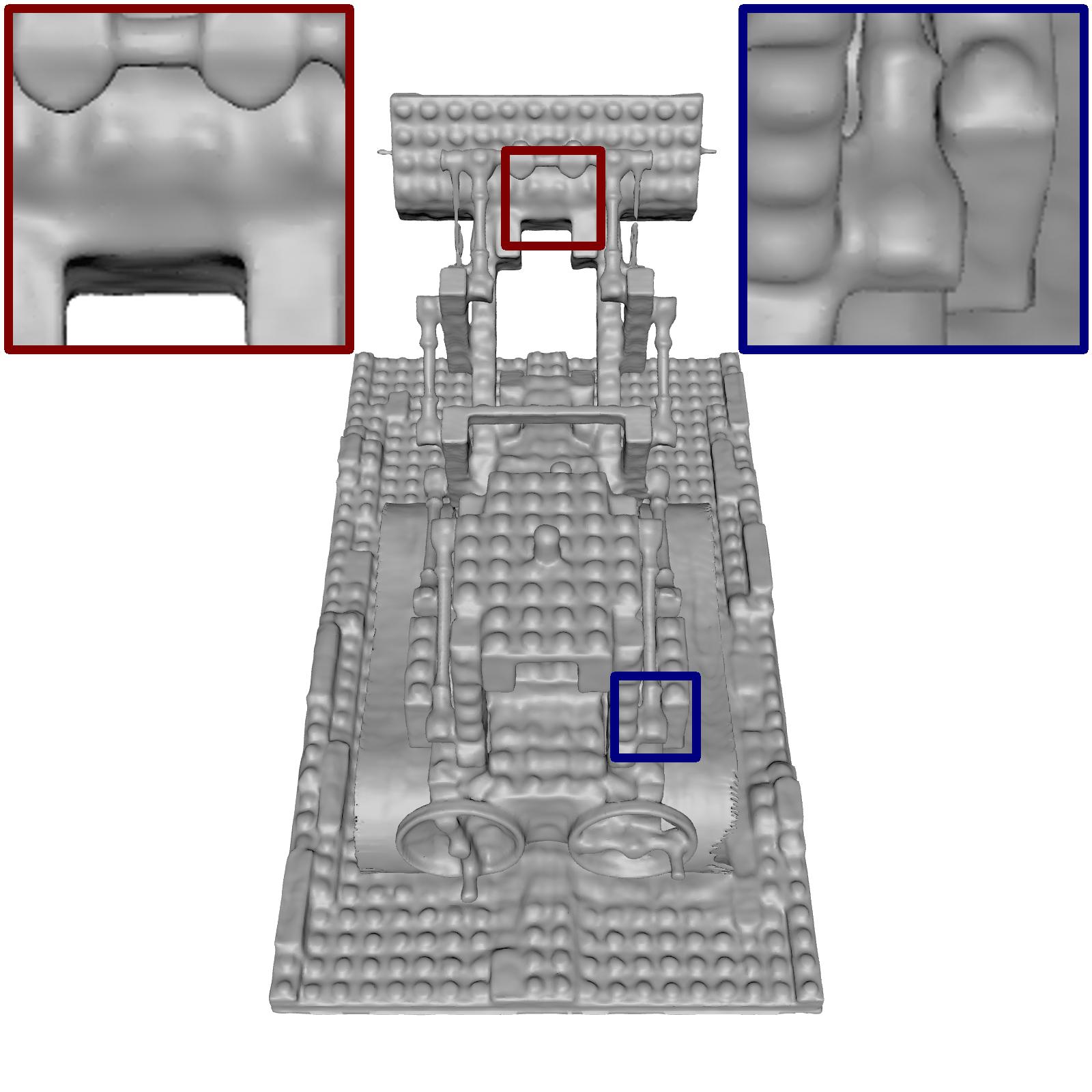} &
        \hspace{\mrg}
        \includegraphics[width=\wid]{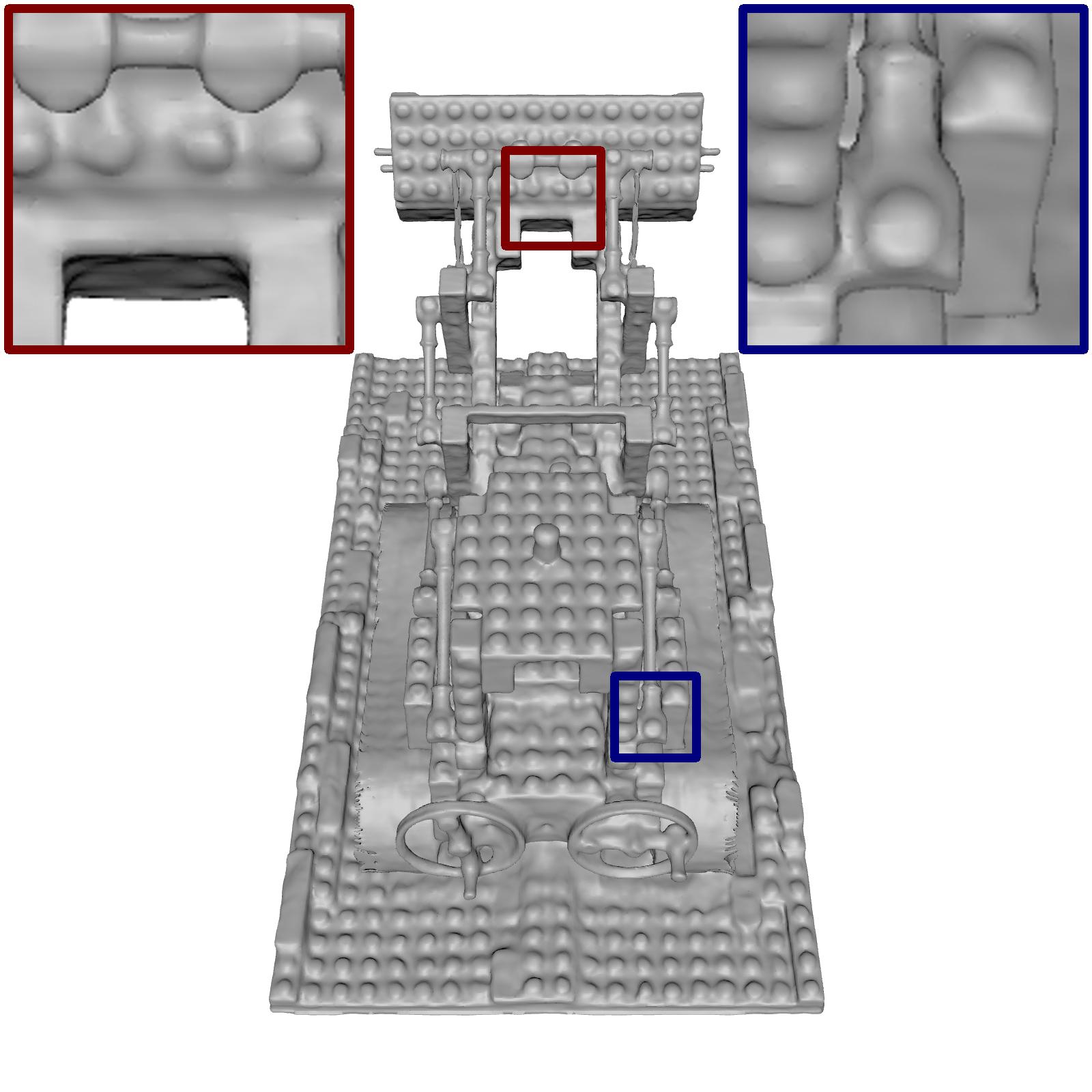}
        \\
        \textbf{Ground truth} & \hspace{\mrg}
        \textbf{NeuralWarp} & \hspace{\mrg}
        \textbf{NeuralWarp (ours)}
    \end{tabular}
    \caption{Additional qualitative results on the Realistic Synthetic 360 dataset~\cite{Mildenhall2020NeRFRS} for NeuralWarp~\cite{darmon2022improving} method.}
    \label{fig:nerf_qual_appendix_nwarp}
\end{figure*}

\begin{figure*}
    \centering    
    \setlength{\wid}{0.23\textwidth}
    \setlength{\mrg}{-0.45cm}
    \setlength{\mrgv}{-0.05cm}
    \begin{tabular}{c cc}
        \vspace{\mrgv}
        \includegraphics[width=\wid]{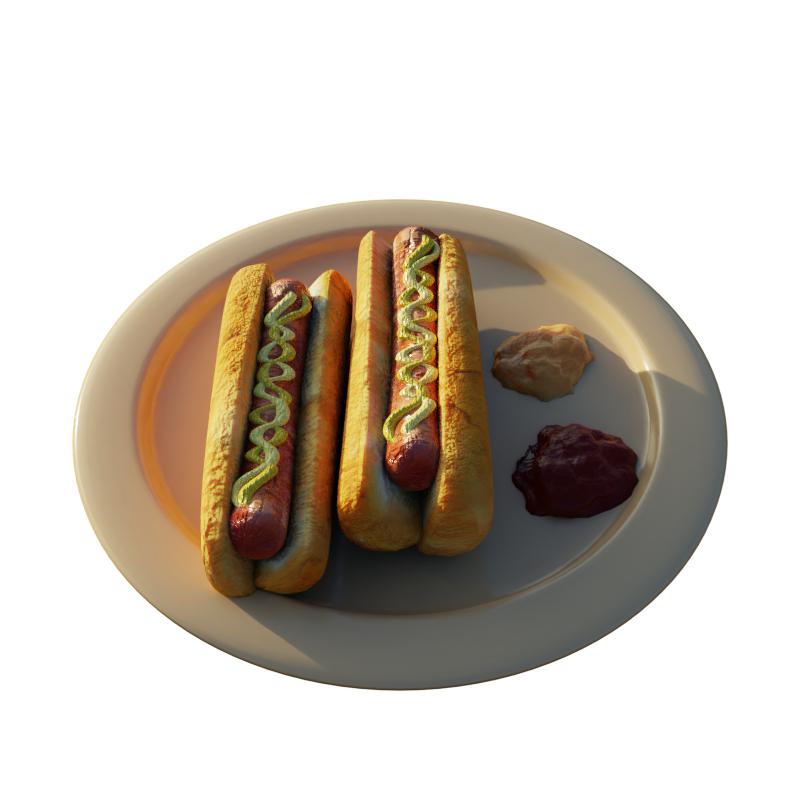} &
        \hspace{\mrg}
        \includegraphics[width=\wid]{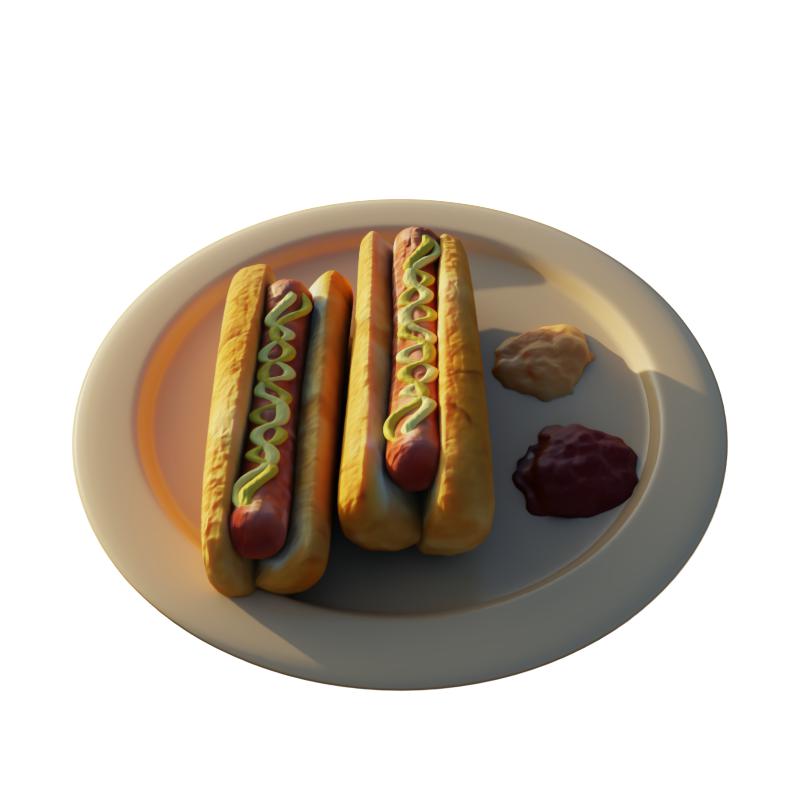} &
        \hspace{\mrg}
        \includegraphics[width=\wid]{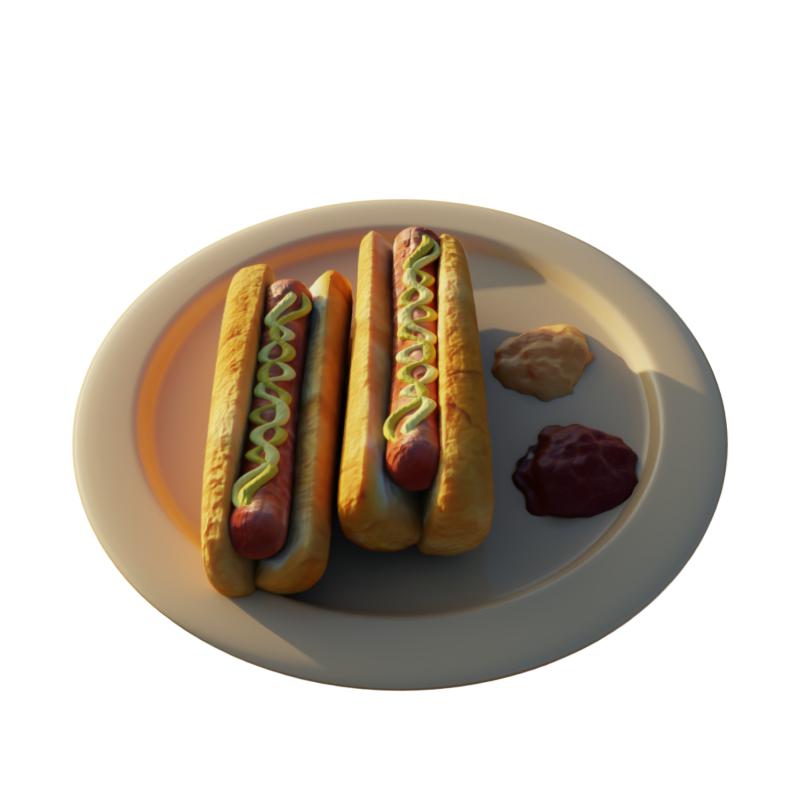}
        \\ 
        \vspace{\mrgv}
        \includegraphics[width=\wid]{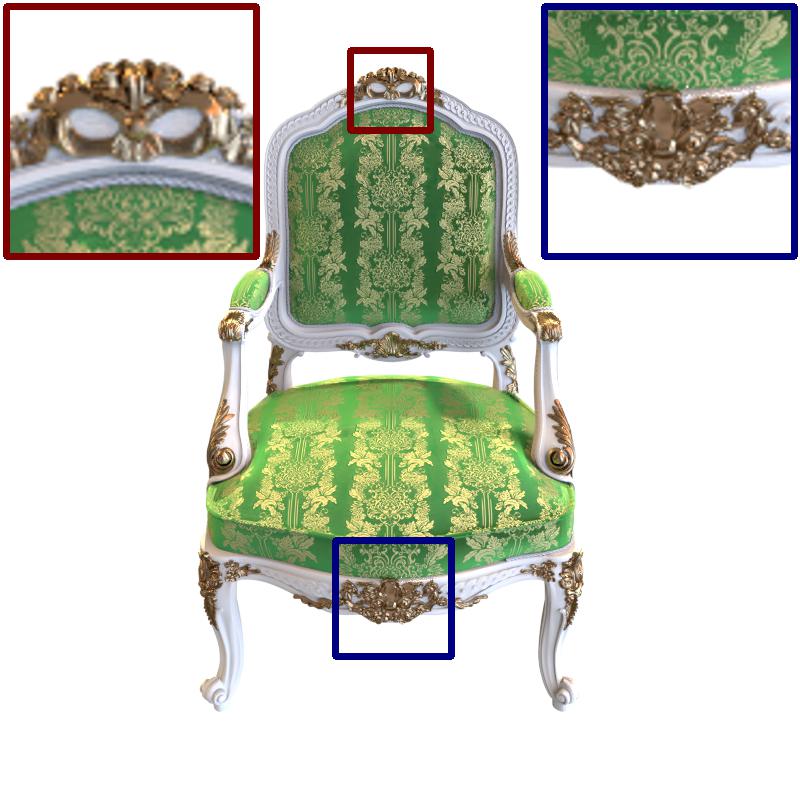} &
        \hspace{\mrg}
        \includegraphics[width=\wid]{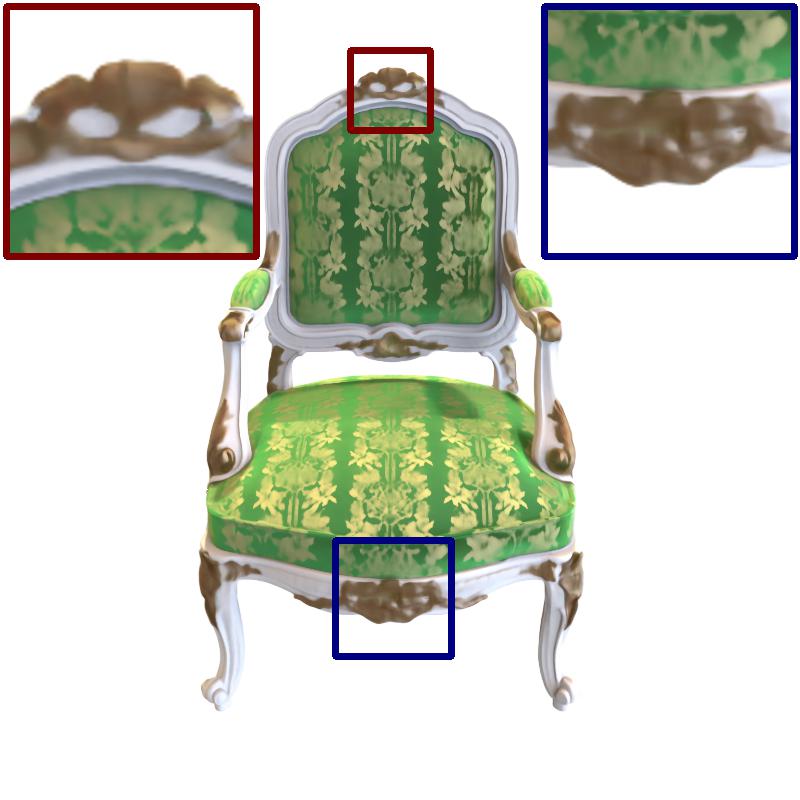} &
        \hspace{\mrg}
        \includegraphics[width=\wid]{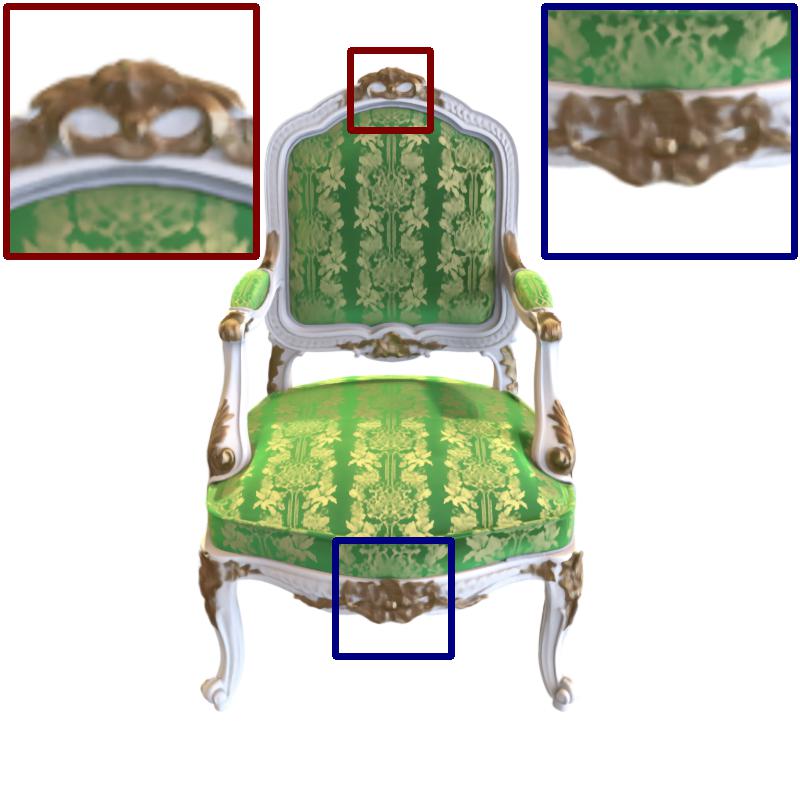}
        \\ 
        \vspace{\mrgv}
        \includegraphics[width=\wid]{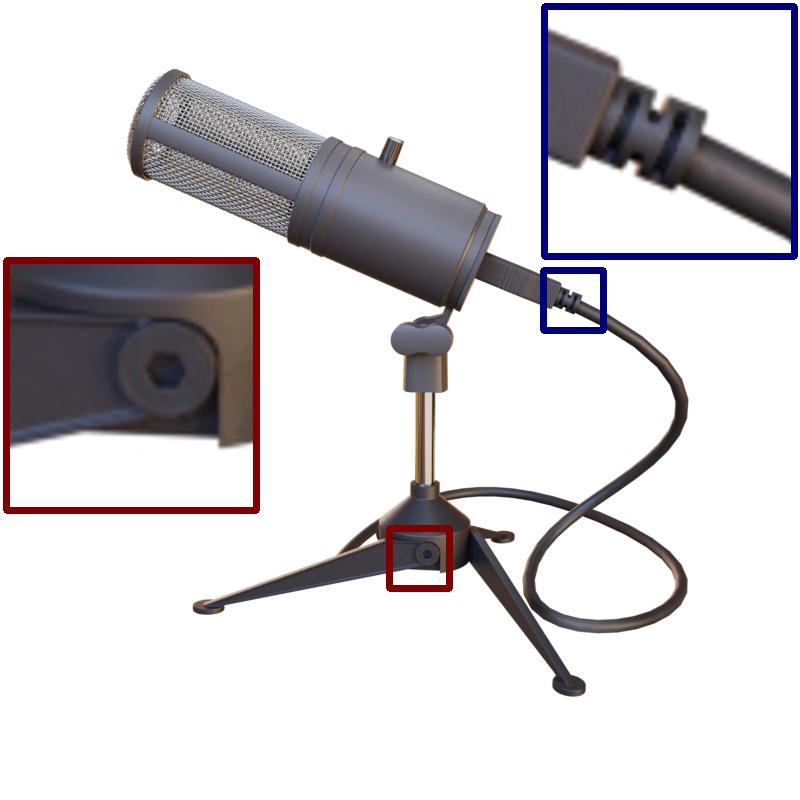} &
        \hspace{\mrg}
        \includegraphics[width=\wid]{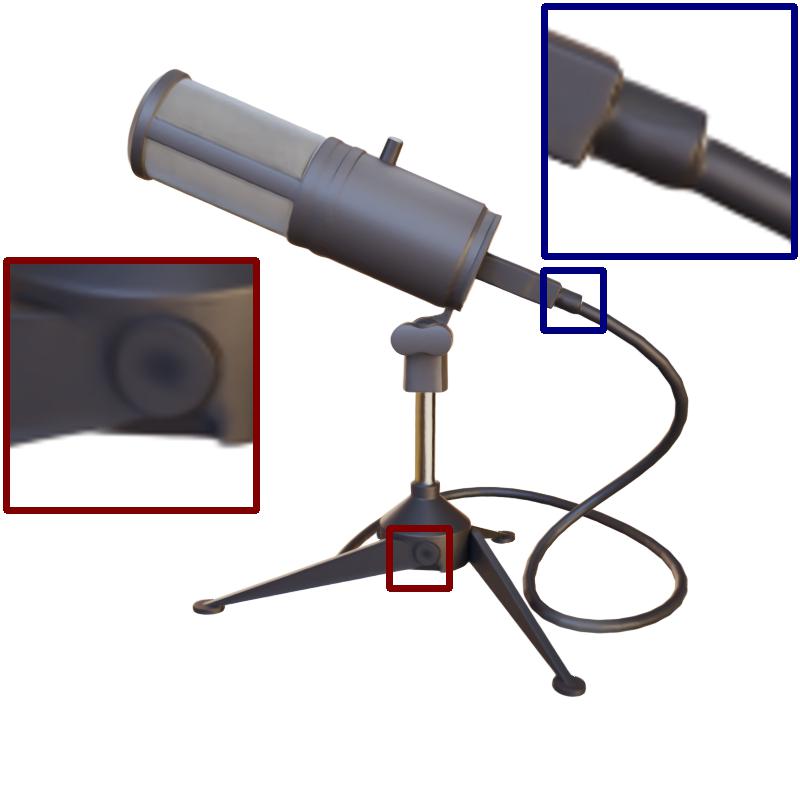} &
        \hspace{\mrg}
        \includegraphics[width=\wid]{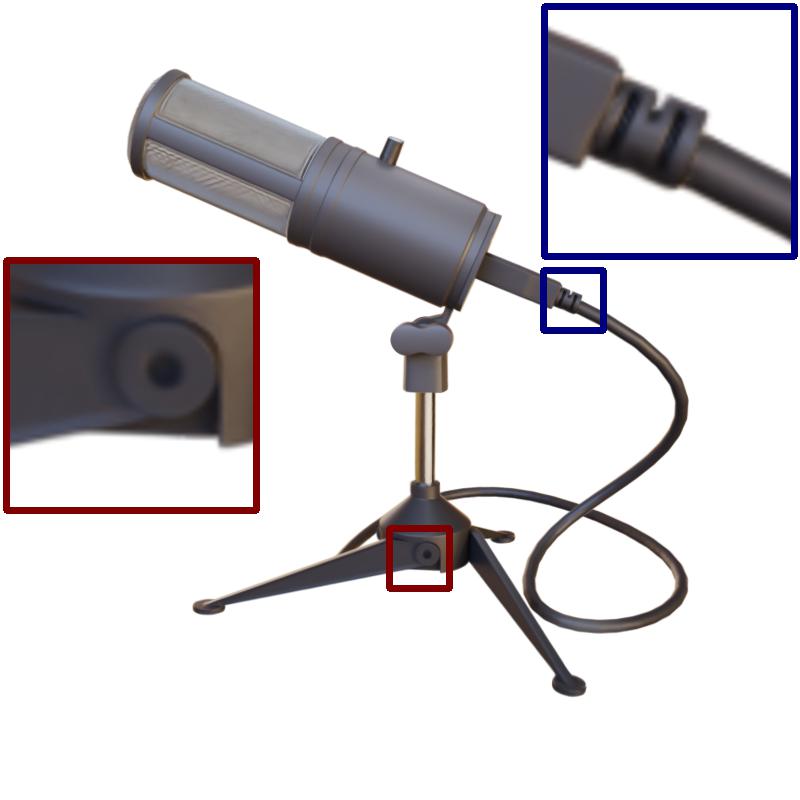}
        \\ 
        \vspace{\mrgv}
        \includegraphics[width=\wid]{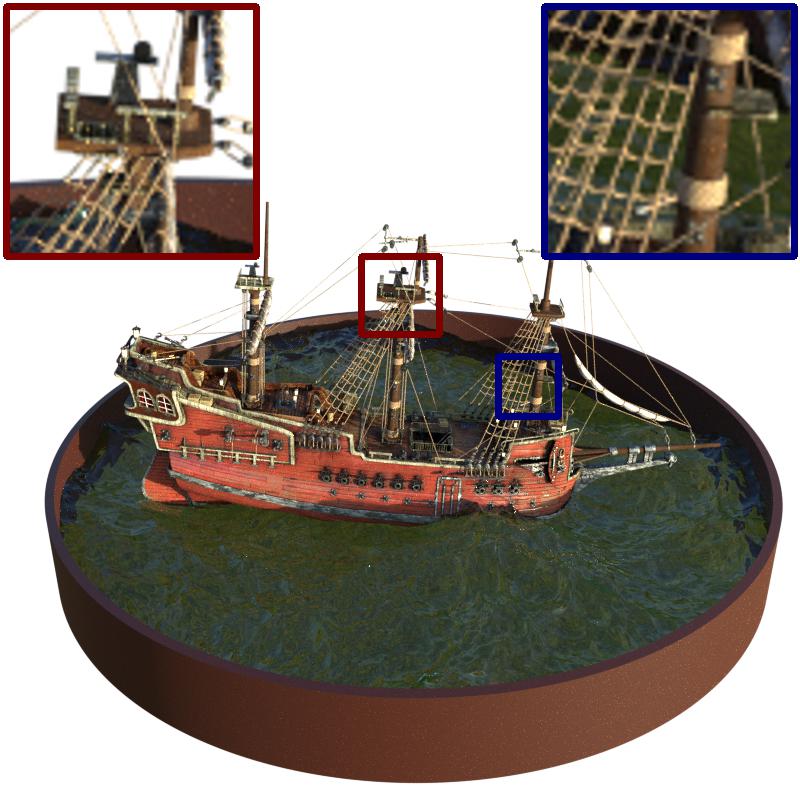} &
        \hspace{\mrg}
        \includegraphics[width=\wid]{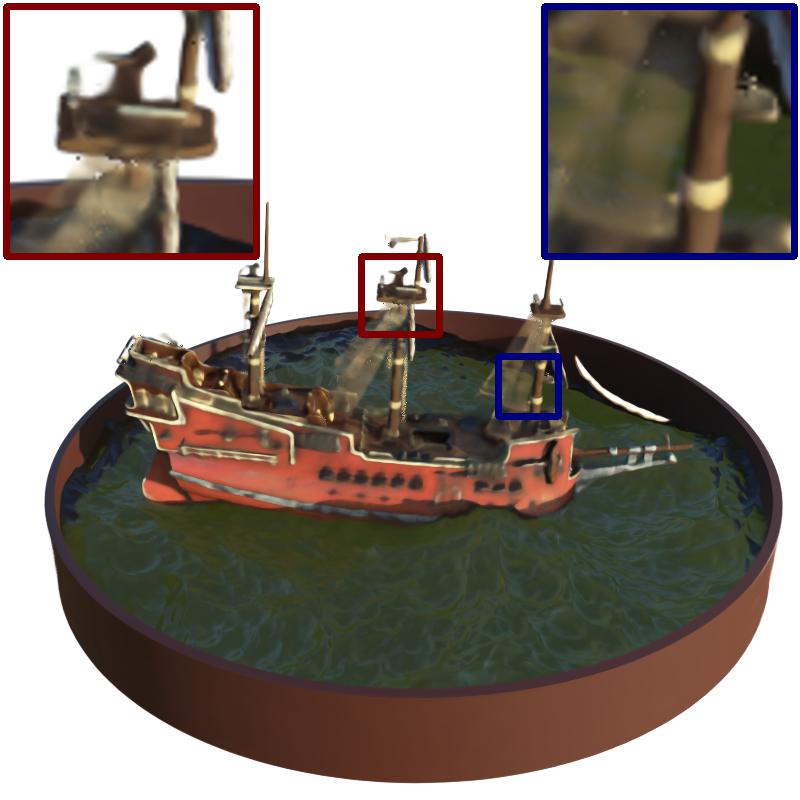} &
        \hspace{\mrg}
        \includegraphics[width=\wid]{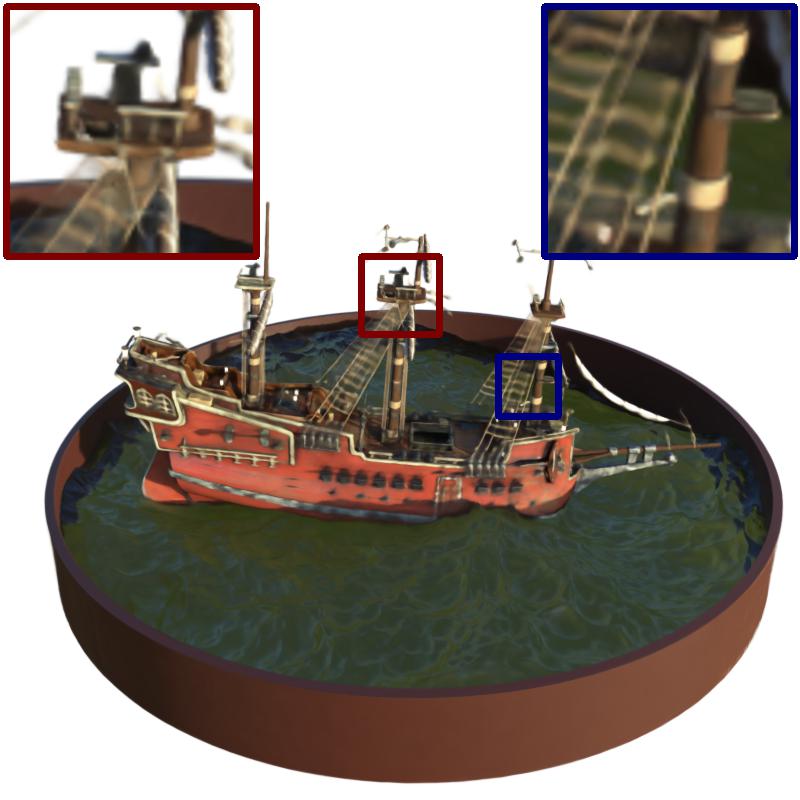}
        \\
        \vspace{\mrgv}
        \includegraphics[width=\wid]{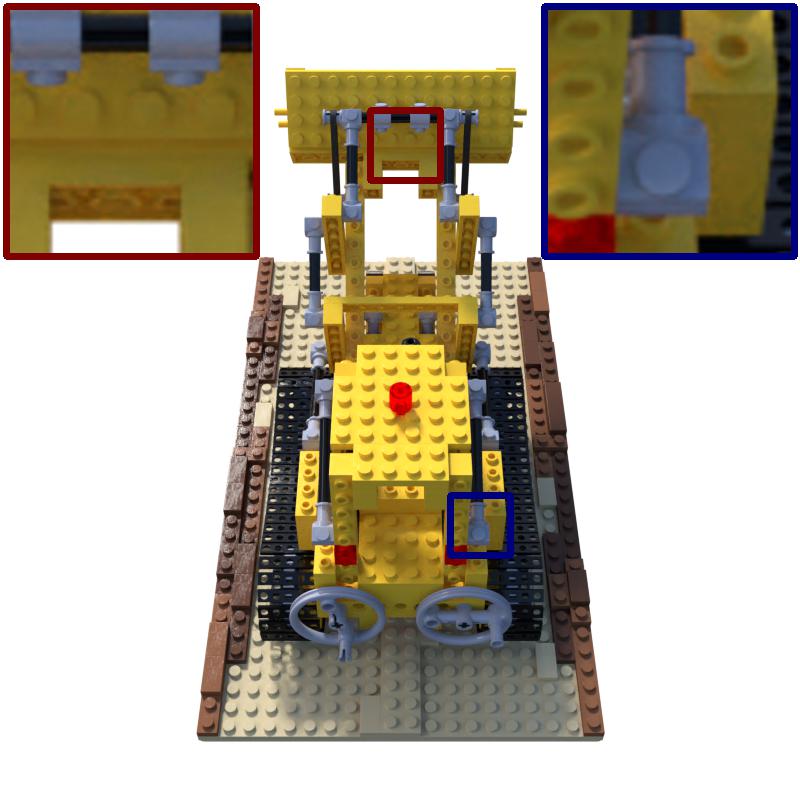} &
        \hspace{\mrg}
        \includegraphics[width=\wid]{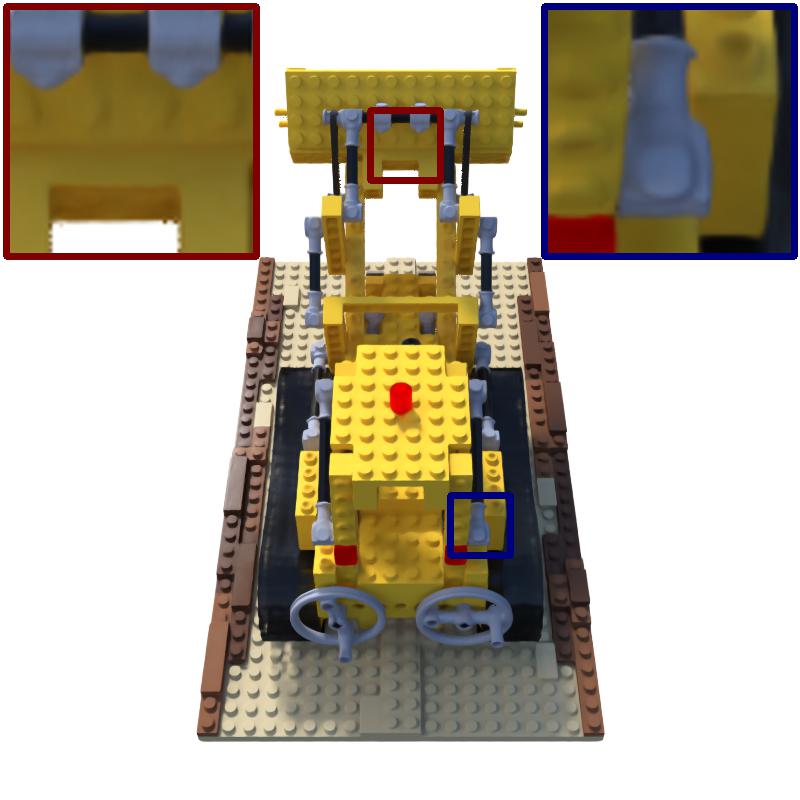} &
        \hspace{\mrg}
        \includegraphics[width=\wid]{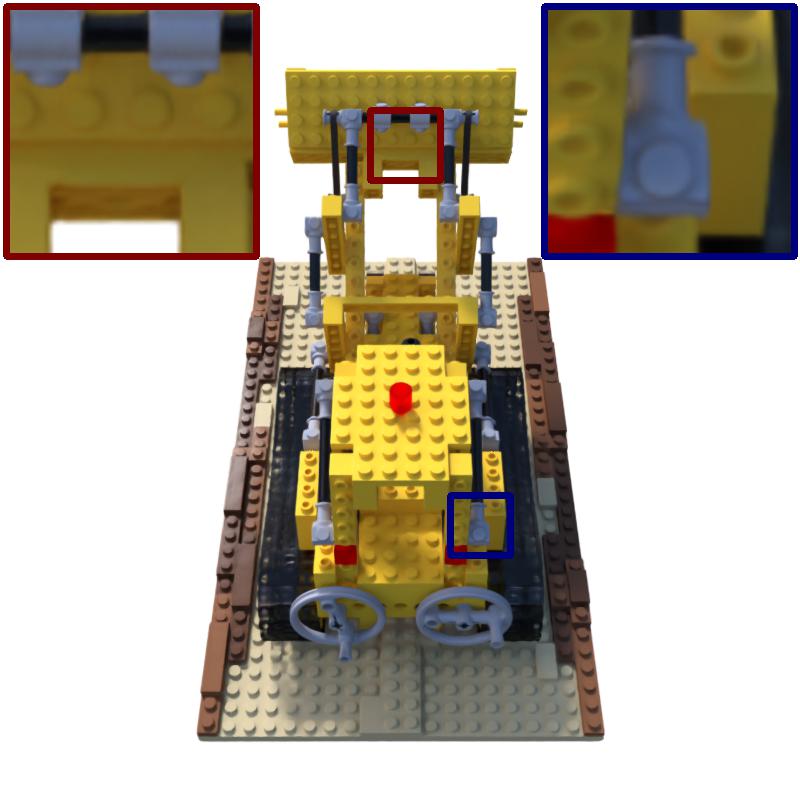}
        \\
        \textbf{Ground truth} & \hspace{\mrg}
        \textbf{NeuS} & \hspace{\mrg}
        \textbf{NeuS (ours)}
    \end{tabular}
    \caption{Rendering results on the Realistic Synthetic 360 dataset~\cite{Mildenhall2020NeRFRS} for NeuS~\cite{Wang2021NeuSLN} method.}
    \label{fig:nerf_qual_renders_renders_appendix_neus}
\end{figure*}

\begin{figure*}
    \centering    
    \setlength{\wid}{0.23\textwidth}
    \setlength{\mrg}{-0.45cm}
    \setlength{\mrgv}{-0.05cm}
    \begin{tabular}{c cc}
        \vspace{\mrgv}
        \includegraphics[width=\wid]{figures/nerf_qual_renders/gt/hotdog.jpg} &
        \hspace{\mrg}
        \includegraphics[width=\wid]{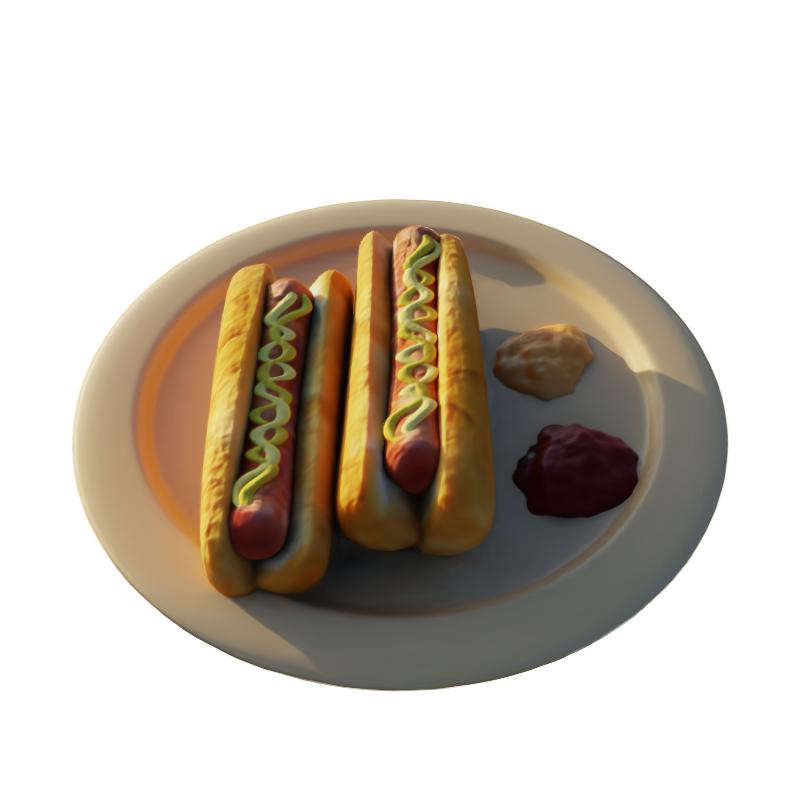} &
        \hspace{\mrg}
        \includegraphics[width=\wid]{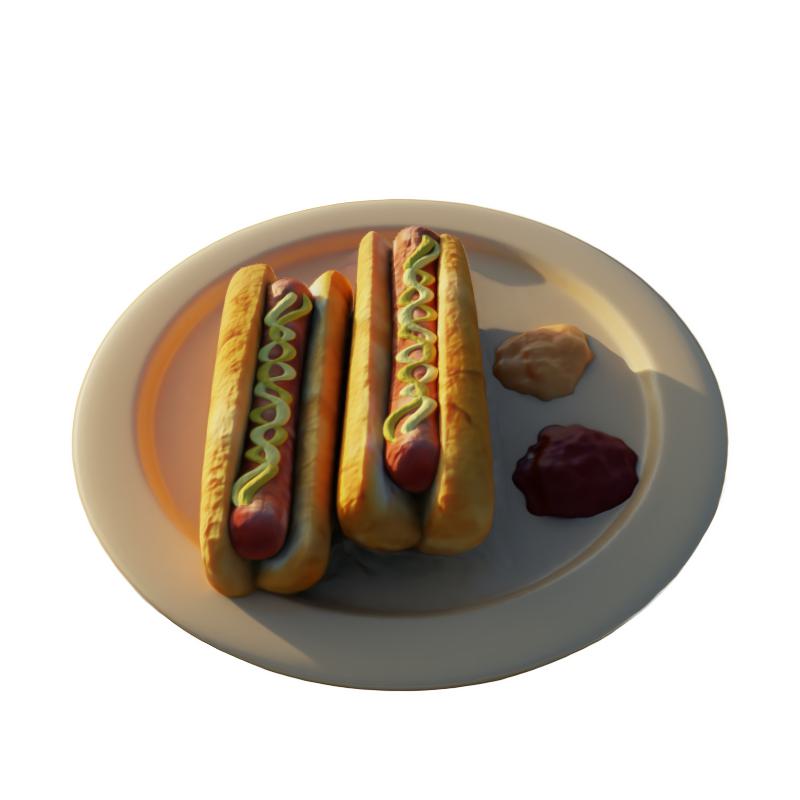}
        \\ 
        \vspace{\mrgv}
        \includegraphics[width=\wid]{figures/nerf_qual_renders/gt/chair.jpg} &
        \hspace{\mrg}
        \includegraphics[width=\wid]{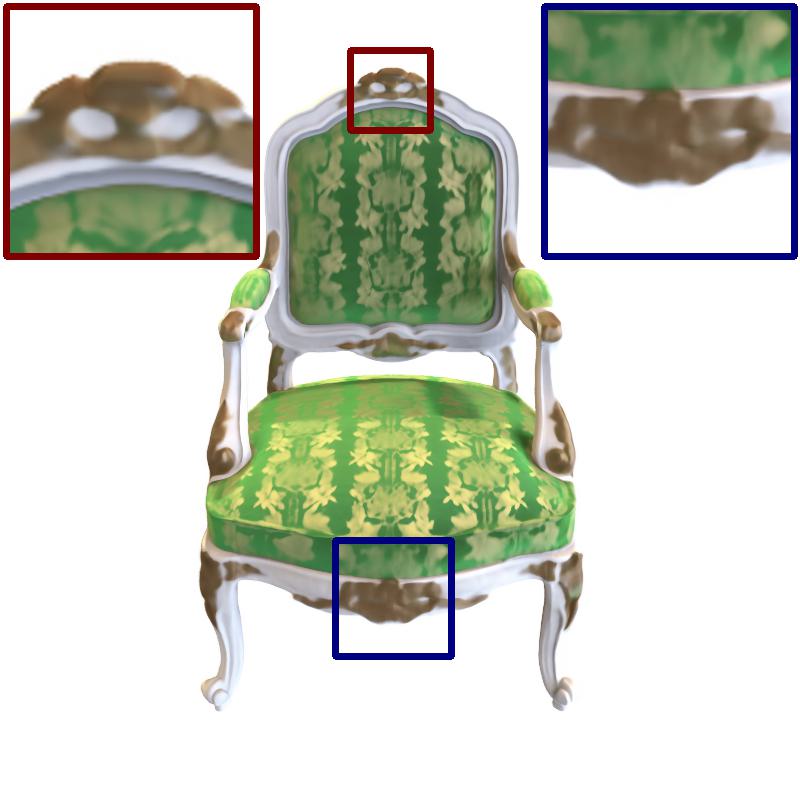} &
        \hspace{\mrg}
        \includegraphics[width=\wid]{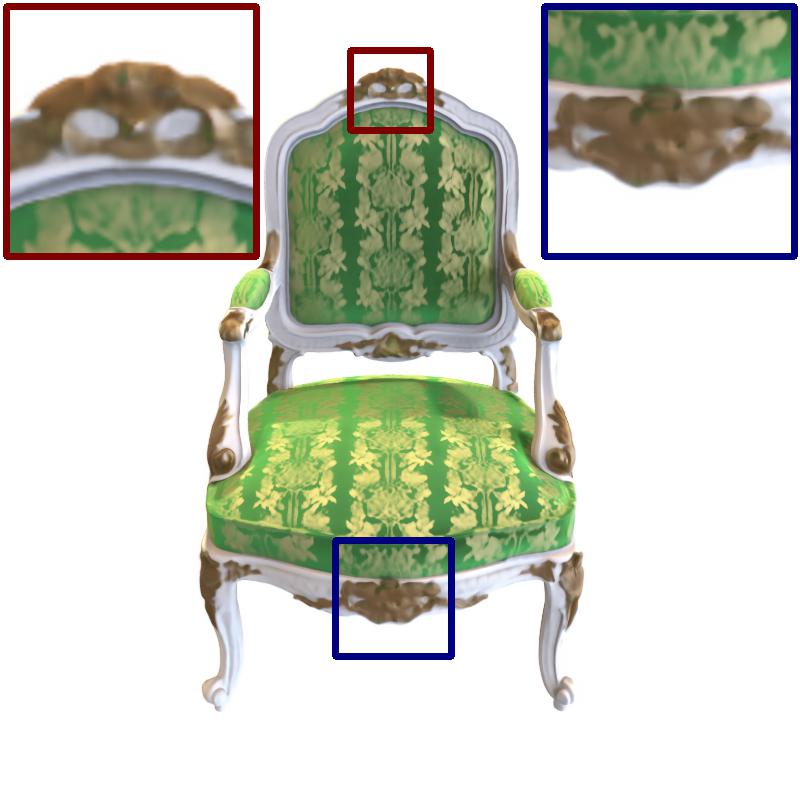}
        \\ 
        \vspace{\mrgv}
        \includegraphics[width=\wid]{figures/nerf_qual_renders/gt/mic.jpg} &
        \hspace{\mrg}
        \includegraphics[width=\wid]{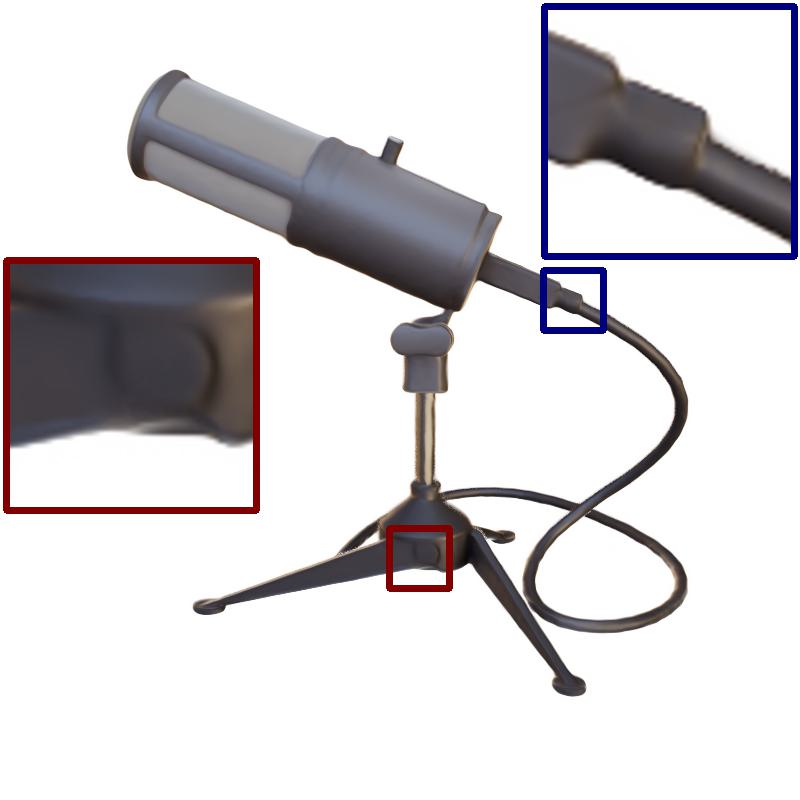} &
        \hspace{\mrg}
        \includegraphics[width=\wid]{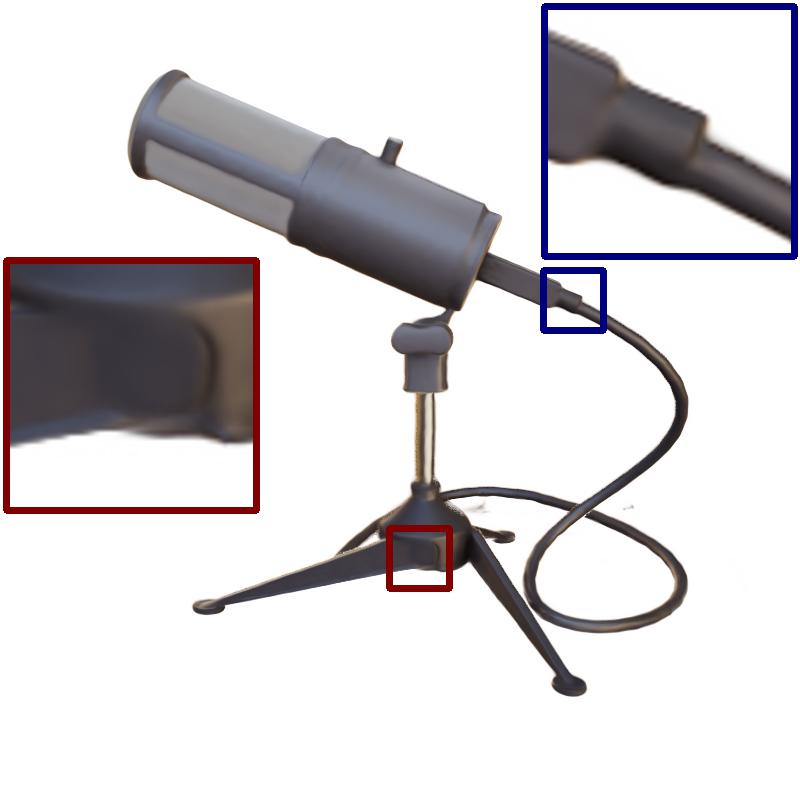}
        \\ 
        \vspace{\mrgv}
        \includegraphics[width=\wid]{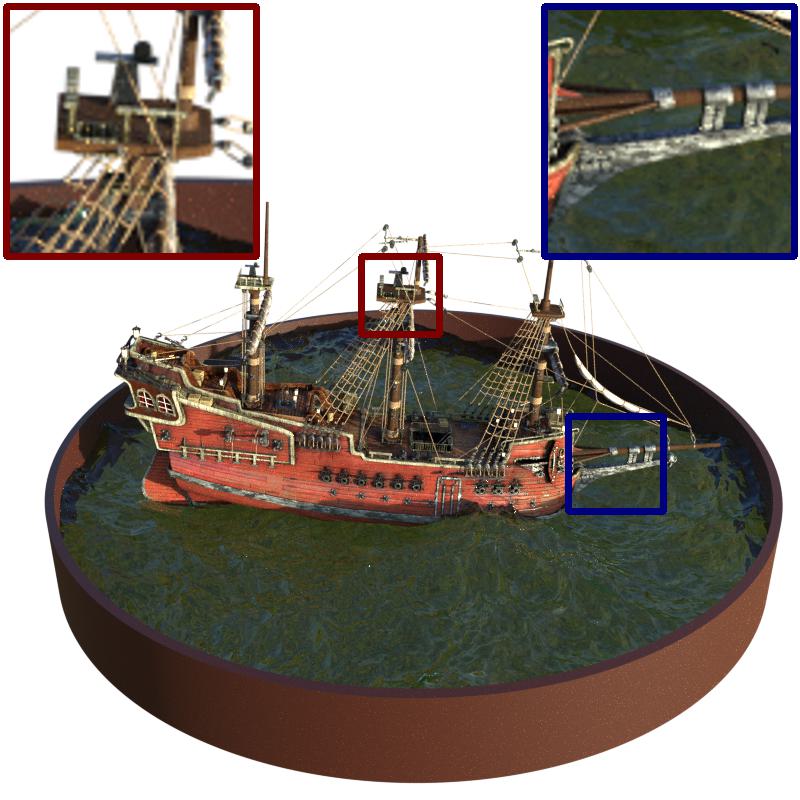} &
        \hspace{\mrg}
        \includegraphics[width=\wid]{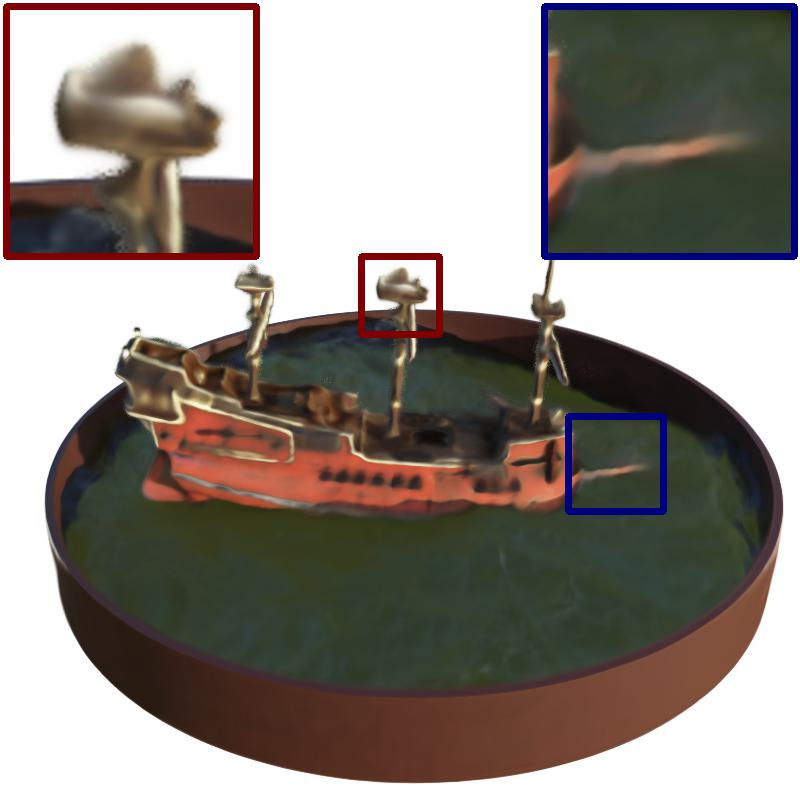} &
        \hspace{\mrg}
        \includegraphics[width=\wid]{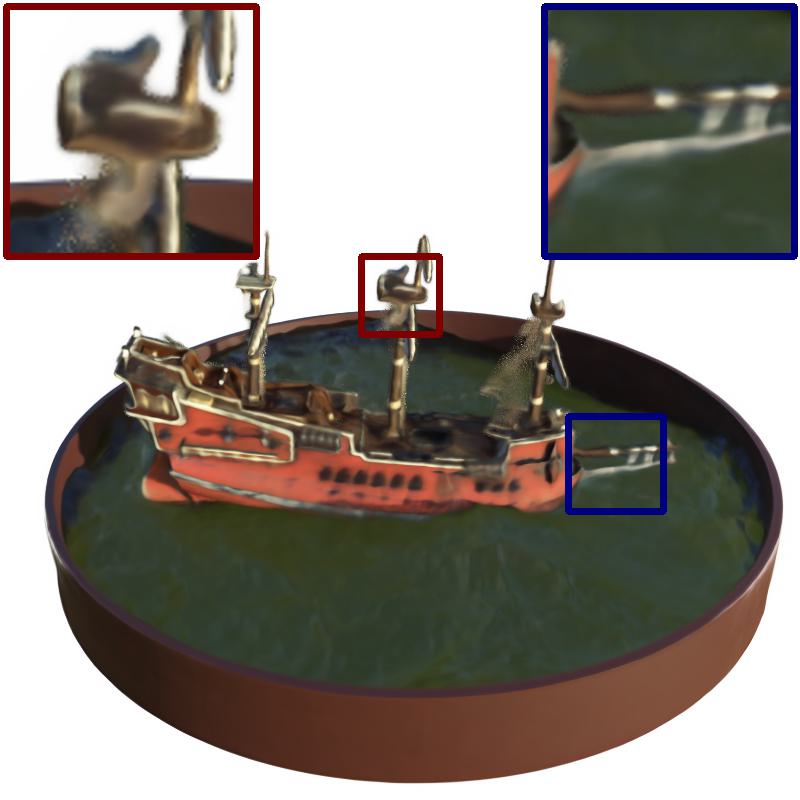}
        \\
        \vspace{\mrgv}
        \includegraphics[width=\wid]{figures/nerf_qual_renders/gt/lego.jpg} &
        \hspace{\mrg}
        \includegraphics[width=\wid]{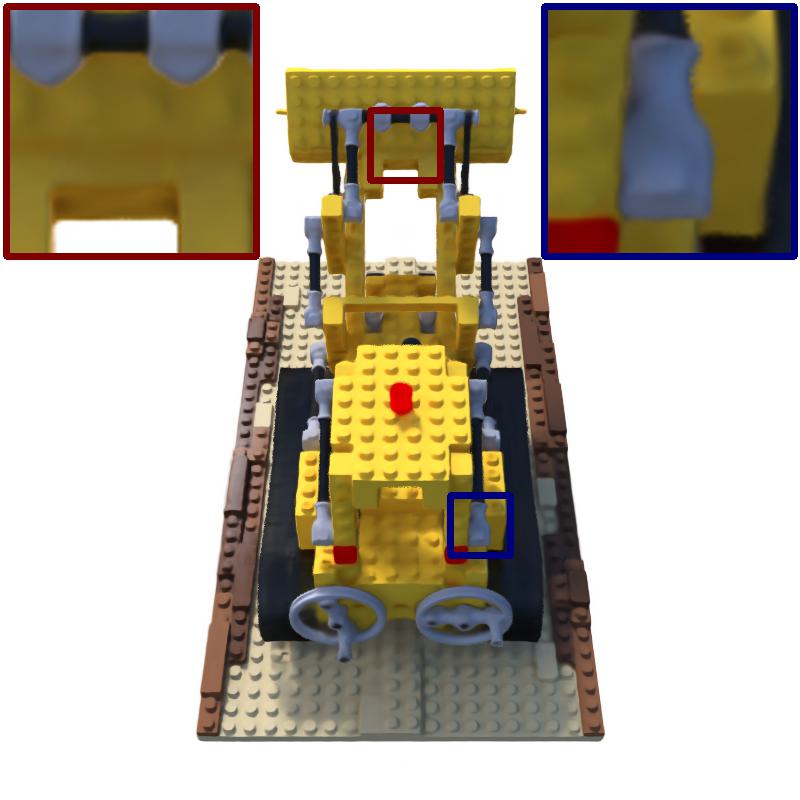} &
        \hspace{\mrg}
        \includegraphics[width=\wid]{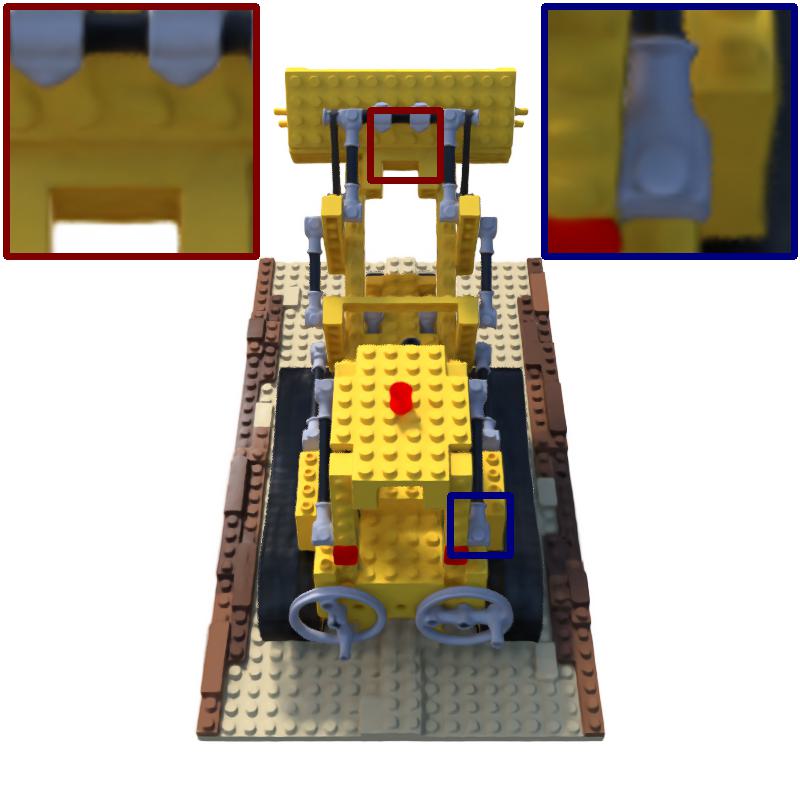}
        \\
        \textbf{Ground truth} & \hspace{\mrg}
        \textbf{NeuralWarp} & \hspace{\mrg}
        \textbf{NeuralWarp (ours)}
    \end{tabular}
    \caption{Rendering results on the Realistic Synthetic 360 dataset~\cite{Mildenhall2020NeRFRS} for NeuralWarp~\cite{darmon2022improving} method.}
    \label{fig:nerf_qual_renders_renders_appendix_nwarp}
\end{figure*}

Next, we include qualitative results for additional scenes of the Realistic Synthetic 360 dataset~\cite{Mildenhall2020NeRFRS} in Figures~\ref{fig:nerf_qual_appendix_neus}-\ref{fig:nerf_qual_appendix_nwarp}, as well as renders in Figures~\ref{fig:nerf_qual_renders_renders_appendix_neus}-\ref{fig:nerf_qual_renders_renders_appendix_nwarp}. We observe that our method not only achieves superior quality of reconstruction but also renders, which is confirmed by the image metrics presented in Table \ref{tab:nerf_quant_render}. We report the PSNR, SSIM, and LPIPS \cite{Zhang2018TheUE} metrics evaluated on the 200 test views provided in the dataset (not seen during training). 

Additionally, in Figure~\ref{fig:convergence} we present the learning curves for both our and the NeuS base model when trained for more iterations on one of the scenes. This experiment confirms that our approach indeed achieves a better local optima, and not just increases the convergence speed.

\paragraph{Ellipse-guided training.} 

We additionally attempted to further increase the sampling efficiency by using a cloud of ellipses instead of spheres to encapsulate the learned surface. We set the ellipses to have their two major axes equal to the scheduled radius and be aligned with the tangent plane, while the remaining minor axis to be aligned with the surface normal. This would allow us to increase the number of samples along the ray with nonzero opacity even further, compared to the sphere-based sampling. However, after running the preliminary experiments with different scaling factors which define the length of the minor axis, we observed no further improvements compared to the sphere-based training. We argue it to be caused by the high efficiency of the combination of our proposed sphere-based training and importance sampling, which is part of the base method.

\end{document}